# The Pacific Knowledge Acquisition Workshop

# 2004

Auckland, New Zealand

August 9 – 10, 2004

Workshop Proceedings

Edited by:
Byeong Ho Kang
Achim Hoffmann
Takahira Yamaguchi
Wai Kiang Yeap

# Preface

Artificial intelligence (AI) research has evolved over the last few decades and knowledge acquisition research is at the core of AI research. PKAW-04 is one of three international knowledge acquisition workshops held in the Pacific-Rim, Canada and Europe over the last two decades. PKAW-04 has a strong emphasis on incremental knowledge acquisition, machine learning, neural nets and active mining.

The proceedings contain 19 papers that were selected by the program committee among 24 submitted papers. All papers were peer reviewed by at least two reviewers. The papers in these proceedings cover the methods and tools as well as the applications related to develop expert systems or knowledge based systems.

The success of a workshop always depends on the support of all the people involved. Therefore, the workshop co-chairs would like to thank all the people who contributed to the success of PKAW-04. First of all, we would like to take this opportunity to thank authors and participants. We wish to thank the program committee members who reviewed the papers and the volunteer students and staff, Yangsok Kim, Sungsik Park and Pauline Mak, in The University of Tasmania, for the administration of the workshop.

# <u>Workshop Committee</u>

**Workshop Honorary Chairs**
*Prof. Paul Compton* (University of New South Wales, Australia)
*Prof. Hiroshi Motoda* (Osaka University, Japan)

**Workshop Chair**
*Dr. Byeong Ho Kang* (University of Tasmania, Australia)

**Workshop Co-Chairs**
*Prof. Achim Hoffmann* (University of New South Wales, Australia)
*Prof. Takahira Yamaguchi* (Keio University, Japan)

**Program Committee**
*Einoshin Suzuki (Yokohama National University, Japan)*
*Takashi Washio (Osaka University, Japan)*
*Derek Sleeman (University of Aberdeen, UK)*
*Shusaku Tsumoto (Shimane Medical University, Japan)*
*Seiji Yamada (National Institute of Informatics, Japan)*
*Yasuhiko Kitamura (Kwansei Gakuin University, Japan)*
*Ashesh Jayantbhai Mahidadia (University of New South Wales, Australia)*
*Takashi Okada (Kwansei Gakuin University, Japan)*
*Enrico Motta (Open University, UK)*
*Rob Kremer (University of Calgary, Canada)*
*Huan Liu (Arizona State University, USA)*
*Rodrigo Martinez (University of Murcia,Okada Spain)*
*Ulrich Reimer (Business Operation Systems,Switzerland)*
*Rose Dieng (INRIA, France)*
*Toshiro Minami (Kyushu Institute of Information Sciences & Kyushu University, Japan)*
*Masayuki Numao (Osaka University, Japan)*
*Mihye Kim (University of New South Wales, Australia)*
*John Debenham (University of Technology, Sydney, Australia)*
*Tim Menzies (NASA, USA)*
*Takao Terano (University of Tsukuba, Japan)*
*Noriaki Izumi (Cyber Assist Research Center, AIST, Japan)*
*Debbie Richards (Macquarie University, Australia)*
*Mike Cameron-Jones (University of Tasmania, Australia)*
*Ray Williams (University of Tasmania, Australia)*
*Peter Vamplew (University of Tasmania, Australia)*
*Rob Colomb (University of Queensland, Australia)*
*Udo Hahn (Freiburg University, Germany)*
*Frank Puppe (University of Wuerzburg, Germany)*
*Masashi Shimbo (Nara Institute of Science and Technology, Japan)*
*Ray Hashemi (Armstrong Atlantic State University, USA)*
*George Macleod Coghill (University of Aberdeen, UK)*

# Workshop Program

**Monday 9th of August**

**Session 1: Active Learning (9:00AM – 10:30AM)**
**Chair: Byeong Ho Kang**

Finding cue expressions for knowledge extraction from scientific text: early results
*Masashi Shimbo, Sayaka Tamamori andYuji Matsumoto,*

Mining Diagnostic Taxonomy for Multi-Stage Medical Diagnosis
*Shusaku Tsumoto*

Constructing Compact Dual Ensembles for Efficient Active Learning
*Huan Liu, Amit Mandvikar and Hiroshi Motoda,*

**Session 2: Incremental KA: RDR (11:00AM-1:00PM)**
**Chair: Hiroshi Motoda**

Generalising Incremental Knowledge Acquisition
*Paul Compton, Tri Minh Cao and Julian Kerr*

Knowledge Acquisition Module for Conversational Agents
*Pauline Mak, Byeong HoKang, Claude Sammut and Waleed Kadous*

Personalized Web Document Classification using MCRDR
*Sungsik Park, Yangsok Kim and ByeongHo Kang*

Achieving Rapid Knowledge Acquisition in a High-Volume Call Centre
*Megan Vazey and Debbie Richards*

Acquisition of ArticulableTacit Knowledge
*Peter Busch and Debbie Richards*

**Session 3: KA Tools (2:00PM-3:30PM)**
**Chair: Achim Hoffman**

Incremental Knowledge Acquisition using RDR for Soccer Simulation.
*Angela Finlayson and Paul Compton*

Domain Ontology Construction with Quality Refinement
*Takeshi Morita, Yoshihiro Shigeta, Naoki Sugiura, Naoki Fukuta, Noriaki Izumi and Takahira Yamaguchi*

Developing an Intelligent Learning Tool for Knowledge Acquisition
on Problem-based Discussion
*Akcell Chiang, Isaac Pak-Wah Fung and R. H. Kemp*

**Session 4: Pannel Discussions (4:00PM-5:00PM)**
**Chair: Paul Compton**
**Title: Knowledge Acquisition for Web Services**

**Session 5: Demonstrations and Posters (5:00PM-5:30PM)**

## Tuesday 10th of August

**Session 1: View of KA/Evaluate KA methods (9:00AM–10:30AM)**
**Chair: Debbie Richard**

Towards Acquiring and Refining Class Hierarchy Design of
Web Application Integration Software
*Naoki Fukuta, Mayumi Ueno, Noriaki Izumi and Takahira Yamaguchi*

Consideration on "Educationality" of Knowledge Acquisition Support Systems
*Toshiro Minami*

A Comparative Study on Cost and Benefit of Capturing Design Rationale
*Yoshikiyo Kato and Koichi Hori*

**Session 2: KA and ML (11:00AM-1:00PM)**
**Chair: Takahira Yamaguchi**

Using Neural Network to Weight the Partial Rules: Application to Classification of
Dopamine Antagonist Molecules
*Sukree Sinthupinyo, Cholwich Nattee, Masayuki Numao, Takashi Okada and Boonserm Kijsirikul,*

Future Directions in Adapting GAs using Knowledge Acquisition
*J.P. Bekmann and Achim Hoffmann*

Incremental Learning of Control Knowledge for Lung Boundary Extraction
*Avishkar Misra, Arcot Sowmya and Paul Compton*

E-Mail Document Categorization Using BayesTH-MCRDR Algorithm:
Empirical Analysis and Comparison with Other Document Classification Methods
*Woo-Chul Cho and Richards, Debbie*

Detecting the Knowledge Frontier: An Error Predicting Knowledge Based System
*Richard Peter Dazeley and Byeong-Ho Kang*

# Table of Contents

## View of Knowledge Acquisition / Evaluate Knowledge Acquisition methods

## Knowledge Acquisition and Machine Learning

# Finding Cue Expressions for Knowledge Extraction from Scientific Text: Early Results

Masashi Shimbo, Sayaka Tamamori, and Yuji Matsumoto

Graduate School of Information Science
Nara Institute of Science and Technology
8916-5 Takayama, Ikoma, Nara 630-0192, Japan
{sayaka-t,shimbo,matsu}@is.naist.jp

**Abstract.** This paper investigates whether and how natural language processing and data mining techniques can be utilized for locating desired knowledge in a large text collection. This task amounts to finding cue words and phrases indicating the location of knowledge, where the challenge is to establish a methodology that can cope with the diversity of expressions. We examine the feasibility of mining cue expressions from the syntactic dependency structure obtained from parsed sentences. As a case study, the (phrasal) expressions concerning a variety of tests related to chronic hepatitis were sought in the Medline abstracts. We observed that dependency analysis helped to narrow down the candidates for verbal expressions, although it was ineffective for other types of expressions.

## 1 Introduction

With the recent growth in the number of text collections available in digital form, there has been increased interest in mining useful knowledge buried in a volume of text. In particular, knowledge extraction from medical literature is appealing from the standpoint of evidence-based medicine (EBM) [13], which practices "integrating individual clinical expertise with the best available external clinical evidence from systematic research" [12]. A source of external evidence is assumed to be clinically relevant research literature, and thus EBM is an immediate application of knowledge extraction from medical text.

The Medline database [16], available from the U. S. National Library of Medicine covers over 11 million bibliographic citations from more than 4 thousand research journals world-wide. It has been a standard corpus for medical knowledge extraction, as a large number of citations contain abstracts as well.

Previous work on knowledge extraction from Medline includes as follows. Blaschke et al. [2] extracted protein interaction relationships using the simple cue patterns of the form 'PROTEIN VERB PROTEIN,' where VERB includes 14 verbs indicating actions, such as *activate*, *bind*, and *suppress*. Rindflesch et al. [11] uses the specific predicate *bind* as a cue, and extracted binding relationships between macromolecules. Khoo et al. [7] attempted to identify the location of causal relationship description using the dependency subtree patterns.

Cue patterns, which work as an indicator of the location of desired knowledge, depend on the domain of text as well as the type of desired knowledge. Hence the first step in knowledge extraction is to find effective cue patterns suitable for the domain and goal at hand.

However, no previous work has, to our knowledge, addressed how cue patterns can be efficiently identified. In all the literature cited above, cue patterns are given a priori, presumably devised by domain experts for the prescribed tasks[1]. It is true that the current technology does not admit finding effective cue patterns without human supervision, yet it should still be possible to narrow the number of candidate patterns from which human experts can sift with less efforts.

Moreover, most previous work views knowledge extraction from text as a cascaded process. In practice, it is rather a process involving a feedback; it iterates the subprocesses of identifying cue patterns, matching them to text, and evaluating the feasibility of matched passages. If cue patterns are too general, they should generate too many irrelevant passages to be inspected by humans; if they are too restrictive to the contrary, they should generate too few. In either case, cue patterns must be revised and the whole process must be reiterated.

Motivated by the above observations, we pursue a methodology to help human experts identify cue patterns effectively. As a case study, the problem of finding cue patterns in the domain of diagnosis tests for hepatitis is addressed.

## 2 Methodology

The major obstacle in collecting cue patterns is the diversity of semantically equivalent expressions. Consider retrieving Medline to see whether gradual increases in ADA level correlate with a certain change in the condition of a patient with chronic hepatitis. It is desirable to know typical expressions used for representing increases in ADA level, because it would reduce the volume of text that should be examined. However, there are a variety of verbs representing value increase in English, such as *increase*, *raise*, and *elevate*, to name a few. In addition, the increase may not be represented with a verb.

Hence, we would like to enumerate as many expressions potentially relevant to the user's objective as possible, yet without imposing on the domain experts a significant increase in the load to sift through the enumerated expressions. This goal leads to a trade-off. Increasing the number of patterns for a better cover rate leads to an enormous number of candidate passages.

Another challenge is how to present the enumerated passages to the domain experts. Since the number of passages are often huge, it is desirable to present only the relevant portion, rather than the whole sentence or abstract. The question remains on how such *relevant* portions can be determined.

To address these issues, we use syntactic dependency structure trees for representing cue patterns. A dependency tree bears information richer than the original sentence viewed simply as a string or a bag of words. Exploiting the structure within the tree allows us a fine-grained control over determining the relevant portions to be presented to the domain experts.

[1] On the other hand, Thomas et al. [15] indicated explicitly that they collected common ways of describing protein interactions through the analysis of 200 abstracts by hand.

In this paper, we concentrate on processing at the sentence and sub-sentence levels, and do not deal with knowledge described over two or more sentences. This decision reflects the reported effectiveness of the text processing units in a text mining task [5].

## 3 Enumerating expressions relevant to hepatitis

### 3.1 Objective and applications

The long-term goal of our project is to help screening the association rules mined separately from time-series data on hepatitis-related tests.

Since data mining techniques typically output a number of association rules most of which do not make sense nor are novel, the cost of sifting these generated rules is often prohibitive. Even if a rule is supported by a vast amount of data, it may just represent a piece of common-sense knowledge, or it may already be known to the public by prior work thus having no novelty [8, 10]. Such knowledge, if obtained from published papers or their abstracts, should make it possible to filter out those ’uninteresting’ rules.

Note that although there is a volume of literature (e.g., [1, 6, 9]) in the data mining community addressing the *interestingness* of mined rules, whether a rule is publicly known or not cannot be detected by these techniques, as they rely on statistical tests based on the same data used for mining rules. Whether a rule is covered by prior work can only be determined with reference to the work, which are generally published in the form of text documents.

### 3.2 Identifying expressions through syntactic dependency structure analysis

The issues discussed in Section 2 are closely related to the language used for representing cue patterns. As we mentioned earlier, we use a syntactic dependency tree but there are other possible choices, including

- Contiguous $n$ words: frequent series of $n$ words.
- Non-contiguous word sequences: frequent sequences of (non-contiguous) words.

However, the contiguous $n$-word representation is inflexible in that it cannot absorb the variations arising from insertion and omission of modifiers, while non-contiguous word sequences are prone to generate meaningless sequential patterns that consist only of individually frequent words, thus requiring extra post-processing to filter out these patterns.

On the other hand, the syntactic dependency tree, which is a form of syntactic parse trees, (i) allows modifiers to be easily removed by exploiting the structure of the tree, and (ii) indirect and direct dependence between words are represented as the locality in the dependency tree, and therefore meaningful portions of the sentences are easier to extract.

Based on these arguments, we use a dependency structure as our language for representing patterns.

## 4 Procedure for identifying cue expressions

We are interested in correlations among the outcomes of clinical tests related to hepatitis and the conditions of the disease. Hence we focus on the pattern of the form '$NP_1$ V $NP_2$' or '$NP_2$ V $NP_1$', where $NP_1$ contains the name of a clinical test, and V is a verbal expression (base verb phrase), and $NP_2$ is another noun phrase, presumably containing other diagnostic tests and the conditions of patients. Syntactic dependency analysis of a sentence admits extraction of these phrases, as it reveals the hierarchical structure among words within a sentence.

Our procedure for identifying cue patterns can be decomposed into four steps:

1. Keyword-based filtering of sentences.
2. Dependency structure analysis of the filtered sentences.
3. Expression extraction from syntactic dependency trees.
4. Filter and rank extracted expressions and hand over to the domain experts for further review.

The rest of this section will delineate each step.

### 4.1 Step 1: keyword filtering

Since syntactic dependency parsing is a computationally intensive task, it is not feasible to apply this process to the whole text collection. Hence we restricted the candidate sentences by using simple keyword matches.

We first filter abstracts containing the word *hepatitis* from the corpus. We then segment these abstracts into sentences, and further filter the sentences containing the names of the diagnostic tests of our interest. The keywords used for filtering are the names of the 660 clinical tests for diagnosing hepatitis, and are the same as the features used in [10] for mining association rules from time-dependent data. They consist of 503 different diagnosis tests[2], such as *glutamic pyruvic transaminaze*, and *glutanic oxalacetic transaminase*. The rest is their synonyms and abbreviations, e.g., *GPT* and *GOT*.

### 4.2 Step 2: dependency structure analysis

There are several ways to obtain syntactic dependency structure trees. In this paper, we take the same method as used in our previous work [14]. We first apply a phrase structure parser to the sentences filtered in Step 1 to obtain phrase structure trees. Charniak parser [3] was used as the phrase structure parser. This parser boasts approximately 90% accuracy at the phrase structure level, when applied to the Wall Street Journal corpus. The dependency structure trees are then obtained by extracting word dependencies from the phrase structure trees.

We illustrate the translation process using the phrase structure subtree in Fig. 1. This tree will eventually be translated into the dependency structure tree depicted in Fig. 3.

Each non-leaf node in a phrase structure tree is labeled with a syntactic category, and each leaf node is labeled with a surface word. In Fig. 1, syntactic categories are typeset in sans serif; e.g., $\mathsf{NP}$ (noun phrase), $\mathsf{VP}$ (verb phrase) and $\mathsf{PP}$ (prepositional phrase). A parent-child relationship in phrase structure trees corresponds to the application of a context-free grammar rule. To be precise, given a parent node and its $n$ children, let $u$ be the syntactic category of a parent node, and $v_1, \ldots, v_n$ be the syntactic categories (or surface words) of the $n$ children. This relationship represents the application of a context-free (CFG) grammar rule $u \rightarrow v_1, \ldots, v_n$.

[2] Provided by courtesy of Chiba University Hospital and Shizuoka University.

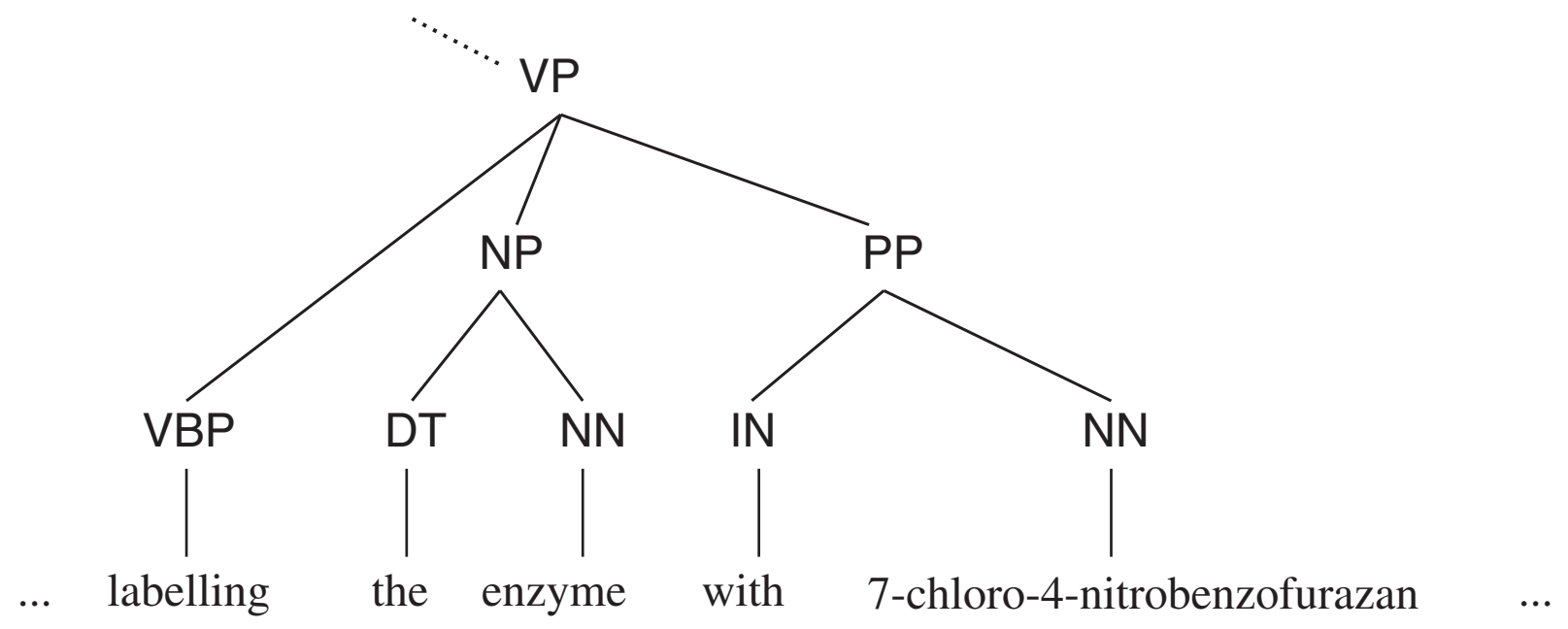

**Fig. 1.** Phrase structure subtree. Leaf nodes correspond to surface words, and each non-leaf node is labeled with a syntax category.

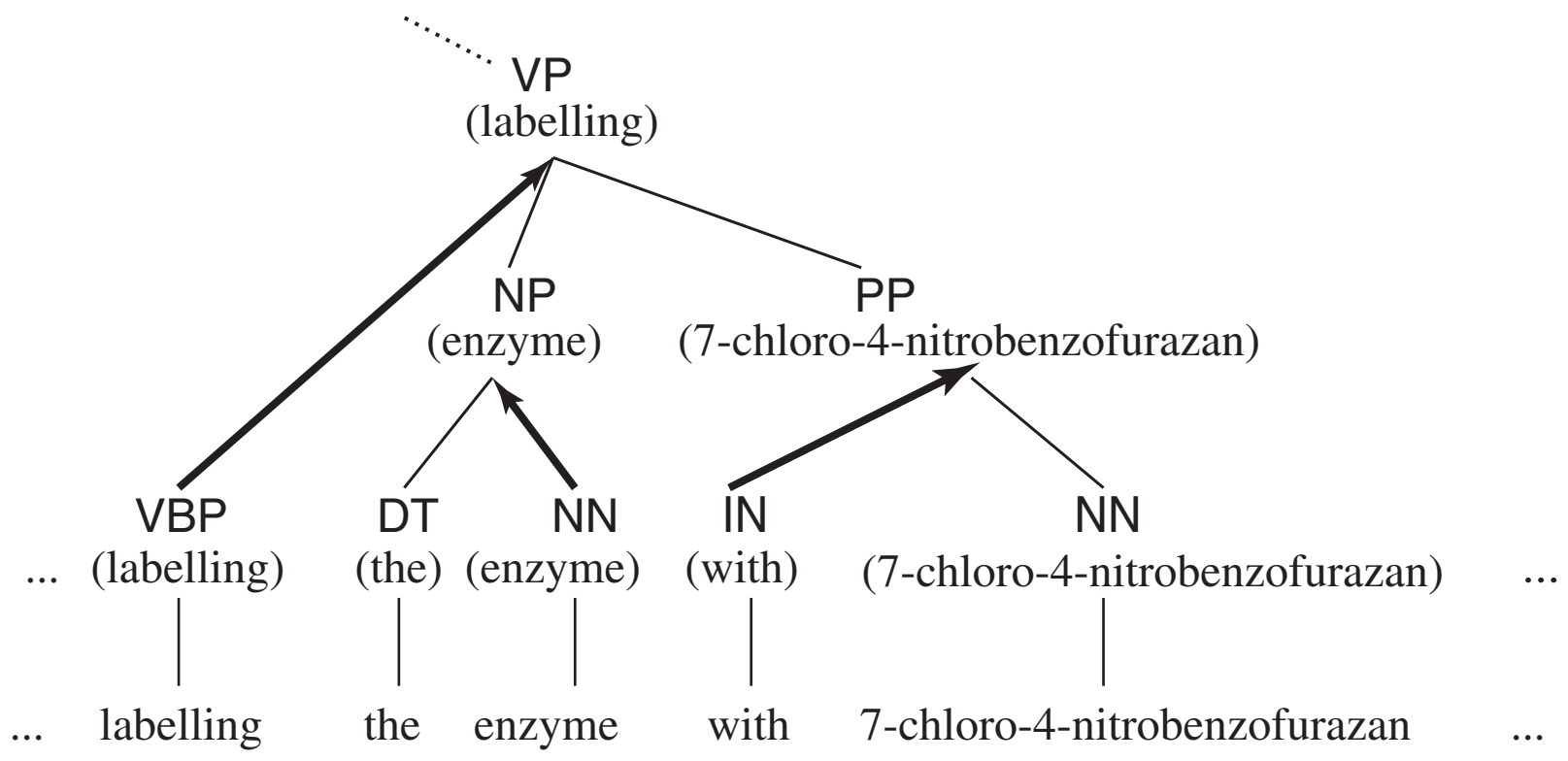

**Fig. 2.** Phrase structure subtree labeled with headwords. Bold arrows depicts the inheritance of head words by the head rules, and inherited head words are shown in parentheses.

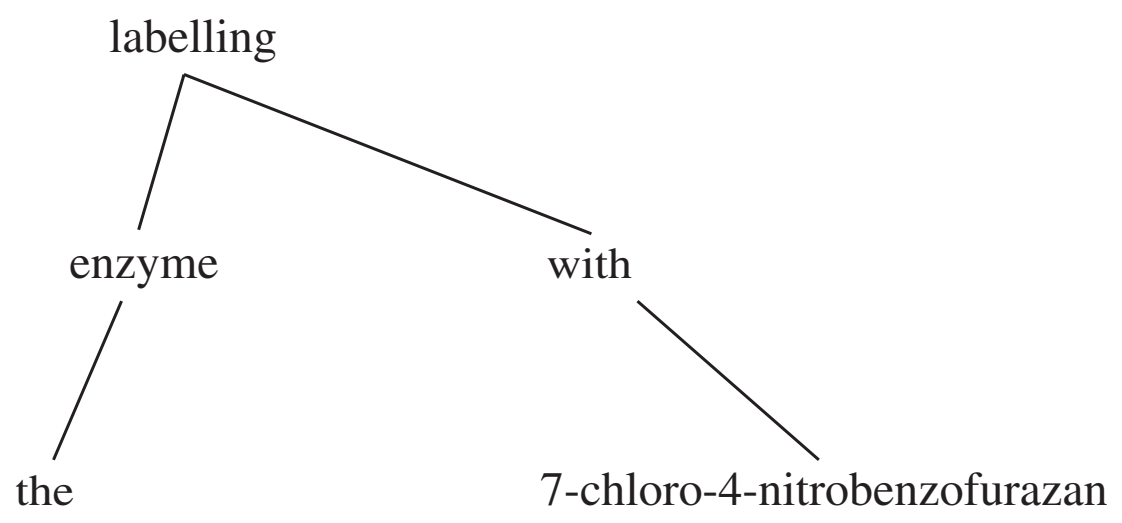

**Fig. 3.** Dependency tree translated from the phrase structure tree in Fig. 1.

To translate a phrase structure tree (PST) into a dependency structure tree, we first label each *non*-leaf node in the PST with a surface word. This makes every nodes in the tree to be associated with a surface word, henceforth called the *head (word)* of the node. The head word of a non-leaf node is inherited from a child of the node. If node $u$ has two or more children, the so-called *head rule*[3] as associated with each CFG rule determines from which one of the $n$ children $v_i$, $1 \leq i \leq n$ the head word should be inherited. The head rule uniquely determines the index $i$ of the children $v_i$ (called the *head constituent*) from which the head word should be inherited to the parent node $u$.

Fig. 2 shows the result of headword labeling scheme applied to the PST in Fig. 1. The inherited head words are shown in parentheses below the syntactic category, and a bold arrow represents the inheritance of a head word. For example, the arrow from $\mathsf{VBP}$ to $\mathsf{VP}$ denotes the head constituent is $\mathsf{VBP}$, but not $\mathsf{NP}$ or $\mathsf{PP}$ for the CFG rule $\mathsf{VP} \rightarrow \mathsf{VBP}, \mathsf{NP}, \mathsf{PP}$.

After all nodes are labeled with head words in the phrase structure tree, its dependency structure tree is extracted by recursively coalescing head constituent nodes with their parents until no more coalescing can be performed.

In Fig. 2, this process corresponds to coalescing every parent-child pair connected with a bold arrow. The node after coalescing inherits the same head word as nodes being coalesced, which should have had the same head given that the child is the head constituent.

The parent-child relationship in the dependency structure tree thus obtained (Fig. 3) represents the head word of a child (directly) depends on the head of the parent; and we say a node $u$ depends *indirectly* on another node $v$, if $v$ is an ancestor of $u$ but is not its parent. For instance, in Fig. 3 the determinant *the* depends directly on *enzyme*, and indirectly on *labeling*.

### 4.3 Step 3: extracting expressions relevant to diagnostic tests

Given the dependency structure trees, we extract the noun phrase containing the names of the clinical tests, the verbal expression, and other phrases depending on the same verbal expression from each tree. We illustrate this step with the dependency tree in Fig. 4. This figure depicts the dependency tree of the sentence ‘A stepwise increase in serum ADA level was observed with increasing severity of liver cirrhosis.’

Given a sentence $S = w_1 w_2 \cdots w_n$ consisting of $n$ words, let $c_i$ ($i = 1, \ldots, n$) be the syntactic category of $w_i$, i.e., the syntactic category assigned to the parent of the leaf node corresponding to $w_i$ in the PST for $S$. Let $T(S)$ denote the dependency tree of $S$. We identify the index $i$ for the $i$-th word in $S$ as the node corresponding to the word in the tree. Let the predefined set of the diagnosis test names be $D$.

[3] We used the head rules due to Collins [4].

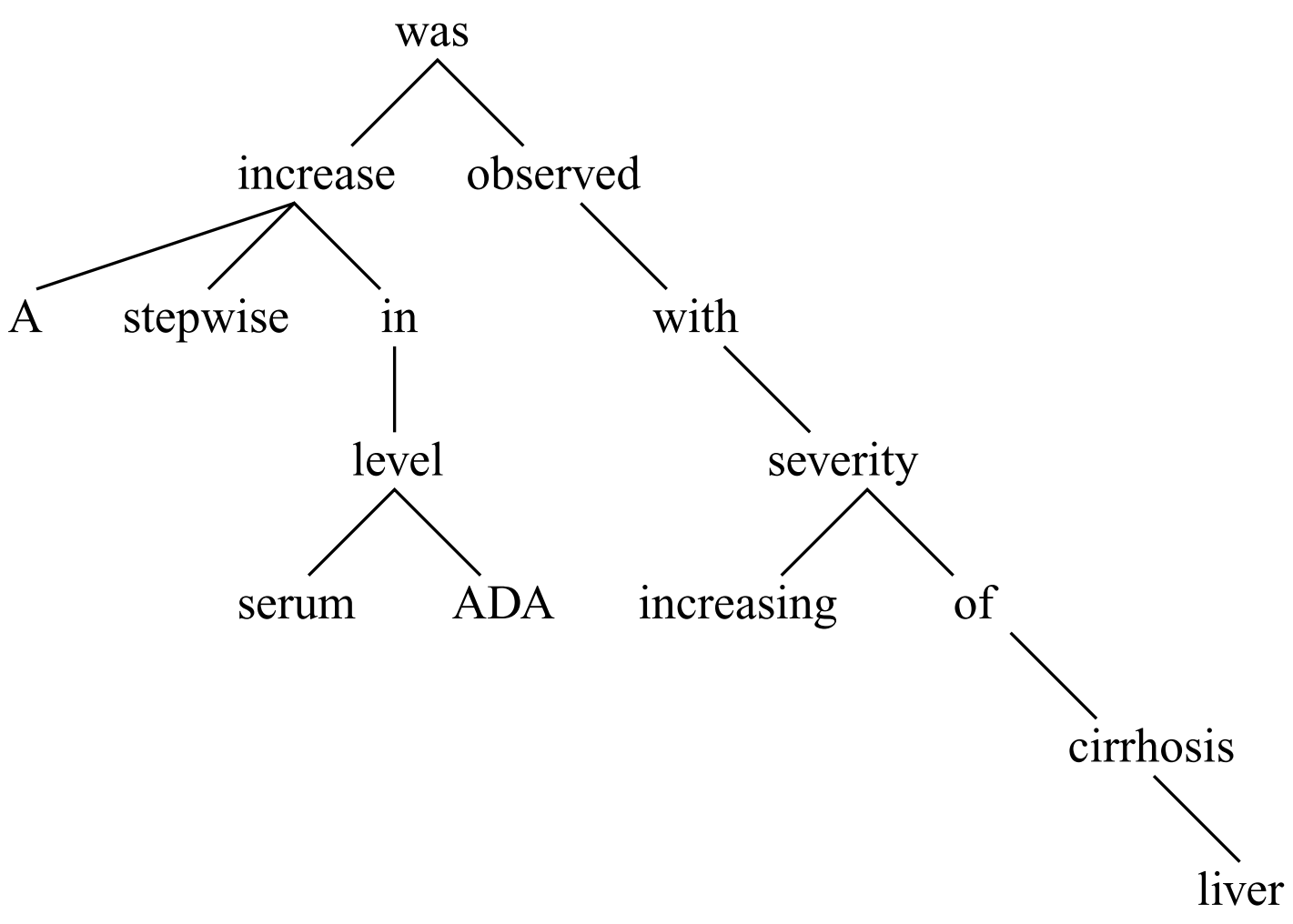

**Fig. 4.** Syntactic dependency tree of ‘A stepwise increase in serum ADA level was observed with increasing severity of liver cirrhosis.’

1. First, locate the names of the tests occurring in a given sentence and register their locations in the set $P$; i.e., $P \leftarrow \{j \mid w_i w_{i+1} \ldots w_j \in D\}$. When the test name consists of multiple words (i.e., $i \neq j$ in the above formula), the location $j$ of the last word is registered. This heuristic reflects the fact that in most cases, the last constituent word is the head of the noun phrase.
   In Fig. 4, we find $w_6 = ADA \in D$ in this step. Hence, $P = \{6\}$.
2. For each node $p \in P$ denoting a test (henceforth called the *pivot*), do:
   (a) Extract noun phrases containing the names of the diagnostic tests.
      i. Let the set $\mathrm{NP}_1 \leftarrow \emptyset$.
      ii. Starting from the pivot $p \in P$, ascend the tree $T(S)$ towards the root until a node is reached such that it is either (1) a verb, (2) an auxiliary verb, or (3) a preposition whose parent node is a verb or an auxiliary verb. Let $v$ be such a node. Put the node encountered along the way to the set $\mathrm{NP}_1$.
      In Fig. 4, we see that the traversal in this step ends when $w_v = was$ is encountered, at which point $\mathrm{NP}_1$ contains the indices for *ADA*, *level*, *in*, and *increase*.
      iii. Rearrange nodes in $\mathrm{NP}_1$ in the order of their indices, concatenate the words corresponding to the nodes in that order.
      In the example, this yields the noun phrase *increase in ADA level.* Note that the modifiers, i.e., *a* and *stepwise*, are excluded from the extracted pattern. This helps to absorb slight difference in modifiers, and also facilitate the reviewing process by the domain experts.
   (b) Extract verbal expressions.
      i. Let node $v$ denotes a verb, auxiliary verb, or preposition that stopped the traversal in Step 2(a)ii. Let the new set VP $\leftarrow \{v\}$. In this example, the index for *was* enters the list.

ii. From $v$, ascend the tree $T(S)$ towards the root while the current node is a verb, an auxiliary verb, or a preposition, and while the node is non-root. Add the encountered nodes along the way in VP. Since *was* is the root of the tree, this processing ends immediately in Fig. 4.

iii. If the reached node is the root of the dependency tree, descend the tree beginning from the child nodes of the root which are either a verb, an auxiliary verb, or a preposition, putting the words encountered in VP. In the figure, two words *observed* and *with* enter VP.

iv. Sort words corresponding to the nodes in VP in the order of their appearance in the original sentence.

This processing yields the verbal expression *was observed with* in the example of Fig. 4.

(c) Extract phrases depending on the verbal expression.

Traverse down from the child nodes that directly depends on the verbal expression. However, we only take into account the children $k$ where the word $w_k$ occurs at the opposite side of the pivot word with respect to the verb $w_v$.

To be precise, let $p$ and $v$ be the indices for the pivot, and the verb located in Step 2(a)ii. if $p < v$, i.e., pivot word $w_p$ occur before the verb $w_v$ in $S$, then traverse only from the children $k \in R$ where $R = \{k \mid \text{Child}(v) \text{ and } v < k\}$. To the contrary, if $i > j$, then traverse only from the elements in $R = \{k \mid \text{Child}(v) \text{ and } k < v\}$.

Let $\text{NP}_2 \leftarrow \emptyset$. For each $k \in R$, traverse all descending paths emanating from $k$, on condition that traversal should be cut off immediately when a conjunction, or an interrogative is encountered. Put all the obtained paths to $\text{NP}_2$.

In effect, two phrasal expressions are obtained from the tree in Fig. 4, i.e., *increasing severity*, and *severity of liver cirrhosis*.

### 4.4 Step 4: sorting obtained expressions for review

Finally, the collected expressions should be ranked and reordered according to some criterion to be subsequently reviewed by the domain experts. In this paper, the extracted expressions are simply sorted by the frequency of occurrences. In the future work, we will pursue the use of more sophisticated ranking methods based on statistical measures, and also to extract frequent sub-patterns in the extracted expressions.

## 5 Experimental results and discussions

We applied the method of Section 4 to the abstracts contained in the Medline 2003 database. In Step 1 of the extraction procedure, 57,987 abstracts contained the word *hepatitis*. From these abstracts, 130,306 sentences were identified as containing the names of the hepatitis-related diagnostic tests. We applied the procedures of Steps 2 and 3 to these sentences and extracted the noun phrases and the verb phrases.

Table 1 shows the 20 most frequent expressions containing the noun phrase containing diagnostic tests, filtered by humans from a total of 91,427 different noun phrases obtained in Step 1 of Section 4.3. Likewise, Table 2 shows the list of the verbal expressions representing some form of relationship, interactions and actions, filtered from the 37,780 verbal expressions obtained through Step 2 of Section 4.3. Finally, Table 3 shows the list of diseases, symptoms and conditions filtered from 251,409 noun phrases that depend on the verbal expressions (Step 3).

**Table 1.** Expressions representing change in the test results

| Rank | Frequency | Phrase |
|---|---|---|
| 52 | 113 | iron overload |
| 59 | 100 | positive for HCV RNA |
| 73 | 84 | detection of HCV RNA |
| 84 | 72 | presence of HBV DNA |
| 121 | 54 | iron concentration |
| 155 | 43 | HCV-RNA negative |
| 157 | 43 | HCV seropositivity |
| 186 | 37 | clearance of HCV RNA |
| 243 | 29 | loss of HCV RNA |
| 261 | 27 | iron deposition |
| 267 | 27 | copper concentrations |
| 288 | 26 | copper accumulation |
| 309 | 25 | dose of interferon |
| 395 | 21 | iron depletion |
| 527 | 16 | copper excretion |
| 575 | 15 | CT findings |
| 586 | 14 | disappearance of HCV-RNA |
| 715 | 11 | low density |
| 717 | 11 | iron reduction |
| 718 | 11 | interferon plus |

**Table 2.** Verbal expressions representing causal relationships

| Rank | Frequency | Expression |
|---|---|---|
| 7 | 1664 | was detected in |
| 13 | 709 | was found in |
| 17 | 615 | revealed |
| 18 | 611 | correlated |
| 24 | 518 | developed |
| 32 | 400 | was associated with |
| 38 | 372 | demonstrated |
| 47 | 316 | was observed in |
| 51 | 294 | occurred |
| 52 | 294 | induced |
| 65 | 235 | show |
| 70 | 223 | suggest |
| 76 | 215 | report |
| 93 | 193 | resulted in |
| 119 | 153 | indicate |
| 129 | 139 | causes |
| 137 | 129 | represents |
| 143 | 126 | seems to be |
| 144 | 126 | performed |
| 149 | 122 | was related to |

**Table 3.** Expressions representing diseases, symptoms, conditions, etc.

| Rank | Frequency | Expression |
|---|---|---|
| 19 | 428 | chronic hepatitis |
| 21 | 397 | HCV infection |
| 48 | 245 | liver disease |
| 75 | 181 | risk factors |
| 133 | 134 | hepatocellular carcinoma |
| 145 | 127 | chronic infection |
| 246 | 89 | liver cirrhosis |
| 252 | 87 | active hepatitis |
| 258 | 86 | acute hepatitis |
| 333 | 71 | anti-HCV positive |
| 384 | 62 | liver damage |
| 427 | 57 | cause of liver disease |
| 433 | 56 | viral hepatitis |
| 545 | 47 | chronic carriers |
| 597 | 43 | non-A hepatitis |
| 601 | 43 | liver injury |
| 811 | 33 | severe disease |
| 818 | 33 | detection of HCV RNA |
| 903 | 30 | chronic hepatitis cirrhosis |
| 1096 | 25 | inhibitory effect |

To summarize, we had to inspect the top 150 extracted verbal expressions to collect 20 expressions of interest (Table 2). On the other hand, for noun phrases containing the test names (Table 1) we needed to inspect more than 700 patterns to collect 20 meaningful patterns, and for diseases, symptoms, and conditions we had to examine over 1000 expressions (Table 3). The latter two cases impose an enormous load to the human inspector.

The possible remedies to further reduce the number of expressions are as follows.

- First filter sentences using verbal expressions as cue, and then extract the rest from the survived sentences.
- Use existing dictionaries and thesauri to restrict the variations in expressions.

As we claimed previously, the subtree representation allows fine-grained control over how the found patterns can be presented to the domain experts (but not fully discussed or demonstrated in this paper). Note however that this claim applies only to inspecting the validity of cue patterns coarsely, but not to the eventual knowledge inspection that should also be conducted by the domain experts. Specifically, although it is possible to present the matched pattern of the form ‘NP VBP NP’ obtained with the method of Section 4.3 omitting the modifiers not dependent on test items, it is not always feasible to show only this portion because important modifier words can be missed out. Consider the dependency tree depicted in Fig. 5. If we apply the method of Section 4.3 to this tree, it is possible to extract the noun phrases, *demolition of ACE by the hepatocyte* and *demolition of ACE by hepatocyte during cellular stress.* However, since the word *failed* does not have a direct or indirect dependency relation with the pivot word *ACE*, it never enters the list of collected noun phrases. Since the omission of *failed* leads to the opposite meaning, the above portion is not an acceptable form of knowledge representation. It is, on the other hand, completely acceptable to omit *failed* when sifting cue patterns is concerned, as addressed in this paper.

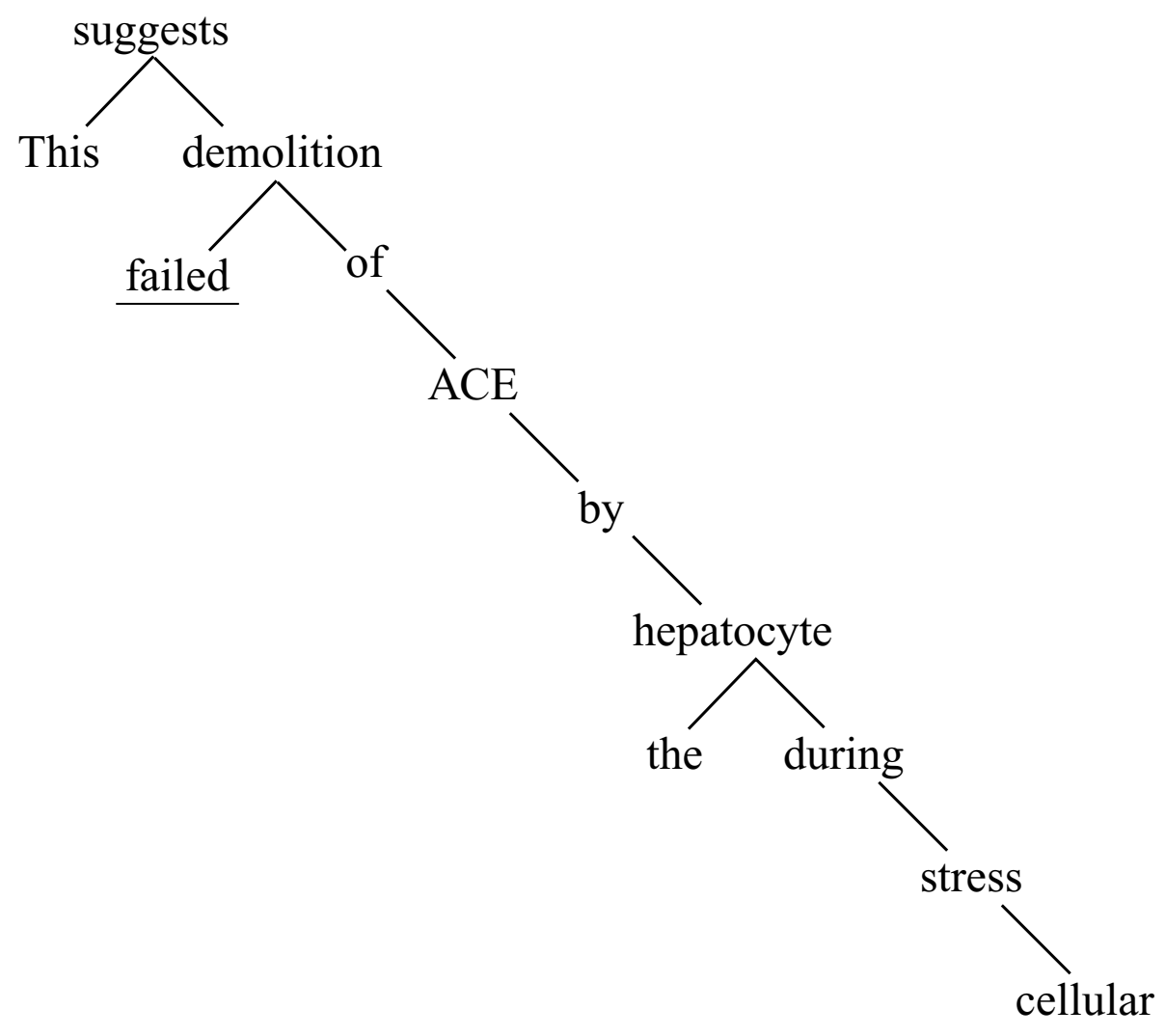

**Fig. 5.** The dependency tree of the sentence "This suggests failed demolition of ACE by the hepatocyte during cellular stress." The adjective *failed* is not an ancestor of the pivot word *ACE*.

## 6 Conclusions

The first step in knowledge extraction from large text data is locating relevant passages. This paper discussed how cue patterns for locating passages can be discovered efficiently. We used syntactic dependency parsing to obtain frequent patterns in the three categories:

1. Noun phrases containing diagnostic tests.
2. Verb expressions representing a relationship, interaction, or action.
3. Symptoms and conditions of hepatitis and other diseases.

The proposed method yielded a better result (a smaller number of candidates) for the second class (verbal expressions), compared with the rest. The first class (non phrases containing diagnostic tests) and the third class (symptoms and conditions of diseases) required a vast amount of human reviews to filter results, and was not satisfactory.

Our future research includes developing efficient methods to further sift through these candidate patterns. We also plan to apply the techniques of collocation identification and tree mining to the extracted expressions, in order to obtain more compact representation of the expressions. Another issue to be addressed is to discriminate words which test names do not directly or indirectly depend on but are still important, such as the adjective *failed* in the example previously mentioned.

# Mining Diagnostic Taxonomy for Multi-Stage Medical Diagnosis

Shusaku Tsumoto

Department of Medical Informatics,
Faculty of Medicine, Shimane University
89-1 Enya-cho Izumo City, Shimane 693-8501 Japan
E-mail: tsumoto@computer.org

**Abstract.** Experts' reasoning selects the final diagnosis from many candidates by using hierarchical differential diagnosis. In other words, candidates gives a sophisticated hiearchical taxonomy, usually described as a tree. In this paper, the characteristics of experts' rules are closely examined from the viewpoint of hierarchical decision steps and and a new approach to rule mining with extraction of diagnostic taxonomy from medical datasets is introduced. The key elements of this approach are calculation of the characterization set of each decision attribute (a given class) and one of the similarities between characterization sets. From the relations between similarities, tree-based taxonomy is obtained, which includes enough information for hierarchical diagnosis. The proposed method was evaluated on three medical datasets, the experimental results of which show that induced rules correctly represent experts' decision processes.

**Keywords** Rough sets, data mining, taxonomy, granular computing.

## 1 Introduction

Rule mining has been applied to many domains. However, empirical results show that the interpretation of extracted rules requires deep understanding for applied domains. One of its reasons is that conventional rule induction methods such as C4.5[6] cannot reflect the type of experts' reasoning. For example, rule induction methods such as AQ15[4], PRIMEROSE[9] induce the following common rule for muscle contraction headache from databases on differential diagnosis of headache:

$[location = whole] \wedge$[Jolt Headache = $no$] $\wedge$[Tenderness of M1 = $yes$]
$\rightarrow$ muscle contraction headache.

This rule is shorter than the following rule given by medical experts.

[Jolt Headache = $no$]
$\wedge$([Tenderness of M0 = $yes$] $\vee$[Tenderness of M1 = $yes$] $\vee$[Tenderness of M2 = $yes$])
$\wedge$[Tenderness of B1 = $no$] $\wedge$[Tenderness of B2 = $no$] $\wedge$[Tenderness of B3 = $no$]
$\wedge$[Tenderness of C1 = $no$] $\wedge$[Tenderness of C2 = $no$] $\wedge$[Tenderness of C3 = $no$]
$\wedge$[Tenderness of C4 = $no$]
$\rightarrow$ muscle contraction headache

where [Tenderness of B1 = *no*] and [Tenderness of C1 = *no*] are added. It is notable that these observation can be found in several medical domains[9].

One of the main reasons why the rules obtained from the dataset are shorter is that these patterns are generated only by a simple criteria, such as high accuracy or high information gain. The comparative studies[9–11] suggest that experts should acquire rules not only by a single criteria but by the usage of several measures.

Those characteristics of medical experts' rules are fully examined not by comparing between those rules for the same class, but by comparing experts' rules with those for another class[9].

For example, the classification rule for muscle contraction headache given in Section 1 is very similar to the following classification rule for disease of cervical spine:

[Jolt Headache = *no*]
$\wedge$([Tenderness of M0 = *yes*] $\vee$[Tenderness of M1 = *yes*] $\vee$[Tenderness of M2 = *yes*])
$\wedge$([Tenderness of B1 = *yes*] $\vee$[Tenderness of B2 = *yes*] $\vee$[Tenderness of B3 = *yes*]
$\vee$[Tenderness of C1 = *yes*] $\vee$[Tenderness of C2 = *yes*] $\vee$[Tenderness of C3 = *yes*]
$\vee$[Tenderness of C4 = *yes*])
$\rightarrow$ disease of cervical spine

The differences between these two rules are attribute-value pairs, from tenderness of B1 to C4. Thus, these two rules are composed of the following three blocks:

$$A_1 \wedge A_2 \wedge \neg A_3 \rightarrow \textit{muscle contraction headache}$$
$$A_1 \wedge A_2 \wedge A_3 \rightarrow \textit{disease of cervical spine},$$

where $A_1$, $A_2$ and $A_3$ are given as the following formulae:
$A_1$ = [Jolt Headache = *no*], $A_2$ = [Tenderness of M0 = *yes*] $\vee$ [Tenderness of $M1$ = *yes*] $\vee$ [Tenderness of M2 = *yes*], and $A_3$ = [Tenderness of C1 = *no*] $\wedge$ [Tenderness of C2 = *no*] $\wedge$ [Tenderness of C3 = *no*] $\wedge$ [Tenderness of C4 = *no*].
The first two blocks ( $A_1$ and $A_2$ ) and the third one ( $A_3$ ) represent the different types of differential diagnosis. The first one $A_1$ shows the discrimination between muscular type and vascular type of headache. Then, the second part shows the differential diagnosis between headaches caused by neck muscles and ones by head muscles. Finally, the third formula $A_3$ is used to make a differential diagnosis between muscle contraction headache and disease of cervical spine. Thus, medical experts first select several diagnostic candidates, which are very similar to each other, from many diseases and then make a final diagnosis from those candidates.

In this paper, the characteristics of experts' rules are closely examined from the viewpoint of hierarchical decision steps. Then, extraction of diagnostic taxonomy from medical datasets is introduced, which consists of the following three procedures. First, the characterization set of each attribute-value pair for a decision attribute(a given class) is extracted from databases. Then, similarities between the characterization sets are calculated. Finally, the concept hierarchy for given classes is calculated from the similarity values.

The paper is organized as follows. Section 2 and 3 introduces rough sets and a characterization set. Section 4 gives an algorithm for extraction of diagnostic taxonomy. Section 5 shows an illustrative example. Section 6 gives how rules are induced after grouping. Finally, Section 7 concludes this paper. The proposed method was evaluated on medical databases, the experimental results of which show that induced rules correctly represent experts' decision processes.

## 2 Rough Set Theory: Preliminaries

In the following sections, we use the following notations introduced by Grzymala-Busse and Skowron[8], which are based on rough set theory[5].

Let $U$ denote a nonempty, finite set called the universe and A denote a nonempty, finite set of conditional attributes, i.e., $a : U \rightarrow V_a$ for $a \in A$, where $V_a$ is called the domain of $a$, respectively. For $A$, $V_A$ denotes a set of the domain of attributes. Then, a decision table is defined as an information system, $A = (U, A \cup \{d\})$, where $\{d\}$ denotes a decision attribute (a set of given classes).

The atomic formulae over $B \subseteq A \cup \{d\}$ and $V_B$ are expressions of the form $[a = v]$, called descriptors over B, where $a \in B$ and $v \in V_a$. The set $F(B, V_B)$ of formulas over B is the least set containing all atomic formulas over $B$ and closed with respect to disjunction, conjunction and negation. For each $f \in F(B, V_B)$, $f_A$ denote the meaning of $f$ in $A$, i.e., the set of all objects in U with property $f$, defined inductively as follows: (1) If $f$ is of the form $[a = v]$ then, $f_A = \{s \in U | a(s) = v\}$ (2) $(f \wedge g)_A = f_A \cap g_A$; $(f \vee g)_A = f_A \vee g_A$; $(\neg f)_A = U - f_a$

By the use of the framework above, classification accuracy and coverage are defined as follows.

**Definition 1.**
*Let $R$ denote a formula in $F(B, V_B)$ and $D$ a set of objects which belong to a decision attribute d. Classification accuracy and coverage(true positive rate) for $R \rightarrow d$ is defined as:*

$$\alpha_R(D) = \frac{|R_A \cap D|}{|R_A|}(= P(D|R)), \text{ and } \kappa_R(D) = \frac{|R_A \cap D|}{|D|}(= P(R|D)),$$

*where $|S|$, $\alpha_R(D)$, $\kappa_R(D)$ and P(S) denote the cardinality of a set $S$, a classification accuracy and coverage of $R$ as to classification of $D$, and probability of $S$, respectively.*

It is notable that $\alpha_R(D)$ measures the degree of the sufficiency of a proposition, $R \rightarrow D$, and that $\kappa_R(D)$ measures the degree of its necessity.

Also, we define partial order of equivalence as follows:

**Definition 2.** *Let $R_i$ and $R_j$ be the formulae in $F(B, V_B)$ and let $A(R_i)$ denote a set whose elements are the attribute-value pairs of the form $[a, v]$ included in $R_i$. If $A(R_i) \subseteq A(R_j)$, then we represent this relation as: $R_i \preceq R_j$.*

Finally, according to the above definitions, probabilistic rules with high accuracy and coverage are defined as:

$$R \stackrel{\alpha,\kappa}{\to} d \ s.t. \ R = \wedge_i [a_i = v_k], \ \alpha_R(D) \geq \delta_\alpha \text{ and } \kappa_R(D) \geq \delta_\kappa,$$

where $\delta_\alpha$ and $\delta_\kappa$ denote given thresholds for accuracy and coverage, respectively.

## 3 Characterization Sets

### 3.1 Characterization Sets

In order to model medical reasoning, a statistical measure, coverage plays an important role in modeling. Let us define a characterization set of $D$, denoted by $L_{\delta_\kappa}(D)$ as a set, each element of which is an elementary attribute-value pair R with coverage being larger than a given threshold, $\delta_\kappa$. That is,

**Definition 3.** *Let $R$ denote a formula in $F(B, V_B)$. Characterization sets of a decision attribute (D) is defined as:*

$$L_{\delta_\kappa}(D) = \{R | R_i = \wedge_i (\vee_j [a_i = v_j]) \text{ and } \kappa_R(D) \geq \delta_\kappa\},$$

Then, three types of relations between characterization sets can be defined as follows: (1) Independent type: $L_{\delta_\kappa}(D_i) \cap L_{\delta_\kappa}(D_j) = \phi$, (2) Overlapped type: $L_{\delta_\kappa}(D_i) \cap L_{\delta_\kappa}(D_j) \neq \phi$, and (3) Subcategory type: $L_{\delta_\kappa}(D_i) \subseteq L_{\delta_\kappa}(D_j)$. All three definitions correspond to the negative region, boundary region, and positive region, respectively, if a set of the whole elementary attribute-value pairs will be taken as the universe of discourse.

Tsumoto focuses on the subcategory type in [10] because $D_i$ and $D_j$ cannot be differentiated by using the characterization set of $D_j$, which suggests that $D_i$ is a generalized disease of $D_j$. Then, Tsumoto generalizes the above rule induction method into the overlapped type, considering rough inclusion[11]. However, both studies assumes two-level diagnostic steps: focusing mechanism and differential diagnosis, where the former selects diagnostic candidates from the whole classes and the latter makes a differential diagnosis between the focused classes.

The proposed method below extends these methods into multi-level steps. In this paper, we consider the special case of characterization sets in which each formulae is given as a conjunctive normal formula and the thresholds of coverage is equal to 1.0: $L_{1.0}(D) = \{R_i | R_i = \wedge_i (\vee_j [a_i = v_j]), \quad \kappa_{R_i(D)} = 1.0\}$ It is notable that this set has several interesting characteristics.

**Theorem 1.** *Let $R_i$ and $R_j$ two conjunctive formulae in $L_{1.0}(D)$ such that $R_i \preceq R_j$. Then, $\alpha_{R_i} \leq \alpha_{R_j}$.*

**Theorem 2.** *Let $R$ be a formula in $L_{1.0}(D)$ such that $R = \vee_j [a_i = v_j]$. Then, $R$ and $\neg R$ gives the coarsest partition for $a_i$, whose $R$ includes $D$.*

**Theorem 3.** *Let $A$ consist of $\{a_1, a_2, \cdots, a_n\}$ and $R_i$ be a formula in $L_{1.0}(D)$ such that $R_i = \vee_j [a_i = v_j]$. Then, a sequence of a conjunctive formula $F(k) = \wedge_{i=1}^k R_i$ gives a sequence which increases the accuracy.*

## 4 Rule Induction with Diagnostic Taxonomy

### 4.1 Intuitive Ideas

As discussed in Section 2, when the coverage of $R$ for a target concept $D$ is equal to 1.0, $R$ is a necessity condition of $D$. That is, a proposition $D \rightarrow R$ holds and its contrapositive $\neg R \rightarrow \neg D$ holds. It means that if $R$ is not observed, $D$ cannot be a candidate of a decision class. If two decision classes have a common formula $R$ whose coverage is equal to 1.0, then $\neg R$ supports the negation of two classes, which means these two concepts belong to the same group. Furthermore, if two target concepts have similar formulae $R_i, R_j \in L_{1.0}(D)$, they are very close to each other with respect to the negation of two classes. In this case, the attribute-value pairs in the intersection of $L_{1.0}(D_i)$ and $L_{1.0}(D_j)$ give a characterization set of the generalized decision class that unifies $D_i$ and $D_j$, $DD_k$. Then, compared with $DD_k$ and other target concepts, classification rules for $DD_k$ can be obtained. When we have a sequence of grouping, classification rules for a given decision classes are defined as a sequence of subrules. From these ideas, a rule induction algorithm with grouping target concepts can be described as a combination of grouping (Figure 1) and rule induction (Figure 2). First, this algorithm first calculates $L_{1.0}(D_i)$ for $\{D_1, D_2, \cdots, D_k\}$. Second, from the list of characterization sets, it calculates the intersection between $L_{1.0}(D_i)$ and $L_{1.0}(D_j)$ and stores it into $L_{id}$. Third, the procedure calculates the similarity (matching number)of the intersections and sorts $L_{id}$ with respect of the similarities. Fourth, the algorithm chooses one intersection $(D_i \cap D_j)$ with maximum similarity (highest matching number) and group $D_i$ and $D_j$ into a concept $DD_i$. These procedures will be continued until all the grouping is considered. The first to fourth steps are described as Figure 1. Finally, rules for each decision class, including grouped ones, are induced. For given decision classes, rules are composed of rules for the upper-level and rules specific to the corresponding given class shown in Figure 2.

### 4.2 Similarity

To measure the similarity between two characterization sets, we can apply several indices of two-way contigency tables. Table 1 gives a contingency table for two rules, $L_{1.0}(D_i)$ and $L_{1.0}(D_j)$. The first cell $a$ (the intersection of the first row and column) shows the number of matched attribute-value pairs. From this table, several kinds of similarity measures can be defined. The best similarity measures in the statistical literature are four measures shown in Table 2[3, 2].

In this paper, we focus on the two similarity measures: one is Simpson's measure: $\frac{a}{min\{(a+b),(a+c)\}}$ and the other is Braun's measure: $\frac{a}{max\{(a+b),(a+c)\}}$.

As discussed in Subsection 4.2, a single-valued similarity becomes low when $L_{1.0}(D_i) \subset L_{1.0}(D_j)$ and $|L_{1.0}(D_i)| << |L_{1.0}(D_j)|$. For example, let us consider when $|L_{1.0}(D_i)| = 1$. Then, match number is equal to 1.0, which is the lowest value of this similarity. In the case of Jaccard's coefficient, the value is $1/1 + b$ or $1/1 + c$: the similarity is very small when $1 << b$ or $1 << c$. Thus, these

**procedure** *Grouping* ;
  **var inputs**
    $L_c$ : *List*; ¿ /* A list of Characterization Sets */
    $L_{id}$ : *List*; ¿ /* A list of Intersection */
    $L_s$ : *List*; ¿ /* A list of Similarity */
  **var outputs**
    $L_{gr}$ : *List*; /* A list of Grouping */
  **var**
    $k$ : *integer*;    $L_g$ : *List*;
  **begin**
    $L_g := \{\}$ ; $L_{gr} := \{\}$;
    $k := n$ /* n: A number of Target Concepts*/
    Sort $L_s$ with respect to similarities;
      Take a set of $(D_i, D_j)$, $L_{max}$ with maximum similarity values;
      k:= k+1;
      **forall** $(D_i, D_j) \in L_{max}$ **do**
        **begin**
          $L_g := \{\}$;
          Group $D_i$ and $D_j$ into $DD_k$;
            $L_c := L_c - \{(D_i, L_{1.0}(D_i)\}$;
              $L_c := L_c - \{(D_j, L_{1.0}(D_j)\}$;
              $L_c := L_c + \{(D_k, L_{1.0}(D_k)\}$;
              Update $L_{id}$ for $DD_k$;
              Update $L_s$;
            $L_g :=$ (Outputs from *Grouping* for $L_c$, $L_{id}$, and $L_s$) ;
          $L_{gr} := L_{gr} + \{\{(DD_k, D_i, D_j), L_g\}\}$;
        **end**
    **return** $L_{gr}$;
  **end** {*Grouping*}

**Fig. 1.** An Algorithm for Grouping

similarities do not reflect the subcategory type. Thus, we should check the difference between $a + b$ and $a + c$ to consider the subcategory type. One solution is to take an interval of maximum and minimum as a similarity, which we call an interval-valued similarity.

For this purpose, we combine Simpson and Braun similarities and define an interval-valued similarity: $\left[\frac{a}{max\{(a+b),(a+c)\}}, \frac{a}{min\{(a+b),(a+c)\}}\right]$ If the difference between two values is large, it would be better not to consider this similarity for grouping in the lower generalization level. For example, when $a + c = 1(a = 1, c = 0)$, the above value will be: $\left[\frac{1}{1+b}, 1\right]$ If $b >> 1$, then this similarity should be kept as the final candidate for the grouping.

The disadvantage is that it is difficult to compare these interval values. In this paper, the maximum value of a given interval is taken as the representative of this similarity when the difference between min and max are not so large.

```
procedure RuleInduction ;
  var inputs
    L_c : List;
    /* A list of Characterization Sets */
    L_id : List; /* A list of Intersection */
    L_g : List; /* A list of grouping*/
    /* {{(D_{n+1},D_i,D_j),{(DD_{n+2},.)...}}} */
    /* n: A number of Target Concepts */
  var
    Q, L_r : List;
  begin
    Q := L_g; L_r := {};
    if  (Q = ∅) then return L_r = {};
    if  (Q ≠ ∅) then do
      begin
        Q := Q − first(Q);
        L_r := Rule Induction (L_c, L_id, Q);
      end
    (DD_k, D_i, D_j) := first(Q);
    if (D_i ∈ L_c and D_j ∈ L_c) then do
      begin
        Induce a Rule r which discriminate
        between D_i and D_j;
        r = {R_i → D_i, R_j → D_j};
      end
    else do
      begin
        Search for L_1.0(D_i) from L_c;
        Search for L_1.0(D_j) from L_c;
         if (i < j) then do
          begin
            r(D_i) := ∨_{R_l ∈ L_1.0(D_j)} ¬R_l → ¬D_j;
            r(D_j) := ∧_{R_l ∈ L_1.0(D_j)} R_l → D_j;
          end
        r := {r(D_i), r(D_j)};
      end
    return L_r := {r, L_r} ;
  end {Rule Induction}
```

**Fig. 2.** An Algorithm for Rule Induction

If the maximum values are equal to the other, then the minimum value will be compared. If the minimum value is larger than the other, the larger one is selected.

**Table 1.** Contingency Table for Similarity

| | | $L_{1.0}(D_j)$ | | |
|---|---|---|---|---|
| | | *Observed* | *Not Observed* | Total |
| $L_{1.0}(D_i)$ | *Observed* | $a$ | $b$ | $a+b$ |
| | *Not observed* | $c$ | $d$ | $c+d$ |
| | Total | $a+c$ | $b+d$ | $a+b+c+d$ |

**Table 2.** A List of Similarity Measures

| | |
|---|---|
| (1) Matching Number | $a$ |
| (2) Jaccard's coefficient | $a/(a+b+c)$ |
| (3) $\chi^2$-statistic | $N(ad-bc)^2/M$ |
| (4) point correlation coefficient | $(ad-bc)/\sqrt{M}$ |
| (5) Kulczynski | $\frac{1}{2}(\frac{a}{a+b}+\frac{a}{a+c})$ |
| (6) Ochiai | $\frac{a}{\sqrt{(a+b)(a+c)}}$ |
| (7) Simpson | $\frac{a}{min\{(a+b),(a+c)\}}$ |
| (8) Braun | $\frac{a}{max\{(a+b),(a+c)\}}$ |

$N=a+b+c+d$, $M=(a+b)(b+c)(c+d)(d+a)$

## 5 Example

Let us consider Table 3 as an example for rule induction. For a similarity function, we use the interval similarity defined in Section 4.2. Since Table 3 has five classes in the decision attribute, an index for grouped concepts, $k$ is set to 6. For extraction of taxonomy, the interval-valued similarity is applied.

### 5.1 Grouping

From this table, the characterization set for each concept is obtained as shown in Fig 3. Then, the intersection between two target concepts are calculated. In the first level, the similarity matrix is generated as shown in Fig. 4.

Since *common* and *classic* have the maximum similarity, these two classes are grouped into one category, $D_6$. Then, the characterization of $D_6$ is obtained as : $D_6 = \{[loc = lateral], [nat = thr], [jolt = 1], [nau = 1], [M1 = 0], [M2 = 0]\}$. In the second iteration, the intersection of $D_6$ and others is considered and the similarity matrix is obtained: as shown in Fig 5. From this matrix, we have to compare three candidates: [2/8,2/4], [3/7,3/6] and [2/7,2/4]. From the minimum values, the middle one: $D_6$ and *i.m.l.* is selected as the second grouping. Thus, $D_7 = \{[jolt = 1], [M1 = 0], [M2 = 0]\}$. In the third iteration, the intersection matrix is calculated as Fig 6 and *m.c.h.* and *psycho* are grouped into $D_8$: $D_8 = \{$ [nat=per], [prod=0] $\}$. Finally, the dendrogram is given as Fig. 7.

**Table 3.** A small example of a database

| No. | loc | nat | his | prod | jolt | nau | M1 | M2 | class |
|---|---|---|---|---|---|---|---|---|---|
| 1 | occular | per | per | 0 | 0 | 0 | 1 | 1 | m.c.h. |
| 2 | whole | per | per | 0 | 0 | 0 | 1 | 1 | m.c.h. |
| 3 | lateral | thr | par | 0 | 1 | 1 | 0 | 0 | common. |
| 4 | lateral | thr | par | 1 | 1 | 1 | 0 | 0 | classic. |
| 5 | occular | per | per | 0 | 0 | 0 | 1 | 1 | psycho. |
| 6 | occular | per | subacute | 0 | 1 | 1 | 0 | 0 | i.m.l. |
| 7 | occular | per | acute | 0 | 1 | 1 | 0 | 0 | psycho. |
| 8 | whole | per | chronic | 0 | 0 | 0 | 0 | 0 | i.m.l. |
| 9 | lateral | thr | per | 0 | 1 | 1 | 0 | 0 | common. |
| 10 | whole | per | per | 0 | 0 | 0 | 1 | 1 | m.c.h. |

Definition. loc: location, nat: nature, his:history,
Definition. prod: prodrome, nau: nausea, jolt: Jolt headache,
M1, M2: tenderness of M1 and M2, 1: Yes, 0: No, per: persistent,
thr: throbbing, par: paroxysmal, m.c.h.: muscle contraction headache,
psycho.: psychogenic pain, i.m.l.: intracranial mass lesion, common.:
common migraine, and classic.: classical migraine.

$$
\begin{aligned}
L_{1.0}(m.c.h.) =\ & \{([loc = occular] \vee [loc = whole]), [nat = per], [his = per], \\
& [prod = 0], [jolt = 0], [nau = 0], [M1 = 1], [M2 = 1]\} \\
L_{1.0}(common) =\ & \{[loc = lateral], [nat = thr], ([his = per] \vee [his = par]), [prod = 0], \\
& [jolt = 1], [nau = 1], [M1 = 0], [M2 = 0]\} \\
L_{1.0}(classic) =\ & \{[loc = lateral], [nat = thr], [his = par], [prod = 1], \\
& [jolt = 1], [nau = 1], [M1 = 0], [M2 = 0]\} \\
L_{1.0}(i.m.l.) =\ & \{([loc = occular] \vee [loc = whole]), [nat = per], \\
& ([his = subacute] \vee [his = chronic]), [prod = 0], \\
& [jolt = 1], [M1 = 0], [M2 = 0]\} \\
L_{1.0}(psycho) =\ & \{[loc = occular], [nat = per], ([his = per] \vee [his = acute]), \\
& [prod = 0]\}
\end{aligned}
$$

**Fig. 3.** Characterization Sets for Table 3

| | m.c.h. | common | classic | i.m.l. | psycho |
|---|---|---|---|---|---|
| m.c.h. | – | [1/8,1/8] | [0,0] | [3/8,3/7] | [2/8,2/4] |
| common | – | – | [6/8,6/8] | [4/8, 4/7] | [1/7,1/4] |
| classic | – | – | – | [3/8, 3/7] | 0 |
| i.m.l. | – | – | – | – | [2/7, 2/4] |

**Fig. 4.** Interval-valued Similarity of Two Characterization Sets (Step 2)

### 5.2 Rule Induction

The grouping obtained from the dataset shows the candidate of the differential diagnosis taxonomy with the given interval-valued similarity. For differential diagnosis, First, this model discriminate between $D_7$(*common*, *classic* and *i.m.l.*)

| | m.c.h. | $D_6$ | i.m.l. | psycho |
|---|---|---|---|---|
| m.c.h. | – | 0 | [3/8, 3/7] | [2/8,2/4] |
| $D_6$ | – | – | [3/7,3/6] | 0 |
| i.m.l. | – | – | – | [2/7,2/4] |

**Fig. 5.** Interval-valued Similarity of Two Characterization Sets after the first Grouping (Step 3)

| | m.c.h. | $D_7$ | psycho |
|---|---|---|---|
| m.c.h. | – | $[0, 0]$ | [2/8,2/4] |
| $D_7$ | – | $[0, 0]$ | [0,0] |

**Fig. 6.** Interval-valued Similarity of Two Characterization Sets after the second Grouping (Step 4)

and $D_8$ (*m.c.h.* and *psycho*). Then, $D_6$ and *i.m.l.* within $D_7$ are differentiated. Finally, *common* and *classic* within $D_7$ are checked. Thus, a classification rule for *common* is composed of two subrules: (discrimination between $D_7$ and $D_8$), (discrimination between $D_6$ and *i.m.l.*), and (discrimination within $D_6$).

The first part can be obtained by the intersection for Figure 6. That is,

$$D_8 \rightarrow [nat = per] \wedge [prod = 0]$$

$$\neg[nat = per] \vee \neg[prod = 0] \rightarrow \neg D_8.$$

Then, the second part can be obtained by the intersection for Figure 5. That is,

$$\neg([loc = occular] \vee [loc = whole]) \vee \neg[nat = per] \\ \vee \neg([his = subacute] \vee [his = chronic]) \\ \vee \neg[prod = 0] \rightarrow \neg i.m.l.$$

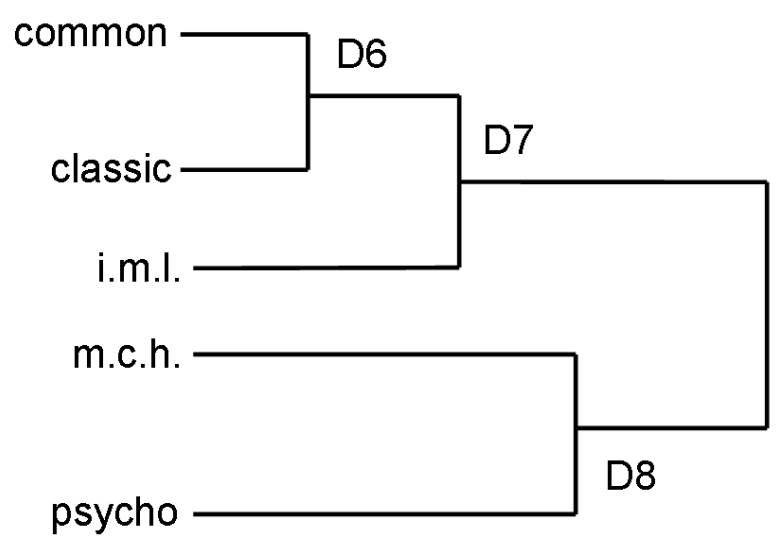

**Fig. 7.** Grouping by Characterization Sets

Finally, the third part of the rule can be obtained by the difference set between $L_{1.0}(common)$ and $L_{1.0}(classic) = \{[prod = 1]\}$.

$$[prod = 0] \rightarrow common.$$

Combining these three parts, the classification rule for *common* is

$$\begin{aligned}
&(\neg[nat = per] \vee \neg[prod = 0]) \\
&\qquad \wedge (\neg([loc = occular] \vee [loc = whole]) \vee \neg[nat = per] \\
&\qquad \vee \neg([his = subacute] \vee [his = chronic]) \vee \neg[prod = 0]) \\
&\qquad\qquad \wedge [prod = 0] \rightarrow common.
\end{aligned}$$

After its simplification, the rule is transformed into:

$$\begin{aligned}
&[nat = thr] \wedge ([loc = lateral] \vee \neg([his = subacute] \vee [his = chronic])) \\
&\qquad\qquad \wedge [prod = 0] \rightarrow common.
\end{aligned}$$

whose accuracy is equal to 2/3.

## 6 Experimental Results

The above rule induction algorithm was implemented in PRIMEROSE5.0 (Probabilistic Rule Induction Method based on Rough Sets Ver 5.0), and was applied to databases on differential diagnosis of headache, meningitis and cerebrovascular diseases (CVD), whose precise information is given in Table 4. In these experiments, $\delta_\alpha$ and $\delta_\kappa$ were set to 0.75 and 0.5, respectively. Also, the threshold for grouping is set to 0.8.[1] This system was compared with PRIMEROSE4.5[11], PRIMEROSE[9] C4.5[6], CN2[1], AQ15[4] with respect to the following points: length of rules, similarities between induced rules and expert's rules and performance of rules.

In this experiment, the length was measured by the number of attribute-value pairs used in an induced rule and Jaccard's coefficient was adopted as a similarity measure for comparison[3]. Concerning the performance of rules, ten-fold cross-validation was applied to estimate classification accuracy.

Table 5 shows the experimental results, which suggest that PRIMEROSE5 outperforms PRIMEROSE4.5 (two-level) and the other four rule induction methods and induces rules very similar to medical experts' ones.

## 7 Discussion

### 7.1 Focusing Mechanism

The readers may wonder why lengthy rules perform better than short rules since lengthy rules suffer from overfitting to a given data. One reason is that a decision attribute gives a partition of datasets: since the number of given classes are 4 to 45, some classes have very low support due to the prevalence of the corresponding diseases. Thus, the disease with the low frequency may not have short-length rules by using the conventional methods. However, since our method is not based on accuracy, but on coverage, we can support the disease with low frequency. Another reason is that this method reflects the reasoning style of domain experts. One of the most important features of medical reasoning is that medical experts finally select one or two diagnostic candidates from many diseases, called focusing mechanism. For example, in differential diagnosis of headache, experts choose one from about 60 diseases. The proposed method models induction of rules which incorporates this mechanism, whose experimental evaluation show that induced rules correctly represent medical experts' rules.

[1] These values are given by medical experts as good thresholds for rules in these three domains.

**Table 4.** Information about Databases

| Domain | Samples | Classes | Attributes |
|---|---|---|---|
| Headache | 52119 | 45 | 147 |
| CVD | 7620 | 22 | 285 |
| Meningitis | 141 | 4 | 41 |

**Table 5.** Experimental Results

| Method | Length | Similarity | Accuracy |
|---|---|---|---|
| | Headache | | |
| PRIMEROSE5.0 | $8.8 \pm 0.27$ | $0.95 \pm 0.08$ | $95.2 \pm 2.7\%$ |
| PRIMEROSE4.5 | $7.3 \pm 0.35$ | $0.74 \pm 0.05$ | $88.3 \pm 3.6\%$ |
| Experts | $9.1 \pm 0.33$ | $1.00 \pm 0.00$ | $98.0 \pm 1.9\%$ |
| PRIMEROSE | $5.3 \pm 0.35$ | $0.54 \pm 0.05$ | $88.3 \pm 3.6\%$ |
| C4.5 | $4.9 \pm 0.39$ | $0.53 \pm 0.10$ | $85.8 \pm 1.9\%$ |
| CN2 | $4.8 \pm 0.34$ | $0.51 \pm 0.08$ | $87.0 \pm 3.1\%$ |
| AQ15 | $4.7 \pm 0.35$ | $0.51 \pm 0.09$ | $86.2 \pm 2.9\%$ |
| | Meningitis | | |
| PRIMEROSE5.0 | $2.6 \pm 0.19$ | $0.91 \pm 0.08$ | $82.0 \pm 3.7\%$ |
| PRIMEROSE4.5 | $2.8 \pm 0.45$ | $0.72 \pm 0.25$ | $81.1 \pm 2.5\%$ |
| Experts | $3.1 \pm 0.32$ | $1.00 \pm 0.00$ | $85.0 \pm 1.9\%$ |
| PRIMEROSE | $1.8 \pm 0.45$ | $0.64 \pm 0.25$ | $72.1 \pm 2.5\%$ |
| C4.5 | $1.9 \pm 0.47$ | $0.63 \pm 0.20$ | $73.8 \pm 2.3\%$ |
| CN2 | $1.8 \pm 0.54$ | $0.62 \pm 0.36$ | $75.0 \pm 3.5\%$ |
| AQ15 | $1.7 \pm 0.44$ | $0.65 \pm 0.19$ | $74.7 \pm 3.3\%$ |
| | CVD | | |
| PRIMEROSE5.0 | $7.6 \pm 0.37$ | $0.89 \pm 0.05$ | $74.3 \pm 3.2\%$ |
| PRIMEROSE4.5 | $5.9 \pm 0.35$ | $0.71 \pm 0.05$ | $72.3 \pm 3.1\%$ |
| Experts | $8.5 \pm 0.43$ | $1.00 \pm 0.00$ | $82.9 \pm 2.8\%$ |
| PRIMEROSE | $4.3 \pm 0.35$ | $0.69 \pm 0.05$ | $74.3 \pm 3.1\%$ |
| C4.5 | $4.0 \pm 0.49$ | $0.65 \pm 0.09$ | $69.7 \pm 2.9\%$ |
| CN2 | $4.1 \pm 0.44$ | $0.64 \pm 0.10$ | $68.7 \pm 3.4\%$ |
| AQ15 | $4.2 \pm 0.47$ | $0.68 \pm 0.08$ | $68.9 \pm 2.3\%$ |

This focusing mechanism is not only specific to medical domain. In a domain in which a few diagnostic conclusions should be selected from many candidates, this mechanism can be applied. For example, fault diagnosis of complicated electronic devices should focus on which components will cause a functional problem: the more complicated devices are, the more sophisticated focusing mechanism is required. In such domain, proposed rule induction method will be useful to induce correct rules from datasets.

### 7.2 Sensitivity to Similarity

The problem with this approach is that several taxonomy trees are obtained when a single-valued similarity is adopted. If Simpson similarity is selected for grouping, two other models are acquired from the small dataset (Fig. 8,9). Although the model shown in Fig. 8 is topologically identical to Fig. 7, the grouping order is different. Thus, when the above rule induction method is applied, rules induced by this model may be different from the above rules. The other model is totally different from those two models, so the obtained rule will be different from the rule in Section 5.

Moreover, if the matching number is selected for grouping, the other model is acquired (Fig. 10).

The selection of the interval-valued similarity is a solution to this problem. However, since this choice may not prevent the multiple model generation in general, it will be our future work to introduce a preference criteria for model selection.

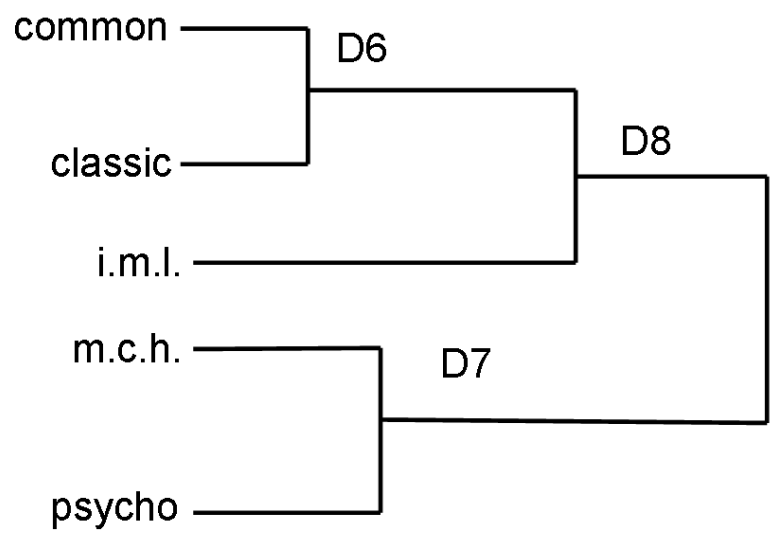

**Fig. 8.** The Second Grouping by Simpson Similarity

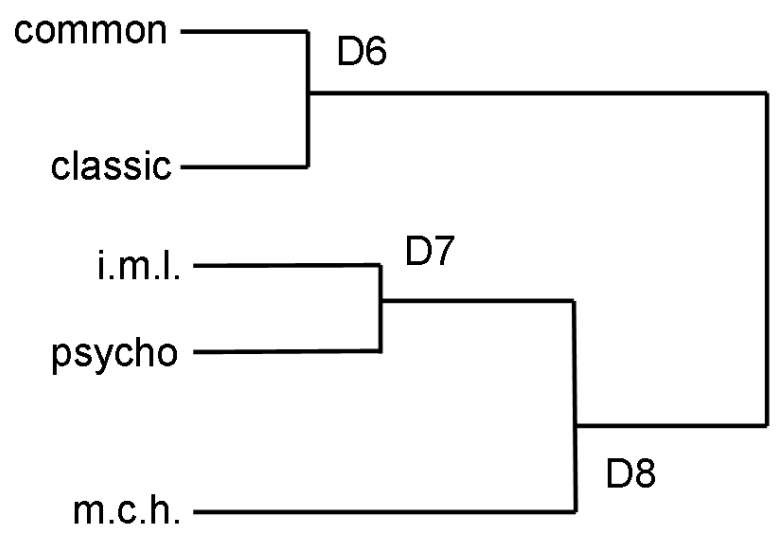

**Fig. 9.** The Third Grouping by Simpson Similarity

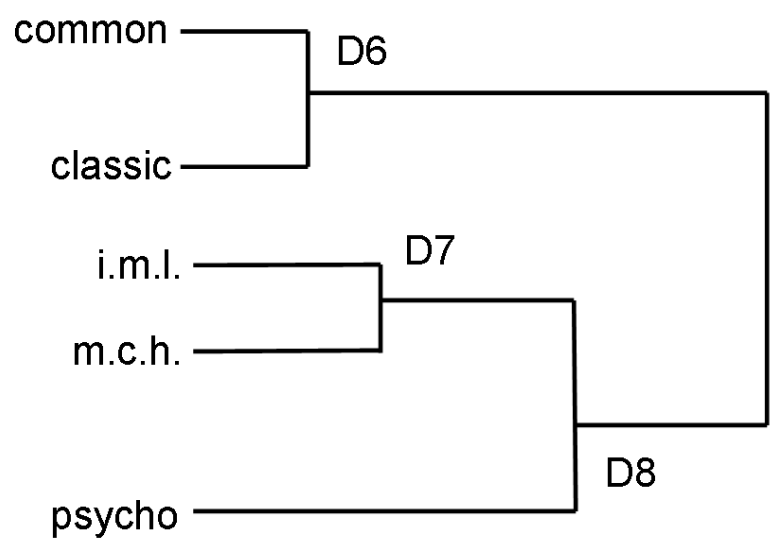

**Fig. 10.** The Second Grouping by Matching Number

## 8 Conclusion

In this paper, the characteristics of experts' rules are closely examined, whose empirical results suggest that grouping of diseases is very important to realize automated acquisition of medical knowledge from clinical databases. Thus, we focus on the role of coverage in focusing mechanisms and propose an algorithm for grouping of diseases by using this measure, which consists of the following three procedures. First, the characterization set of each attribute-value pair for a decision class(a given class) is extracted from databases. Then, similarities between the characterization sets are calculated. Finally, the concept hierarchy for given classes is calculated from the similarity values. The proposed method was evaluated on three medical datasets, the experimental results of which show that induced rules correctly represent experts' decision processes.

Although the proposed method gives a good performance with diagnostic taxonomy, it is possible that the method outputs multiple models. This observa-tion is dependent on the selection of the similarity measure. It will be our future work to solve this problem.

## Acknowledgements

This work was supported by the Grant-in-Aid for Scientific Research (13131208) on Priority Areas (No.759) "Implementation of Active Mining in the Era of Information Flood" by the Ministry of Education, Science, Culture, Sports, Science and Technology of Japan.

# Constructing Compact Dual Ensembles for Efficient Active Learning

Huan Liu[1], Amit Mandvikar[1], and Hiroshi Motoda[2]

[1]Department of Computer Science and Engineering, Arizona State University
[2] Institute of Scientific and Industrial Research, Osaka University
{huanliu,amit}@asu.edu, motoda@ar.sanken.osaka-u.ac.jp

**Abstract.** A good ensemble is one whose members are both *accurate* and *diverse*. Active learning requires a small number of *highly* accurate classifiers so that they will not disagree with each other too often. Ensemble method, however, are not good candidates for active learning because of their different design purposes. In this paper, we propose to use *dual ensembles* for active learning in binary-class domains, and investigate how to use the diversity of the member classifiers of an ensemble for efficient active learning. As active learning requires iterative training of the member classifiers in an ensemble, it is imperative to maintain a *small* number of classifiers in an ensemble for learning efficiency. We empirically show using benchmark data that (1) number of classifiers varies for different data sets to achieve a good (stable) ensemble; (2) feature selection can be applied to classifier selection to construct compact ensembles with high performance. A real-world application is used to demonstrate the effectiveness of the proposed approach.

## 1 Introduction

Active learning is a framework in which the learner has the freedom to select which data points are added to its training set [22]. An active learner may begin with a small number of labeled instances, carefully select a few additional instances for which it requests labels, learn from the result of that request, and then using its newly-gained knowledge, carefully choose which instances to request next. More often than not, data in forms of text (including emails), image, multi-media are unlabeled, yet many supervised learning tasks need to be performed [2, 18] in real-world applications. Active learning can significantly decrease the number of required labeled instances for effective learning, thus greatly reduce expert involvement in labeling and allow a vast body of supervised learning algorithms to be applied to mainly unlabeled data. In recent years, there has been considerable interest in ensemble methods [6, 11, 21]. Ensemble methods are learning algorithms that construct a set of classifiers and then classify new instances by taking a weighted or unweighted vote of their predictions. An ensemble often has smaller expected loss or error rate than any of the $n$ individual (member) classifiers. A good ensemble is one whose members are both *accurate* and *diverse* [7, 12].

On the first glimpse, it seems straightforward that ensemble methods can be employed to build classifiers for active learning. A closer look suggests otherwise. This work explores the relationship between the two learning frameworks, attempts to take advantage of the good learning performance of ensemble methods for active learning in a real-world application, and studies how to construct an ensemble for effective active learning. In the following, we will first study the relationship between the two in detail in Section 2, propose to use dual ensembles for active learning in Section 3, next discuss the diversity issue of ensemble learning with respect to ensemble size - the number of member classifiers in an ensemble as well as empirical results on the benchmark data sets in Section 4, and then go into details of selecting the necessary and diverse member classifiers for an ensemble in Section 5. The experimental results and discussions of active learning with dual ensembles are presented in Section 6. The work is concluded in Section 7.

## 2 Ensembles and Active Learning

Active learning aims to reach high performance using as few labeled instances as possible. It can be very useful where there are limited resources for labeling data, and obtaining these labels is time-consuming or difficult [22]. There exist widely used active learning methods. Some examples are: Uncertainty sampling [15] selects the instance on which the current learner has lowest certainty; Pool-based sampling [17] selects the best instances from the entire pool of unlabeled instances; and Query-by-Committee [10, 23] selects instances that have high classification variance themselves. Query-by-Committee (QBC) measures the variance indirectly, by examining the disagreement among class labels assigned by a set of classifier variants, sampled from the probability distribution of classifiers that results from the labeled training instances. Now let us turn to ensemble methods that also involve building a set of classifiers.

Studying methods for constructing good ensembles of classifiers has been one of the most active areas of research in supervised learning [7]. The main discovery is that ensembles are often much more accurate than the member classifiers that make them up. A necessary and sufficient condition for an ensemble to be more accurate than any of its members is that the member classifiers are accurate and diverse [12]. An accurate classifier is one that has an error rate of better than random guessing on new instances; more specifically, each member classifier should have its error rate below 0.5. Two classifiers are diverse if they make different (or uncorrelated) errors on new data points. In reality, the errors made by member classifiers will never be completely independent of each other, unless the predictions themselves are completely random (in which case the error rate will be greater than 0.5) [11]. However, so long as each member's error rate is below 0.5, with a sufficient number of members in an ensemble making somewhat uncorrelated errors, the ensemble's error rate can be very small as a result of voting. Many methods for constructing ensembles have been developed such as Bagging [3], Boosting [9], and Error-correction Output Coding [8]. We consider Bagging in this work as it is the most straightforward way of manipulating the training data [7]. Bagging relies on bootstrap replicates of the original training data to generate multiple classifiers that form an ensemble. Each bootstrap replicate contains, on the average, 63.2% of the original data, with several instances appearing multiple times.

After reviewing an active learning method QBC and an ensemble method Bagging, we notice that both employ a set of classifiers of the same type: active learning uses the set of classifiers to find instances that the classifiers disagree about their predictions, but ensemble learning is to use the set of classifiers to increase diversity in order to achieve high predictive accuracy. Both count on disagreement or diversity of classifiers. Disagreement is closely associated with diversity. Classifiers that do not disagree are not diverse, in other words, only diverse classifiers will possibly disagree. Accuracy and diversity are, however, contradictory goals: diverse classifiers have to make errors on different instances; and accurate classifiers will agree with each other [11]. For example, if a classifier is 100% accurate, other equally accurate classifiers are impossible to disagree, no matter how many of them are generated.

Disagreement or diversity of classifiers are used for different purposes for the two learning frameworks: in ensemble learning, diversity of classifiers is used to ensure high accuracy by voting; in active learning, disagreement of classifiers is used to identify critical instances for labeling. For the former, we want as high diversity as possible; for the latter, disagreement should not occur too often as frequent disagreement requires more manual labeling. In order for active learning to work effectively, we need a *small*[1] number of *highly* accurate classifiers so that they will disagree with each other, but not too often (this is determined by the nature of highly accurate classifiers). Otherwise, the purpose of active learning to learn with as *few* instances as possible cannot be achieved. For ensemble learning to work, however, one should shun highly accurate classifiers in order to achieve high diversity - weak learners can exhibit high diversity as we discussed earlier - with a *large* number of classifiers. Another essential difference between the two is that active learning is an iterative process and ensemble learning is not. Hence, ensemble learning such as Bagging cannot be simply employed for active learning like QBC.

Since ensemble methods have shown their *robustness* in producing *highly accurate* classifiers and each of member classifiers such as decision trees [5, 4, 19] can be *very efficient* in training and testing, we investigate below (1) how we can employ ensembles in active learning and (2) how we can build *compact* ensembles for efficient active learning.

## 3 Dual Ensembles for Active Learning

Dual ensembles are class-specific: one ensemble is built for each class in a binary class domain. For a single ensemble to be used in active learning, we need to determine two thresholds: $\delta_0$ and $\delta_1$ to define the majority for classes 0 and 1. That is to define what a majority of prediction is for 0 or 1 separately: if the number of "1" predictions is $> \delta_1$, the ensemble outputs 1; else if the number of "0" predictions is $> \delta_0$, then the ensemble outputs 0; otherwise, the ensemble is uncertain about its prediction. In addition, there could be many ways to define $\delta_0$ and $\delta_1$ for a reasonably large ensemble size. The dual ensembles only need one threshold for each ensemble to define majority which is easy to define: given $M$ classifiers, the threshold is $\lfloor (M+1)/2 \rfloor$. The above difference is illustrated in Figure 1. When dual ensembles $(E_1, E_0)$ disagree, uncertain predictions ensue. The disagreement between $E_1$ and $E_0$ occurs when both are certain but suggest different outcomes, or both are uncertain. Since ensembles $E_1$ and $E_0$ are highly accurate themselves, we do not expect that they frequently disagree. We use $E_1$ and $E_0$ to classify testing data set and select the uncertain instances by disagreement. We then ask the expert to label these instances and add the labeled instances to the training data. We continue this until there is not adequate performance increase in subsequent iterations.

[1] A small ensemble size will make iterative learning more efficient, other things being equal.

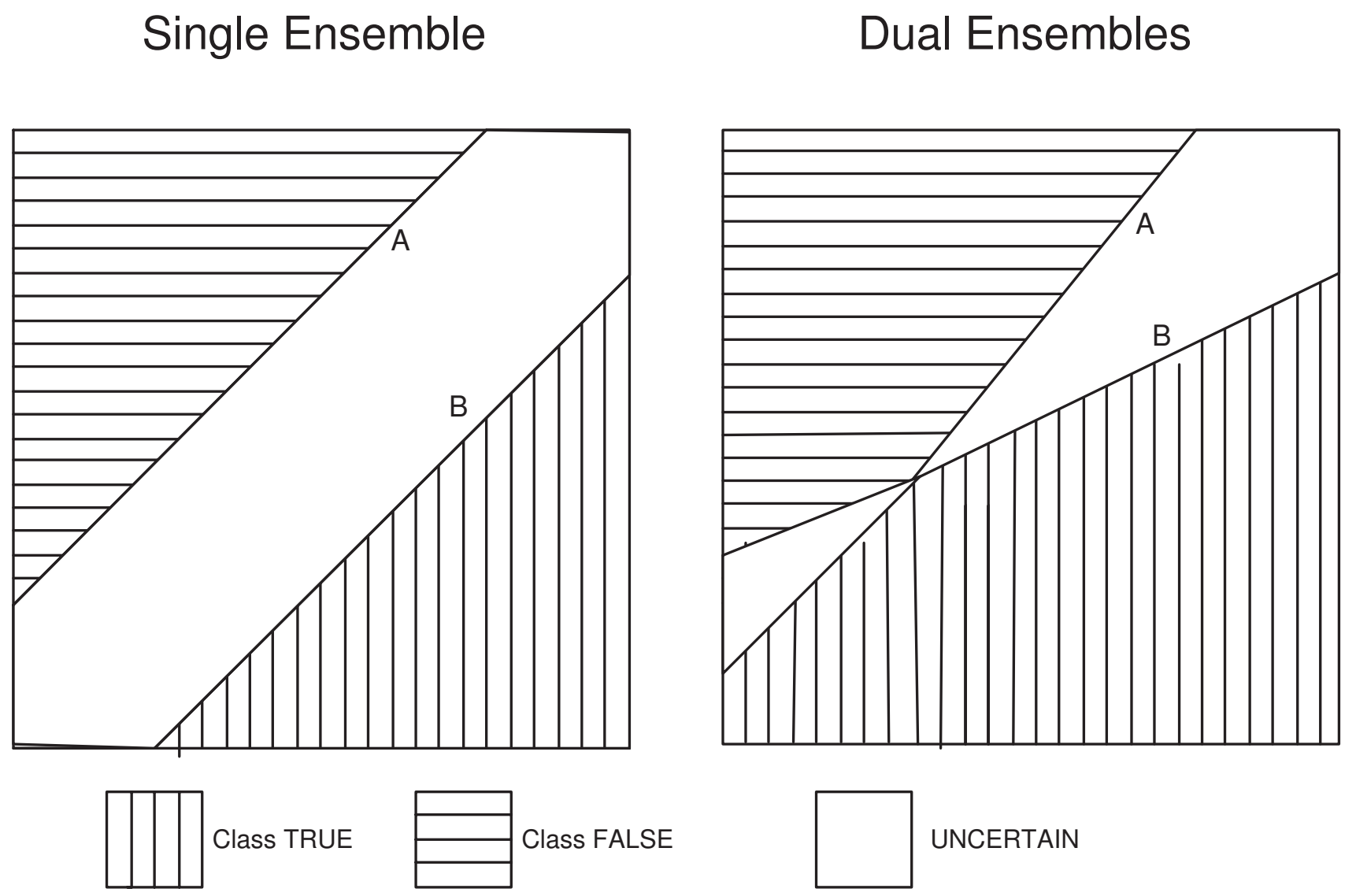

**Fig. 1.** Difference between single and dual ensembles. Classification is defined over the attribute space. A and B define decision boundaries

Active learning is an iterative process, hence using ensembles in active learning imposes an additional constraint: only a necessary number of member classifiers should be used and the number should be kept small so long as accuracy and diversity are maintained. This is because a large number of member classifiers will incur large (re)training cost for active learning. We present the procedure of building dual ensembles in Figure 2. The use of feature selection is discussed in Section 5.

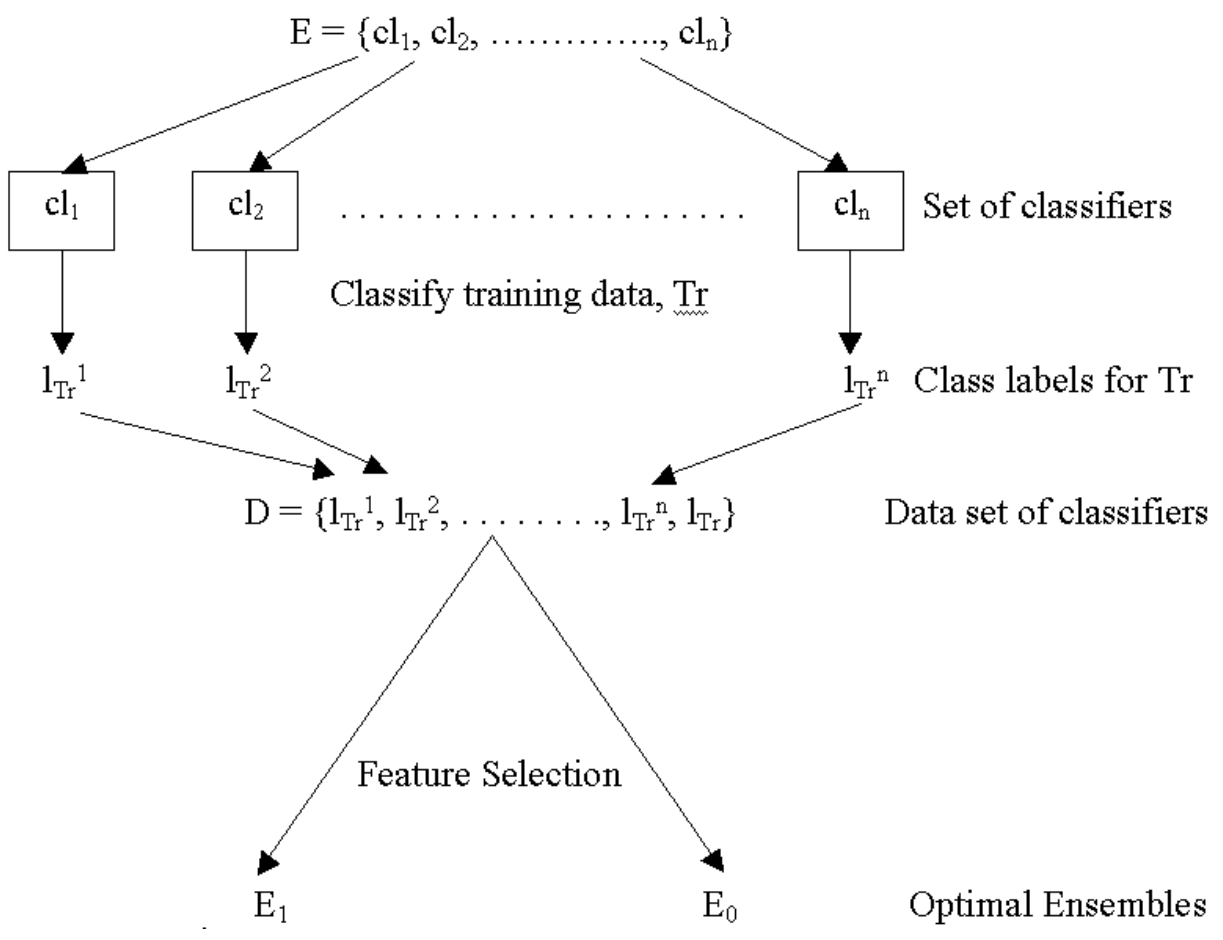

**Fig. 2.** Procedure to build dual ensembles.

We empirically investigate next whether it is possible to find compact ensembles with good performance.

## 4 Accuracy and Diversity of Ensembles

Intuitively, ensemble size required for ensemble learning mainly hinges on the complexity of the training data. For a fixed type of classifier (say, decision trees), the more complex the underlying function of the data is, the more members an ensemble needs. The complexity of the function can always be compensated by increasing the number of members for a given type of classifier until the error rate converges [4, 9]. As we mentioned earlier, an ensemble's goodness can be measured by accuracy and diversity. Following [11], let $\hat{Y}(x) = \hat{y}_1(x), ...\hat{y}_n(x)$ the set of the predictions made by member classifiers $C_1, ..., C_n$ of ensemble $E$ on instance $\langle x, y\rangle$ where $x$ is input, and $y$ is the true class. We give some definitions below.

**Definition 1.** *The* **ensemble prediction** *of a uniform voting ensemble for input $x$ under loss function $l$ is $\hat{y}(x) = arg\min_{y\in Y} E_{c\in C}[l(\hat{y}_c(x), y]$.*

The ensemble prediction is the one that minimizes the expected loss between the ensemble prediction and the predictions made by each member classifier $c$ for the instance $\langle x, y\rangle$.

**Definition 2.** *The* **loss** *of an ensemble on instance $\langle x, y\rangle$ under loss function $l$ is given by $L(\langle x, y\rangle) = l(\hat{y}(x), y)$.*

The error rate of a data set with $N$ instances can be calculated as $e = \frac{1}{N}\sum_1^N L_i$ where $L_i$ is the loss for instance $x_i$. **Accuracy** of ensemble $E$ is $1 - e$.

**Definition 3.** *The* **diversity** *of an ensemble on input $x$ under loss function $l$ is given by* $\overline{D} = E_{c \in C}[l(\hat{y}_c(x), \hat{y}(x))]$.

The diversity is the expected loss incurred by the predictions of the member classifiers relative to the ensemble prediction. Commonly used loss functions include square loss ($l_2(\hat{y}, y) = (\hat{y} - y)^2$), absolute loss ($l_{||}(\hat{y}, y) = |\hat{y} - y|$), and zero-one loss ($l_{01}(\hat{y}, y) = 0$ iff $\hat{y} = y$; $l_{01}(\hat{y}, y) = 1$ otherwise). In case of a binary classification problem, these give the same result. We proceed to conduct experiments below.

### 4.1 Experiments on Benchmark Data Sets

The purpose of the experiments in this section is to observe how diversity and error rate change as ensemble size increases. We use benchmark data sets [1] in the experiments. These data sets have different numbers of classes, different types of attributes and are from different application domains.

We used Weka [24] implementation of Bagging [3] as the ensemble generation method and used J4.8 [24](the Weka's implementation of C4.5) without pruning as the base learning algorithm in the experiments. For each data set, we run Bagging with increasing ensemble sizes from 5 to 151 and record each ensemble's error rate $e$ and diversity $\underline{D}$. We run 10-fold cross validation and the average values for $\overline{e}$ and $\overline{D}$ are calculated.

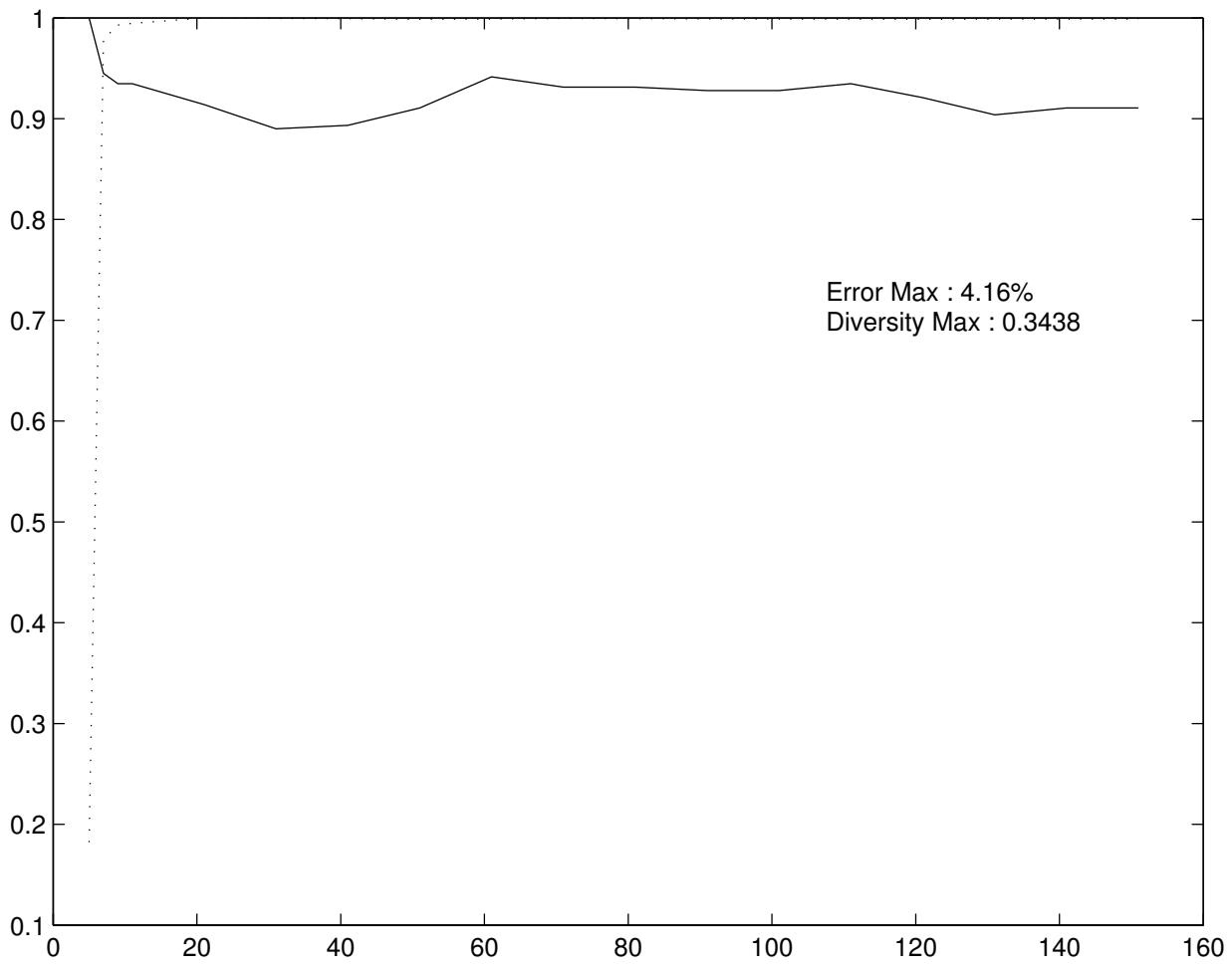

**Fig. 3.** Normalized diversity and Error plots for breast data. "1" corresponds to given Max values.

**Table 1.** Ensemble diversity and error rates for different ensemble sizes on various benchmark data sets.

| Dataset | Diversity $\overline{D}$ | | | | | | Error Rate (%) $\overline{e}$ | | | | | |
|---|---|---|---|---|---|---|---|---|---|---|---|---|
| | 5 | 9 | 21 | 61 | 101 | 141 | 5 | 9 | 21 | 61 | 101 | 141 |
| anneal | 0.228 | 0.236 | 0.237 | 0.237 | 0.237 | 0.237 | 1.103 | 1.225 | 1.180 | 1.136 | 1.169 | 1.203 |
| audiology | 0.378 | 0.701 | 0.699 | 0.732 | 0.739 | 0.741 | 18.938 | 18.230 | 16.947 | 16.460 | 16.726 | 16.593 |
| autos | 0.354 | 0.604 | 0.640 | 0.664 | 0.671 | 0.674 | 21.073 | 18.098 | 15.659 | 15.561 | 14.927 | 14.878 |
| balance | 0.182 | 0.455 | 0.482 | 0.508 | 0.508 | 0.514 | 18.256 | 17.456 | 16.832 | 16.736 | 16.672 | 16.656 |
| breast | 0.063 | 0.341 | 0.344 | 0.343 | 0.343 | 0.343 | 4.163 | 3.891 | 3.805 | 3.920 | 3.863 | 3.791 |
| breast-c | 0.161 | 0.150 | 0.159 | 0.163 | 0.162 | 0.164 | 27.867 | 27.378 | 27.028 | 26.818 | 26.573 | 26.469 |
| colic | 0.280 | 0.310 | 0.328 | 0.328 | 0.328 | 0.328 | 14.783 | 14.565 | 14.375 | 14.212 | 14.158 | 14.185 |
| colic-orig | 0.099 | 0.100 | 0.117 | 0.110 | 0.104 | 0.101 | 33.696 | 33.696 | 33.696 | 33.696 | 33.696 | 33.696 |
| credit-a | 0.357 | 0.398 | 0.431 | 0.442 | 0.443 | 0.444 | 14.261 | 13.957 | 14.000 | 13.725 | 13.681 | 13.739 |
| credit-g | 0.234 | 0.252 | 0.264 | 0.265 | 0.265 | 0.265 | 27.590 | 26.450 | 25.790 | 25.210 | 24.930 | 24.950 |
| diabetes | 0.279 | 0.302 | 0.338 | 0.342 | 0.341 | 0.341 | 25.690 | 24.609 | 23.620 | 23.242 | 23.034 | 23.073 |
| glass | 0.476 | 0.544 | 0.592 | 0.625 | 0.625 | 0.637 | 27.196 | 25.467 | 23.084 | 23.505 | 22.897 | 22.710 |
| heart-c | 0.300 | 0.353 | 0.405 | 0.432 | 0.442 | 0.447 | 19.175 | 19.175 | 17.921 | 16.898 | 16.106 | 16.139 |
| heart-h | 0.319 | 0.343 | 0.346 | 0.352 | 0.351 | 0.352 | 20.034 | 20.374 | 20.000 | 20.306 | 20.578 | 20.578 |
| heart-st | 0.329 | 0.352 | 0.394 | 0.426 | 0.436 | 0.437 | 21.407 | 20.889 | 20.667 | 19.593 | 19.815 | 19.889 |
| hepatitis | 0.168 | 0.181 | 0.180 | 0.182 | 0.180 | 0.180 | 17.290 | 17.742 | 16.774 | 16.129 | 16.258 | 16.129 |
| ionosphere | 0.116 | 0.321 | 0.327 | 0.327 | 0.327 | 0.327 | 8.319 | 7.749 | 7.464 | 7.550 | 7.407 | 7.379 |
| iris | 0.059 | 0.611 | 0.624 | 0.636 | 0.649 | 0.652 | 5.267 | 5.400 | 5.667 | 5.200 | 5.200 | 5.200 |
| kr | 0.398 | 0.438 | 0.473 | 0.477 | 0.477 | 0.477 | 0.626 | 0.620 | 0.645 | 0.576 | 0.582 | 0.563 |
| labor | 0.234 | 0.297 | 0.325 | 0.323 | 0.321 | 0.319 | 14.211 | 13.860 | 12.281 | 11.579 | 11.930 | 11.754 |
| lymph | 0.231 | 0.396 | 0.425 | 0.438 | 0.440 | 0.441 | 21.554 | 20.473 | 19.595 | 20.068 | 19.932 | 19.392 |
| mushroom | 0.352 | 0.397 | 0.431 | 0.459 | 0.465 | 0.472 | 0.000 | 0.000 | 0.000 | 0.000 | 0.000 | 0.000 |
| prim-tumor | 0.526 | 0.700 | 0.739 | 0.751 | 0.752 | 0.752 | 58.289 | 56.873 | 55.310 | 54.100 | 54.366 | 54.366 |
| sonar | 0.358 | 0.402 | 0.435 | 0.452 | 0.457 | 0.459 | 24.904 | 21.875 | 21.539 | 21.298 | 21.587 | 21.154 |
| soybean | 0.772 | 0.775 | 0.824 | 0.845 | 0.850 | 0.853 | 8.258 | 7.599 | 7.291 | 7.072 | 6.969 | 6.838 |
| vehicle | 0.231 | 0.396 | 0.425 | 0.438 | 0.440 | 0.441 | 27.589 | 26.891 | 26.868 | 26.277 | 26.277 | 26.525 |
| vote | 0.068 | 0.380 | 0.385 | 0.384 | 0.384 | 0.384 | 3.609 | 3.494 | 3.333 | 3.287 | 3.241 | 3.218 |
| zoo | 0.302 | 0.579 | 0.588 | 0.593 | 0.593 | 0.593 | 7.129 | 7.228 | 6.931 | 7.525 | 7.723 | 8.020 |
| image | 0.318 | 0.368 | 0.416 | 0.442 | 0.460 | 0.470 | 0.093 | 0.093 | 0.093 | 0.095 | 0.0951 | 0.0951 |

### 4.2 Results and Discussion

We report diversity and error rates of the sample ensemble sizes (5, 9, 21, 61, 101, 141) in Table 1. The last data set (Image) is from our application domain to be explained later. We have run experiments with 18 ensemble sizes (5, 7, 9, 11, 21, 31, 41, 51, 61, 71, 81, 91, 101, 111, 121, 131, 141, and 151) with 10-fold cross validation for each data set (29 sets in total). Note that for mushroom dataset the error rates are all 0 whereas the diversities are not zero. This is because the error rate becomes 0 if the majority of the member classifiers gives a correct class, even if all of them are not necessarily the same. In Figures 3 and 4, two sets of curves are demonstrated. Both diversity values (**dashed lines**) and error rates (**solid lines**) are normalized for plotting purposes. The vertical axis shows percentage ($p$). The max values of diversity and error rate are given in each figure. We can derive absolute values for diversity and error rates following $Max \times p$. The trends of diversity and error rates are of our interest. We can observe a general trend that diversity values increase and approach to the maximum, and error rates decrease and become stable as ensemble size increases.

The results show that smaller ensembles (with around 30-70 classifiers) can achieve accuracy and diversity values similar to those of larger ensembles. In the following section, we will show a procedure for selecting compact dual ensembles and use these findings for a real-world application on image classification with unlabeled data and propose a novel feature selection approach to choose member classifiers.

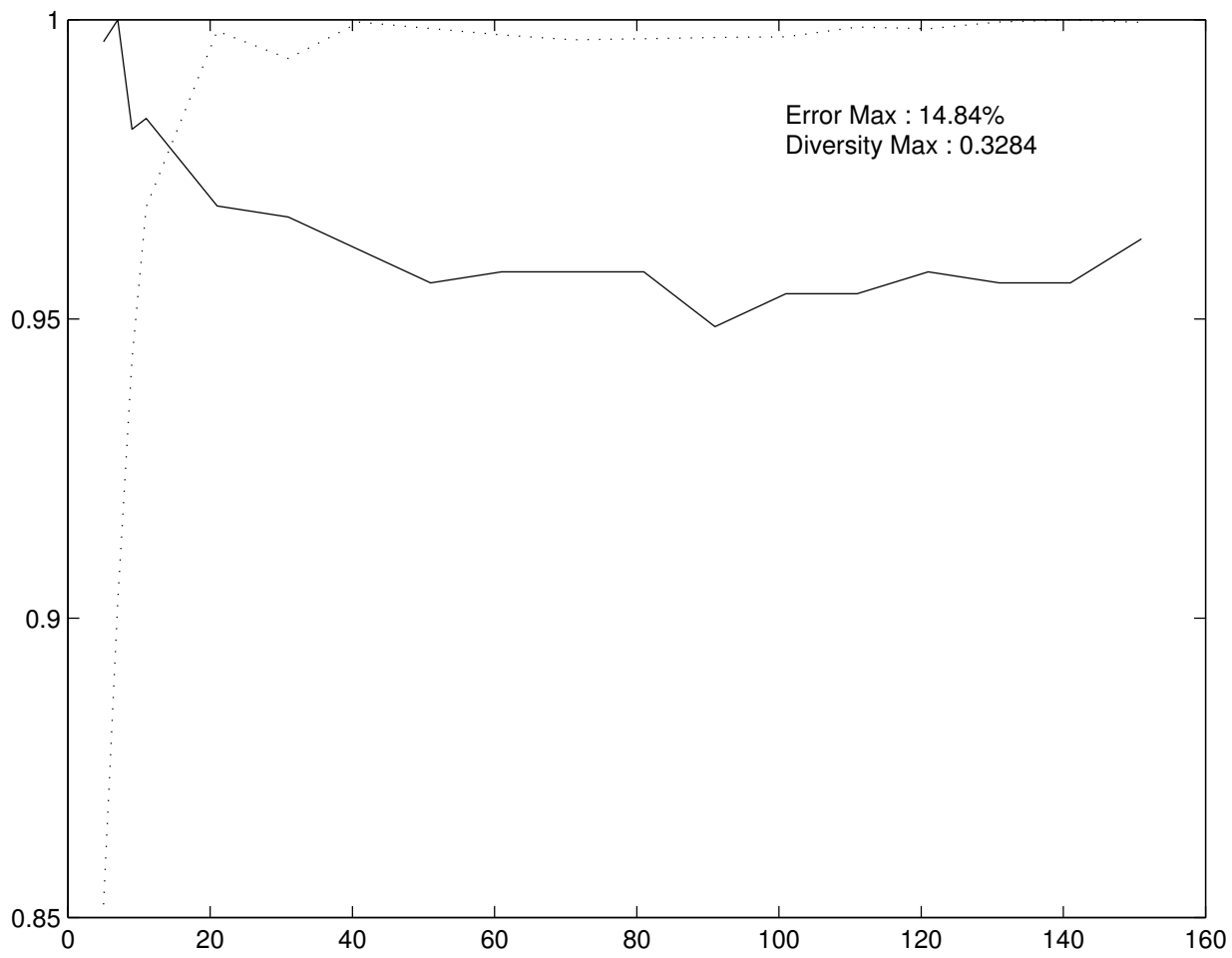

**Fig. 4.** Normalized diversity and Error plots for colic data. "1" corresponds to given Max values.

## 5 Selecting Compact Dual Ensembles via Feature Selection

The experiments with the benchmark data sets show that one can find an ensemble with a small number of member classifiers that can maintain similar accuracy and diversity to those of larger ensembles. Effectively selecting a small number of such classifiers will facilitate the building of dual ensembles for active learning. We show now how feature selection can be applied to classifier selection for compact ensembles. Conventional feature selection methods [13, 14, 16] select features by optimizing one single criterion (e.g., accuracy, consistency, dependency, correlation). In this case, we need to select features with two criteria (accuracy and diversity). In addition, features actually represent member classifiers, therefore we also need to consider this special nature for feature selection. In the following, we first briefly introduce the application domain - classification of *unlabeled* images, then introduce how a training data set is constructed based on the predictions of member classifiers, and propose a feature selection algorithm that is designed for selecting classifiers based on accuracy and diversity in order to effectively build dual ensembles.

### 5.1 Active learning in image domain

The real-world problem we face is to classify Egeria Densa in images. Egeria is an exotic submerged aquatic weed causing navigation and reservoir-pumping problems in the west coast of the USA. As a part of a control program to manage Egeria, classification of Egeria regions in aerial images is required. This task can be stated more specifically as one of classifying massive data *without class labels.* Relying on human experts for labeling Egeria regions is not only time-consuming and costly, but also inconsistent in their performance of labeling. Massive manual classification becomes impractical when images are complex with many different objects (e.g., water, land, Egeria) under varying picture-taking conditions (e.g., deep water, sun glint). In order to automate Egeria classification, we need to ask experts to label images, but want to minimize the task. Active learning is employed to reduce expert involvement in labeling images. The idea is to let experts label some instances of Egeria and non-Egeria regions, learn from these labeled instances, and then apply the active learner to new images. New instances will be recommended by the active learner for labeling, but the number of such instances is expected to be significantly less than labeling all instances in new images. Since experts are still involved in the process of active learning, the retraining with recently requested labeled instances has to be fast so the expert can be actively engaged in the process for high performance classification. Therefore, we need to employ very strong learners (such as ensembles) in order to learn with as few labeled instances as possible. We discuss how to construct dual ensembles for this purpose. Each image consists of 5329 instances ($73 \times 73$ regions) represented by 13 attributes of color, texture and edge.

### 5.2 Training data for classifier selection

Often 50-100 member classifiers are used to generate ensembles [4, 20]. They work well for a variety of data sets, as also shown in our benchmark data experiments. Since the initial training of ensembles for active learning is off-line, we can afford to choose a larger number. We build our starting ensemble $E_{max}$ by setting $max = 100$ member classifiers in this work. The essential problem can be rephrased as: given an ensemble $E_{max}$ with 100 member classifiers, *efficiently find* a compact ensemble $E_M$ composed of $M$ classifiers, with $M$ being the *smallest* number of member classifiers that can have similar error rate and diversity of $E_{max}$.

To generate a training set for the task of selecting member classifiers, we first perform Bagging with 100 member classifiers. We then use the learned classifiers ($C_k$) to generate predictions for instance $\langle x_i, y_i \rangle : \hat{y}_i^k = C_k(x_i)$. The resulting data set consists of instances of the form $((\hat{y}_i^1, ..., \hat{y}_i^K), y_i)$. After this data set is constructed, the problem of selecting member classifiers becomes one of feature selection.

### 5.3 Algorithm to efficiently determine ensemble size

Using Bagging, we employ only one learning algorithm - decision trees, so each member classifier should be equally good. That is, we should not expect any one classifier to be significantly superior to the others. However, when the ensemble size ($M$) is sufficiently large, accuracy of the members can remain high via voting. Likewise, diversity of an ensemble is also determined by $M$: an ensemble with a single member has diversity value 0 according to Definition 3. Evidence in the experiments on benchmark data sets suggests that there exists a necessary ensemble size beyond which the performance improvement as the ensemble size increases is not significant.

**DualE: selecting compact dual ensembles**

**input:** $Tr$ : Training data,
$FSet$ : Full set of classifiers in $E_{max}$,
$N$ : size of $FSet$ i.e., $max$,

**output:** $E_1$ : Optimal ensemble for class=1,
$E_0$ : Optimal ensemble for class=0;

```
01 begin
02     Generate N classifiers from T_r with Bagging;
03     Tr_1 ← Instances(Tr) with class label= 1;
04     Tr_0 ← Instances(Tr) with class label= 0;
05     Calculate diversity, D_0 and error rate, e_0 for E_max
       on Tr_1;
06     U ← N;
07     L ← 0;
08     M ← (U+L)/2;
09     while |U − M| > 1
10         Pick M classifiers from FSet to form E';
11         Calculate diversity, D' and error rate, e' for E'
           on Tr_1;
12         if ((D_0−D')/D_0 < 1%) and ((e'−e_0)/e_0 < 1%)
13             U ← M;
14             M ← M - (M−L)/2;
15         else
16             L ← M;
17             M ← M + (U−M)/2;
18         end;
19     end;
20     E_1 ← E';
21     Repeat steps 5 to 19 for Tr_0;
22     E_0 ← E';
23 end;
```

**Fig. 5.** Algorithm for selecting classifiers

**Table 2.** Comparison between selected dual ensembles with $E_{max}$ for Breast data

| | Dual $E_s$ | | Dual $E_r$ | | | | $E_{max}$ | |
|---|---|---|---|---|---|---|---|---|
| | Acc% | #UC | Acc% | #UC | Acc Gain% | UC Incr% | Acc% | Acc Gain% |
| Fold 1 | 95.9227 | 3 | 94.0773 | 13.6 | -1.9238 | 353.33 | 96.1373 | 0.2237 |
| Fold 2 | 97.2103 | 5 | 94.4206 | 15.2 | -2.8698 | 204.00 | 96.9957 | -0.2208 |
| Fold 3 | 94.8498 | 12 | 93.5193 | 8.7 | -1.4027 | -27.50 | 94.4206 | -0.4525 |
| Average | 95.9943 | 6.67 | 94.0057 | 12.5 | -2.0655 | 176.61 | 95.8512 | -0.1498 |

Therefore, we only need to determine ensemble size $M$ which is the smallest and can keep similar accuracy and diversity of $E_{max}$. We design an algorithm **DualE** that takes $O(\log max)$ to determine $M$ where $max$ is the size of the starting ensemble (e.g., 100)[2]. In other words, we test an ensemble $E_M$ with size $M$ which is between upper and lower bounds $U$ and $L$ (initialized as $max$ and 0 respectively). If $E_M$'s performance is similar to that of $E_{max}$, we set $U = M$ and $M = (L + M)/2$ ; otherwise, set $L = M$ and $M = (M + U)/2$. The details are in Figure 5. What still remains is the definition of performance similarity between two ensembles. The performance is defined by error rate $e$ and diversity $D$. The diversity values of the two ensembles are similar if $\frac{D_0 - D'}{D_0} \leq p$ where $p$ is a user defined number ($0 < p < 1$) for defining similarity (the smaller it is, the more similar) and $D_0$ is of the reference ensemble. In the same spirit, the error rates of the two ensembles are similar if $\frac{e' - e_0}{e_0} \leq p$ where $e_0$ is of the reference ensemble.

## 6 Experiments

Two sets of experiments are conducted with **DualE**: one is on a benchmark data set and the other is on the image data. The purpose is to examine if the compact dual ensembles selected by **DualE** can work as expected. When dual ensembles are used, it is possible that they give different class labels to some instances. These instances are called *uncertain* instances. In the context of active learning, the uncertain instances will be given to an expert for labeling. Therefore, the number of uncertain instances is reported in the experiments below in addition to accuracy. For ensemble $E_{max}$, the prediction of $E_{max}$ is the majority of the predictions of the member classifiers, and there is no disagreement. So for $E_{max}$ only the accuracy is reported and there are no uncertain instances.

### 6.1 Benchmark data experiment

The classic 10-fold cross validation results of benchmark data sets are in Table 1. We design a new 3-fold cross validation scheme here, which uses 1-fold for training, the remaining 2 folds for testing. This is repeated for all the 3 folds of the training data. In addition to comparing with $E_{max}$, we also randomly select member classifiers to form dual ensembles. We do so 10 times and use their average accuracy and number of uncertain instances in comparison. The results are shown in Table 2. Average values for each column are also given. Gain (and Incr) is calculated against $E_s$ as $(V' - V_{E_s})/V_{E_s} \times 100$. Dual $E_s$ are the selected ensembles using **DualE** to ensure that diversity and accuracy of a compact ensemble are similar to $E_{max}$. Dual $E_r$ are randomly selected ensembles. Their results averaged over 10 such ensembles are shown in the table. Ensemble sizes of $E_1$ and $E_0$ for $E_S$ are 10, 5 for Fold 1; 5, 10 for Fold 2; and 11, 5 for Fold 3, respectively. Ensemble sizes of $E_1$ and $E_0$ for $E_r$ are the same as the ones in $E_s$ for the corresponding folds. The reduction from 100 to the range of 10 is significant.

[2] This design assumes that one can build an initial ensemble with very large $max$.

Comparing dual $E_s$ and dual $E_r$, we notice the differences: dual $E_r$ exhibit lower accuracy and higher number of uncertain instances, which manifest the importance of maintaining high accuracy and diversity in building compact ensembles. Comparing dual $E_s$ and $E_{max}$, we observe no significant change in accuracy. This is consistent with what we tried to do in **DualE** (maintaining both accuracy and diversity). Therefore, selected dual ensembles ($E_s$) can be used for active learning. The sizes of selected dual ensembles are much smaller than 100 - the size of $E_{max}$.

### 6.2 Image data experiment

For the image set, there are 17 images already labeled by experts. One image is used for training and the rest for testing. The training results (diversity and error rate) of 10-fold cross validation have been shown in Table 1 (last row). From the viewpoint of active learning, we want to have the training set as small as possible so that in practice, an expert does not need to label too many instances in order to obtain a training data set. The following benchmark data experiment is designed with this purpose in mind. We wish to see if what is learned from one training image can be applied to the remaining images. We first train an initial ensemble $E_{max}$ with $max = 100$ on the training image, then obtain accuracy of $E_{max}$ for the 17 testing images. As seen in the last row of Table 1, $E_{max}$ is very accurate in terms of 10-fold cross validation. Although images are aerial photos about Egeria, they were shot at different places and times. In other words, these images are similar, but do have their differences from the training image. The idea is to let the learned dual ensembles take care of the majority of the regions of the test images and only recommend the uncertain regions to an expert for labeling, and the labeled instances are used to adapt the dual ensembles. **DualE** found $E_1$ and $E_0$ of sizes 10 and 5, respectively. Again, they are significantly smaller than 100. The results are shown in Table 3. It clearly shows that accuracy of dual $E_s$ is similar to that of $E_{max}$. The number of uncertain regions is also relatively small (the smallest is 0, the largest is 88, the average is about 18). This clearly demonstrates the effectiveness of using dual ensembles for active learning in reducing the expert involvement for manual labeling.

**Table 3.** Selected dual ensembles vs. $E_{max}$ for Image data

| Image | $E_s$ | | $E_{max}$ | |
|---|---|---|---|---|
| | Acc% | #UC | Acc% | Acc Gain% |
| 1 | 81.91 | 1 | 81.90 | -0.0122 |
| 2 | 90.00 | 0 | 90.00 | 0.0000 |
| 3 | 78.28 | 38 | 79.28 | 1.2775 |
| 4 | 87.09 | 34 | 86.47 | -0.7119 |
| 5 | 79.41 | 0 | 79.73 | 0.4029 |
| 6 | 84.51 | 88 | 84.77 | 0.3076 |
| 7 | 85.00 | 3 | 85.41 | 0.4823 |
| 8 | 85.95 | 18 | 86.6 | 0.7562 |
| 9 | 71.46 | 0 | 72.32 | 1.2035 |
| 10 | 91.08 | 2 | 90.8 | -0.3074 |
| 11 | 89.15 | 31 | 88.82 | -0.3702 |
| 12 | 75.91 | 0 | 76.02 | 0.1449 |
| 13 | 66.84 | 0 | 67.38 | 0.8079 |
| 14 | 73.06 | 49 | 73.73 | 0.9170 |
| 15 | 83.1 | 1 | 83.24 | 0.1684 |
| 16 | 76.57 | 14 | 76.82 | 0.3265 |
| 17 | 87.67 | 31 | 88.42 | 0.8555 |
| Average | 81.58 | 18.24 | 81.86 | 0.3676 |

## 7 Conclusions

Ensemble methods such as Bagging can achieve good learning performance by increasing ensemble size for high diversity. They have been proven an efficient approach to classification problems. In this work, we point out that (1) ensemble methods are not suitable for active learning because active learning is an iterative process that interacts with a user for instance labeling; (2) dual ensembles are very good for active learning if we can build compact ensembles. Our empirical study suggests that there exist compact ensembles. We continue to propose **DualE** that can find compact ensembles with good performance via feature selection. Experiments on the benchmark data and image data exhibit the effectiveness of dual ensembles for active learning. We plan to extend dual ensembles to multiple ensembles to handle multi-class classification problems in our future work.

## Acknowledgments

We greatly appreciate Jigar Mody's help in implementing some code and running some experiments and Dr. Patricia Foschi's assistance and advice in getting data and interpreting them in this work.

# Generalising Incremental Knowledge Acquisition

Paul Compton, Tri M. Cao and Julian Kerr
compton,tmc,juliank@cse.unsw.edu.au

School of Computer Science and Engineering
University of New South Wales and
Smart Internet Technology Co-operative Research Centre

**Abstract.** We outline an approach to building knowledge-based system based on tightly controlling the order of evaluation of the knowledge components of the system. The order of evaluation is based on two relations, sequence and correction that correspond to the changes that an expert may wish to make to a knowledge base and knowledge acquisition is structured so that new knowledge is added having one of these relations with existing knowledge in the system. We further propose that the knowledge components added might be any knowledge-based systems or programs rather than rules. This proposal is a generalisation of the Ripple-Down Rule incremental approach to building knowledge-based systems.

## 1 Introduction

A rule based system can be described a set of rules, an inference engine to enable the rules to be evaluated, mechanisms to control the order in which rules are evaluated and the working memory which contains input data plus any output from rules that fire. The control information for how the rules are evaluated can be encoded as a *dependency graph* [ABW88,Col99] or implicitly stated in the software that implements the inference engine. Even though there has been some move towards second generation expert systems (which contain a conceptual model of the domain), the majority of expert systems developed to date have been rule based.

The importance of controlling the order of rule evaluation should be noted. It is often assumed that declarative programming avoids the programmer/knowledge engineer having to think about the order of evaluation. This is certainly not the case in rule-based systems and is probably not true for any declarative programming. Rule-based systems include control mechanisms such as conflict resolution strategies to decide which rule should be fired (its conclusion added to working memory) first if a number of rules are satisfied by the data. These strategies are essentially heuristic; e.g, fire the rule with greatest number of conditions first, so that there will be circumstance where the ordering is inappropriate. The knowledge engineer then changes the rules to make the conflict resolution strategy give the right conclusion. Conflict resolution applies to the order in which rule fires; the inference engine also evaluates rules in a particular order, which will need to be taken into account. For example, the MYCIN backward chaining inference engine evaluates rule conditions in the order in which the conditions occur in the rule. The MYCIN knowledge engineers therefore organised the order of conditions in rules so that when the system asks the user about a particular rule condition, it does this in an order that is appropriate for the domain. Much of the following is about developing a very explicit ordering for evaluation and firing so that they cannot and do not need to be controlled by the person adding rules. That is, knowledge acquisition becomes more declarative if firing and evaluation and firing ordering becomes more procedural.

This description of a rule-based system can be generalised if we consider that a single rule plus inference engine can be considered as a program: that is, given some input the rule evaluation may result in some output (if the rule fires). A rule-based system can thus be generalised to a set of programs, mechanisms to control the order in which these programs are run and working memory where the original input data plus any output from the various programs is stored. Of course what makes such a system a knowledge- based is that the programs (or rules) are added to capture the preferences and beliefs of the owner/supervisor/teacher of the system in some sort of knowledge acquisition process.

In such systems there would be two situations where knowledge acquisition was required: firstly the system's output is correct but incomplete and secondly that part of whole of the output is wrong. If we assume that we cannot look inside the system to fix it and that the system has a lot of value so that we do not want to discard it, then these errors must be fixed by adding a second KBS so that the output of the first is passed to the second to have extra information added or to have the output from the first system replaced in the specific circumstances where it has been found to be wrong.

This is what must happen at the atomic rule level, except hidden by the editing that occurs.

– If we have a system with a single rule then we add an extra rule to deal with some new circumstance.
– If the rule fires inappropriately we add extra conditions to it to restrict the circumstance in which it fires and add a new rule to give the correct conclusion in the circumstances. Despite the two step editing, what is happening logically is that the new rule specifies the circumstances in which its conclusion should be given rather than the previous rule. That is if both rules fire the conclusion of the later rule replaces that of the initial rule.

If we cannot edit the rule, (or the black box that has produced the incorrect output), all we can do is to add a program that replaces the output of the first program in certain circumstances.

Cases that are used to test the knowledge base are of central importance. The only possible of way of characterising, testing or evaluating a knowledge base is via a set of cases. As will be discussed, these cases may not test every possible behaviour of the KBS; the testing is only as good as the cases available. However, regardless of the quality of the cases, they are the only way of characterising the system.

This can be seen in the example of a single rule: If the single rule is an overgeneralisation and gives the wrong conclusion for a case, it is equally that the new rule that replaces the conclusion of the first rule, will also be an overgeneralisation. If the expert overgeneralised the first time, how can he or she be relied upon to do better next time. If the new rule is to result in an improvement there must be a case or cases which correctly fire the first rule and are assumed to specify its scope, which need to be tested to see that they are still handled correctly by the first rule rather than the correction rule. The philosophical arguments for why an expert can never be relied on are outlined in [CE90].

A central insight of this work is that is that the two tasks of adding knowledge to add conclusions or replace conclusions do not have to be implicit in knowledge engineering. Rather systems can be recursively structured so that all knowledge acquisition is explicitly achieved by adding a KBS/program/rule to augment or replace output. Secondly, we hypothesise that there can be no advantage in carrying out these tasks in an implicit way by editing the knowledge base. Rather, there is risk or introducing other errors in such editing. The assumption here is that the circumstances in which knowledge is appropriate are never fully defined (and cannot be) so that it is inappropriate to hope for a perfect fix for such knowledge, the fix will never be complete. Ultimately the only way to handle errors is to provide knowledge of the circumstances where output has to be added to or replaced. And this knowledge will also need to be augmented or replaced.

Rule-level versions of these ideas are known as Ripple-Down Rules (RDR). Various RDR approaches have been developed and applied to a range of domain including: pathology [EC93], configuration [CRP$^+$98], control [SS97], heuristic search [BH00], document management using multiple classification [KYMC97], and resource allocation [RC98] with considerable success. The generalisation here is an extension of a previous generalisation applying only to rules [CR00]. Beydoun and Hoffman have also generalised RDR with a multiple RDR knowledge base approach, Nested RDR (NRDR). However, their linking between knowledge bases is via intermediate conclusions of concepts as used in heuristic classification [Cla85]. The generalisation here attempts to control the linking between knowledge-bases itself in an RDR-like fashion. An example of this is an image processing system which links decision trees developed by machine learning in an refinement structure [KC03].

## 2 Basic Concepts

### 2.1 Input

The inputs of each system are called cases. A case is the data, relationships in the data and any theory that may be provided as input to the KBS plus the output of the KBS. Note that a KBS cannot change the case it is provided with; it can only add to it. That is a KBS linked to a blackboard could not delete information from the blackboard, it could only add information. This does not mean the KBS cannot decide there is something wrong with the information it is provided. It means that if the case were rerun after being processed, the same output would be provided again. The the output would be the original input plus some sort of statement that there is something wrong with the input.

Informally, we define a case as a finite list of atomic objects which can be originally given or as output from evaluation. The atomic object representation will depend on the underlying language used. For example, if the language used is a first order language, the objects will be the set of atomic formulas. For a configuration task, the objects here are variable assignments. The order of the list is important because it keeps track of the order of evaluation.

### 2.2 Output

Output can be either information of some type that is added to the case, or a request for some other agent to add information to the case. If information is added, it may be a classification, a design or theory or it may be adding relations to the data; e.g. in a resource allocation problem resources are assigned to users of those resources, and perhaps temporal or spatial relations are also added specifying the temporal sequence and relative locations of resources.

If the output is a request for another agent to add information, this agent may be a human who provides other information or a program for example carrying out a calculation, another KBS of the type defined above or another kind of knowledge based system.

In particular, a special symbol, called $\mathrm{NO_OUTPUT}$ is used when no output can be derived from the current input. That is the system does not have knowledge relevant to the case in hand.

### 2.3 Primary rules

A primary rule (or a clausal rule) is a formula of the form

$$O \longleftarrow L_1, L_2, \ldots, L_m$$

where the $L_i$'s are conditions that refer to information in the case and the condition will be true or false depending on whether the relevant object is in the case. $O$ is either an atomic object or a request for a additional information or the special symbol, $\mathrm{NO_OUTPUT}$.

### 2.4 RDR Agent

An RDR agent manages how control is passed between the various programs used and how data is posted to the blackboard and passed to programs. Note that the programs called may themselves be other RDR agents who may have their own blackboards and programs which they call. The RDR agent is simply a blackboard controller, but one which organises how programs are called. The RDR agent does not have any explicit human knowledge, as the control structure it learns is determined simply by whether a correction or an additional knowledge base is invoked.

### 2.5 Control mechanism

The control mechanisms here are very simple and are of three types. One type of control mechanism handles requests for specific programs (see output above), the other organises the sequence of KBS independent of requests for specific agents. The sequence is determined by two types of relations between KBS: sibling and correction relations. These two relations are determined by the knowledge acquisition process.

For completeness there is also a general control mechanism that after each addition to the case (output posted to the blackboard), the whole reasoning process restarts with the first KBS. This is not a strict requirement, but the reasons for this will be outlined under knowledge acquisition.

**Information requests** If the output from a KBS indicates that a request be posted to another agent to provide information to the blackboard, this is acted on immediately. If the agent is unable to respond or does not respond quickly enough, then the answer NO_OUTPUT is posted. That is, the output from the KBS is the result of the action suggested and the KBS is only considered to have completed its task when a final response is posted.

**Sibling relation** A case is input to the first KBS which then produces output which is added to the case. This enhanced case is now passed to the second knowledge base and further output is added. Any case passed to the first KBS must be passed to the second KBS. That is, the original KBS is replaced by a sequence of KBS. There can be any number of KBS in a sequence of KBS, but the sequence replaces the original KBS (and each consequent sequence). As described below, extra KBS are added in a sibling relationship when knowledge inadequacies are discovered. Since they are added over time, the sibling sequence is always ordered by age or time of addition. Note again that the output from any later KBS cannot change the case it is provided with; i.e. the input plus output from the earlier sequence.

**Correction relation** A case is input to the first KBS, but is then passed to a second KBS before the output is added to the case (i.e. posted to the blackboard). If the second KBS provides output, this output from the second KBS is added to the case, not the output from the first KBS. If the second KBS does not provide any output then the output from the first KBS is added to the case. Any case passed to the first KBS which produces output, is passed to the second KBS. That is, the original KBS is replaced by a correction sequence of KBS. There can be any number of KBS in a correction sequence of KBS, but the sequence replaces the original KBS (and each consequent sequence).

More than one correction KBS can be added to correct a KBS. In this case, the correction KBSs have a sibling relation. That is, the case is initially passed to the first correction KBS, are then passed to the second correction KBS. If the first correction KBS adds output to the case, the second correction KBS acts as a conventional sibling KBS. However, if the first KBS does not add any output, the output from the original KBS is not immediately added to the case, rather the case is passed to the second correction KBS which may add output to replace the output of the original KBS, or if it too fails to add any output then the original output of the first KBS is added. (The circumstances in which more than one correction may apply will be discussed below)

A KBS can also be any combination of both types of sequences, resulting in a recursive structure, with these two types of relationships possible at every level. Note that a correction rule may be added to a KBS which is a sibling sequence of KBS. In this case all the output which is produced by the sibling sequence is replaced. Alternatively the correction may be added to the particular KBS that caused the error. The knowledge acquisition issues which determine which approach is used will be discussed. However, it should be noted that if a correction KBS replaces a specific KBS rather than a sequence, a case must be passed to a correction KBS before being passed to a sibling KBS to determine the output from the first KBS. That is the evaluation is depth first rather than breadth first.

**Repeat inference** After a piece of output is added to a case, control is passed back to the first KBS and inference starts again but with the enhanced case. This process is repeated until the case passes through the system with no more output being added . The reason for repeat inference is that some features of a case may be provided in the initial case on some occasions but on other occasions these same features will be generated as output from a KBS. If KBS which uses these features was developed before the KBS that produces the features, then the first KBS using the features will not be effectively used without repeat inference. The reason inference is repeated as soon as output is generated, is that the supervisor/owner decides extra output is required in the context that the previous repeat inference has been completed. Hence the new KBS, should only be used in the same circumstances.

## 3 Semi-formal Specification

In this section, we would like to give a semi-formal description of the proposed system. First, we look at the knowledge base representation. Second, we describe the control mechanism through the evaluation functions.

### 3.1 Representation

A knowledge base $K$ is one of three forms: a primary rule, or is composed from two component knowledge bases by sibling or correction relations. In addition, each knowledge base is associated with a set of input cases, named cornerstone cases. In a more formal way, we can define $K$ recursively:

$$K = \begin{cases} (R, D) \\ (\text{Sib}(K_1, K_2), D) \\ (\text{Cor}(K_1, K_2), D) \end{cases}$$

where $R$ is a primary rule defined above, $K_1, K_2$ are knowledge bases and $D$ is the cornerstone case set. Note that the cornerstone case set is attached to both primary rules and composite knowledge bases, they correspond to two different knowledge acquisition techniques described later: global refinement and local refinement.

Knowledge bases can communicate through special $Request$ objects. A $Request$ object contains the address of the agent which will carry out the request and the input data that had been passed to the knowledge base.

### 3.2 Evaluation

The evaluation function $\text{Eval}(K, d)$ can be defined recursively as follows

- If $d$ is the case passed to $K$ and $K$ is a primary rule $R$, which is of the form $A \longleftarrow L_1, L_2, \ldots, L_m$ then $Eval(K, d) = A$.
- If $K = \text{Sib}(K_1, K_2)$, let $o_1 = \text{Eval}(K_1), o_2 = \text{Eval}(K_2)$
  - if $o_1$ is not NO_OUTPUT and $o_1$ is not in $d$ then $\text{Eval}(K, d) = o_1$,
  - otherwise $\text{Eval}(K, d) = o_2$ (note that $o_2$ can also be NO_OUTPUT).

– If $K = \text{Cor}(K_1, K_2)$, let $o_1 = \text{Eval}(K_1), o_2 = \text{Eval}(K_2)$
  - if $o_1$ is NO_OUTPUT then $\text{Eval}(K, d) = \text{NO_OUTPUT}$, otherwise
  - if $o_1$ is not NO_OUTPUT and $o_2$ is NO_OUTPUT then $\text{Eval}(K, d) = o_1$,
  - otherwise $\text{Eval}(K, d) = o_2$.

From the definition, we can see that the evaluation function returns as soon as there is an output that is to be added to the case. The returned output is the conclusion of a knowledge base where none of its corrections applies to the current input. The repeated evaluation function $\text{RepeatedEval}(K, d)$ can be defined as applying Eval to the data until the output does not change. The following algorithm will show $\text{RepeatedEval}(K, d)$ is computed.

```
o := {}
do
    d := d ∪ o
    o := Eval(K, d)
    if o = Request then o := getRequestedInformation fi
    if o = d then
            RepeatedEval(K, d) = o;
            exit
    fi
od
```

The operation of $Request$ can be seen from the algorithm. As the external agent does not have the same control mechanism as RDR agent, we simply send the current data and assign the result to the output.

## 4 Knowledge Acquisition

The fundamental strategy for knowledge acquisition is to add knowledge when and if a case is handled incorrectly. This means that knowledge is added for real cases in real circumstances. Secondly since the cost of knowledge acquisition is effectively constant with knowledge base size, knowledge can be added while the system is in actual use and becomes a small but interesting extension to normal work or activity flow.

Of particular importance: it can be noted that since no information is removed from the blackboard there is an implicit assumption that solutions to all problems can be assembled linearly. That is, there is no need for any backtracking; information initially added does not need to be removed. This seems to be a plausible assumption in that although a human may use a propose-and-revise or similar approach to developing a solution, they can provide a linear sequence of justification when they are explaining how they reach a conclusion . The broad knowledge acquisition strategies outlined in the introduction then apply as follows.

In the following knowledge acquisition, we consider the special function $Request$ and the special symbol NO_OUTPUT to be the same as the other conclusion objects when constructing the knowledge base.

### 4.1 Global Refinement

The simplest case is to add an extra KBS or to add a replacement KBS which applies to the entire previous KBS. That is, no matter how complex the previous output, it will be replaced by other output in some circumstances. The cornerstone cases are then checked to see if their output is changed and if so whether this is appropriate. If any cornerstone cases have had their output changed inappropriately the application of the added KBS is made more restrictive. (If the extra KBS is a single rule, the user adds further conditions which apply to the case in hand, but not to the cornerstone case.) If the user is to replace some component from the output, we have the following operator

$$K' = (\mathrm{Cor}(K, R), D)$$

where $K$ is the original knowledge base, $R$ is the refinement knowledge base and $D$ is the cornerstone case set associated with the new knowledge base $K'$. $D$ is the union of the the cornerstone cases from $K$ and $R$. Similarly, if the user chooses to add further components to the output, we have

$$K' = (\mathrm{Sib}(K, R), D).$$

### 4.2 Local Refinement

The user looks at the sequence of output and decides that one of the outputs in the sequence have to be replaced. A KBS is added to do this. The case then has to be rerun as some of the later outputs may be missing or wrong, and perhaps a series of changes need to be made to the case to get all the components right. This generalises to the idea that when the output for a case is being fixed one corrects whatever outputs need correcting in the sequence in which they are provided. If the corrections cause further errors in the sequence, these too are corrected in sequence. The following algorithm shows how this is done. Suppose the input data $d = \{d_1, d_2, \ldots, d_m\}$, we have the output $\mathrm{RepeatedEval}(K, d) = \{o_1, o_2, \ldots, o_n\}$:

1. the expert identifies the first wrong output component $o_i$
2. the expert identifies the component knowledge base $K$ which 7 $o_i$ (from the list provided by the system)
3. do a global refinement to $K$
4. rerun the input data with respect to the new knowledge base
5. if output is correct, stop the process, otherwise, go to step 1.

In each step, the newly added component will only affect the performance of the local knowledge base.

## 5 Conclusion

Previous work on RDR has been explicit about attaching a rule to another rule using a correction relation and been explicit about the use of cases. However, it has been less explicit about the sequence relationship except for [CR00]. Because of the lack of focus on the sequence relationship some RDR systems have included other control mechanisms such as conflict resolution strategies. In this paper we have reduced all inference control to the two relations of sequence and correction and elaborated these relations.

We have further proposed that these relations can be used between knowledge-based systems or other programs as well as between rules. Again cases are used to initiate and guide knowledge acquisition. We suggest that this generalisation should enable extremely powerful RDR systems to be developed.

We have also suggested that perhaps all knowledge acquisition can be reduced to correcting or adding to knowledge using these two relations and that perhaps the success of RDR comes from explicitly ensuring that knowledge is added to existing knowledge using one or other of these relations, rather than allowing essentially uncontrolled editing. We are not able to prove such a conjecture at this stage, but would suggest that an RDR approach does seem to facilitate easier knowledge-based system development than free editing.

Our hope is that the generalisation outlined here will lead to far more sophisticated systems being assembled from more complex components, but that this is incrementally with similar ease to rule-based RDR.

## Acknowledgement

The support of the Smart Internet Technology Co-operative Research Centre is gratefully acknowledged.

# Knowledge Acquisition Module for Conversational Agents

Pauline Mak[1], Byeong-Ho Kang[1], Claude Sammut[2], Waleed Kadous[2]

University of Tasmania, Tasmania, Australia,
{Pauline.Mak,Byeong.Kang}@utas.edu.au
[2] University of New South Wales, Sydney, Australia,
{claude,waleed}@cse.unsw.edu.aubhkang@utas.edu.au

**Abstract.** The focus of traditional conversational agents is placed on natural language processing and understanding the needs of the user. These agents are typically implemented for specific domains such that domain knowledge and conversations are built manually. Domain knowledge of these agents are encoded as conversational content which causes a poor separation between the two distinct types of knowledge. This problem is magnified by the lack of knowledge acquisition tools and, as a result, agents find it difficult to adapt to different domains and to update existing knowledge bases.

The framework proposed in this paper aims to rectify this problem by building a module to handle knowledge acquisition. The module acquires knowledge through a case-based methodology called Ripple Down Rules (RDR); a technique that has been employed successfully across a host of expert system.

## 1 Background

Intelligent conversational agents have been the ultimate goal in the realm of human computer interactions. This has motivated a large number of agents of varying complexity to surface in recent years [7]. One of the earliest, and perhaps most well known, example of such agents is Eliza [12]. Eliza engages a conversation through simple scripts that prompt the speaker with questions derived from the latest utterance. While this approach allows one to converse with Eliza, the resulting dialogue becomes a series of repetitious questions. For an agent to be of any use, it should be able to provide some kind of service, such as giving an expert advice in a field of specialisation. This is the type of agent that this paper will focus on and in particular, the customisation of such agents.

Contemporary conversational agents look to the fields of linguistics and memory for more realistic models of agents [4]. From this, many different types of parser and speech generation systems are built. For instance, the Phillips train system [1] is a conversational agent that is connected to a public telephone network which can provide train timetabling information through conversation. This agent essentially treats the user input as a query into the timetable database. A similar system by Zelle and Mooney [13] can learn to translate sentences into queries for obtaining geographical information about the United States. Both of these system have a predefined database that stores the knowledge of the agent. While this is good for systems where most facts are atomic, this is not suitable for storing expert knowledge. This problem solving knowledge is based on heuristics which cannot be suitably expressed in a flat database.

There are even more ambitious agents that do not only incorporate speech, but also body language, such as emotions and spatial awareness, as exemplified by Rea [3] and Max [6]. These agents are able to convey subtleties such as facial expressions which are important non-verbal cues for expressing the context in which spoken information is given.

Recent developments in the Cyc project allows expert to update the knowledge base through a web-based application [2]. The system has tools such as dictionaries and analogies to introduce new facts into the knowledge base. The system is also sufficiently intelligent to generate related facts which are implied by the user input. While this system provides one of the most innovative method of knowledge acquisition, it only applies to adding facts in a common-sense knowledge base. While this can provide a solid back-end for general knowledge, the architecture is not suitable for retaining specific domain knowledge.

In summary, contemporary conversational agents are focused on the problem of providing a natural and adaptive way to interact with the end user. Little has been done on simplifying the process of updating the knowledge of the agents. This paper proposes a knowledge acquisition module that will allow conversational agents to learn domain specific knowledge, thus, creating a flexible framework for customising specialised agents. The subsequent sections will demonstrate how this structure works.

## 2 Framework

The module builds on an existing conversational agent and knowledge acquisition methodology. A special data structure known as frame is fundamental to the development of the module. The following sections shall introduce these elements briefly.

### 2.1 Conversational Agent

The module for knowledge acquisition is built on existing conversational agent technologies. First and foremost, the conversational agent that is augmented with the module is Probot. It is an extension to the programming language iProlog, an enhanced version of standard Prolog [10].

Probot is a turn-based conversational agent, where the user and the agent alternate speaking. There is no natural language processing in Probot, instead user input is matched against patterns defined in files known as scripts. Context of these patterns are separated through the use of topics; this allows the same patterns with varying meanings to coexist within the agent without ambiguity [11]. It is possible to define multiple topics for the agent, however, the agent can only handle one topic at a time.

### 2.2 Frames

Frames are structures (first proposed by Marvin Minsky [8]) used for describing a typical situation, such as entering a room. Each aspect of the situation is represented by a slot. Minsky believes that one understands dialogue through finding frames that describe the content most appropriately. He also believes that there is a hierarchy of frames, where each subsequent level in the chain of inheritance contains increasing amount of details about the topic of interest, much like subclasses in the object oriented paradigm. A frame is also able to maintain a list of questions that can be asked when certain values are needed.

There are two types of knowledge that operate in a conversational agent. One is conversation knowledge and the other, domain knowledge. These types of knowledge are modular, and use a single data structure known as a frames though these are maintained through separate procedures.

Frames can also react to events through the use of daemons. These are similar to event handlers in languages such as Java. iProlog has defined a number of daemons that are called when a variety of events, such as the instantiation of a frame, occurs.

Frames are ideal data structures for the knowledge acquisition module as it supports flexible data storage as well as containing elements of conversation in a modular manner.

### 2.3 Knowledge Acquisition

The knowledge acquisition methodology adopted for this project is the Ripple Down Rule (RDR) approach. It was first devised by Paul Compton [5] as a consistent approach to updating large knowledge bases [9].

The core assumption of this methodology is that experts make judgments on a case by some implicit reasoning. Different conclusions are given as a result of variations in these cases. Experts are better at identifying the variations in cases that distinguishes one from another.

Knowledge is stored within a rule tree, where each node is a conclusion as well as a set of conditions that it must satisfy. As a new case is given to the knowledge base, it will be tested against rules within the tree. Conclusion is given by the last node that the case completely satisfies. If the expert disagrees with the conclusion, new rules can be written to correct the knowledge base. New rules branches from the last correct node, that is, the node that had the incorrect conclusion. As the rest of the tree is not affect by this change, RDR gives a consistent approach to incremental learning.

Updates are only required when the expert disagrees with the conclusion the knowledge base has drawn. In practice, each update requires the specification of a case, a set of rules that identifies the case and a new conclusion that is different from the conclusion generated by the knowledge base. These elements will form parts of the conversation that the agent must undertake for extracting knowledge from the expert.

## 3 Module

The module is consisted of two knowledge acquisition components; domain and conversational knowledge. Both uses the frame hierarchy to store information required by the knowledge acquisition process and the details of which will be examined in detailed in the following sections.

### 3.1 Knowledge Acquisition Frame Hierarchy

The knowledge acquisition frame (KA frame) is used for storing information required for a single update. There are a number of fixed elements that a KA frame as dictated by RDR and is discussed previously in Section 2.3. KA frame should contain three slots, where each item is represented by a frame and are essentially subframes. As a result, a KA frame will have the following subframes: case frame, conclusion frame and condition frame, as illustrated in Figure 1. Shaded boxes indicates that the frame is domain dependent.

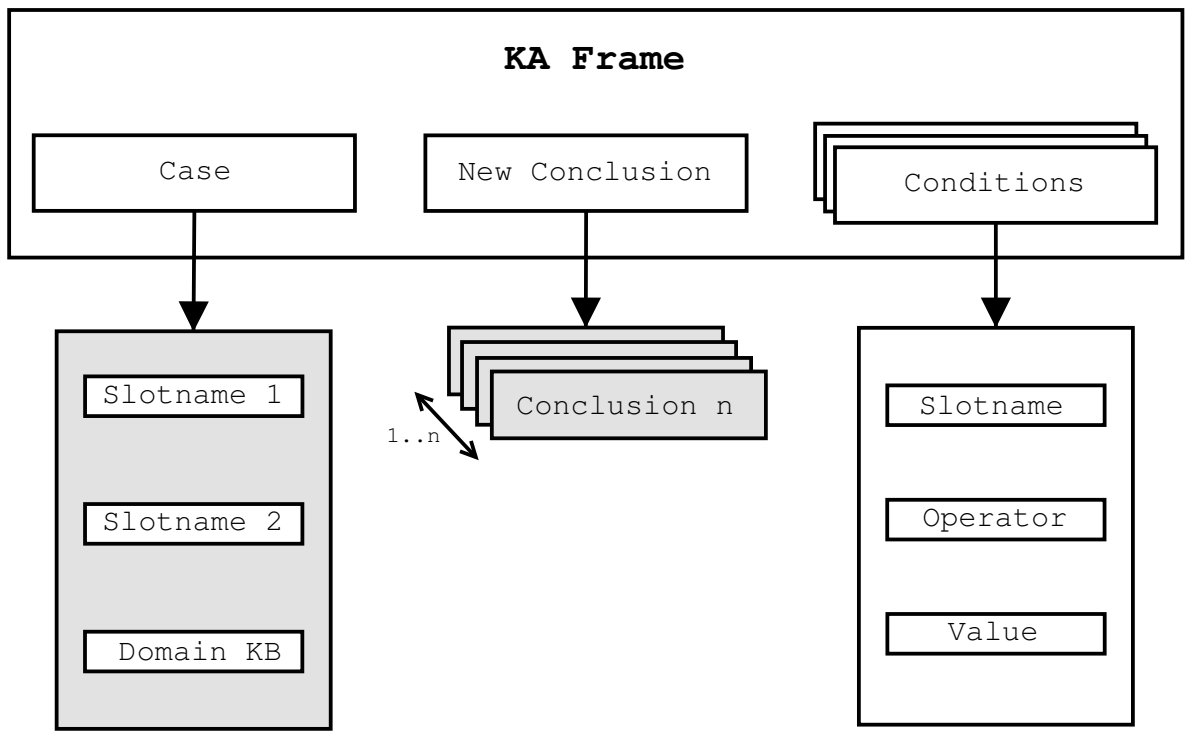

**Fig. 1.** KA Frame

The case frame describes features of a situation from which an expert can draw a conclusion, such as symptoms of a patient. This frame is domain dependent, and must be remodeled for every knowledge base the agent is introduced to. This task is the responsibility of the developer that is adapting the agent to the new domain, of course, with sufficient consultation with the experts.

The case frame also houses the domain knowledge base. This is a logical choice as the knowledge base must have access to the features of a case in order to draw conclusions from it. Due to daemon inheritance, the knowledge base will only need to be defined in the generic case frame and all instance frames will have access to it. Therefore, any new cases created will also use the most recent version of the knowledge base. This inheritance structure complements RDR;

consistency allows all existing and future cases to share the same knowledge base without the need to restructure.

Conclusions are possible classifications that an expert may give to a case. This set of values can be stored as flat values, however in some cases, it may be represented more adequately by frames. For example, the agent may learn how to select cars for a given customer. Conclusions in this case would be better represented by a car frame rather than just the model of the car. This frame is domain and implementation independent, and must be recreated for each new knowledge base.

Conditions are responsible for holding rules related information where rules specify distinguishing properties of a case. These properties are predicates that will either evaluate to true or false. Due to the frame structure, these predicates are required to know which slot in the case frame is to be tested. Therefore, a condition frame needs to have three slots: slot name, operator and comparison value. In the simplest case, a slot will be verify whether it holds a specific value. However, there are other data types that requires more than just equality, such as numeric and date comparisons. It is possible to create new definitions of comparison frames if these new frames contain the three slot (slot name, operator and comparison value.)

It is possible for an expert to define multiple rules for the same case. This is made possible by allowing the KA frame to take a list of condition frames. When added to the knowledge base, the list will be coalesced with the boolean function AND.

Once all the data is collected, knowledge can be updated in the knowledge base by calling a predefined function *add_ rdr*. This will add the new rule at the appropriate location in the RDR tree.

This hierarchy covers the basics for knowledge acquisition, however, consideration should be given to understanding expert terminology for specifying these rules and cases. The next section shall explain in detail how these terminologies can be obtained by the agent and how these two hierarchies cooperate.

### 3.2 Conversation Acquisition Frame Hierarchy

The conversation acquisition frame hierarchy allows an agent to translate what an expert has said into an internal representation. Previously in Probot, users had to create scripts for recognising patterns in speech. This was a simple yet functional approach. Unfortunately, each slot is interpreted differently and requires a new set of patterns. As frames are populated with more slots, scripts have to be written for each new slot; resulting in a proliferation of scripts. The hierarchy proposed here is a simple solution for solving this problem by merging scripts and internal data representations into the frame structure.

Knowledge acquisition in this module uses RDR for storing patterns that an agent may need to recognise and it is similar to the hierarchy described in the previous section. However, this is restricted to conversational knowledge which gives greater control over how the knowledge is stored and manipulated and as a result more fixed elements can be defined. As specified by the RDR paradigm, the conversational knowledge acquisition hierarchy should contain a case, rule and conclusion. Any utterance from the user is treated as a case. Rules are patterns that can be matched over the case. Finally conclusions are translated values for slots in the frame.

Any slot that requires user speech input needs to have the infrastructure for accessing utterances from the user as well as a location to store rules that are applied to the input. An additional slot is added to the frame structure, *utterance*, which stores user input and is updated with every utterance. Presently, the *utterance* slots for every frame throughout the hierarchy will have the same sentence to interpret. However, it is possible to include some text manipulation as the input is interpreted by the framework and will be an area for future exploration. The knowledge base is stored in another fixed slot, *interpret* in which, the rules for interpretation are stored here.

The infrastructure alone is not enough for specifying the rules; there needs to be a way to qualify the patterns. A new function, *u_match*, is defined in iProlog that will allow the following type of patterns to be matched, based on a list of given words, *a*, to match against:

- *all*: the utterance must have every word in the list *a*.
- *any*: the utterance must contain at least one word that is in the list *a*.
- *has_phrase*: the utterance must contain the phrase *a*.

This is not intended to be a comprehensive list of operation that can be performed on speech text, though they are acceptable for most circumstances, particularly in keyword matching. Further operations can be added to this function by adding new definitions to *u_match* in iProlog.

The acquisition of these patterns is done manually, by which, the user has to explicitly initiate the process. However, logically it should also be possible for the process to be instantiated when the system finds that none of its frames can interpret the user input.

Part of the vision of creating this module is to allow frames to be reusable. As each frame is responsible for interpreting data for itself, it is possible to reuse the same conversational knowledge base. For example, gender may be a subframe of another frame, such as person. The agent then learns some rules for defining the gender of a person. The agent may need to be adapted to other domains at some point later, which requires information on interpreting the gender, say, for animals. There are commonalities in the language used between the domains and it is here that modularity of the proposed hierarchy is key. Rules learned from the previous domain can be applied to the new domain. Of course, new rules have to be created for domain specific terminologies, in addition to incorrect conclusions as frames are used in different contexts. RDR excels in incremental knowledge acquisition, which reinforces its necessity in this module.

## 4 Dialogue Control

The goal of the dialogue control component is to make expert-agent conversations natural. Most customer inquiry phone lines require users to say or enter specific answers when promoted with a fixed set of questions, which leads to a menu driven dialogue rather than a conversation.

It is important to first examine what kind of interaction the agent must engage in for knowledge acquisition. It is essentially a task for filling in data, which, when completed, the agent is able to update the knowledge base with. As described in previous sections, the agent needs a fixed number of elements. However, it is not necessary for the users to give all of the information in a fixed order. Probot uses scripts to control the direction and the content of a conversation. This is not very practical if flexibility is a requirement, as it is not possible to anticipate all combinations of which order a user may specify facts. As a result, previous implementations of the knowledge acquisition module follows the menu driven dialogue model.

A more adaptive approach is to allow users to specify slot values in no fixed order. It is imperative for the old script system to change. As the conversation is centered around extracting information from the user to fill in the frame, a frame is now bounded to a script. The purpose of a script is to direct the overall conversation, such as, how the agent should react once all the necessary information is given, whereas frames now handle the translation of dialogue to some internal knowledge representation. The top-most frame of the hierarchy will be associated with the script. Since some information may be stored within subframes, it is necessary to pass information down the structure such that subframes can also interpret the user input for values. In order to do so, the frame must know which slots are subframes. This leads to the inclusion of an additional slot called *subframes* where the slot name and the subframe type is paired. In the future, slots that uses frame will automatically tagged as a subframe and will eliminate the need for the extra slot. Scripts can then use this information to create subtopics. Subtopics link a subframe to the top-most frame, and thus, provide a channel of communication for passing user input. Every input is filtered through the frame hierarchy. As each frame is able to process user input, slots can be filled in any order.

Scripts must be able to verify that an entire frame has been filled successfully. There may be elements in frames that are optional and it is therefore necessary to specify which slot is a requisite. A new daemon, *required*, is introduced which uses any iProlog predicate to check whether a value has fulfilled the requirements. This works in conjunction with a new function, *has_req*, which applies the predicate defined in *required* to determine whether a frame is complete. When applied to subframes, this will allow the entire hierarchy to be checked. In terms of knowledge acquisition, this will guarantee that the user has provided enough information for the knowledge base to be updated. This forms the last element needed for flexible conversation.

## 5 Conclusion

Current conversational agents are primarily focused on realistic simulation of human interactions, such as the use of body language and behaviour models.

However, development into agent customisation has been negligible. This paper has demonstrated a framework that is a structured method for agents to learn domain and conversational knowledge.

It is made possible through the separation of conversational and domain knowledge to provide a consistent approach for customising an agent. Complementing this architecture is the use of RDR. The methodology supports incremental learning in a consistent manner. This is significant as the system is aimed to be adaptive and can gain knowledge while in use.

A crucial part of the module is the use of frames. This structure encapsulates data in a manner that allows knowledge to be reusable. The cost to adapt an agent to a similar domain will be reduced. Frames also contribute to flexible conversation flow, which would otherwise become a menu driven dialogue.

In summary, the module detailed in this paper is a novel approach to solving the problem of customising conversational agents. It is capable of learning conversational and domain knowledge in a flexible and reusable manner.

## 6 Future Work

This framework has all the elements required for knowledge acquisition for both domain and conversational knowledge, though, the work thus far is of a preliminary nature.

Gauging the success of a conversational agent is a difficult manner, as there are no standard metrics for measuring the performance. As RDR has already proven to be a successful methodology for knowledge acquisition, there is no need to test whether the module can acquire knowledge. Instead, further work needs to be done on testing how the entire framework cooperates when adapting to a specific domain. Aside from testing on one domain, there is also a need to adapt the agent to multiple domains - this way, the re-usability of the framework can be revealed.

## Acknowledgments

This research has been funded by the Smart Internet Technology CRC and the authors would like to express thanks for the support received throughout the production of this paper. Also, thanks to Adam Berry for proof reading this paper.

# Personalized Web Document Classification using MCRDR

Sung Sik Park, Yang Sok Kim, and Byeong Ho Kang

School of Computing, University of Tasmania
Hobart, Tasmania, 7001, Australia

{sspark,yangsokk,bhkang}@utas.edu.au

**Abstract.** This paper focuses on real world Web document classification problem. Real world Web documents classification has different problems compare to experimental based classification. Web documents have been continually increased and their themes also have been continually changed. Furthermore, domain users' knowledge is not fixed apart from classification environments. They learn from classification experience, broaden their knowledge, and tend to reclassify pre-classified Web documents according to newly obtained knowledge to fit various contexts. To handle these kinds of problems, we use Multiple Classification Ripple-Down Rules (MCRDR) knowledge acquisition method. The MCRDR based document classification enables domain users to elicit their domain knowledge incrementally and revise their knowledge base (KB), and consequently reclassify pre-classified documents according to context changes. Our experiment results show MCRDR document classifier performs these tasks successfully in the real world.

## 1 Introduction

The size of available documents to be handled has grown rapidly since the Internet was introduced. For example, Pierre [1] estimates the number of pages available on the Web is around 1 billion with almost another 1.5 million added per day and some Internet search service companies reported that they cover around 3 billion pages [2]. Many Web document management systems have been developed because Web documents are now considered as one of the major knowledge resources.

Before computer technology was introduced, people mainly relied on manual classification such as library catalogue systems. In the early stages of the computerized classification development, computer engineers moved this catalogue system into the computer systems. However, as the size of available Web documents grows rapidly and people have to handle them within limited time, automated classification becomes more important.

Machine learning (ML) based classifiers have been widely used for automatic document classification and there are various approaches such as clustering, support vector machine, probabilistic classifier, decision tree classifier, decision rule classifier, and so on[3]. But they have some problems when they are applied to real world applications because they capture only a certain aspect of the content and tend to learn in a way that items similar to the already seen items (training data) are recommended (predefined categories) [4]. However, it is difficult to collect well defined training data sets because Web documents (e.g., news articles, academic publications, and bulletin board messages) are continually created by distributed world-wide users and the number of document categories also continually increases. To manage this problem, the document classifiers should support incremental knowledge acquisition without training data. Though some ML techniques such as clustering techniques [5-7] are suggested as solutions for incremental classification, they do not sufficiently support personalized knowledge acquisition (KA). Document classifiers in the real world should support personalized classification because classification itself is a subjective activity [1]. To be successful personalized document classifiers, they should allow users to manage classification knowledge (e.g., create, modify, delete classification rules) based on their decision. But it is very difficult when users use ML classifiers because understanding their compiled knowledge is very difficult and their knowledge is so strongly coupled with the knowledge of training data sets that it is not easily changed without deliberate changing them.

Rule-based approach is a more favorable solution for the incremental and personalized classification task because the classification rules in knowledge base (KB) can be personalized, understood, and managed by users very easily. But rule-based systems are rarely used to construct an automatic text categorization classifiers since the '90s because of the knowledge acquisition (KA) bottleneck problem [3, 8]. We used Multiple Classification Ripple-Down Rules (MCRDR), an incremental KA methodology, because it suggests a way that overcomes the KA problem and enables us to use the benefits of rule-based approach. A more detail explanation will be suggested in section 2.

Our research focuses on the personalized Web document classifier that is implemented with the MCRDR method. In section 2, we will explain causes of the KA problem and how MCRDR can solve that problem. In section 3, we will explain how our system implemented in accordance with MCRDR method. In section 4, we will show empirical evaluation, which is performed three different ways. In section 5, we will conclude our research and suggest further works

## 2 Knowledge Acquisition Problems and MCRDR

KA problems are caused by cognitive, linguistic and knowledge representational barriers [8]. Therefore, the promising solution for the KA must suggest the methodology and KA tools that overcome these problems.

**Cognitive Barrier.** Because knowledge is unorganized and often hidden by compiled or ***tacit*** knowledge and it is highly interrelated and is retrieved based on the situation or some other external trigger, knowledge acquisition is discovery process. Therefore, knowledge often requires correction and refinement - the further knowledge acquisition delves into compiled knowledge and areas of judgment, the more important the correction process becomes [9]. From the GARVAN-ES1 experience, Compton et al [10] provide an example of an individual rule that has increased four fold in size during maintenance and there are many examples of rules splitting into three or four different rules as the systems' knowledge was refined. Compton and Jensen[11] also proposed that knowledge is always given in context and so can only be relied on to be true in that context. MCRDR focuses on ensuring incremental addition of validated knowledge as mistakes are discovered in the multiple independent classification problems [12, 13].

**Linguistic Barrier.** Communication difficulties between knowledge engineers and domain experts are also one of the main deterrents of knowledge acquisition. Traditionally, knowledge is said to flow from the domain expert to the knowledge engineer to the computer and the performance of knowledge base depends on the effectiveness of the knowledge engineer as an intermediary [8]. During the maintenance phase, knowledge acquisition becomes more difficult not only because the knowledge base is becoming more complex, but because the experts and knowledge engineers are no longer closely familiar with the knowledge communicated during the prototype phase [11]. Domain knowledge usually differs from the experts and contexts. Shaw[14] illustrates that experts have different knowledge structures concerning the same domain and Compton and Jansen[11] show that even the knowledge provided by a single expert changes as the context in which this knowledge is required changes. For these reason, MCRDR shift the development emphasis to maintenance by blurring the distinction between initial development and maintenance and knowledge acquisition is performed by domain experts without helping the knowledge engineer[1] [13].

**Knowledge Representation Barrier.** The form in which knowledge is available from people is different from the form in which knowledge is represented in knowledge systems. The difference between them, called representation mismatch, is central to the problem of KA. In order to automate KA, one must provide a method for overcoming representation mismatch [15]. KA research has been aimed to replace the knowledge engineer with a program that assists in the direct "transfer of expertise" from experts to knowledge bases [16]. Mediating representation facilitate communication between domain expert and knowledge engineer. Intermediate representations provide an integrating structure for the various mediating representations and can form a bridge to the knowledge base[17]. We used ***folder structure user interface***, which is largely used for manual document classification in traditional document management application, as ***mediating representation method*** and ***difference lists*** and ***cornerstone cases*** as ***intermediating representation***. Folder manipulations are interrelated with the MCRDR KA activities in our system.

[1] This does not mean MCRDR needs no help from knowledge engineer or programmer. Rather, they are required for the initial data modeling (Kang, B. H., Compton, P., Preston, P., 1996).

## 3 Real World Web Document Classifier with MCRDR

The system, a text classification system for Web documents, is a component of the Personalized Web Information Management System (PWIMS) System [18]and is implemented with C++ program language and the MCRDR methodology. It is used to construct both Web document classification and personalized Web portal.

### 3.1 Folder Structures as a Mediating Representation

The choice of representation can have an enormous impact on human problem-solving performance [19, 20]. The term mediating representation is used to convey the sense of coming to understand through the representation and it should be optimized for human understanding rather than for machine efficiency. It is suggested to improve the KA process by developing and improving representational devices available to the expert and knowledge engineer. Therefore, it can provide a medium for experts to model their valuable knowledge in terms of an explicit external form [17]. We use traditional folder structures as a mediating representation because users can easily build a conceptual domain model for the document classification by using folder manipulation. Our approach differs from the traditional knowledge engineering approach because we assume there is no mediate person (knowledge engineer). Rather the domain experts or users directly accumulate their knowledge by using KA tools [12].

### 3.2 Inference with MCRDR document Classifier

A classification recommendation (conclusion) is provided by the last rule satisfied in a pathway. All children of satisfied parent rule are evaluated, allowing for multiple conclusions. The conclusion of the parent rule is only given if none of the children are satisfied [13, 21, 22]. For example, the current document has a set of keywords with {a, b, c, d, e, f, g}.

1. *The system evaluates all the rules in the first level of the tree for the given WL (rules 1, 2, 3 and 5 in Fig. 1.). Then, it evaluates the rules at the next level which are refinements of the rule satisfied at the top level and so on.*
2. *The process stops when there are no more children to evaluate or when none of these rules can be satisfied by the WL in hand. In this instance, there exist 4 rule paths and 3 classifications (classes 2, 5, and 6).*
3. *The system classifies into the storage folder structures (SFS)' relevant nodes (F_2, F_5, and F_6) according to the inference results.*
4. *When the expert finds the classification mistakes or wants to create the new classifications, he updates the classification knowledge via the knowledge acquisition interface.*

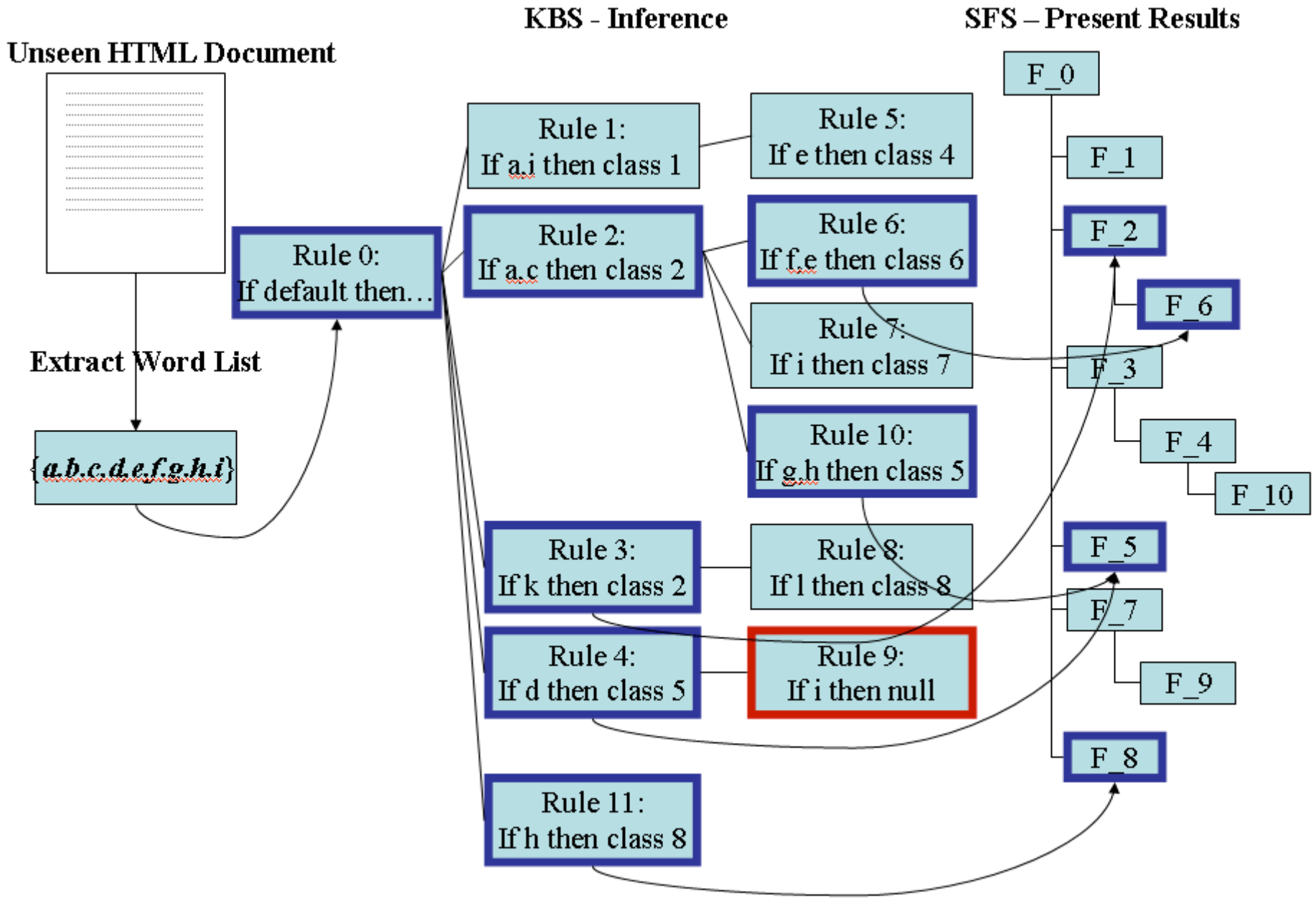

Fig. 1. Inference for the Web document classification

### 3.3 Knowledge Acquisition and Intermediate Representation

KA and inference are inextricably linked in the MCRDR method, so some KA steps depend on the inference and vice versa [13, 23, 24]. The KA process consists of the following sub-tasks: 1) initiating KA process, 2) deciding KA method, and 3) validating new rules.

**Initiating KA Process.** KA process is initialized by users when they dissatisfy the system's inference result. Kelly [25] suggested that "every construct has a specific range of convenience, which compromise all things to which the user would find its application useful." The range of convenience of each construct defines its extension in terms of a single aspect of a limited domain of events [17]. The users' decision for initializing new KA processes depends on the range of convenience. There are two different kinds of KA initialization: the KA process begins when the system recommends incorrect class or no class [23] and users initiate it (human initiated KA) and users move or copy some pre-classified documents to another folder (system initiated KA).

**Deciding KA Methods.** There are three kinds of KA methods: refinement KA, stopping KA, and ground-breaking KA.

- ***Refinement KA***: If the user thinks that the current document should be classified into the sub folder (may not exist) of the recommend folder, the user selects (or creates and selects) the sub folder of the folder recommended by the system. The new rule should be added under the current classification rule as the child rule, because it refines current rule. For example, if a certain document that contains keyword "a" and "c", it will be classified into folder F_2 in Fig. 1. But users may want to classify this document to folder F_6 (this folder may not exist when this document classified) because it contains keyword "f" and "e". In this case, the new refinement rule is created under the rule 2 and its conclusion is class 6.

- ***Stopping KA***: If the current inference result is obviously incorrect and the users do not want to classify incoming documents into this folder, he/she makes stopping rules with certain condition keyword/keywords. The new stopping rule won't have any recommendation for a folder. For example, if a certain document that contain keyword "d", it will be classified folder F_5 in Fig. 1. But users may not want to classify this document to folder F_2 because it contains keyword "i". In this case, the new rule with condition "i" is added under the rule 4 and its conclusion is "null".
- ***Ground-breaking KA***: For example, if a certain document that contains keyword "k", it will be classified folder F_2 in Fig. 1. But domain experts may not want to classify this document to folder F_2 because it contains keyword "h" and they want to make new classification. In this case new rule is added under the root node (e.g. rule 11).

The KA process is initiated by system when users copy or move pre-classified documents to other folder/folders. Its KA method depends on the action types. If the action is moving, the stopping KA and ground breaking KA are needed. For example, if users want to move some documents in F_6 to F_1, they must select keywords that make stopping rules and ground breaking rules such as "***t***". In this example, new rule conditions will be "a" and "c" and "f" and "e" and "***t***". If the action is copying, only the ground breaking rule is automatically created by the system. Its condition is the same as the original rule but it has a different conclusion.

**Validating with Cornerstone Case and Difference List.** Bain [26] proposed that the primary attributes of intellect are consciousness of difference, consciousness of agreement, and retentiveness and every properly intellectual function involves one or more of these attributes and nothing else. Kelly[25] stated "A person's construction system is composed of a finite number of dichotomous construct." Gaines and Shaw[27] suggested KA tools that are based on the notion that human intelligence should be used for identifying differences rather than trying to create definitions. In our system, the experts must make domain decisions about the differences and similarities between objects to validate new rule. Our system supports users with ***cornerstone case*** and ***difference list*** [12, 13, 21, 23, 24]. As shown in Fig. 1, an n-ary tree is used for knowledge base (internal schema). MCRDR uses a "***rules-with-exceptions***" knowledge representation scheme because the context in the MCRDR is defined as the sequence of rules that were evaluated leading to a wrong conclusion or no conclusion with existing knowledge base [13]. Though users can see the whole knowledge base (internal schema) in our system, it is not directly used for KA. Instead, MCRDR uses ***difference list*** and ***cornerstone case*** for intermediate representation. The documents are used for the rule creation are called "**cornerstone cases**" and saved with the rules. Each folder may have multiple rules and cornerstone cases. When users make refinement rule or stopping rule, all related rules must be validated but we do not want for users to make a rule that will be valid afterward. Rather we want to present the users with a list of conditions (called "***difference lists***") to choose from which will ensure a valid rule. The difference between the intersection of the cornerstone cases which can reach the rule and the new case cannot be used [12]. Cases which can be reclassified by the new rule appear in the system. The users may subsequently select more conditions from the different keywords lists to exclude these cases. Any case which is left in this list is supposed to classify the new folder by the new rule. A prior study shows that this guarantees low cost knowledge maintenance [13, 23].

## 4 Experiment

The goal of our research is to develop personalized Web document classifiers with MCRDR. The experiments are designed to the performance evaluation in the various classification situations. We consider three different cases: 1) document classification without domain change by single user, 2) document classification with domain changes by single user, and 3) document classification within single domain by two users.

**Data Sets.** We uses three different data sets: health information domain, IT information domain (English), and IT and finance domain (Korea), which are collected by our Web monitoring system for one month[18]. Table 1 represents the data sets that are used for our experiments.

Table 1. Inference for the Web document classification

| Data | Domain | Source | User | Articles |
|---|---|---|---|---|
| Data Set 1 | Health | BBC, CNN, Australian, IntelliHealth, ABC (US), WebMD, MedicalBreak-throughs | 1 | 1,738 |
| Data Set 2 | IT (English) | Australian, ZDNet, CNN, CNet, BBC, TechWEB, New Zealand Herald | 2 | 1,451 |
| Data Set 3 | IT/Finance (Korean) | JungAng, ChoSun, DongA, Financial News, HanKyeung, MaeKyeung, Digital Times, iNews24 | 1 | 1,246 |

**Results.** Classification effectiveness can be usually measured in terms of precision and recall. Generally two measures combined to measure the effectiveness. However, we only use precision measure because our system is a real world application and there is no pre-defined training data set. Fig. 2 shows the experiment's results. In each figure, horizontal axis represents the cases, left vertical axis represents the precision rates and right vertical axis represents the number of rules.

**Experiment 1.** This experiment is performed by a single user without domain changes in the health news domain. The user classified 1,738 articles with 348 rules. Though there are some fluctuations of the precision rate and rule numbers, there exist obvious trends: the precision rate gradually increases and the number of rules gradually decreases as the cases increase. Precision rate sharply increases from starting point to a certain precision level (around 90%) and is very stable after that point. This is caused by the fact that the domain knowledge continually change and as the user knows the domain, the more classification knowledge is needed.

**Experiment 2.** This experiment is performed by a single user in IT and Finance news domain (Korean). Totally, 1,246 articles are classified and 316 rules are created by the users. At first, user classifies IT articles from the business relationship view (e.g. customers, competitors and solution providers). New view point for the domain (technical view) is added when user classifies 550 cases and new domain (finance) added when user classifies 800 cases. When the view point changed, the precision rate went down from 90% to 60% but precision rate recovers around 80% by classifying a small additional amount of cases. When the new domain (financial news) is added to the current domain (IT news), the precision rate sharply decreases to 10% and the rule creation goes up 30 but a very small number of cases is needed to recover 80% precision. This result shows that our document classifier can work efficiently with domain changes.

**Experiment 3.** This experiment is performed by two users in the same IT news domain (English). In total, they classified 1,451 articles with 311 rules: User 1 classifies 1,066 articles with 228 rules and user 2 classifies 432 articles with 83 rules. The classification result is shown in Fig. 2 (c). When user 1 classified 500 articles, the precision of classifier reached around 90%. After that point, new rules are gradually created and the precision rate is slightly improved until user 1 classifies 1000. When a different user (user 2) starts to classify, the precision rate shapely down to 60% and many new rules are created. But small articles are needed to get a similar precision rate. This result means that our classifier can be adaptively applied when different users classify.

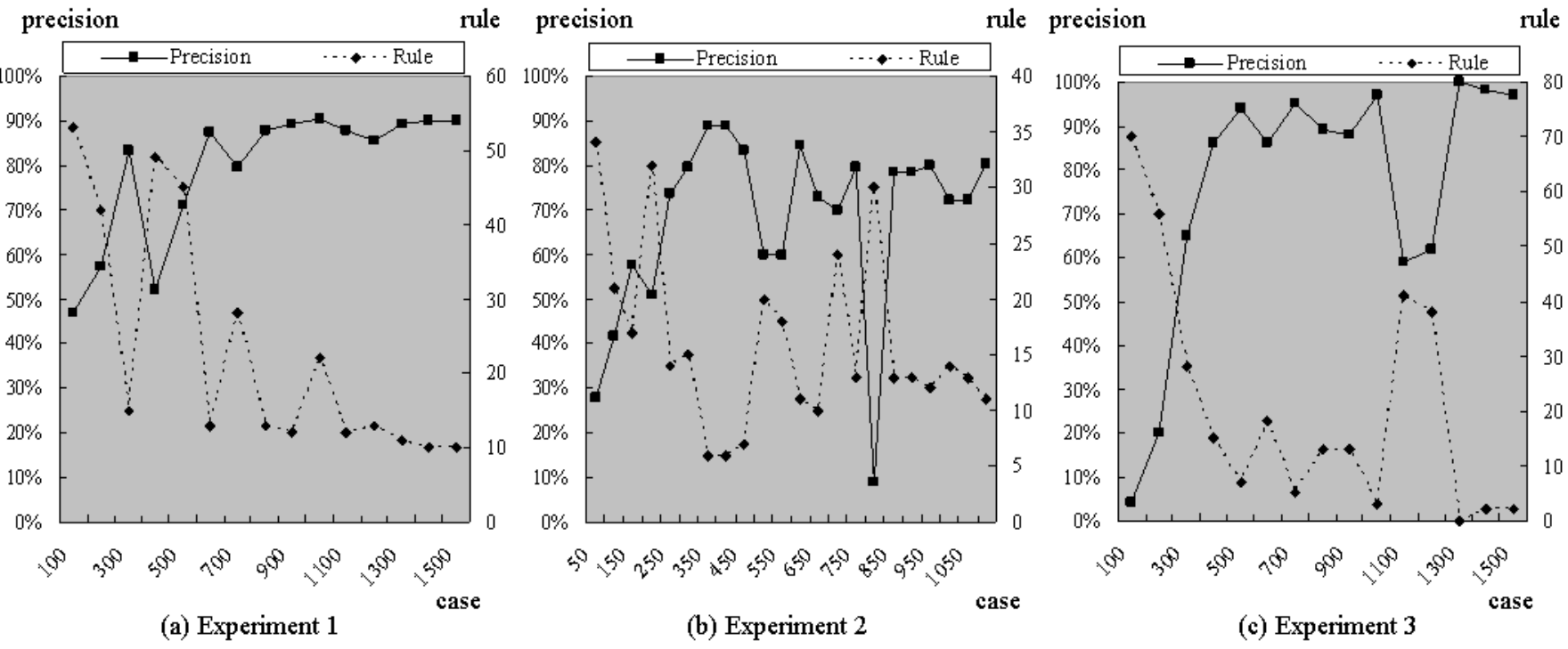

Fig. 2 Classification Results

## 5 Conclusion

We suggested the MCRDR based document classifier. MCRDR is an incremental KA method and is used to overcome the traditional KA problem. Our classifier used the traditional folder structures as a mediating representation. Users can construct their conceptual document classification structures by using an MCRDR based classifier. In our system, the KA and inference process is inextricably linked, so some KA steps depend on the inference and vice versa. The KA process begins when the classifier suggest no folder or incorrect folders or users activate some function in folders such as copying or moving some cases. There are three different KA methods – refinement KA, stopping KA, and ground-breaking KA. In the validation process, we used corner cases and difference list as an intermediate representation. Experiment results show that users can create their document classifier very easily with small cases and our system successfully supports incremental and robust document classification. An incremental KA based classification works well in a certain domain where the information continually increases and the creation of training set for machine learning is hard.

However, this attitude does not deny the machine learning research works. Rather we view our approach can be a collaborator of machine learning technique. Wada et al. suggest integration inductive learning with RDR [28], Suryanto and Compton suggest a reduced KA with decision tree [29]. Especially we view our approach can help construct a fine training data set with cost efficiency in the initial stage. Research for the combining incremental KA approach with machine learning techniques will be our further work.

# Achieving Rapid Knowledge Acquisition in a High-Volume Call Centre

Megan Vazey and Debbie Richards

Department of Computing
Division of Information and Communications Sciences
Macquarie University
Sydney, Australia
Email: megan@excelan.com.au, richards@ics.mq.edu.au

**Abstract**

Ripple Down Rules (RDR) has been applied to a number of domains. In this paper we consider a new application area that presents a number of new challenges. Our application is a high-volume call centre that provides a service / help desk function in a complex problem domain. We propose that the combined use of multiple classification ripple-down-rules (MCRDR) together with a web-enabled hyperlink-rich browser front-end will provide an effective tool to help call-centre knowledge workers cut through the potential information overload presented by both intra- and inter-nets; speed up the processes of knowledge acquisition and re-use; and assist with decision support and problem resolution.

We consider the implementation issues faced by corporations in their transition from a simple call / defect-tracking call centre model to a much enriched knowledge-centered model and we examine the role MCRDR can play in the call-centre context including workflow integration, accessibility, usability, and incentives. In order to improve the fit with our application area, we suggest a number of variations to the MCRDR theme. Our implementation and evaluation of these ideas is ongoing.

## 1. Introduction

The Internet revolution of the mid 1990s has bought an unprecedented level of global knowledge and opinion to homes and offices alike. On both the academic and commercial front, it has prompted an enormous amount of interest and investment in knowledge management. The simple hyperlink has been a stunning force for change in the way we now perceive and work with knowledge. It has also created countless opportunities to track, review and comprehend the paths that users take through knowledge mazes presented by inter- and intra-nets alike.

One corporate sphere with an enormous thirst for knowledge is the customer call-centre. The last decade has seen globally explosive growth in call-centres providing both Customer Relationship Management (CRM) and help-desk[1] functions: advising customers, answering queries, and resolving customer problems. In this domain, rapid access to appropriate, accurate and concise knowledge is paramount.

[1] also termed 'service-desk'

### 1.1 Call / Defect Tracking Software - Problem Ticketing

Historically, call-centers have placed call / defect tracking at the core of their query receipt and resolution process.

Typically, incoming calls are logged in a call / defect tracking database. Basic features of the incoming case are logged such as date, time, client name, and query summary. More specific details may also be included such as the name, model and / or version of any defective product (e.g. hardware or software) together with a query description.

Problem tickets may be machine generated e.g. with problem equipment ringing or emailing through the problem tickets. They may also be entered directly by customers, for example via a user-driven web interface. More traditionally they originate with front-line customer service personnel taking the first call.

The problem ticket passes through several states in which it moves between workers at various levels in the organization, for example from machine-generated, web-created, or front-line customer service personnel (state := *new*) to first or second tier customer service or technical support personnel (state := *assigned* then *opened* then *resolved*) then on to a team leader or even back to the customer (state := *closed*).

### 1.2 What Knowledge?

Somewhere between the *opened* state and the *resolved* state is where the real magic and art of problem solving by human experts is called upon. Here is where the customer service or technical support personnel go to their procedures manuals, training handouts, technical support home pages, knowledge databases, web search engines and inter / intra-nets to commence the often labor intensive process of finding out how to resolve the customer's query.

More experienced call-centre personnel create their own cheat-sheets, HTML link-rich home pages, and *uncovered* series to keep tabs on sources of knowledge likely to help them with their daily grind.

The problem, as we see it, is a perception of which knowledge is most relevant to the call-centre. And sometimes we can't see the wood for the trees…

In the call-centre context, one of the most important sources of knowledge is the knowledge of *how similar problems were solved in the past*. We believe that acquisition and re-use of this type of knowledge in the call-centre will deliver enormous benefit to customers and employees, and drastic bottom-line improvements for call-centers and their help/service desks.

Imagine, when a new problem ticket comes in, the customer service personnel is presented with a set of refinement queries enabling them to more specifically describe the type of problem being observed by the customer. Immediately that the new information is entered, the history of how *similar problems were solved in the past* is presented to the user – which internet links proved useful, and which knowledge-base references helped.

We believe that a simple extension to the problem-ticketing paradigm described above is all that is required.

### 1.3 An Expert System Approach

We believe expert system technology can be applied to record the decision making intelligence that call-centre and help / service desk personnel use to determine the resources (knowledge base, web, document or otherwise) that best assist with particular types of customer query.

We intend to augment the multiple-classification-ripple-down-rules (MCRDR) algorithm introduced by Kang, Compton and Preston (1995). MCRDR is a variation of the ripple-down-rules (RDR) algorithm developed by Compton and Jansen (1990) and described further in section 0.

We believe that the strengths of the MCRDR approach which includes easy knowledge acquisition, maintenance and validation performed directly by the domain expert, the customisability of the knowledge to suit the local environment, the use of cases to contextualise and assist knowledge acquisition are key reasons why MCRDR can offer a valuable solution in this area.

Closing the loop between the answers a user is searching for, and the quality of the answers retrieved is a vital step in training the system to excel in matching solutions to problems.

As time goes on, the cumulative effect of presenting more and more cases to our system is that the system gets trained and refined to achieve high levels of accuracy in matching solution resources to problem types. This will obviously be of huge benefit to the call center helpdesk - no more fumbling around with search engines, local web pages, or existing knowledge bases to find the relevant information. Experience with the MCRDR algorithm in other applications such as pathology (Edwards et al 1993) and the experiments of Kang (1996) suggest that such an expert system will grow rapidly in its level of matching accuracy as cases are added.

In our system, each MCRDR classification will be a set of one or more intra- or inter-net hyperlink references, where each reference points to web content that will assist with troubleshooting the current case.

MCRDR can be considered as a variant to the Case-Based Reasoning approach (discussed further in section **Error! Reference source not found.**). CBR is appropriate where there is no formalized knowledge in the domain or where it is difficult for the expert to express their expertise in the format of rules (Kang et. al, 1996). However, MCRDR is more than just a case-based reasoner. MCRDR is also a rule based approach in that it uses rules to index the cases, thus addressing a problem associated with CBR systems that often require manual indexing. The cases motivate and assist rule development and the rules provide structure for storage and retrieval of the cases as we will see in section 3.

### 1.4 Challenges for MCRDR

The call-center help-desk context under consideration has a number of properties that present new challenges for the MCRDR algorithm:

- the system must interact with a legacy ticketing system and legacy knowledge base
- the system needs to deal with numerous cases (in the order of 50 per day locally, and 300 per day globally)

- the volume of cases being dealt with means that the workflow must inherently deal with system maintenance i.e. system maintenance must be in-circuit
- initial problem descriptions are sparse – the case definition matures as the customer service personnel interacts with the customer and *works the case*.
- while most cases are resolved promptly, a number of cases are open for days or even weeks
- problem receipt and resolution is asynchronous since there is a time delay (up to a day) between when the system receives a problem case, and when a customer service representative can attend to it.
- archived cases and the conclusions registered to them need to be available for several years (perhaps 10 years for some cases) into the future
- old cases may be edited
- multiple users will use and update the system, but a limited subset of privileged users will approve their updates
- old conclusions may be edited
- the granularity of conclusions may vary widely and conclusions that are web links may expire

Further this project is concerned with solving various call center problems using MCRDR for a real organization which imposes further characteristics on the problem and constraints on the solution.

- There are very many attributes which will vary across cases and new attributes will frequently need to be added.
- The range of values possible for those attributes is also very large and the dependencies between these A-V pairs may be very strong. For example, we are dealing with troubleshooting across multiple systems, platforms, vendors, versions, etc.
- We don't have control over the cases, which are stored in the parent company's database.

In the next section we describe a number of previous MCRDR approaches to the Help Desk domain and how the scope of our project differs from these. In Section 3 we introduce the basic ideas behind MCRDR. Section 4 presents our solution, Interactive Recursive MCRDR (IR-MCRDR), describing the key issues and extensions that are needed to address the complex Call Centre environment. Our conclusions are given in the final section.

## 2. MCRDR-based Help Desk Approaches

MCRDR has been explored in the help desk environment by Kang, Yoshida, Motoda, Compton in 1996; Kim, Compton and Kang in 1999; and again by Kim in 2003. We note in the final subsection the key differences in the application we are developing.

### 2.1 A Help Desk System with Intelligent Interface

The prototype described by Kang, Yoshida, Motoda, Compton (1996) combined a keyword search with Case-Based Reasoning indexing techniques to provide a guided MCRDR interaction that was able to quickly steer users to appropriate help information on the internet. Their system considered updates by a single expert only.

As noted by Kang et. al. (1996) the MCRDR engine has two problems as an information retrieval engine. The first one is the number of conditions that are to be reviewed by the user. The second one is the number of interactions between the user and the system.

Their prototype attempted to minimise this problem by allowing users to apply a keyword search to effectively pre-filter the rule tree to only include those cases that satisfied the keyword search criteria. The user could then interact with a minimised MCRDR rule tree to select the relevant cases and update the knowledge accordingly.

The idea is compelling and may prove to be a useful adjunct for browsing the knowledge in our system.

### 2.2 Incremental Development of a Web Based Help Desk System

The system described in section 2.1 implemented an information retrieval function for users. The prototype described by Kim, Compton, and Kang (1999) extended this prototype by allowing an expert user to also build and maintain the help desk document knowledge base by applying keywords to help documents.

### 2.3 Document Management and Retrieval for Specialised Domains: An Evolutionary User-Based Approach.

In her PhD thesis, Kim (2003) re-evaluated the prototype described in section 2.2 and applied the concept lattice from Formal Concept Analysis (FCA) (Wille 1992) to generate a browsing structure to assist users in navigating the knowledge base.

### 2.4 A Very Different Help-Desk Application

Our system differs considerably in scope from the previous RDR help-desk systems described above, in particular:

- The A-V pairs and rules in our system are not simple keywords, and simple tests for existence of keywords. Rather, the attributes may be any type e.g. integer, float, string, enumerated type, or free-text; they may be single valued, one of a set, or some of a set; and tests may include tests for range such as 'installation date > 2001/01/30'; for existence (indicated as ?) such as '? patch 3.6.5'; for containment e.g. 'case description contains *machine generated*' or for equivalence e.g. 'version == 3.2'.
- Multiple users will describe the cases through an interactive question-answer interface to the system that will assign the relevant A-V pairs to the case.
- Our system will be maintained by multiple users, not just a single user.
- Knowledge acquisition and system maintenance needs to be in-circuit – with 50 cases per day to handle locally, users won't wait for a knowledge engineer to *get back to them.*
- Our system needs to fit smoothly into the workflow of a bustling call centre – expediency, efficiency and accuracy will be key to the system's success.

## 3. Introducing Multiple Classification Ripple Down Rules

The MCRDR algorithm (Kang, Compton and Preston, 1995) is now a decade old and in that time it has seen numerous implementations. In a programmatic sense, the algorithm is clearly and concisely explained in its founding paper, however, we include our own brief description of the MCRDR decision tree in Table 1 below. An example MCRDR decision tree is provided in Figure 1.

**Table 1: Description of the MCRDR Decision Tree**

| | |
|---|---|
| 1. | There is an N-ary Tree of RuleNodes. |
| 2. | Each RuleNode has a *rule* and a *conclusion*. |
| 3. | The topmost RuleNode in the tree evaluates to `TRUE` for every case in the system. |
| 4. | Cases comprise of *attribute-value* pairs eg in the case of a frog, *'voice'* == *'croaks*, *'movement'* == *'hops'*. |
| 5. | The *rule* at each RuleNode tests the attribute properties eg that *'movement'* == *'flies'*. |
| 6. | Each case is evaluated against the topmost parent RuleNode and then successively down the tree for each child RuleNode. |
| 7. | If the result is `TRUE` for a parent Rulenode, the case is recursively evaluated against all of its child RuleNodes. |
| 8. | The live conclusion list for a Case includes the conclusions from the last `TRUE` RuleNode in every path down the RuleNode Tree. |

As an example, in the case of a bird in Figure 1 where ('movement' == 'flies') and ('voice' == 'sings'), the last TRUE RuleNode in every path down the RuleNode Tree gives the conclusions 'sweet sounding', 'bird'.

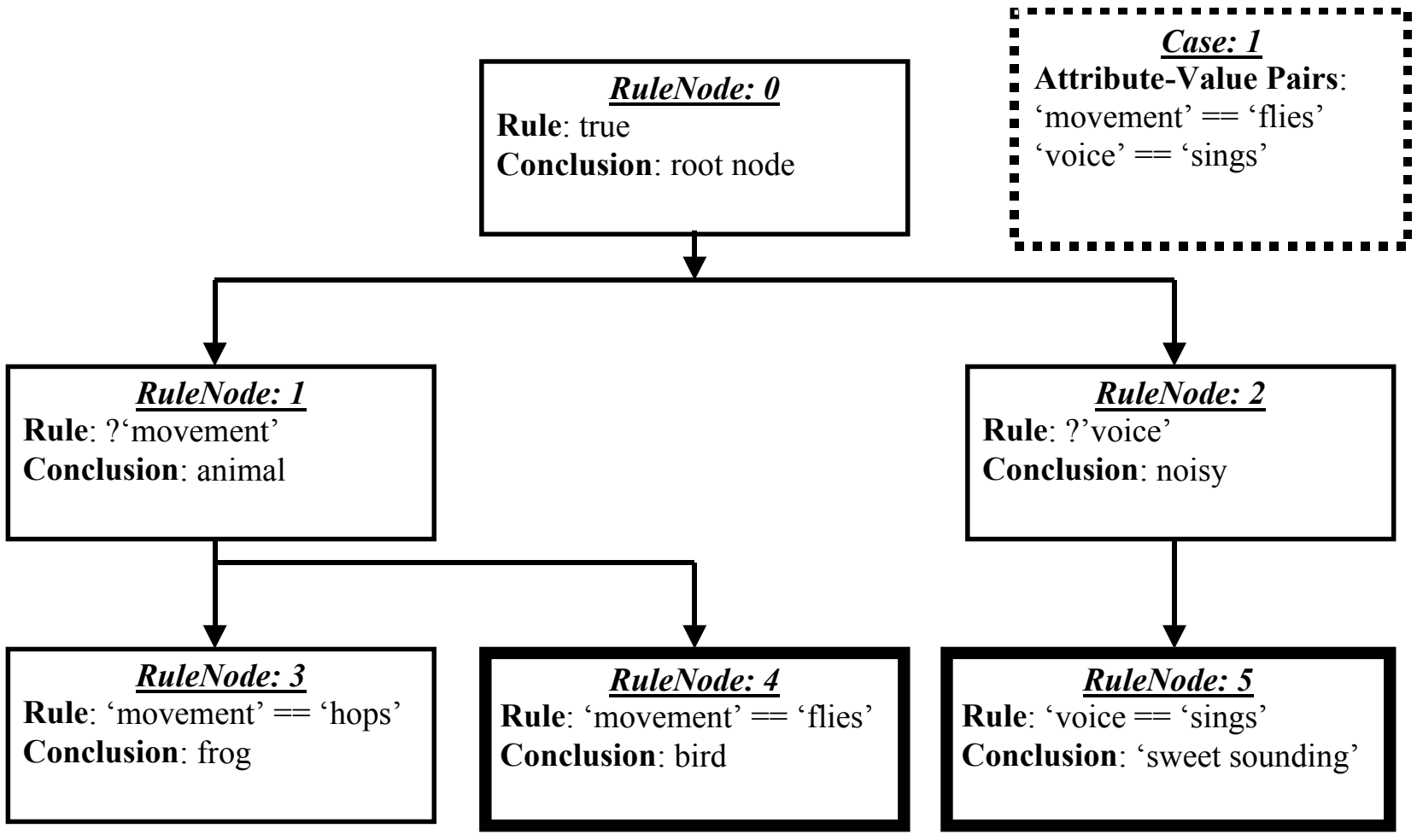

**Figure 1: MCRDR Decision Tree (Case == Bird)**

When a new case is added to the system, the user can choose to accept a given conclusion or alternatively reject it by creating a differentiating rule with an alternate conclusion. In that case:

- The new rule must be a valid boolean expression which is able to be evaluated by the MCRDR engine.The rule for the new RuleNode should be different from the rules of its ancestor RuleNodes.
- The rule for the new RuleNode may optionally be restricted to a single test eg that ('movement' == 'flies'), rather than a conjunction of tests[2].
- The new RuleNode must have either a different conclusion, or a different rule compared to it's sibling RuleNodes.
- The new RuleNode must test for some feature of the Review Case and must evaluate to TRUE for the Review Case.
- The new RuleNode must distinguish between the Review Case and all of the Cornerstone Cases for the parent RuleNode.

New RuleNodes can generally[3] be placed at one of two places in the tree (Kang, Compton and Preston, 1995):

- At the top of the RuleTree to provide a new independent conclusion.
- Beneath the current RuleNode as a replacement conclusion or as a stopping conclusion.

Further details regarding our implementation are provided in section 4.

---

[2] Where more complicated conjunctions of tests are allowed, the new RuleNode is more likely to be added to the top of the tree and a stopping rule used at the end of the path – the overall result is a flatter rule-tree structure (Kim, 2003).

[3] Actually, there is a possibility that new RuleNodes could be placed in the path between the topmost RuleNode and the current RuleNode by asking the user to identify the minimum set of rules in the current path that the case must satisfy for the new RuleNode.

## 4. A Solution for Call-Centre Service-Desks

Keeping in mind the challenges for MCRDR described in section 1.4, we intend to explore the following variations to the MCRDR theme.

### 4.1 Legacy Problem Ticketing System and Knowledge Base

Call-centres have made and continue to make enormous investments in evaluating and purchasing workflow software. Quite separate from the financial investment in software is the investment in organizational learning - "*the way we do things around here*". This extends beyond the training of front-line personnel to an investment in custom reports and metrics to assist in performance management of the call-centre.

We intend to develop a MCRDR-based expert system that can operate independently and that can be used to augment any legacy ticketing or workflow system that the call-centre may have already invested in.

### 4.2 Background Processing

To deal with the large volume of cases and minimise system-processing delays for the user, our system will do the initial evaluation of case conclusions as soon as a case is received, rather than when a customer service representative chooses to open it. In other words, the initial determination of which conclusions apply to a case will be done in background mode.

Whenever a new RuleNode is added, or a case or conclusion is edited, the system will update its internal references to maintain database integrity.

### 4.3 Maintenance

Self-maintenance of the knowledge-base is central to the design of the MCRDR system such that when revised knowledge comes to hand, the system is immediately updated to reflect the new understanding.

As well, when completely new problem domains are added, the system immediately starts training itself towards coverage of the new domain. Actually, the system can be configured to identify areas where knowledge is lacking and it can prompt users accordingly.

### 4.4 IR-MCRDR

Expediency, accuracy, and efficiency are key performance criteria for the call-centre. This means that our solution needs to be designed for minimal user-decisions.

One problem presented by existing defect tracking and knowledge management solutions in the call-centre is that users are presented with long lists of A-V pairs, many of which are irrelevant to the problem on-hand. It is left to the user to apply a mental filter for each new case when filling in these A-V pairs.

Importantly, we aim to reduce the decision burden for users of the system, and thereby speed up the process of problem determination as well as reduce the risk of information overload to the user. In this endeavor, we intend to apply and extend two variations to the RDR theme: Recursive RRDR (Mulholland et al 1993) that involved repeated inference cycles using the single classification RDR structure; and Interactive RDR (IRDR) which was a technique that allowed the RDR system to prompt the user for more information when required. Hence we propose the development of IR-MCRDR.

Our idea is to use IR-MCRDR in a configuration sense. Our system will assist the user in honing the problem definition by using the current case context to prompt the user for more detail about the specific problem being considered. Only relevant A-V entries will be requested of the user, depending on the current case context.

In addition, as more A-V pairs are gathered to define the case, the case will invoke conclusions that lie deeper down the rule paths of the decision tree, and the conclusions displayed will become more specific to the particular problem being observed.

Our hope is that this approach will quickly guide users to the most relevant conclusions for the problem case under consideration.

### 4.5 Relaxing the Case Differentiation Test for new RuleNodes

With previous implementations of MCRDR, a new RuleNode can only be added when the new rule differentiates the case in question from all the cornerstone cases at the node above. This means that when a new RuleNode is added, all of the cases that previously tested TRUE for the parent node are now evaluated at the new child node.

In our system we must retain references to old cases. Therefore, there could be hundreds, if not thousands of cases at that node.

We propose that the user need only differentiate against a limited number of cases at the parent node, namely those that present a unique set of A-V pairs, and possibly those that meet a certain expiration threshold, for example less than 1 year old.

### 4.6 Separation of Live and Registered Conclusions

We propose a separation between the *live* conclusions that the system currently evaluates for a given case, and the *registered* conclusions that users have previously confirmed as being true (that is successful solutions) for a given case. This mechanism guards against the undesirable scenario where a user may examine a case that they dealt with in the past, only to find that the system generated conclusions have changed compared to the conclusions that the user themselves registered for that case.

This proposal gives rise to the decision scenarios shown in Table 2. For example, the system evaluates an old case and shows the user that there is a registered conclusion that is no longer live – perhaps the case has been edited or the case now falls down to a new conclusion – the user can either reject the conclusion (010) to *Remove this Rule from the Registered List*, or accept the conclusion (011) to *Create a New Rule to Accept this Conclusion.*

**Table 2: Decision Scenarios, a Truth-Table**

| Live Conclusion | Registered Conclusion | Accept (1) or Reject (0) Conclusion | Action |
|---|---|---|---|
| 0 | 0 | 0 | Do nothing |
| 0 | 0 | 1 | Create a New Rule |
| 0 | 1 | 0 | Remove this Rule from the Registered List |
| 0 | 1 | 1 | Create a New Rule to Accept this Conclusion |
| 1 | 0 | 0 | Create a New Rule to Reject this Conclusion |
| 1 | 0 | 1 | Register this conclusion |
| 1 | 1 | 0 | Create a New Rule to Reject this Conclusion |
| 1 | 1 | 1 | Do nothing |

### 4.7 Managing Approvals

Obviously it is vital to the success of our tool that it actually gets used – *the more it is used, the more useful it will prove to be.* In this vein, our implementation will allow multiple users to dynamically update the knowledge.

User's confidence in the presented solutions will also play a large part in users wanting to use the system. We intend to provide for an approval mechanism that allows one or more trusted experts to indicate their approval of a given solution.

Therefore, a third conclusion state entitled *approved conclusion* will be provided. An *approved* conclusion is one that was initially live, was subsequently *registered* by a user, and finally *approved* by a trusted expert for a given case. If further down the track the case is edited, or the rule tree modified in such a way that the *live conclusions* no longer match the *registered conclusions*, then that case may be subjected to *re-registration* and *re-approval.*

It may also prove useful to allow RuleNodes themselves to have an approval status attached to them. For example, approved by *username* on *yyyy/mm/dd* per RuleNode. It is anticipated that such a mechanism may help to build credibility and hence increase the confidence that users are willing to place in system-generated conclusions.

Table 3 shows a proposed presentation format for the live, registered and approved conclusion states, and the user options to accept or reject each conclusion.

**Table 3: Live, Registered, and Approved Conclusion States for Case 1: A Sweet Sounding Bird**

| RuleNode Details | | | | | | Recommendations | |
|---|---|---|---|---|---|---|---|
| Rule Node | Rule | Conclusion | Live | Registered | Approved | Accept | Reject |
| 4 | 'movement'=='flies' | bird | yes | meganv 2004/03/30 | richards 2004/05/06 | ◉ | ○ |
| 5 | 'voice'=='sings' | sweet sounding | yes | meganv 2004/04/01 | - | ○ | ◉ |

### 4.8 Recording a Change History per Case and per RuleNode

Given that old and expired cases may *fall through* to new conclusions, or that old cases or even old conclusions may be edited and modified, a change history can be kept per case showing how the conclusion list has evolved over time.

Similarly, given that conclusions containing web references may expire and require update, a change history per RuleNode is also required.

### 4.9 Building Credibility

We intend to satisfy user-demand for credibility by keeping a record at each RuleNode of how many cases presently refer to that RuleNode for conclusions that are both live and registered.

As well, for each RuleNode, the system will record the case that first caused the RuleNode to be created, and all the other cases that presently refer to it.

Users typically wouldn't need to know about RuleNodes at all. They would simply see the information presented as approved conclusions together with a credibility statement for each conclusion.

For each conclusion, a trace of the path through the rule tree is also provided. We see this as fulfilling the role of an explanation of worthiness for system generated conclusions. As discussed by Doyle, Tsymbal, and Cunningham (2003) the use of explanations increases user acceptance of the predictions offered by knowledge based systems. They help to justify predictions in complicated domains where the domain is not fully understood or complicated heuristics are used.

Table 4 shows the type of information that could be presented to advanced users for RuleNodes in the rule tree.

Wherever a RuleNode (or conclusion) is presented, a hyperlink to that RuleNode's system (or conclusion's) details could be provided.

As well, links to other places in the rule tree where the same conclusion appears may be fruitful.

**Table 4: RuleNode 4**

| **RuleNode** | **4** |
|---|---|
| ParentNode: | 1 |
| RHS SiblingNode | none |
| Rule | 'movement' == 'flies' |
| Conclusion | bird |
| **Approval** | **richards 2004/05/06** |
| **Credibility Statement** | **Registered and Live for 3 cases** |
| Registered Case list | 1,2,3 |
| Live Case list | 1,2,3,4,5,6 |
| Live Path Case list | 1,2,3,4,5,6 |
| Approved Case list | 1 |

The registered, live and approved case lists record the manner in which existing cases are presently linked to this RuleNode. The Live Path Case list is explained in section 4.10.

### 4.10 Handling Numerous Cases – Optimising the Rule Tree

Given the large number of cases that need to be handled, the underlying data structure is critical. In our implementation, the following optimised tree structure is used:

- RuleNodes are evaluated on a once-only as-needs basis for each case (unless the case itself is modified). Most of this processing is done in the background when the case is first received. The result is recorded in the rule tree.
- The live N to N relationship between RuleNodes and Cases is recorded via the 'Live Case list' at each RuleNode and via the 'Live Rule list' within each Case.
- When a new RuleNode is added, the system only evaluates cases that are `TRUE` for the immediate parent node - the 'Live Path Case list' at each RuleNode records cases which are `TRUE` at the given RuleNode. This saves significantly in processing overhead by reducing the number of cases examined.
- The user can re-open old cases, for example those that may have dropped-through. In that instance, the case is re-evaluated from the top of the RuleTree down.
- The 'Registered Case list' and 'Registered Rule list' records the relationships between a case and its registered RuleNodes.

## 5. Conclusions

There are a number of other issues we have considered concerning feedback and collaboration, managing incentives to users to encourage the entry of good conclusions, the handling of conclusion granularity and expiration, accessibility to the system globally and continuously, usability and the possible inclusion of data mining to automatically extract knowledge from previously recorded decision data. We do not have space here to discuss our proposed strategies to these issues but acknowledge that the success of the system will only be possible if they are adequately addressed.

We have presented in this paper an overview of the call-centre domain and outlined some of the challenges it raises. We propose the use of a back-end MCRDR knowledge base supported by a web browser front-end to assist knowledge workers solve the many problem reports they receive daily. Due to the features of this domain, particularly the evolving nature of the cases themselves, we have suggested a number of modifications to standard MCRDR and have proposed Interactive Recursive MCRDR.

On the implementation side a prototype has already been developed. The key obstacles we currently face are access to the existing databases containing problem cases and solutions (in separate databases and formats) and having to develop a system that will not only interface with current systems but fit into the corporate structure and culture. These softer issues have always been the real challenges in knowledge acquisition.

**Acknowledgement.**

This work has been generously funded via the Australian Research Council Linkage Grant LP0347995. We wish to thank our industry partner for their support and assistance.

# Acquisition of ArticulableTacit Knowledge

Peter Busch, Debbie Richards

Department of Computing
Division of Information and Communications Sciences
Macquarie University
Sydney, Australia
Email:busch, richards@ics.mq.edu.au

**Abstract.** The knowledge economy is recognizing tacit knowledge as a resource even more valuable than their codified knowledge stocks. However, while much discussion is contained in the organizational and managerial literature, there are few technological solutions to assist its capture. In this paper we first consider the nature of tacit knowledge, the difficulties associated with its acquisition and the codification process. We then offer our technology-supported approaches, one from the knowledge acquisition community and the other from the knowledge management/information systems community. We compare these two approaches with each other and other related work from these fields.

## 1. Introduction to Knowledge

Organisations to date have been generally successful at creating and maintaining their codified knowledge stocks, but the tacit component is a phenomenon that is only just now starting to receive serious attention. It has for example been shown [2], that whilst codified knowledge has always permitted managerial decisions to be *planned*, it was the tacit knowledge component that was often called upon in emergency situations to provide decisions in a fast changing situation. As an aside, the structures of organisations [19] themselves may also affect transfer [15].

Tacit knowledge (TK) in itself is clearly the opposite of codified knowledge. Codified knowledge exists in print or electronic form and tends to be available to some degree either freely or for sale, or perhaps in the form of patent and classified documentation. What we often refer to as codified knowledge is however not necessarily knowledge, but information. In other words it does not become knowledge until the receiver understands what it is they are receiving. Technically speaking tacit knowledge on the other hand *is* knowledge, not data or information, insofar as the term tends to be used to describe knowledge that is far more heavily based on personal understanding or experience.

Strictly speaking tacit knowledge cannot be codified, rather what passes for tacit knowledge is actually the implicit knowledge that we as individuals all make use of to greater or lesser degrees of success. What is meant by implicit knowledge is that component that is not necessarily written anywhere, but we *tacitly* understand that using such knowledge is likely to lead to greater personal success. Stated another way, tacit knowledge is "knowledge that usually is not openly expressed or taught … by our use of tacit in the present context we do not wish to imply that this knowledge is inaccessible to conscious awareness, unspeakable, or unteachable, but merely that it is not taught directly to most of us" [35 :436, 439]. Or as Baumard [1] differentiates, "on the one hand it is implicit knowledge, that is something we might know, but we do not wish to express. On the other hand, it is tacit knowledge, that is something that we know but cannot express" (:2).

An important factor in any knowledge discussion is that of its 'stickiness'. Stickiness refers to the way in which knowledge adheres to particular individuals or contexts. Codified knowledge tends to be far less sticky than tacit knowledge, to which end tacit knowledge almost always requires human contact for transfer.

We see knowledge as being a manifestation of skills and means expressed by humans, making use of both data and information. Sveiby [34] states that "knowledge cannot be described in words because it is mainly tacit … it is also dynamic and static", furthermore, "information and knowledge should be seen as distinctly different. Information is entropic (chaotic); knowledge is nonentropic. The receiver of the information – not the sender – gives it meaning. Information as such is meaningless" (:38, 49). Although we realise that data is the most basic representation of information and that organised information requires a component of knowledge, if we take this reasoning one step further, we may envisage a knowledge hierarchy as illustrated in Figure 1. What begins as TK (Stage 1) (components of which may never be articulated), ultimately becomes separated from that which is able to be articulated (Stage 2), and eventually is so (Stage 3). In due course knowledge becomes categorised (Stage 4) and thereafter codified into rule sets (stage 5). The definitive examples of codification include mathematical, chemical or other scientific formulae. Finally, but not absolutely, the formulae are based on the axioms of the mathematics, which cannot be both complete and consistent [9], and on the decision that the interpretation of the axioms is valid in the domain in which they are being applied. Codification rests ultimately on continuing agreement to decisions previously made – no absolute or complete articulation is therefore ever possible.

Sternberg [33] notes that TK "is acquired [in the face of] low environmental support", meaning we do not receive much help as individuals in acquiring this knowledge. If the knowledge is difficult to acquire it is also difficult to transfer. Certainly a major proportion of tacit knowledge research is focused on attempting to make tacit knowledge explicit, a process that Nonaka, Takeuchi and Umemoto [22] refer to as externalisation. Broadly speaking however, tacit knowledge is gained either through (a.) personal experience over time and perhaps place or (b.) by serving in an apprenticeship fashion with someone who is senior and able to pass the knowledge on to the 'trainee' [10]. The important point to note is that tacit knowledge cannot by its very nature be passed in written format, as at this stage the knowledge is no longer tacit, but explicit. In contrast, Articulate Knowledge is acquired through formal education, writings, books, rule sets, legal code to name but a few examples.

It is important to realise that a proportion of tacit knowledge can never actually be articulated, for "much of it is not introspectable or verbally articulable (relevant examples of the latter would include our tacit knowledge of grammatical or logical rules, or even of most social conventions)" [23 :603]. Social conventions such as etiquette sets or what constitutes a proof, become codified over time as a practical matter, because the parties involved accept, agree or submit to the conventions, rules, laws (or the means of arriving at them) as the case may be. Such examples are all very contextual and ultimately tacit of course.

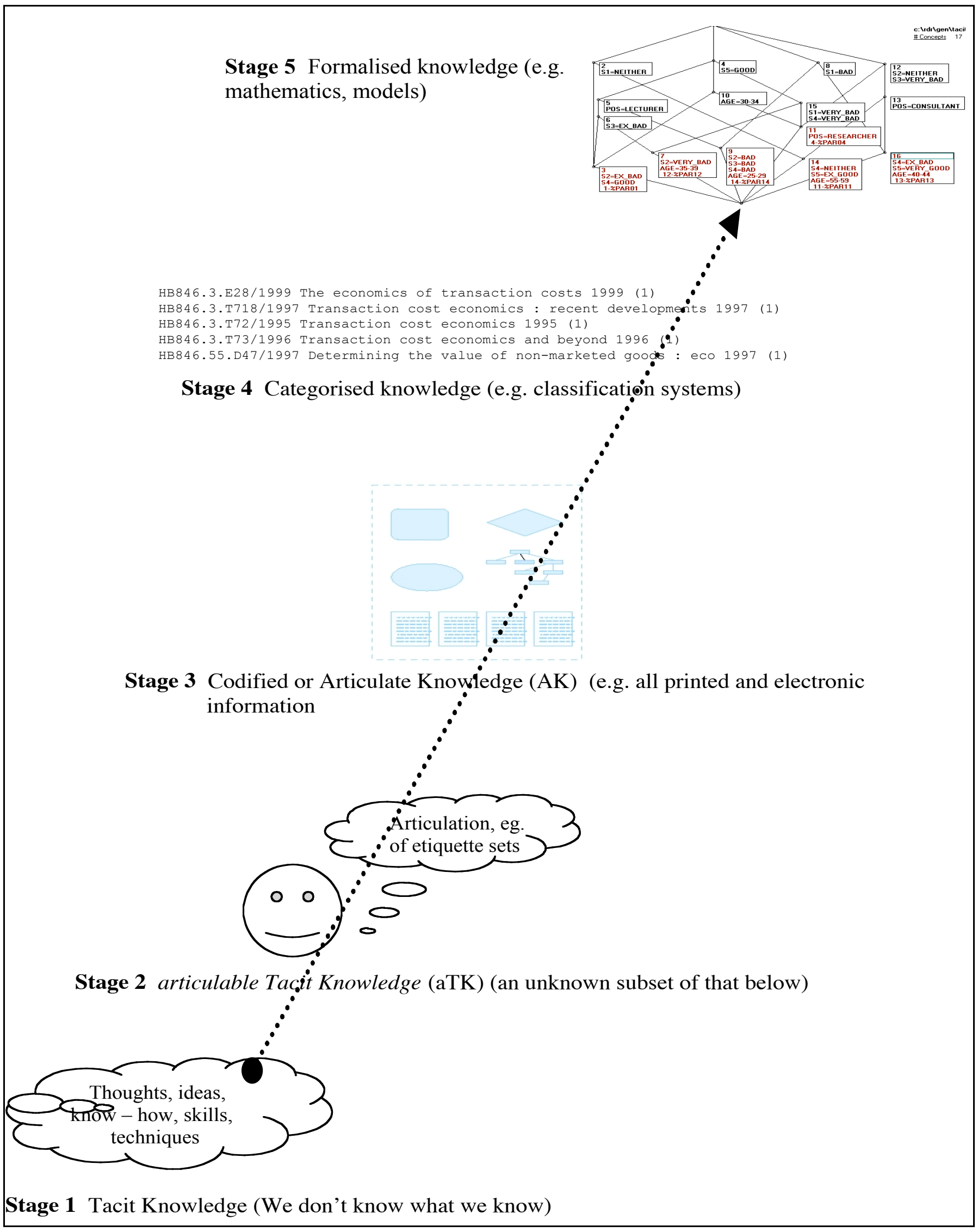

**Figure 1: The knowledge hierarchy**

MacKay [18] had, as early as 1974 alluded to the differences between articulable and inarticulable tacit knowing:

1) The "tacit" aspect of knowledge, as Polanyi himself has pointed out, is what we have in common with lower animals, presumably all of their "knowing" is tacit.

2) Therefore, we must distinguish between what *we* can say we know, and what a suitably equipped *observer* could say we know; between what *we cannot* put into words, and what *cannot be* put into words.

3) It is scientifically inappropriate to regard knowledge we can express in words as paradigmatic, and tacit knowledge as a peculiar special case. What we need from the outset is a methodology that can cope with tacit knowledge, taking verbalisable knowledge as a special case (:94)

Certainly such instances tie in with Polanyi's concepts of tacit knowledge being related to "know[ing] more than we can tell", or "knowledge that cannot be articulated", however we realise now that only a subset, even if a large subset, of tacit knowledge is truly not articulable. And that this subset is typically representing physical skill sets which simply do not lend themselves to codification, but can only be transferred through the 'indwelling' of the individual learning the new skill for themselves.

## 2. The tacit knowledge conversion process

While it has been shown "…….. that *new* tacit knowledge is generated as former tacit knowledge becomes codified" [29 :104], if we examine this process more closely, we feel in actual fact the transition to codified knowledge is not so sudden. What begins as an initial process of socialisation as pointed out by Nonaka [22], characteristic of experts showing novices 'the ropes', turns into a gradual codification process. A graphical interpretation of this principle is provided in fig. 2.

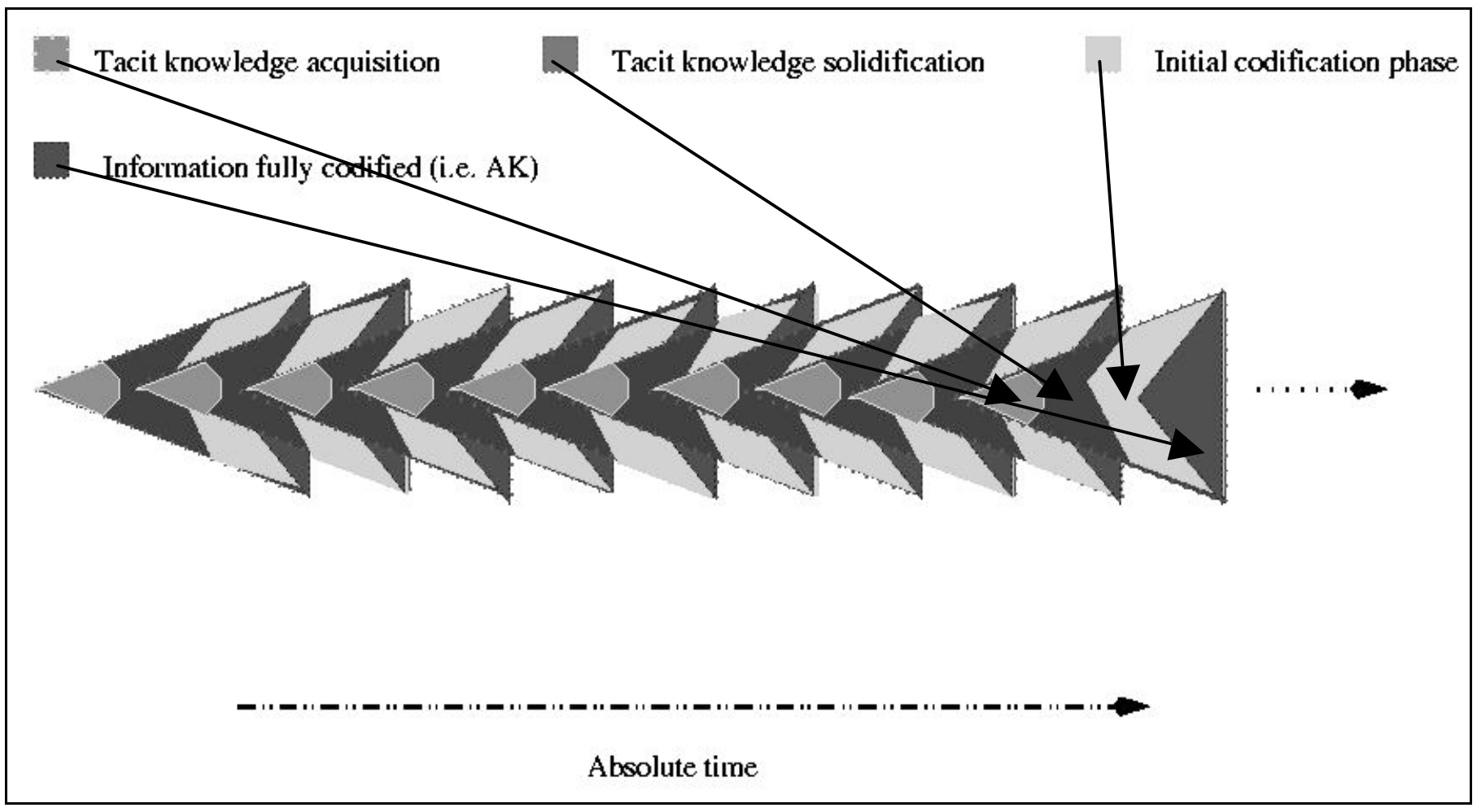

**Figure 2: The tacit knowledge codification cycle**

As shown in Figure 2, before the full codification phase takes place tacit knowledge is initially formalised by systems typically in ‘unspoken rules’, that nevertheless exist within the organisational sphere. We may term these ‘etiquette sets’. Over time even etiquette becomes codified. We find codified examples of such rule sets in almost every society which dictate how behaviour should be conducted in all manner of situations (often social), from dining behaviour to what may be deemed acceptable relationships between the sexes. The partial codification phase characterises an environment where notes are available but not in any ‘official capacity’. Examples would include ‘research in progress’, ‘draft documents’, material which is ‘not to be quoted’ and so on. Such material is far from being tacit, however fully codified it is not. Full codification is widespread, and includes all manner of printed and electronic material.

Let us bear in mind several other points. First of all tacit knowledge transference between individuals is often thought to take place whereby this knowledge becomes codified over time as for example in figure 2. We see this process as cyclical rather than strictly sequential as depicted in the figure by phases overlapping and at times occurring concurrently. In other words although tacit knowledge becomes chronologically codified, the transference from one individual to another does not take place equally. Senior people generally tend to teach junior people tacit knowledge, or experts tend to teach novices. A novice may however be senior and the expert junior, especially in the sciences and technology where young people may be more up to date technologically.

Eraut [8] provides an interesting insight into tacit knowledge elicitation problems chiefly those of bias likely within the respondents to any testing approach:

1. our series of encounters with another person are unlikely to provide a typical sample of his or her behaviour: the reasons and circumstances for the meetings will largely determine the nature of those encounters, and our own presence is also likely to affect what happens;
2. we are most likely to remember events within those encounters that demand our attention, i.e., those that are most ‘memorable’ rather than those which are most common;
3. preconceptions, created by earlier encounters, affect both parties’ behaviour on later occasions, so the sample is not constructed from genuinely independent events;
4. people develop personal constructs [12], or ways of construing their environment, as a result of their life experiences; and these affect their understanding of, and hence behaviour towards, those whom they meet (:121 – 122).

Nonetheless even given such criticisms, few alternative approaches remain for attempting to explicate and in some way measure this pervasive but all too often underestimated source of intelligence, other than that proposed by Sternberg’s Yale University research group. The work we present in section 4.2 builds on this work and will be discussed again there.

## 3. Primary reasons for undertaking tacit knowledge based research

Despite the difficulties associated with the measurement and/or capture of this elusive resource, there a numerous reasons for undertaking tacit knowledge related research particularly from an organisation and improved workplace performance perspective.

From the workplace point of view, a study of tacit knowledge is usually but not necessarily concerned with the area that has come to be known as Knowledge Management (KM). The capturing of tacit knowledge has been noted as being fundamental to such management. Indeed it was noted that "through 2001, more than 50 percent of the effort to implement knowledge management will be spent on cultural change and motivating knowledge sharing (0.8 probability)", which Casonato and Harris [5] had envisaged as including the more effective utilisation of tacit knowledge.

Tuomi [35] in relation to the Information Technology environment has summed up the importance of tacit knowledge management:

> If the design principles and methodology cannot address the tacit component, it cannot tell us where and how much we should invest in the explication of knowledge. In general, it can be argued that there has been too little emphasis on the sense - making aspects of information systems. This is becoming an increasingly important issue as information systems are increasingly used for collective meaning processing (:111).

Indeed the increasing sophistication of information systems has been a major factor in a number of organisational movements for example the migration from technology management to human based knowledge management. Another is the move from an information based view to a knowledge based one. A further example concerns the move from a hierarchical organisational view to a work activity view, for example the use of people on short term teams, based not upon their hierarchy in the organisation but the skills they bring to the team. One final example is that information systems are now not just information processing machines, rather they are now being geared towards providing a means of knowledge transfer, as in the example of Lotus Notes systems [34, 24].

The relationship of tacit knowledge to the workplace need not surprise us. Reasons for studying this phenomenon include maximising usage of organisational intellectual capital [7]. Another commonly cited reason relates to capturing the expertise of professionals, the most notable examples occurring within the *sensu latu* medical domains [31, 10, 28]. The capturing of professional expertise usually means articulating tacit knowledge in the form of generalisable principles so that these principles may then be transferred to others [28]. In other words novices will ideally be in a position to gain from a more experienced, yet perhaps not always present mentor. The expertise of a mentor often permits knowledge to be formulated and entered into an expert system, or at the very least a Lotus Notes system as for example at Roche [3]. Granted such knowledge has been explicated, but it was often tacit to begin with.

One major factor encouraging the study of tacit knowledge relates to the overall economic benefit it brings. The very issue of the economics of tacit knowledge is debateable and researchers tend to differ in their interpretations of tacit knowledge along philosophical lines, from the holism of system sciences to the methodological individualism adopted by economists. While, as noted, strictly speaking tacit knowledge by its very nature cannot be articulated [16], it is interesting to note that economists arguing in reductionist terms consider that cost is the factor preventing its complete codification. A more extreme economic interpretation is “that tacit knowledge is just knowledge not codified (but potentially codifiable)” (Cowan, David and Foray 2000 in [16]).

The need for organisations to provide environments which support tacit knowledge transfer will have an impact on work practices. For instance it has been noted that telecommuting has had a detrimental effect on tacit knowledge transfer as far as junior employees are concerned as they are unable to pick up many of the workplace cues they require for on the job success [24]. In turn, those with more marketable skills (both articulable and tacit knowledge) are more likely to find employment at a salary that satisfies them.

Thus we see that individuals need to ensure that they are in positions to acquire tacit knowledge and organisations need to find economically viable means to facilitate individuals in this endeavour by providing environments conducive to its flow and also retention in the organisation. Let us consider next how capture can occur and how its flow can be measured and modelled.

## 4. Approaches to Tacit Knowledge Capture

The overwhelming majority of research to-date has focussed on the explicit (stage 3) or above stages of knowledge. Expert systems themselves can be viewed as mechanisms for categorising knowledge and thus reside at the fourth level. Current KBS research is predominantly concerned with the development of ontologies as a way of acquiring domain and task structures (e.g. [11]). Ontologies provide a formal model and thus fit into the fifth stage in our knowledge hierarchy in Figure 1. Similarly, the previous focus on the development of general problem solving methods (PSM) also fits in the final stage. We suspect this focus on stage three or above types of knowledge is due to the apparently increasing difficulty in capturing knowledge as we move down the levels. In support of this claim, we note that the shift to developing ontologies and general PSMs was a response to the problems associated with getting experts to articulate their knowledge into expert systems. These knowledge-level modelling [20] approaches were aimed at providing a structured means of acquiring and organising knowledge. Further they aimed to support the reuse and sharing of knowledge as another means of alleviating the knowledge acquisition bottleneck. While modelling and formalisation of knowledge has been a key focus of traditional KBS research and has offered numerous computational solutions, KM research has stressed the importance of implicit or tacit knowledge but offers few technological solutions for its identification and transfer. In both communities tacit knowledge is often treated as that knowledge which can’t be captured.

In this section we consider two approaches, one from the KBS community and the other from the KM community, that offer supporting technology to capture tacit knowledge. Both approaches focus on the behaviour of experts rather than getting experts to describe what they know. In keeping with Sternberg’s observation that tacit knowledge is transferred without the assistance of others both approaches elicit the behaviour directly from the expert (or novice) rather than through an intermediary such as a knowledge engineer. However, in contrast to Sternberg's further observation that there tends to be low environmental support for acquisition of tacit knowledge, these approaches do offer some assistance.

### 4.1 Tacit Knowledge Acquisition and Modelling: A KBS Approach

First we very briefly introduce the Ripple Down Rules (RDR) [6] knowledge acquisition and representation technique. RDR is based on a situated view of knowledge. The situated view rejects the notion, that knowledge, including tacit knowledge, is stored in memory and simply needs to be retrieved in the appropriate circumstances. Instead, knowledge is seen to evolve and to be "made-up" to fit each situation. Thus, a situated view of knowledge places great emphasis on incremental techniques that allow change, capture context and which acquire knowledge without relying on a human to state or codify that knowledge.

RDR offers a way of capturing knowledge, both tacit and explicit, because it does not attempt to distinguish between the different types of knowledge but captures knowledge while the domain expert exercises his or her expertise. The domain expert is not asked to develop models of the domain or to offer explanations of their reasoning process/es. RDR facilitates articulation and performs the codification of the behaviour of the expert which is based on their already codified knowledge plus their tacit knowledge.

Knowledge acquisition involves running a case against the current knowledge base resulting in one or more conclusions being offered by the system. The user reviews each of the conclusions. If a conclusion is missing they can add a new conclusion. If they disagree with any conclusions they may choose to override each incorrect conclusion with a new one by assigning a new conclusion in its place. This forms the rule conclusion. The user is shown the case/s associated with the rule that gave the incorrect conclusion and the user must pick features in the current case that differentiate the current case from the cases previously seen by this rule. The features chosen are attribute-value pairs found in the current case, such as age=27 or age=young_adult. By picking features which differentiate between the current and previous cases, the approach ensures that the new knowledge does not invalidate prior knowledge. Thus addressing the maintenance and validation problems associated with traditional rule-based systems.

The update of the knowledge base occurs without the user being aware of the structure of the knowledge, the knowledge representation or that the conclusion chosen and case features comprise the rule conclusion and conditions, respectively. From the user's viewpoint, the process is simply one of: run a case, review system's conclusion, if they agree go on to the next case. If they don't agree, the user states what conclusion would be appropriate and why in the context of the features of the case. This is what experts do naturally.

Knowledge transference occurs when another individual uses the knowledge base (KB). Transfer is further assisted through the use of Formal Concept Analysis (FCA) [38] which allows retrospective and automatic development of knowledge models that the user can explore. In this technique FCA takes the RDR KB as input and generates a set of concepts which are ordered into a complete lattice. When lattices from multiple experts are combined [25] the resulting lattice can be viewed as an ontology because the lattice provides a specification of a shared conceptualisation. By capturing knowledge in action, we support codification (stage 3) of articulable tacit know-how (stage 2) and acquire (and generate) knowledge at stages 4 and 5.

In summary, RDR is a hybrid case-based and rule-based approach. The cases provide the context in which the knowledge applies, and the rules, together with the use of an exception structure for knowledge representation, provide the indexes for storage and retrieval of the relevant cases. Context is also critical in the FCA technique and captured in what is known as a formal context. When we automatically convert the RDR KB to a formal context we can then visualise the knowledge in a line diagram. The combined use of RDR and FCA approach thus supports all types of knowledge in the knowledge hierarchy.

### 4.2 Tacit Knowledge Measurement, Modelling and Diffusion: A KM Approach

We now consider a different approach that is not concerned directly with the capture of tacit knowledge at all, but rather the identification of its existence (stage 1 to 2) and the transference process (stages 2 to 3). As noted, very little work has been conducted at these levels, with stages 4 and 5 already well researched in most disciplines. The knowledge captured is a side-effect which could be applied in a more traditional way to assist with decision making and knowledge transfer or even to determine whether someone is highly employable or not. This somewhat unusual motivation contrasts with KBS research and has come about because the focus is on tacit knowledge which by its very definition does not lend itself easily to articulation. Given that knowledge is highly contextual and to a large extent in the "eyes of the beholders", measurement of its existence is in many cases as relevant as, and certainly a first step in, its actual capture.

While with RDR we did not attempt to define any of the types of knowledge being captured, in keeping with the goals of minimal modelling and effort, for this tacit knowledge work we needed to further refine our notion of tacit knowledge and define what was being measured and/or acquired. For the *practical* purposes of this research conducted in the Information Technology (IT) domain, tacit knowledge was defined to comprise the *articulable implicit IT managerial knowledge* that IT practitioners draw upon when conducting the "management of themselves, others, and their careers" [36]. This approach to the IT managerial nature of articulable tacit knowledge follows closely along the lines of [1]. When such tacit knowledge is shared from mutual experience and culture it gains a dimension within an organisation.

The details of the research goals, methodology and case studies are given in [4]. The essence of the work was development, deployment and detailed analysis of a survey conducted within a number of organisation. The questionnaire included an inventory of 16 IT workplace scenarios that sought to test how experts (as nominated by their collegues as part of the survey) responded to these typical scenarios. See Figure 3 for an sample scenario and answer option. This approach to tacit knowledge testing follows along the lines developed by Sternberg [33] from the field of psychology. The responses of the peer-identified experts are treated as the tacit knowledge oracle and compared against the responses of novices to measure who has and how much tacit knowledge within the organisation To determine if there are differences between population groups (age, gender, ethnicity, educational background, employment tenure) and the levels of tacit knowledge present within the groups, and whether this knowledge is likely to be passed from and among these different groups we also gathered biographical data via the survey. The responses of the experts and novices, together with their biographical characteristics were analysed using statistical methods and modelled qualitatively using Formal Concept Analysis [38], which permitted more fine grained analysis in a graphical form.

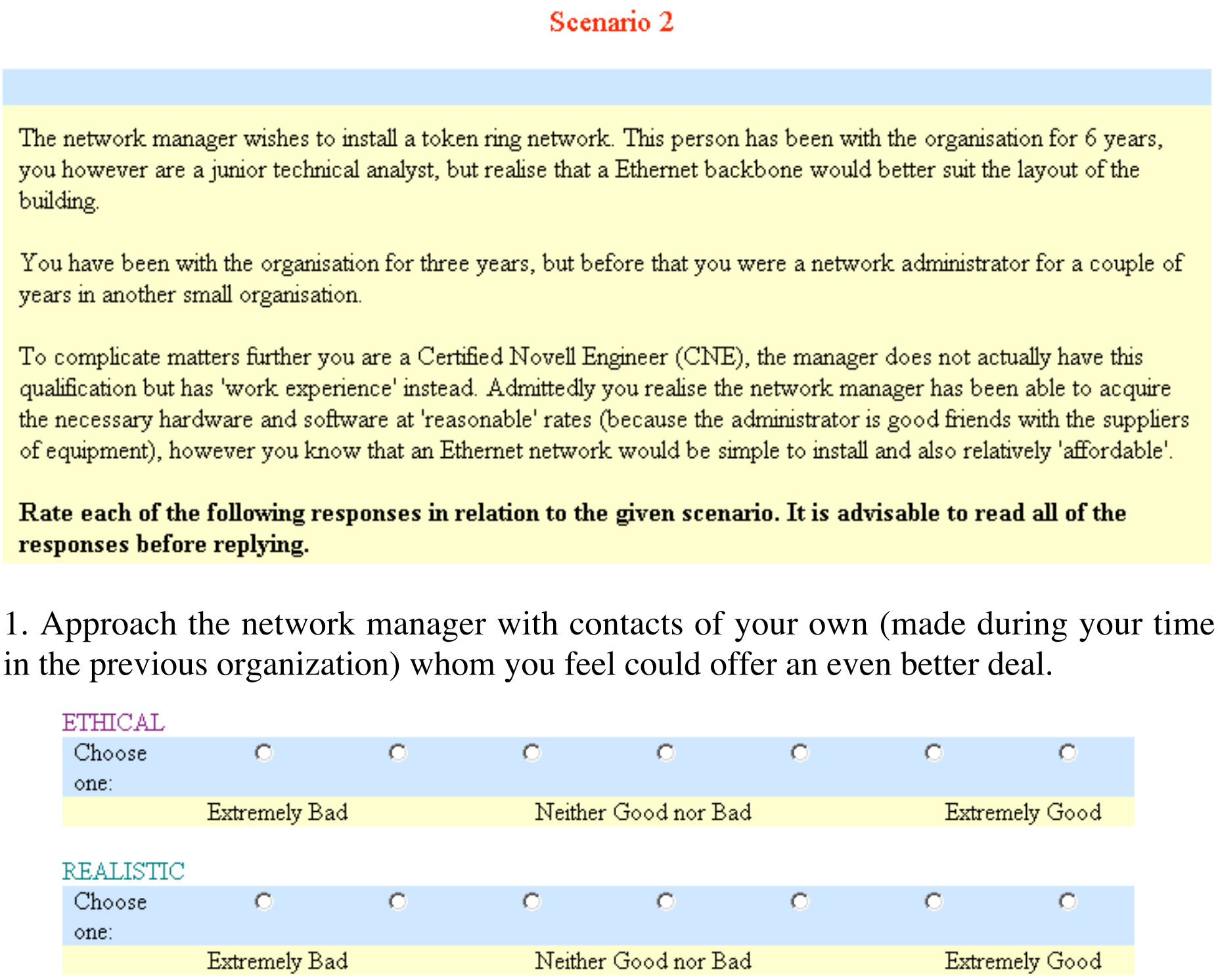

Scenario 2

The network manager wishes to install a token ring network. This person has been with the organisation for 6 years, you however are a junior technical analyst, but realise that a Ethernet backbone would better suit the layout of the building.

You have been with the organisation for three years, but before that you were a network administrator for a couple of years in another small organisation.

To complicate matters further you are a Certified Novell Engineer (CNE), the manager does not actually have this qualification but has 'work experience' instead. Admittedly you realise the network manager has been able to acquire the necessary hardware and software at 'reasonable' rates (because the administrator is good friends with the suppliers of equipment), however you know that an Ethernet network would be simple to install and also relatively 'affordable'.

**Rate each of the following responses in relation to the given scenario. It is advisable to read all of the responses before replying.**

1. Approach the network manager with contacts of your own (made during your time in the previous organization) whom you feel could offer an even better deal.

ETHICAL

Choose one:

Extremely Bad | Neither Good nor Bad | Extremely Good

REALISTIC

Choose one:

Extremely Bad | Neither Good nor Bad | Extremely Good

**Figure 3: Scenario 2 from the tacit knowledge inventory and Answer Option 1**

In addition to modelling the knowledge and the features of the knowledge holders, we sought to map the likelihood of intra-organisational diffusion of aTK among information technology personnel. The term likelihood is used here, because absolute knowledge transfer is difficult to prove other than through the ability of reading another's mind. In order to gain an insight into knowledge flows, we need to be able to map the social relationships that take place between employees. The application of Social Network Analysis (SNA) [27] permits us to illustrate such relationship patterns in the form of questions answering who is seen, how frequently, the meeting importance and the formality of the meeting (for example, a chance meeting at the coffee machine vs. a formally organised and conducted meeting).

Our research involved 128 participants across 3 organisations of size small, medium and large. While at this stage our extrapolations are restricted to the organisation studied, we were able to determine that:

- experts gave significantly different responses to certain (types of) questions as compared to novices (via formal concept analysis combined with statistical analysis of the survey);
- as a follow on from the previous point, we were able to conclude a number of general behavioural characteristics of experts as compared to novices such as experts being prepared to say they are overcommitted and are less likely to "pass the buck".
- some individuals in the organisation were behaving like experts but had not been recognised as such by their peers (we called these the expert non-experts) (via FCA),
- biographical parameters did not play a *significant* role with regard to tacit knowledge utilisation and information technology personnel, that is, experts did not belong to a certain demographic (via SNA and statistical analysis).
- tacit knowledge flow from experts and expert non-experts to novices was best achieved in the small-sized organisation (via SNA),
- tacit knowledge bottlenecking could be seen to exist particularly in the largest firm (via SNA)
- the characteristics of the optimal firm include a single clique arrangement, a lack of widespread use of electronic forms of communication, a dense communication pattern insofar as daily meetings involve all staff, and meetings held are largely informal (via SNA).

In conclusion we note the following similarities and differences between this work and the work described in the previous subsection. In both approaches: knowledge is acquired using grounded examples in the form of cases or scenarios; experts are identified by their peers; and FCA is used to model the captured knowledge. In contrast, the knowledge captured via RDR is in the form of rules which support deductive reasoning and there is an explicit attempt to articulate knowledge. The tacit measurement and diffusion work captures a range of responses which are seen as alternative solutions of varying suitability and identification of who has tacit knowledge rather than being concerned with what that knowledge looks like. Thus the approach can be used to determine unidentified experts. The knowledge acquired via RDR is closer to the traditional expert system approach where the KB is based on the view of a single expert, though FCA has been employed to support knowledge comparison and integration [25]. On the other hand, the tacit knowledge measurement and diffusion work is focused on comparison of experts and novices and the likelihood of tacit knowledge diffusion through the use of SNA. RDR or FCA do not consider knowledge flows. The key similarities that make them suitable for tacit knowledge acquisition is that both are grounded in cases/scenarios and concerned with knowledge in action.

## 5. Related Research and Conclusion

Within the KM literature the work by Noh et al. [21] bears many similarities as it is also case-based and uses cognitive maps which are in some respects like the FCA concept lattice. However, just as we have found in our review of other work in the KBS area, Noh *et al.*'s approach begins with a formalisation phase in which the user is required to develop a cognitive map. The cognitive maps are stored in a case base. Given the difficulties associated with acquiring and, even more so, validating models we have some reservations with starting with formalisation by the user. We also have a reservation regarding the cognitive maps themselves based on our experience into causal modelling which found that getting experts to formalise causal knowledge was extremely difficult since this knowledge was often unknown. A better approach was to automatically generate possible causal links and allow the user to review and revise these [17]. Kolodner [14] suggests the use of cases as the starting point in domains where causal models are not well understood. However, in the approach by Noh et al. 2000, causal knowledge must first be acquired from which cases are developed. From Kolodner's remarks we could conclude that the knowledge being captured is actually explicit and codified knowledge rather than tacit knowledge. Following the formalisation phase in [21] is the reuse phase. In this phase the case base is adapted to fit the new situation using fitting and garbage ratios to retrieve appropriate cognitive maps from the case base. Indexing, retrieval and adaptation of cases are not simple tasks. To overcome these difficulties, the RDR approach uses rules specified by the expert in the course of problem solving as the indexes to our case-base. The final phase of [21] is problem solving where the adapted cognitive model is applied to the new problem and then stored in the case base. The two approaches we offered begin with the user performing problem solving on cases/scenarios and formalisation of the acquired knowledge into a concept lattice is handled by the system rather than the user.

The incremental, action-driven and context-based nature of our work is also found in the work of [26] who have developed a knowledge-enhanced email system known as kMail. When a user sends an email they can include links to organisational memories such as databases or websites which results in a memory-concept association being developed. As in our approaches, knowledge acquisition/maintenance is performed by the user and occurs when the user deems the context to be appropriate. Knowledge in action is captured incrementally without the need for the user to prespecify knowledge models. The simple nature of interaction in kMail is another feature that we share and commend. KMail demonstrates that if you allow knowledge to be captured in action, the distinction between explicit and tacit knowledge becomes irrelevant. Despite these fundamental similarities which demonstrate the importance of handling knowledge in context and getting the human computer interaction side of the system right, the kMail system differs in the knowledge acquisition technique, the knowledge resources, the nature of the problem and the purpose of the systems.

The combined RDR/FCA approach is novel within the KBS community. There are other approaches which emphasise the role of the user, (e.g. the Protégé family of tools [11]) or which do not ask the expert to describe their knowledge but allow the knowledge to emerge through various interactions (e.g. tools based on personal construct psychology [30]. However, in the first case there is still reliance on the user to define the knowledge models up front. In the second case, the techniques are not incremental in that user must consider the whole domain and specify the context at the start. The RDR approach is incremental and the context evolves as new cases are seen.

The triangulated KM methodology offers more than a unique combination of three existing approaches. Previous work into the measurement of tacit knowledge has been primarily within the field of psychology. The Sternberg means of testing for tacit knowledge is considered to be the most practical approach for undertaking research in the organisational domain. Busch differs from the work of Sternberg in that the tacit knowledge inventory (the workplace scenarios) were based in the IT domain, rather than business management, and the questionnaire approach is combined with FCA and SNA. In Sternberg's work participants tended to be students or military personnel and thus he was able to select the appropriate population size. Our study however was based in actual organisations of varying sizes with varying levels of access to employees given to us. To complicate the access issue was the fact that since we needed to gather relationship data for the social network analysis component we were unable to make the questionnaires anonymous and this further reduced the number of willing participants. Given that psychological approaches to testing tend to rely heavily on both descriptive and analytical statistics and we were faced with (sometimes small) sample sizes beyond our control, FCA became a valuable alternative means of data analysis. The application of FCA to the visualisation of questionnaire data is quite novel, with the exception of the work by Kollewe [13] which represented the data in an alternative way and Spangenberg and Wolf [32] which also used FCA for displaying the results of likert scales. FCA has proven useful not only in interpreting biographical and tacit knowledge inventory results specifically but also in the identification of those who behaved similarly to experts but who had not been deemed by their peers to be experts. Through identification of this third group using FCA we were able to obtain statistically significant differences between the novices and experts (peer-identified plus FCA-identified). The other major component of the research which also distinguishes it from the work of Sternberg was to assess the soft knowledge flows within three specific organisations.

In summary, this paper has considered the nature of tacit knowledge and where it fits into a hierarchy of knowledge, noting the difficulties associated with capturing tacit knowledge and in conducting tacit knowledge research. We speculated on the process by which tacit knowledge becomes codified and offered two quite different approaches: Ripple Down Rules and Tacit Knowledge Measurement, Modelling and Diffusion. The latter was explicitly concerned with the tacit knowledge component but not necessarily its capture and the former explicitly concerned with its capture but not just the tacit knowledge. Nevertheless, both managed to acquire tacit knowledge through a focus on the behaviour of experts as they interacted with situations rather than the more mainstream approaches to knowledge acquisition which require its articulation by a domain expert typically via a knowledge engineer. It is through this focus on knowledge in action, that we go beyond the capture of codified (explicit book) knowledge to also achieve capture of tacit (implicit know-how) knowledge.

# Incremental Knowledge Acquisition Using RDR for Soccer Simulation

Angela Finlayson and Paul Compton

Department of Artificial Intelligence,School of Computer Science and Engineering,
The University of New South Wales, Sydney 2052, Australia

**Abstract.** This paper describes a system which aims to allow soccer coaches to specify the behaviour of agents for the Robocup simulation domain. The system focuses on an incremental approach to building multi-agent teams, supporting modification and maintenance of strategies over time. The technique that has been used is an adaptation of the incremental knowledge acquisition methodology called Ripple-Down Rules (RDR)[1]. RDR has been comprehensively evaluated for classification problems and has been extended for use in other domains, but has not been explored in the area of co-operation and planning in a multi-agent environment. Preliminary results suggest that the use of this incremental approach for multi-agent systems is promising and indicate the potential for future work in this area.

## 1 Introduction

The Robocup soccer server [13] is a useful simulation tool for developing and evaluating different approaches in the field of Artificial Intelligence, robotics and multi-agent systems. This domain has been of particular interest due to the underlying challenges of multi-agent co-ordination in a complex real-time environment with limited communication. More established teams such as CMU [15] and Uva Trilearn [16] have released code samples, providing solutions to many of the low-level challenges such as server synchronisation, individual skills such as kicking, intercepting the ball, dribbling and world model creation. These code samples serve as a great start for many new teams. As a result, much of the current research focuses on higher level strategies such as planning, team co-ordination [18, 19], and opponent modelling [12]. With the introduction of the online coach [14] to the soccer server, which acts as an external advice-giving agent there has also been increased interest in the generation and integration of advice [11].

Most of the research so far has involved hand coding of these higher level strategies or adapting machine learning techniques such as reinforcement learning [17]. The drawback with many of these techniques is that often the tactics and strategies are buried within the code or the system. This can make modifications to strategies time-consuming, error-prone and difficult to debug [20]. Another problem is that specification of behaviours is limited to computer programmers rather than domain experts. Although domain experts can be consulted, much is lost in the translation. It would be desirable for domain experts(i.e. soccer coaches) to have a primary role in knowledge acquisition and to be able to directly interact with the system to produce teams and be able to add and refine knowledge to develop a team over time. Teams such as the Dirty Dozen [20] and the Headless Chickens [21] have attempted to address this issue. The Dirty Dozen developed SFLS, a rule based language that allows representation of team strategies in a rule base that can be modified easily by humans. However there is still the issue of added or modified rules interacting with existing rules which can result in undesired and inexplicable behaviours. This is the core maintenance problem of any rule-based system [9]. We believe it can only be worse where there is not a natural link between human descriptions of soccer behaviour and machine data representations and manipulations.

The Headless Chickens focused on allowing domain experts using a graphical user interface to specify individual and team strategies. The interface allows the user to specify general strategies about player formations and passing and dribbling directions during different play modes. Styles could also be chosen such as whether a player has a preference for dribbling or passing the ball. These chosen styles determine priority levels for the different kinds of behaviours during a game. Using this system, the user has limited control over the agent making it hard to predict agent behaviour. It is also difficult to determine reasons for unwanted behaviour [21].

The aim of this paper is to explore the possibility of using an incremental knowledge acquisition technique to elicit knowledge of soccer strategies from a non-computer science soccer coach. This system would aim to provide an easy to use interface that allows an expert to monitor and refine strategies of soccer. The critical difference from the work above is the focus on gradual incremental refinement. Our motivation is to develop such techniques for complex interactive multi-agent planning environments with soccer as a challenging example.

The paper is organised as follows: The next section provides an overview of the knowledge acquisition methodology ripple-down rules(RDR)[1] giving examples of how it can be used by domain experts to easily create knowledge bases, application areas that it has been applied to and its inference structure. In section 3, we discuss our framework for adapting RDR to the domain of robot soccer simulation. Section 4 describes our evaluation of the system, section 5 discusses possibilities for future work and in section 6 we give our conclusions.

## 2 RDR Background

RDR is a tool that was developed to facilitate incremental knowledge acquisition by domain experts without the aid of a knowledge engineer. RDR was inspired by the observation that experts don't tend to give comprehensive explanations for their decision making. Rather they justify their conclusion given the context of the situation [9]. Based on this philosophy, certain features of RDR have emerged. The system gradually evolves over time while in use and validates any rules added to ensure that the addition of new knowledge does not degrade the previous knowledge base. Rules are added to the knowledge base to deal with specific cases where the system has made an error. These cases that prompted the addition of the new rule are stored along with the rule and are called cornerstone cases. The addition of new knowledge only requires the expert to identify features in a case that distinguish it from other cornerstone cases retrieved by the system and new knowledge is organised by the system, rather than the expert. These features provide the central difference between RDR and other Knowledge Acquisition approaches [2]. The advantage of this system is that the cumulative refinement over time allows the system to develop a high level of expertise, with the expert only expected to deal with individual errors. This approach contrasts with the intensive knowledge modelling approaches used in other knowledge acquisition systems.

### 2.1 RDR in practice

RDR systems have been developed for a range of application areas and tasks. The first industrial demonstration of this approach was the PEIRS system, which provided clinical interpretations for pathology testing [30]. The approach has also been adapted and used for a number of tasks such as multiple classification [4], control [5], heuristic search [8], document management [10], configuration [6] and resource allocation [7]. There has also been work done in the area of combining machine learning techniques with RDR to reduce the amount of knowledge acquisition needed from an expert [3].

Studies comparing RDR to machine learning techniques have shown that RDR systems converge and end up with similar sized knowledge bases as those created by machine learning techniques [4, 22] and that they cannot be compressed much by simple reorganisation [23]. The exception structure of RDR has been used for representation for machine learning [24–26], it was found to tend to produce more compact KBs than other representations [27]. However machine learning systems depend on well classified examples in sufficient numbers whereas an expert can provide a rule for a single case and a working system will start to evolve. Experts can also deal more successfully with single rare cases [3].

Overall these studies demonstrate that the application of RDR techniques can be used to provide simple and effective knowledge acquisition environments in a range of areas. There is now significant commercial experience of RDR confirming the efficiency of the approach. One company, Pacific Knowledge Systems supplies tools for pathologists to build systems to provide interpretative comments for medical Chemical Pathology reports. One of their customers now processes up to 13,000 patient reports per day through their RDR knowledge bases and have built about 10,000 rules, giving very highly patient-specific comments. They have a high level of satisfaction from their general practitioner clients and from the pathologists who keep on building more rules, or rather who keep on identifying distinguishing features to provide subtle and clinically valuable comments. A pathologist generally requires less than one days training and rule addition is a minor addition to their normal duties of checking reports, taking at most a few minutes per rule. (Pacific Knowledge Systems, personal communication)

In this paper we explore the possibility of applying this technique to the Robocup simulation domain and more generally areas that involve multi-agent strategies and co-ordination.

### 2.2 Knowledge Acquisition with RDR

An empty RDR knowledge base starts of with one rule which classifies all cases with a default classification. Over time the expert interacts with the system, adding rules to correct classifications on a case by case basis. Example 1 shows the general structure of a single classification ripple-down rule knowledge base.

*Example 1 (Structure of a single classification RDR knowledge base).*

**if** true **then** default classification **except**
    **if** a **and** b **then** c1 **because** case1:[a,b,q] **except**
        **if** c **then** c2 **because** case2:[a,b,c]
    **else if** d **and** e **then** c3 **because** case3:[a,c,d,e]
    **else if** f **then** c4 **because** case4:[a,c,f]

RDR exception structures can also be viewed as binary trees where each node has a rule condition, a conclusion and two branches. One branch represents the exception branch and the other represents the alternative branch. Figure 1 shows an alternative representation to the resulting knowledge base shown in Example 1.

In this example there have been four rules added to the knowledge base along with the original default rule. If we used this knowledge base to classify a case where a and b were true then we would conclude c1. However if in our case a, b and c were true we would conclude c2. This illustrates the exception structure used in RDR. If we had a case where a and b were not true, the other alternative rules would be tried in order. In this case if d and e were true, then we would conclude c3, or if f was true we would conclude c4. If the case satisfied none of the alternative rules, then the default classification would be given.

Along with conditions, each rule contains a cornerstone case (case1, case2, case3 and case4). These represent the cases that were classified incorrectly and prompted the creation of the new rule they are associated with. When a case is misclassified, the cornerstone case associated with the last rule that fired, is used to build a new rule. The expert must choose the relevant attributes that differentiate the new case from the corner stone case, and these are used in the condition of the new rule. The new rule is automatically added to the RDR knowledge base structure. For example if we had case5 with attributes [a,b,f,g,q] this would be classified by our example above, as c1. If the expert decided that this was incorrect and should really be classified as c3, a new rule would be added to the knowledge base as an alternative on the exception branch of Rule 1. The expert would choose the features in the new case that differentiated it from the cornerstone case (in this example case1). In this example f or g could be chosen to differentiate between the new case and the old case on their own or in combination with other attributes in the new case. Attribute q could not be chosen on its own as it appears in the cornerstone case and the current case and would not differentiate between the two. Figure2 shows the resulting RDR knowledge base if the expert chose g as the attribute to differentiate case5 and case1.

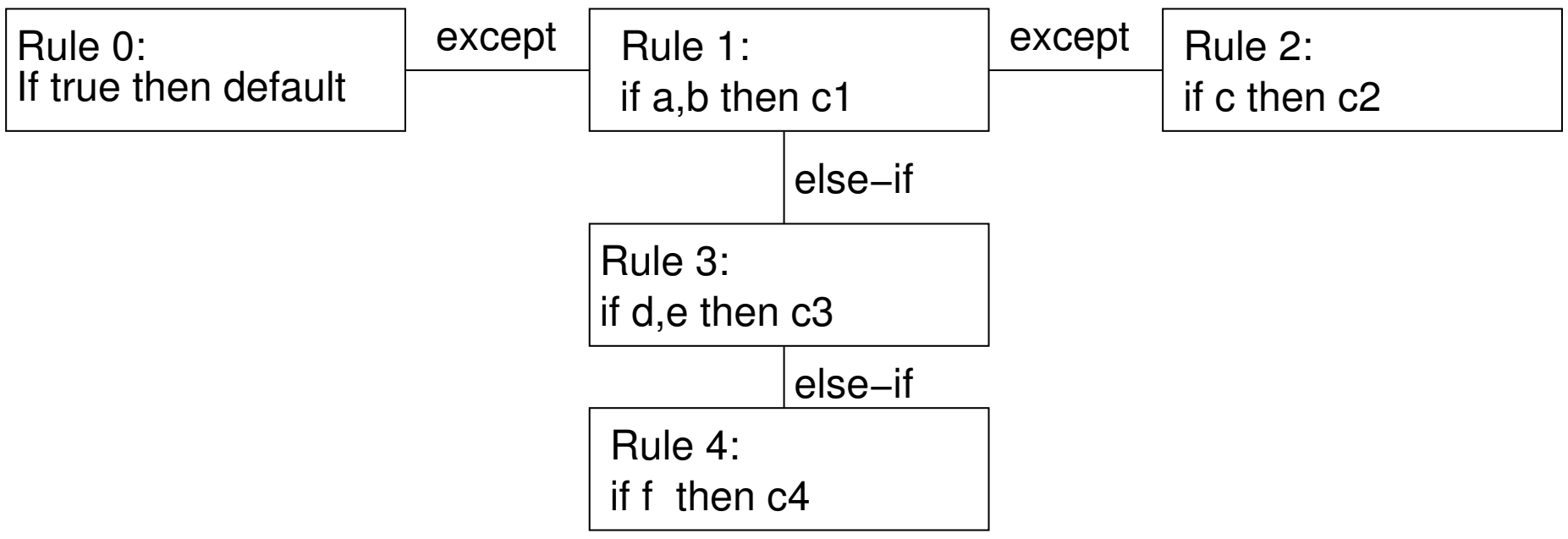

**Fig. 1.** Tree representation of the RDR in Example 1

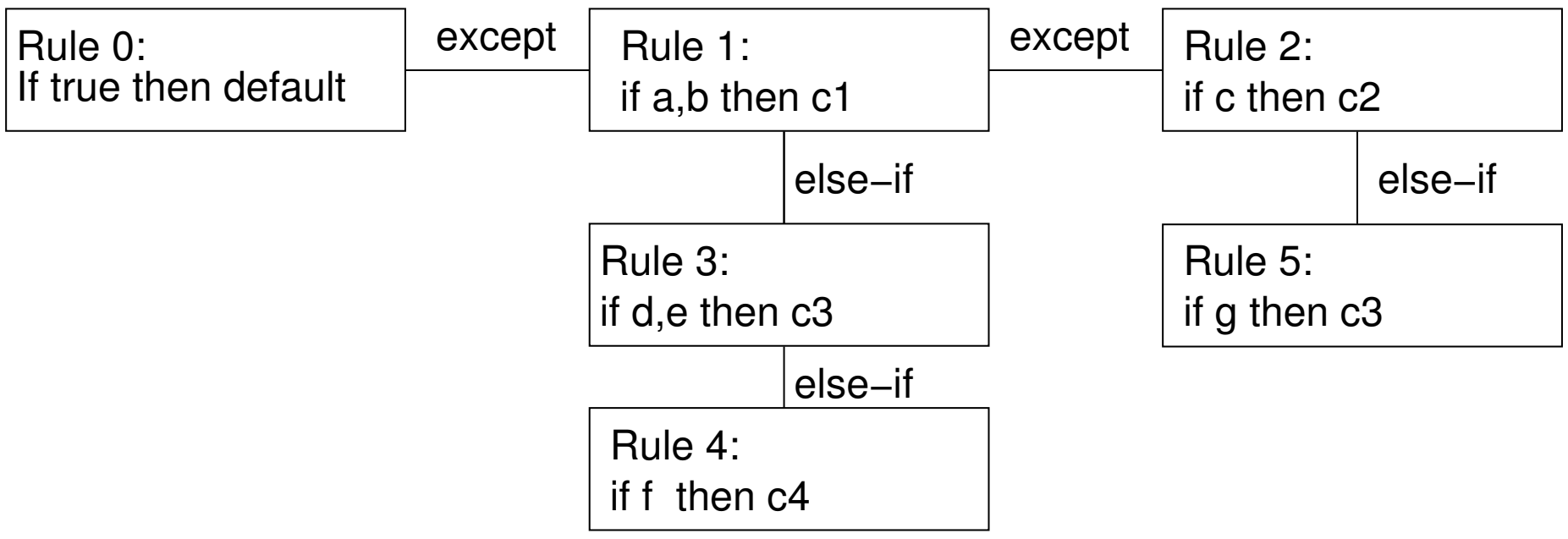

**Fig. 2.** Tree representation of the RDR in Example 2

## 3 A framework for adapting RDR to the soccer simulation domain

A framework for developing RDR soccer agents (RDRAs) has been developed. RDRAs are soccer agents that incrementally acquire knowledge from a human expert. This requires building two components, an agent with an RDR structure that controls its behaviour and an interface to allow the expert to update knowledge in the RDR knowledge base used by the agents.

### 3.1 RDR agent

Soccer playing clients were created where an RDR inference structure was used to decide what behaviour to undertake at each clock cycle. This has been done using a single classification RDR structure. The RDR knowledge base consists of rules that map attributes that represent the world model of the agent to an appropriate behaviour. The world model of the agent includes beliefs about the state of the game such as the location of the ball, the location of other teammates, the location of the agent itself and past actions. Behaviours include actions that a human player might take to play a soccer game eg. pass to player ahead, intercept the ball etc. The collection of low level behaviours and underlying world model and synchronisation mode was primarily developed from rUNSWiftII [28] which was based on the CMU99 low level code release [15].

For simplicity our system did not include all the available features from the soccer server. There was no use of the online coach facility or heterogeneous players, and communication between players was limited to that of taking turns of sharing their world models. AttentionTo and PointTo features were not explored either. However our incremental approach allows for easy maintenance and addition of other features at a later stage. The testing focused on creating a knowledge base which is shared throughout the team. The knowledge base we concentrated on creating was for the normal play mode for all players except the goalie. Code from rUNSWiftII team was used for other play-modes and the goalie. In later work we will provide knowledge acquisition for these other play modes. This knowledge could be incorporated into the existing knowledge base or separate knowledge bases.

The players were given different general roles such as defender, midfielder, attacker and sweeper. These roles were also broken down further to uniquely identify each player i.e. left defender, middle defender, right defender. The question arose as to whether each role would have its own knowledge base or a shared knowledge base. It was decided that since there would be some behaviour that would be common to all roles, that one knowledge base would be created for all players. The roles were made available as attributes to allow the expert to specialise behaviours within the shared knowledge base. However the other approach could also be tried.

The RDR knowledge base starts off with nothing but a rule specifying the default behaviour of the player. This is the behaviour that the player uses in the absence of any other information. In our implementation the default behaviour was to do nothing. However other default behaviours could be used such as facing the ball or running to the ball. Even the simple high level code distributed in code releases [15] [16] could be used as the default behaviour. When a game is played, snapshots of the agents world models and behaviours chosen for each cycle are logged. This is used as the data for the knowledge acquisition phase. As the knowledge base grows, rules are added and the overall strategy is incrementally refined.

An example of the kind of rules that can be added to the knowledge base is given below. In this example attributes roleType, inRegion and ballKickable are used and the behaviour chosen is passToPlayerInRegion. Both attributes and behaviours can have arguments. The attribute inRegion and passToPlayerInRegion both require arguments representing regions of the field. These are regions of the field that have been pre-defined. In future versions the end user will be able to define regions of their own.

*Example 2 (An example of an RDRA rule).*

**if** roleType == Defender **and** inRegion(LEFTMIDDLE) **and** ballKickable
**then** passToPlayerInRegion(LEFTATTACKING)

### 3.2 Knowledge acquisition and inference from user perspective

The process of knowledge acquisition aims to map the state of the field observed by the agent to reasons to justify an appropriate behaviour. The interface allows the expert to watch or step through a replay of a game and find incorrect behaviour. The game can be paused at the appropriate cycle where the expert believes a player should be behaving in a different way. The expert can then add new rules to change future behaviour of agents in similar situations. A display panel displays the world models of each agent as this is what the agent must base its decision upon. The world models are logged to file during the original game to make this facility possible. The display panel also displays the behaviour that the agent decided upon for each cycle. This facility not only provides an interface for knowledge acquisition but is valuable for debugging low level behaviours and world models.

The cycle of knowledge acquisition from the users perspective is as follows:

1. A game (or partial game) is played with a team of RDRAs against another opponent team. This creates a log of the game and also logs of the individual players world models from the RDRA.
2. The expert watches the replay of the game on the RDR Monitor, stopping it when it gets to a cycle where a player is behaving in a way that the expert considers to be incorrect.
3. The expert chooses the appropriate behaviour that the player should be doing at that point in time.
4. The expert must then justify why this particular case requires a different behaviour. To do this the expert must simply choose features that differentiate the current case from the cornerstone case associated with the last rule that was satisfied. These justifications are used to create the condition for the new rule and validate the addition of the new rule. Fig3.
5. The new rule is created mapping the features chosen by the expert to the appropriate behaviour, and added to the knowledge base. It is located automatically as a refinement on the rule which caused the misclassification.

At this stage one cycle of knowledge acquisition has been completed. The new knowledge base is now used for making decisions for the agent, and the cycle starts again. This process is repeated, gradually refining agents behaviours.

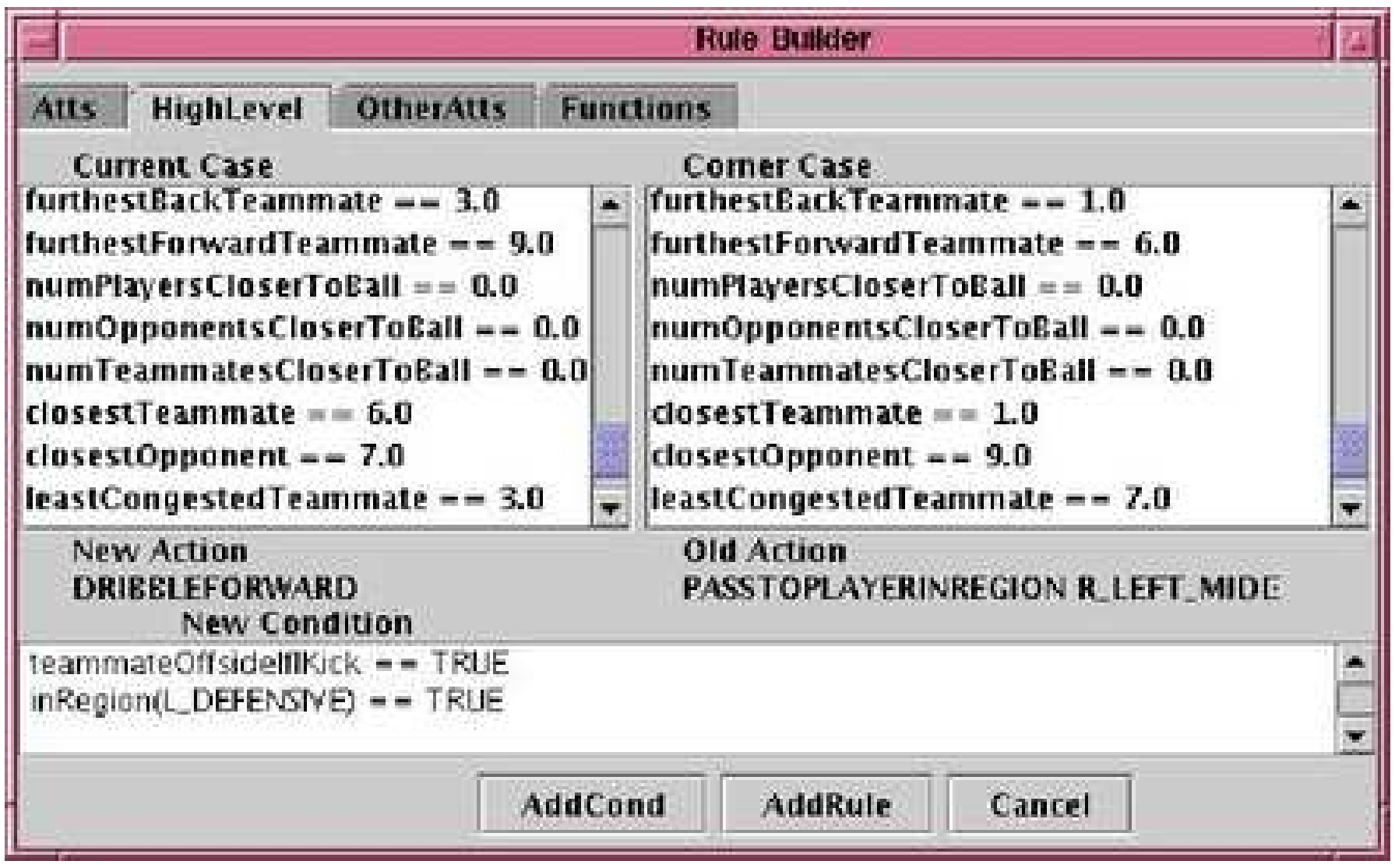

**Fig. 3.** Screen shot of the Rule Builder Window

It is important to note that at any given moment an agents world model is not accurate.Thus an agent may appear to be behaving incorrectly according to the replay, but when looking at the world-model of the agent, it may be behaving correctly according to its belief of the state of the world. Knowledge acquisition by the experts does not attempt to fix behaviours based on an incorrect world model or noise in their effectors - other than being able to control the agents body and/or neck angle by turning. Thus it is important for agents to log their world models and for these to be displayed in the user interface as well as the action chosen by the agent. This helps the user to be able to identify these different sources of incorrect behaviour, and the logs are used as the data for training.

**Validation of new rules.** In every RDR system, the expert chooses a list of relevant features from a difference list. The difference list is the set of differences between the cornerstone case and the current case that has been incorrectly classified. However the conditions for rules in this system are not all simple propositional valued attributes. So a difference list cannot be created for all possible attributes. In our current system parameterised attributes can be chosen from a list and automatic comparisons can be made with different arguments between the current case and the corner case. This helps an expert decide whether this attribute is relevant in distinguishing these cases from one another.

In future versions of the system it would be desirable to make it easier for experts to choose relevant attributes, as there are currently over 100 attributes available. The addition of a visual display of both corner stone and current case, as well as a priority list of attributes that are likely to be of interest to an expert in a given situation may potentially ease this process for the expert. The priority lists could be given default values or be created by the expert or be based on usage patterns and have the potential to dynamically change. This may help experts choose from the large number of attributes available.

## 4 Testing and Evaluation

An initial version of the system was released for testing. The aim of this was not only to test the feasibility of our incremental approach but also to get feedback about the set of features and behaviours provided for knowledge acquisition and the design of the user interface. The system was tested using four volunteers. The subjects consisted of three postgraduate students and one postdoctoral researcher in computer science. The subjects received a short half hour tutorial on how to use the system and were given a demonstration. They were then able to use the system over a period of two weeks. Ideally, we would like to test our RDRAs using real soccer coaches and plan to do so in future studies. However since the aim is to investigate the possibility of incrementally transferring knowledge to RDRAs to represent soccer strategies and team co-ordination, we considered anyone with an understanding and interest in soccer to have some level of expertise.

The users were all able to create rule bases for players and train and play against each other and teams from the Robocup simulation league [29]. One of the users was able to create a team that was able to play at a similar standard to that of the UVA Trilearn benchmark team [16]. This is the team that was used as an opponent for qualification to Robocup simulation league in 2003[29]. This is promising as the user was able to do this in a short frame of time, creating a rule base with approximately 90 rules. Each rule took on average a minute or so to add, once a cycle was found where a players behaviour was inappropriate. The knowledge base created by this user was further evaluated to measure the performance of the team at various points of the rule base creation. Each result is the score against the UVA Trilearn benchmark team. The results are shown in Table 1. Unfortunately only the results of one of the volunteers are presented as the other volunteers were unable to spend enough time on the task due to other commitments.

**Table 1.** Results - RDRA Team vs UVA Trilearn

| | Default | 10 Rules | 20 Rules | 40 Rules | Final |
|---|---|---|---|---|---|
| Game1 | 0-2 | 0-11 | 0-6 | 0-3 | 1-1 |
| Game2 | 0-3 | 0-10 | 0-7 | 0-2 | 0-1 |
| Game3 | 0-2 | 0-9 | 0-6 | 0-1 | 1-0 |
| Game4 | 0-2 | 0-11 | 0-7 | 0-3 | 0-1 |
| Game5 | 0-3 | 0-8 | 0-7 | 0-3 | 2-1 |

The results show the progression of the teams performance as more rules are added by the user to the knowledge base. The first set of results used the default rule base with the default behaviour of doing nothing. In these games, the soccer players except for the hand-coded goalie, just stood in their initial stationary positions, repositioning only in other play modes such as free kicks. It is interesting to note that this behaviour of doing nothing is somewhat of a defensive strategy as it tended to keep the other team away from the goal by exploiting the off-side rule. However this is obviously not a desired team strategy. The next set of results show the teams performance after 10 rules were added by the user to the knowledge base. This team exhibited behaviours of following the ball and repositioning. Despite the fact that this team 'looked' like they were playing a better game of soccer than the default team, this was not represented by the score line. This seemed to mainly be because the opposition team was able to move further down the field without being off-side, to score more goals, resulting in losses by a larger margin. The addition of further rules shows a reduction in goals scored by the opposition as illustrated by the results at the 20 and 40 rule mark. Finally the rule base with 91 rules shows a team that is able to score goals and perform at a similar level to the bench mark team.

## 5 Discussion and Future Work

Our results suggest that our incremental approach is promising as rules are able to be added incrementally to fine tune players performance without corrupting the previous knowledge base. However these are only preliminary results so it is not clear yet as to whether this incremental approach will always result in incremental improvement. Future work also needs to be done to see the effect of further addition of rules to see what level of performance could be reached. However much work needs to be done to allow soccer teams to be created that can perform at a competitive level in the Robocup simulation league.

One of the main problems with the system is providing the soccer coaches with an appropriate set of features and behaviours to specify the strategies they desire. The appropriate level of abstraction is needed to provide the experts with enough control without overloading them with unnecessary complexity. The current implementation provides low-level attributes as well as pre-defined higher level attributes and behaviours for the expert to use. During the evaluation stage it was identified that at times the users felt unable to express their strategies with the features and behaviours provided and future work would clearly need to expand the language provided to the expert. However the users also reported being over-whelmed by the large number of choices available in the system already. Thus future work needs to focus on not only expanding the number of features and behaviours available to the user but presenting them in a more intuitive way to reduce the frustration and cognitive load on the user in searching through long lists of suitable behaviours and features for justification. Features and behaviours that are not useful should also be identified and removed.

One way to address the issue of the large number of attributes is to provide priority lists, as discussed in Sec.3.2, to help the expert choose relevant features to differentiate the current situation from the cornerstone case. This could be developed in combination with a more graphically driven user interface which would involve displaying graphically the attributes of the cornerstone cases and the current case, allowing the user to click on relevant parts of the graphical display to select relevant features and actions. Ideally the soccer coach would be able to graphically indicate what a play should be by circling areas and drawing arrows on the field display on the monitor, similar to the interface described in [21]. These could be used to suggest conditions to the user that would satisfy these situations and behaviours to create a rule.

Attempting to comprehensively model the domain was a difficult task, so it would be desirable for the user to be able to express features in their own language and create their own higher level attributes and actions during the knowledge acquisition phase. Earlier work on NRDR[8] explored this idea and was developed to allow users to create their own higher level attributes during the knowledge acquisition phase. This would be an interesting area for future research and could involve creating a system that would allow users to create higher level actions as well as attributes.

One of the main bottle-necks of the knowledge acquisition process identified during the evaluation phase on the current process, was finding an appropriate place in the game to add a new rule. In the current system this involves watching a replay of a game to find a situation where a players' behaviour should be corrected. This has shown to be one of the time consuming parts of the knowledge acquisition process. It would be useful to provide a tool which could allow an automated search through a log of a game for cycles that are of particular interest, for example when the experts team has the ball and is in a shooting position. Other areas for future work include further exploring the approach described in [3] to automatically generate intermediate features to reduce the amount of knowledge acquisition needed by the user.

Future work should also create rule bases for the goalie and other modes of play to make a more cohesive team and also make use of the additional features of the soccer server such as the heterogeneous players. Our RDR approach could also be used for the creation of an online coach. Integration of external advice from the online coach into the RDR framework of the player would also be an interesting area to explore. Finally, the system should be evaluated more extensively and the teams created should be tested against some of the top teams in the Robocup competition.

## 6 Conclusion

Incremental KA using RDR has been comprehensively evaluated for classification problems and has been extended to other areas such as multiple-classification tasks. We developed a preliminary system for RDR agents in the soccer simulation domain. Our results suggest that it is possible to build a Robocup soccer simulation team at the user level using an incremental knowledge acquisition technique. Future work is needed to further develop the system and a more comprehensive evaluation is needed.

## Acknowledgements

We would like to thank Ken Nguyen, Victor Jauregui, Son Bao Pham and above all Rex Kwok for their invaluable contributions to this research.

# Domain Ontology Construction with Quality Refinement

Takeshi Morita[1], Yoshihiro Shigeta[1], Naoki Sugiura[1], Naoki Fukuta[1], Noriaki Izumi[2], and Takahira Yamaguchi[3]

[1] Shizuoka University, 3-5-1 Johoku, Hamamatsu, Shizuoka 432-8011, Japan,
morita@ks.cs.inf.shizuoka.ac.jp,
http://mmm.semanticweb.org

[2] National Institute of AIST, 2-41-6, Aomi, Koto-ku, Tokyo, Japan

[3] Keio University, 4-1-1 Hiyoshi, Kohoku-ku, Yokohama-shi, Kanagawa, Japan

**Abstract.** In this paper, we propose an ontology development support environment for the Semantic Web. The environment supports the user by two steps. First, an initial ontology is generated semi-automatically. Then the environment supports the user to refine the ontology by providing candidates of relationships to be added and modified. The advantage of our environment is focusing the quality refinement phase of ontology construction. Through interactive support for refining the initial ontology, OWL-Lite level ontology, which consists of taxonomic relationships (class - sub class relationship) and non-taxonomic relationships (defined as property), is constructed effectively.

## 1 Introduction

As the scale of the Web becomes huge, it is becoming more difficult to find appropriate information on it. When a user uses a search engine, there are many Web pages or Web services which are syntactically matched with user's input words but semantically incorrect and not suitable for user's intention. In order to defeat this situation, the Semantic Web [1] is now gathering attentions from researchers in wide area. Adding semantics (meta-data) to the Web contents, software agents are able to understand and even infer Web resources. To realize such paradigm, the role of ontologies [2] is important in terms of sharing common understanding among both people and software agents [3]. In knowledge engineering field ontologies have been developed for particular knowledge system mainly to reuse domain knowledge. On the other hand, for the Semantic Web, ontologies are constructed in distributed places or domain, and then mapped each other. For this purpose, it is an important task to realize a software environment for rapid construction of ontologies for each domain. Towards the on-the-fly ontology construction, many researches are focusing on automatic ontology construction from existing Web resources, such as dictionaries, by machine processing with concept extraction algorithms. However, depending on domains (a law domain etc.), the important concepts which doesn't occur frequently in the resources may be required to be added by hand for ontology construction. In such a domain, if a user doesn't intervene, constructing ontologies cannot readily be done. Considering such situation, we believe that the most important aspect of the on-the-fly ontology construction is that how efficiently the user is able to complete making the ontology for the Semantic Web contents available to the public. For this reason, ontologies should be constructed not fully automatically, but through interactive support by software environment from the early stage of ontology construction. Although it may seem to be contradiction in terms of efficiency, the total cost of ontology construction would become less than automatic construction because if the ontology is constructed with careful interaction between the system and the user, less miss-construction will be happened. It also means that high-quality ontology would be constructed.

In this paper, we propose a software environment for user-centered on-the-fly ontology construction named DODDLE-R (Domain Ontology rapiD DeveLopment Environment - RDF [4] extension). The architecture of DODDLE-R is re-designed based on DODDLE-II [5], the former version of DODDLE-R. DODDLE-R has the following five modules: Input Module, Construction Module, Refinement Module, Visualization Module, and Translation Module. Especially, to realize the user-centered environment, DODDLE-R dedicates to the Refinement Module. It enables us to develop ontologies with interactive indication of which part of ontology should be refined. DODDLE-R supports the construction of both taxonomic relationships and non-taxonomic relationships in ontologies. In addition to co-occurrency based methods, we exploit learning concept pairs by syntactic analysis and concept abstraction by using domain-specific taxonomy. We assumed that it was possible to construct more refined ontologies by combining these indicators and experimented. Additionally, because DODDLE-II has been built for ontology construction not for the Semantic Web but for typical knowledge systems, it needs some extensions for the Semantic Web such as OWL (Web Ontology Language) [6] export facility. DODDLE-R contributes the evolution of ontology construction and the Semantic Web.

## 2 The DODDLE-R Architecture

Figure 1 shows the overview of DODDLE-R. The main feature of DODDLE-R has the following five modules: Input Module, Construction Module, Refinement Module, Visualization Module, and Translation Module. Input Module, Hierarchy Construction Module, and Hierarchy Refinement Module are included in DODDLE-I to support a user to construct taxonomic relationship. Relationship Construction Module and Relationship Refinement Module were added on DODDLE-II development that supports constructing both taxonomic and non-taxonomic relationships. Visualization Module and Translation Module are newly developed in DODDLE-R to extend functionalities for the Semantic Web such as exporting constructed ontology in OWL format. First, the user selects some terms as the input in Input Module. In Construction Module, DODDLE-R generates the basis of the ontology, an initial concept hierarchy and set of concept pairs, by referring to WordNet [7] as an MRD (Machine Readable Dic-tionary) and documents. An initial concept hierarchy is constructed as an IS-A hierarchy of terms. Set of concept pairs are extracted by using co-occurrency based statistic methods. These pairs are considered to be closely related and that will be used as candidates to refine and add non-taxonomic relations. The user identifies some relationship between concepts in the pairs. In Refinement Module, the initial ontology produced by Construction Module is refined by the user through interactive support by DODDLE-R. In order to refine the initial ontology, we manage concept drift and evaluate set of concept pairs. Because the initial concept hierarchy is constructed from a general ontology, we need to adjust the initial concept hierarchy to the specific domain considering an issue called Concept Drift. It means that the position of particular concepts changes depending on the domain. For concept drift management, DODDLE-R applies two strategies: Matched Result Analysis and Trimmed Result Analysis. These strategies are described in our former study [5]. At the construction phase of concept specification template from set of concept pairs generated by Construction Module, DODDLE-R needs a criterion to evaluate significant concept pairs. In [5], two statistics based methods are investigated: the value of context similarity by WordSpace method [8] and the value of confidence by the association rule learner [9]. Those methods and values based on co-occurrency of concepts work well in terms of wide use (do not depend on some particular domains), its cost of preparation. In addition to co-occurrency, in this paper, we investigate two learning algorithms to get concept pairs based on syntactic analysis and concept abstraction by using domain-specific taxonomy. We describe these methods below. The ontology constructed by DODDLE-R can be exported with the representation of OWL. Finally, $MR^3$ (Meta-Model Management based on RDF(S) [10] Revision Reflection) [11] is connected with DODDLE-R and works with an RDF(S) graphical editor.

### 2.1 Learning Concept Pairs by Syntactic Analysis

In the set of statistically generated concept pairs, some of them are existing as a pair of concepts in noun phrase and verb phrase in terms of structure of a sentence. DODDLE-R adds extra value to them for the evaluation of significant concept pairs. The relation value of concept pair (consists of concept $x$ and concept $y$) in noun and verb phrase ($nvrelation(x, y)$) is calculated based on frequency of appearance in a text corpus(formula 1):

$$nvrelation(x, y) = \frac{nvfrequency(x, y)}{\sum nvpair} \tag{1}$$

$nvfrequency(x, y)$ is a frequency of concept $x$ and concept $y$ to appear in a pair of noun and verb phrase. Let $nvpair$ be the number of all concept pairs which appear in a noun and verb pair in a text corpus.

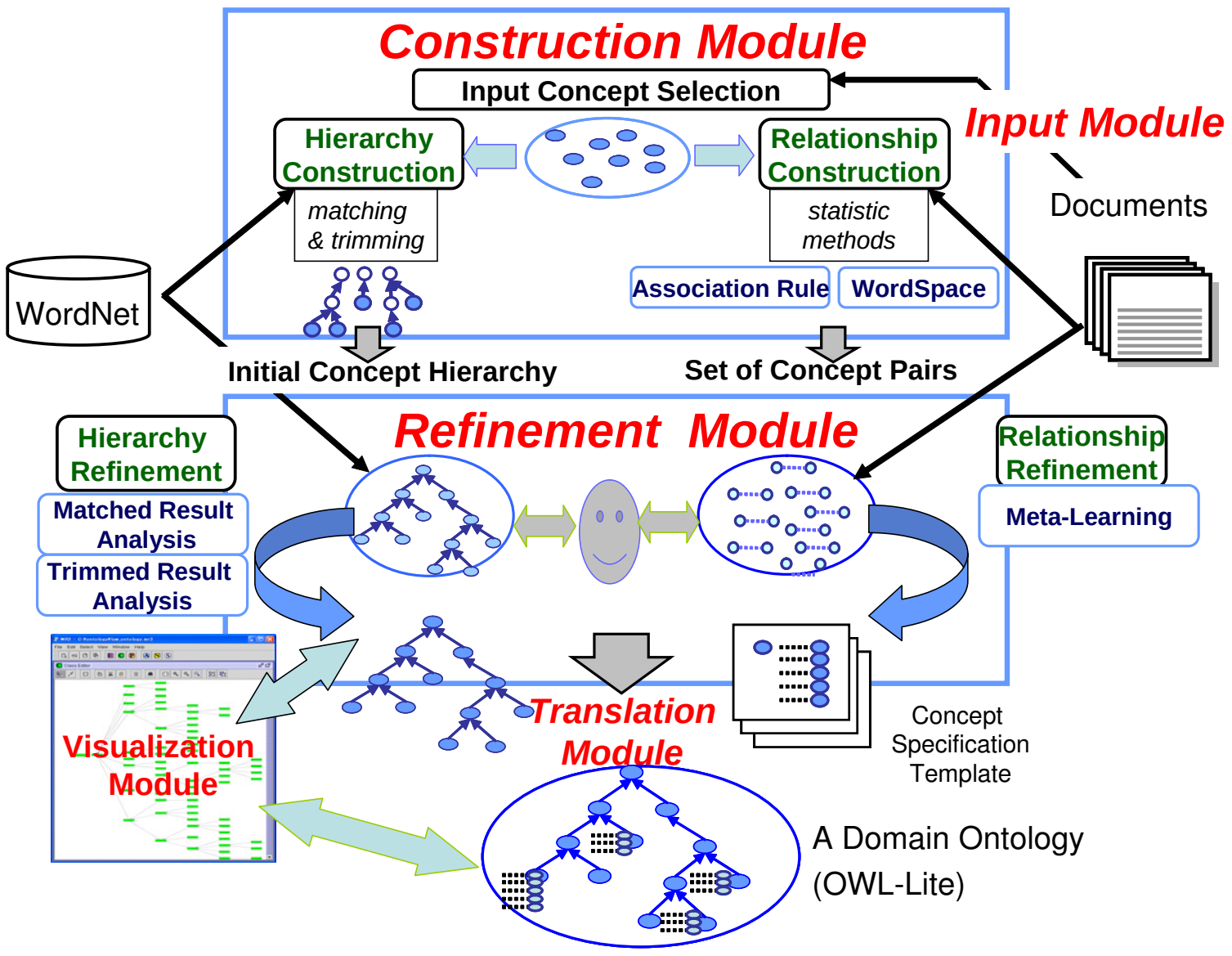

**Fig. 1.** DODDLE-R overview

### 2.2 Concept Abstraction by Using Domain-Specific Taxonomy

In the domain specific text corpus, there are some concepts which have not been defined as the input concepts, but more or less related to them. Figure 2 shows an example of such situation. There is a concept "buyer" in the text corpus, and "person" in the domain specific taxonomy which has already been constructed by the user at this point. By referring to an MRD (WordNet), we can see that "buyer" is a kind of "person" in common sense. We believe that it is important to involve such concepts in the text corpus because it means to consider "common" knowledge which is usually not focused on the domain ontology construction. The same can be said to the concept "seller".

Figure 3 shows the abstraction from a concept in the text corpus to an input domain concept. There are input concepts (concept A and concept B). A concept X, located in a sentence in the text corpus, has the concept A and concept B as the upper concept. In such case, DODDLE-R abstracts the concept X to the concept B (more specific concept than the concept A).

## 3 Implementation

In this section, we describe the system architecture from the aspect of system implementation. DODDLE-R is realized in conjunction with $MR^3$ [11]. $MR^3$ is an RDF(S) graphical editor with meta-model management facility such as consistency checking of classes and a model in which these classes are used as the type of instances. Figure 4 shows the relationship between DODDLE-R and $MR^3$ in terms of system implementation. Both $MR^3$ and DODDLE-R are implemented in Java language (works on Java 2 or higher). $MR^3$ is implemented using JGraph [12] for RDF(S) graph visualization, and Jena 2 Semantic Web Framework [13] for enabling the use of Semantic Web standards such as RDF, RDFS, N-triple and OWL. By using these libraries, $MR^3$ is implemented as an environment for graphical representation of the Semantic Web contents. Additionally, $MR^3$ also has plug-in facility to extend its functionality.

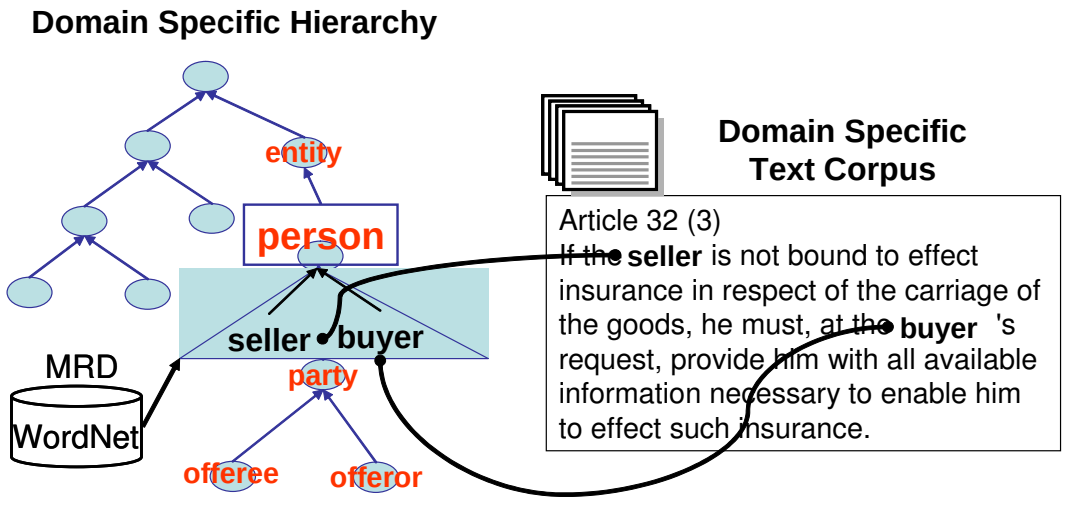

**Fig. 2.** An example of the concept abstraction.

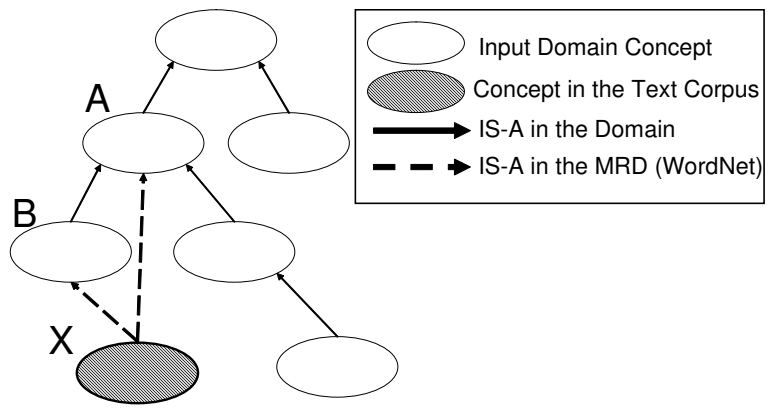

**Fig. 3.** The abstraction from a concept in the domain specific text corpus to a concep in the domain specific taxonomy

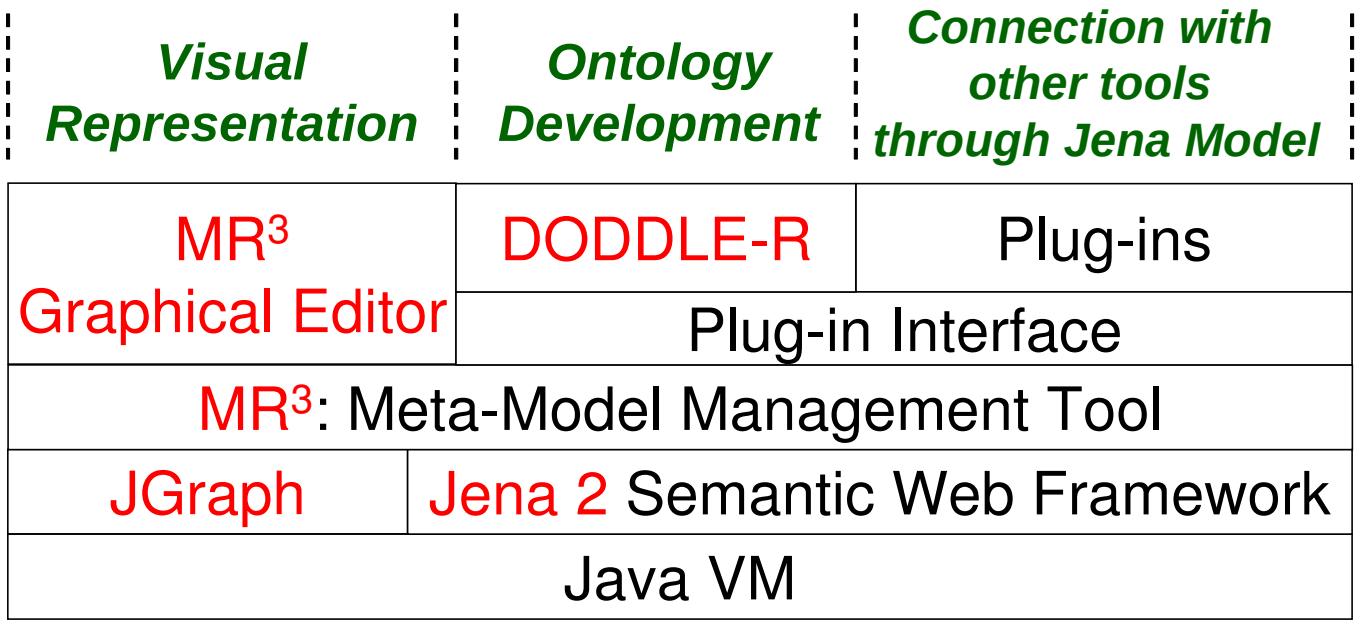

**Fig. 4.** DODDLE-R architecture

Figure 5 shows a typical usage of DODDLE-R. DODDLE-R's user interface consists of Input Module, Construction & Refinement Modules for Hierarchy, Construction & Refinement Modules for Relationships, Visualization Module $MR^3$ , and Translation Module into OWL-Lite. First, the user selects some terms as the input in Input Module ((1) in Figure 5). As input of DODDLE-R, the user associates those terms with concepts by referring "the WordNet concepts" in (1) of Figure 5. For example, the user decide which "concept" (i.e. synset in WordNet) is suitable for the term "party". By referring to the synset and term's definition, the user selects an appropriate concept for the word "party".

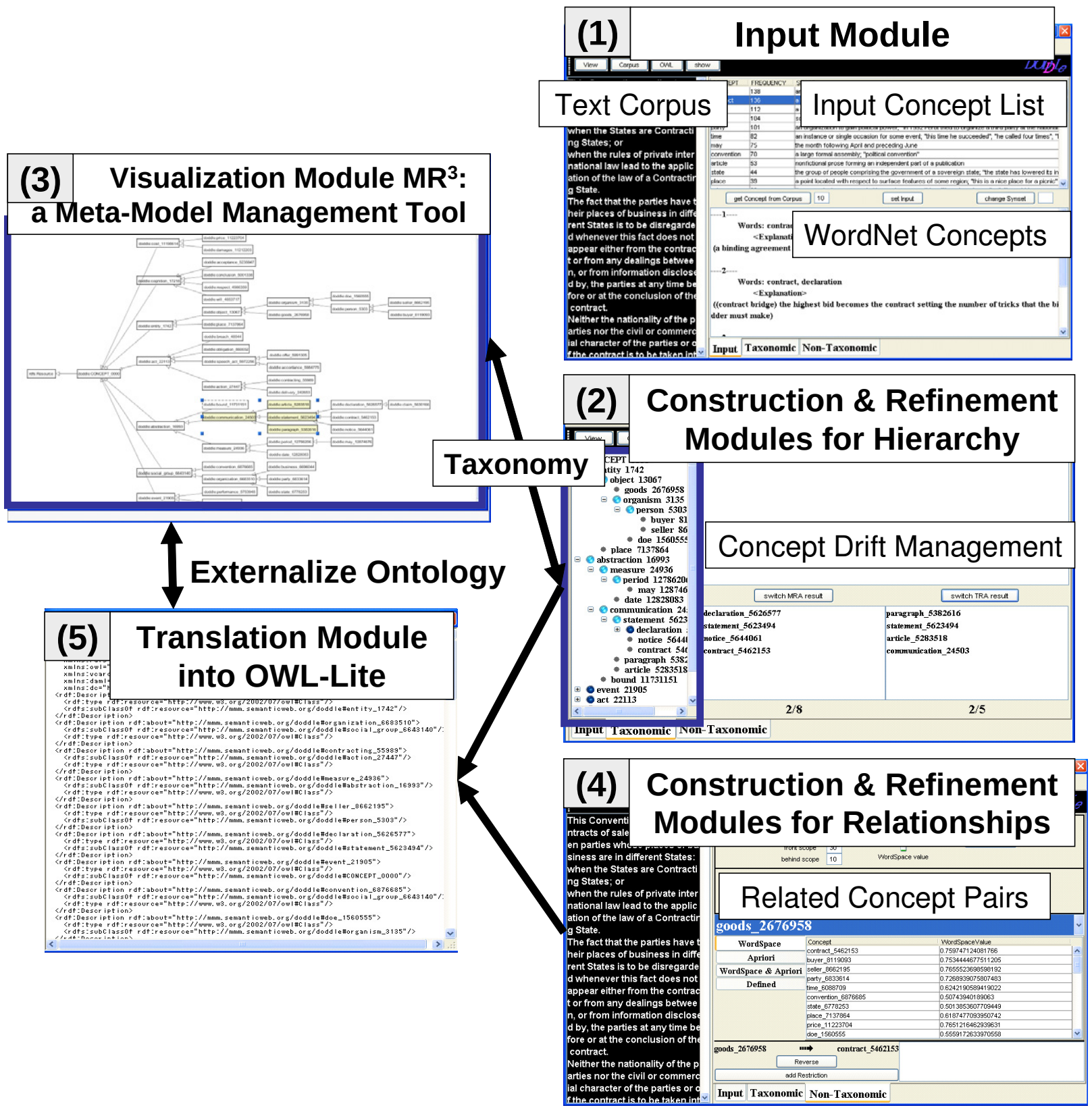

**Fig. 5.** A typical usage of DODDLE-R

After mapping terms and their synsets, an initial concept hierarchy is produced. Also set of concept pairs are extracted by statistic methods such as WordSpace method and the association rule learner by default parameter. (2) of Figure 5 shows the Construction & Refinement Modules for Hierarchy. This module indicates some groups of concepts in the taxonomy so that the user can decide which part should be refined. (3) of Figure 5 shows the display of concept drift management in Visualization Module $MR^3$ . (4) of Figure 5 shows Construction & Refinement Modules for Relationships. This module is used for setting parameters used in the WordSpace method and the association rule learner to apply to documents in order to generate significantly related concept pairs. In WordSpace method, there are parameters such as the gram number (default gram number is four), minimum N-gram count (to extract high-frequency grams only), front scope and behind scope in the text. In the association rule learner, minimum confidence and minimum support are set by the user. As a result, the user got a domain ontology as (5) in Figure 5.

## 4 Case Studies in a Law Domain

In order to evaluate the two learning algorithms proposed in section 2.1 and 2.2, we have done an experiment in the particular law field, called "the Contracts for the International Sale of Goods" (CISG) [14]. CISG document contains some long noun phrase and verb phrase in terms of structure of a sentence and some of significant concept pairs are existing in the noun phrase and verb phrase. The text of the contract consists of approximately 10,000 words. This experiment has been done by using the domain specific taxonomy which had been constructed in our former study [5]. In this paper, we focus on concept evaluation phase in Refinement Module. $value_{base}(x,y)$ is the criterion to rank significant concept pairs in DODDLE-II. $value_{base}(x,y)$ is the average of "context similarity" by WordSpace method and "confidence" by the association rule learner. Details about WordSpace method and association rule learner are explained in [5]. In this case study, we prepare $value_{nv}(x,y)$ and $value_{nv+abst}(x,y)$ as new criteria. $value_{nv}(x,y)$ is added the effect of noun phrase-verb phrase relation value (nvrelation(x,y)) to $value_{base}(x,y)$. $value_{nv+abst}(x,y)$ is added the effect of $value_{abst}$ ($value_{nv}(x,y)$ using domain specific taxonomy) to $value_{nv}(x,y)$, considering the abstraction of concepts. We prepare $value_{nv+abst}(x,y)$ in order to evaluate how the concept abstraction by using domain-specific taxonomy is useful to extract significant concept pairs. The numerical formulas of $value_{nv}(x,y)$ and $value_{nv+abst}(x,y)$ are shown as below.

$$value_{nv}(x,y) = \frac{1}{2}(value_{base}(x,y) + nvrelation(x,y)) \tag{2}$$

$$value_{nv+abst}(x,y) = \frac{1}{3}(2 \cdot value_{nv} + value_{abst}(x,y)) \tag{3}$$

As the case study, we have calculated the precision and the recall. In this case study, DODDLE-R sorts the list of concept pairs by the descending order of each value, and extracts concept pairs from the top of the list. The solution set of significant concept pairs is 41 pairs which have been selected by domain experts in advance. If the precision and the recall marks high score in the early stage of the concept pair extraction (left side of each graph), it means the method is efficient. Figure 6 describes the precision of extracted concept pairs. As in the graph, $value_{nv}$ is better than $value_{base}$. It means that the information of syntactic analysis (the relationship between noun phrase and verb phrase) works appropriately. On the other hand, in the case of using co-occurrency, syntactic analysis and domain specific taxonomy ($value_{nv+abst}$), overall result is lower than $value_{nv}$. Figure 7 shows the recall of extracted concept pairs. Especially in the early stage, $value_{nv}$ and $value_{nv+abst}$ is higher than $value_{base}$, and $value_{nv}$ is slightly higher than $value_{nv+abst}$. Entirely, $value_{nv+abst}$ marked low score regardless of our intention to involve lower concepts in the text corpus. In this experiment, there is bias in some abstraction from concepts in the text corpus to input concepts in the domain specific taxonomy. Some abstraction to particular concepts occurred much more than others, and it caused explosion of incorrect concept pairs. However, the overall score of $value_{nv}$ is higher than others because the syntactic analysis worked well.

According to the experimental results, it is important to select combinations of algorithms to get better candidate of concept pairs. In order to construct the sophisticated domain ontology, the user should combine appropriate algorithms for the ontology. If DODDLE-R provides various kinds of algorithms and the presentation of the combination possibility of algorithms, and can combine algorithms easily in Refinement Module, DODDLE-R will support much more parts on constructing more sophisticated ontologies.

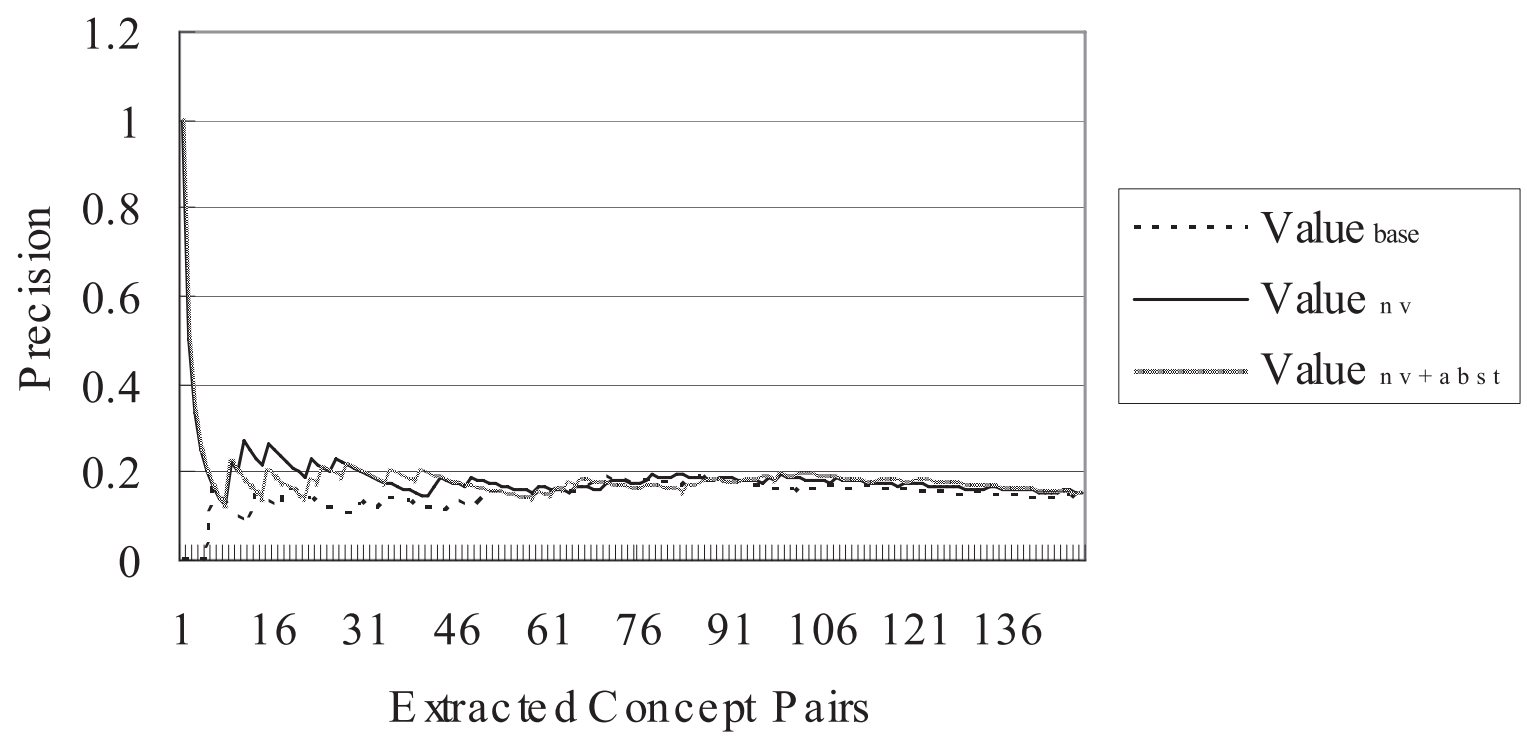

**Fig. 6.** The transition of precision

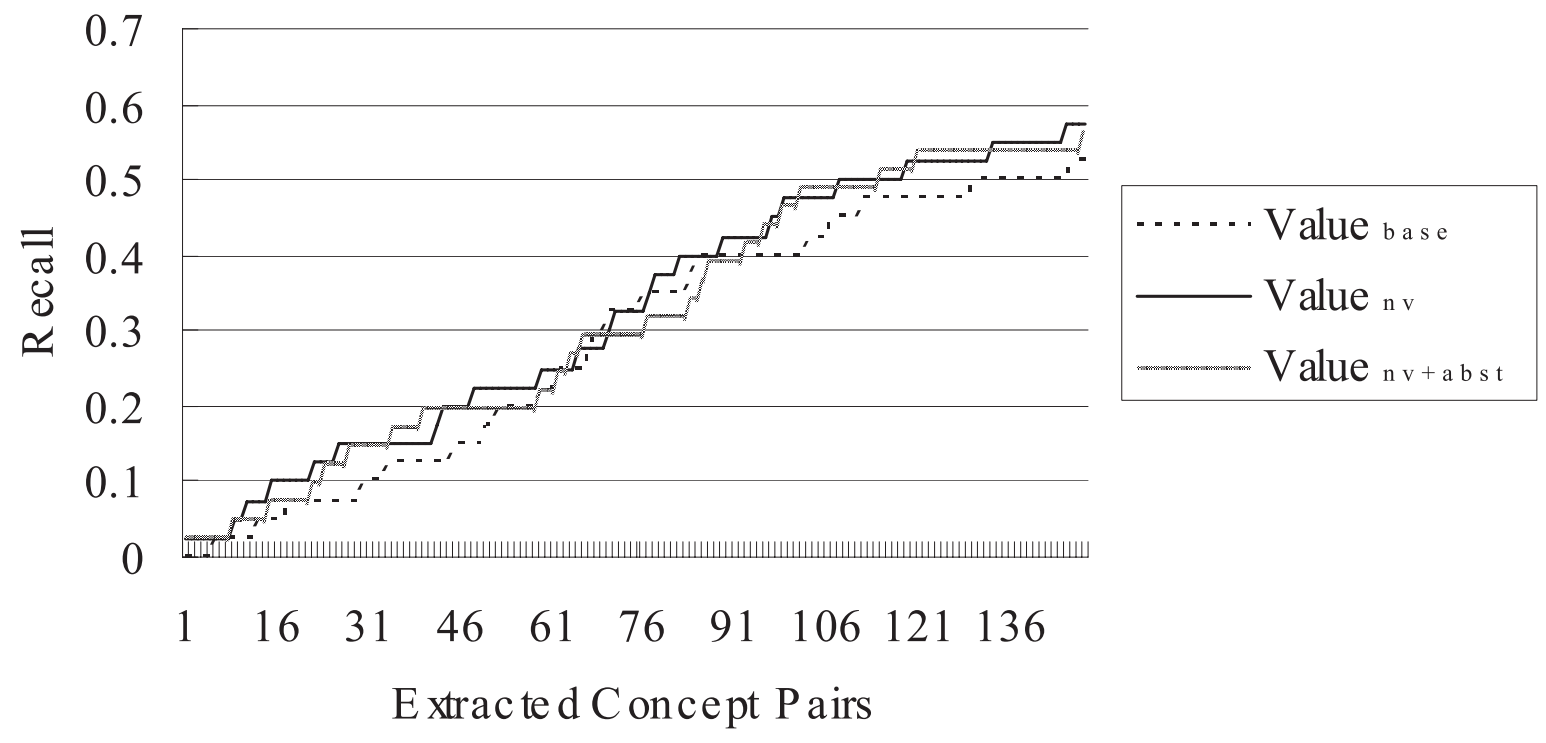

**Fig. 7.** The transition of recall

## 5 Related Work

Navigli et,al. proposed OntoLearn [15], that supports domain ontology construction by using existing ontologies and natural language processing techniques. In their approach, existing concepts from WordNet are enriched and pruned to fit the domain concepts by using NLP (Natural Language Processing) techniques. They argue that the automatically constructed ontologies are practically usable in the case study of a terminology translation application. However, they did not show any evaluations of the generated ontologies themselves that might be done by domain experts. Although a lot of useful information is in the machine readable dictionaries and documents in the application domain, some essential concepts and knowledge are still in the minds of domain experts. We did not generate the ontologies themselves automatically, but suggests relevant alternatives to the human experts interactively while the experts' construction of domain ontologies. In another case study [16], we had an experience that even if the concepts are in the MRD (Machine Readable Dictionary), they are not sufficient to use. In the case study, some parts of hierarchical relations are counterchanged between the generic ontology (WordNet) and the domain ontology, which are called "Concept Drift". In that case, presenting automatically generated ontology that contains concept drifts may cause confusion of domain experts. We argue that the initiative should be kept not on the machine, but on the hand of the domain experts at the domain ontology construction phase. This is the difference between our approach and Navigli's. Our human-centered approach enabled us to cooperate with human experts tightly.

From the technological viewpoint, there are two different related research areas. In the research using verb-oriented method, the relation of a verb and nouns modified with it is described, and the concept definition is constructed from this information (e.g. [17]). In [18], taxonomic relationships and Subcategorization Frame of verbs (SF) are extracted from technical texts using a machine learning method. The nouns in two or more kinds of different SF with the same frame-name and slot-name are gathered as one concept, base class. And ontology with only taxonomic relationships is built by carrying out clustering of the base class further. Moreover, in parallel, Restriction of Selection (RS) which is slot-value in SF is also replaced with the concept with which it is satisfied instantiated SF. However, proper evaluation is not yet done. Since SF represents the syntactic relationships between verb and noun, the step for the conversion to non-taxonomic relationships is necessary.

On the other hand, in ontology learning using data-mining method, discovering non-taxonomic relationships using an association rule algorithm is proposed by [19]. They extract concept pairs based on the modification information between terms selected with parsing, and made the concept pairs a transaction.

By using heuristics with shallow text processing, the generation of a transaction more reflects the syntax of texts. Moreover, RLA, which is their original learning accuracy of non-taxonomic relationships using the existing taxonomic relations, is proposed. The concept pair extraction method in our paper does not need parsing, and it can also run off context similarity between the terms appeared apart each other in texts or not mediated by the same verb.

## 6 Conclusion

In this paper, we presented a support environment for ontology construction named DODDLE-R, which aims at a total support environment for user-centered on-the-fly ontology construction. Its main principle is that high-level support for users through interaction. First, the user selects some terms as the input in Input Module. Then, Construction Module generates the basis of ontology in the forms of an initial concept hierarchy and set of concept pairs, by referring to WordNet as an MRD and a document. Refinement Module provides management facilities for concept drift in the taxonomy and identifying significant set of concept pairs in extracted related concept pairs. In this paper, we investigate two learning algorithms to get concept pairs based on syntactic analysis and concept abstruction by using domain-specific taxonomy. We assumed that it was possible to construct more refined ontologies by combining these indicators and experimented. According to the experimental results, it is important to select combinations of algorithms to get better candidate of set of concept pairs. Finally, Translation Module produces an OWL-Lite file, which is able to put on public as a Semantic Web ontology.

We think that meta-learning scheme can be applied to Refinement Module of DODDLE-R. CAMLET [20], a constructive meta-learning scheme has been proposed that can reconstruct learning algorithms from method level. This approach will help to determine which learning algorithm to use on extracting set of concept pairs on each domain.

# Developing an Intelligent Learning Tool for Knowledge Acquisition on Problem-based Discussion

Akcell Chiang, Isaac Pak-Wah Fung, R. H. Kemp

Information Sciences and Technology, Massey University,
Private Bag 11222, Palmerston North, New Zealand
{C.C.Chiang, P.W.Fung, R.Kemp}@massey.ac.nz

**Abstract.** In this paper, we report MALESAbrain, a model built on the notions of threshold and knowledge-weight from the discipline of machine learning in building up an intelligent supporting tool for a group of learners. It requires participants to judge or criticize the solutions posted by others on the forum before exploring or chatting further knowledge-content. The system then sums up the judgment scores as its knowledge-weight to pass the thresholds set up for ranking/arranging the learning issues. This constraint design for judgment, therefore, becomes a mechanism for critical thinking. It helps transform forum and chat room into a learning discussion platform for cooperative learning and knowledge acquisition.

## 1 Introduction

This paper presents a problem-based learning (PBL) discipline model, developed by the authors, to facilitate the process of cooperative knowledge-building, obtained from several people learning together. The model integrates notions from education of critical thinking [1] and machine learning [2], and has been embedded in a problem-based learning system. Accordingly, the "critical thinking" design of the model takes into account different perspectives of the issue. The "machine learning" design of the model then contributes to help the participants to approach a commonly agreed solution [3].

Discussion-based Internet forums or interactive chat rooms are effective help systems for on-line participant problem-solving. Many companies, particularly computer venders [4-6], adopt this technology to provide product training, learning or Q&A on their web sites. Nevertheless, the current state of chat room technology only provides a platform for information exchange and organization. It is still unable to stimulate the users to learn from looking at the problem from different perspectives. This paper pushes the edge of this technology further by incorporating critical thinking stimulation in teamwork discussions. We claim that such a method would sharpen knowledge acquisition in general and web-based educational services in particular. Separating the discussion topic from instant chat would further enhance the system into a comprehensive problem-based learning environment.

Problem-based learning [7] and critical thinking [1] skills have been used widely in learning education. These features are particularly noticeable in nursing education [8, 9]. This problem-based discipline can also apply to vocational education programs, such as, computer troubleshooting, which could be taught as a case study through Internet discussion. As we know, a trainee in a professional discipline, such as a computer technician, is more than just a passive learner but also an active problem-solver in real world situation. Therefore, if the problem is encountered first in the learning process and serves as a focus or stimulus for the development of problem solving or reasoning skills; it would give user-centered learners more control over their learning. They would appreciate this learning discussion and embrace this added responsibility [8].

In the on-line learning workshop, participants play the role of problem-solvers and the educator plays the role of facilitator or tutor. The educator needs only to constrain the participants' discussions within the learning domain. During learning, the educator can observe the on-line learning progression. In the role of facilitator/tutor, the educator may, on occasion, post a guiding problem, or post a hint-suggestion or even interactively join chat room discussion. At the end, the educator can obtain the participants' knowledge from the individuals' contributions.

## 2. The Model design for Problem-based Discussion

We have developed an intelligent tool, name MALESAbrain[1], to help participants think critically and learn topics like computer troubleshooting through an Internet workshop. The designed model takes an active role in sharpening the participants' contributions towards for viewpoints on the discussion issues. In discussion, the tool would highlight the importance of those issues which help the participants pay more attention to consensus solutions for better discussion and problem solution. The model consists of three main stages (Fig. 1) to facilitate participants in problem-based discussion:

i) "Critical thinking" [8]. This stage stimulates the participants to think about alternative aspects of the problem. They need to judge others' posted solutions by
   - giving personal preference or judgment on solutions posted by others;
   - contributing personal problem-solution suggestions for the feedback of preference from the judgments of others.
ii) Pay "attention" [10] to important issues.
   - Participants need to pay attention and think about why certain issues accumulate higher scores than others.
   - The highly-scored issues are highlighted by the system to stimulate more discussions on them.
   - Those extensively discussed issues therefore end up with more meaningful contents to help solve the problems.

---

[1] The acronym for "**Ma**chine-**L**earning-**E**xpert-**S**ystem **A**lgorithm for brainstorming"

iii) Improve the "learning-rate" [11] of the discussion-problem.

- An indicator shown in the learning tool helps the educator and participants perceive what percentage of the discussion-problems has resulted in consensus.
- one figure indicates how many discussion-problems did not result in consensus.
- another figure indicates how many discussion-problems did result in consensus results

The learning-rate includes a percentage indicator and two figures, which enable the educator and participants to understand the progress of the discussion in the workshop during learning.

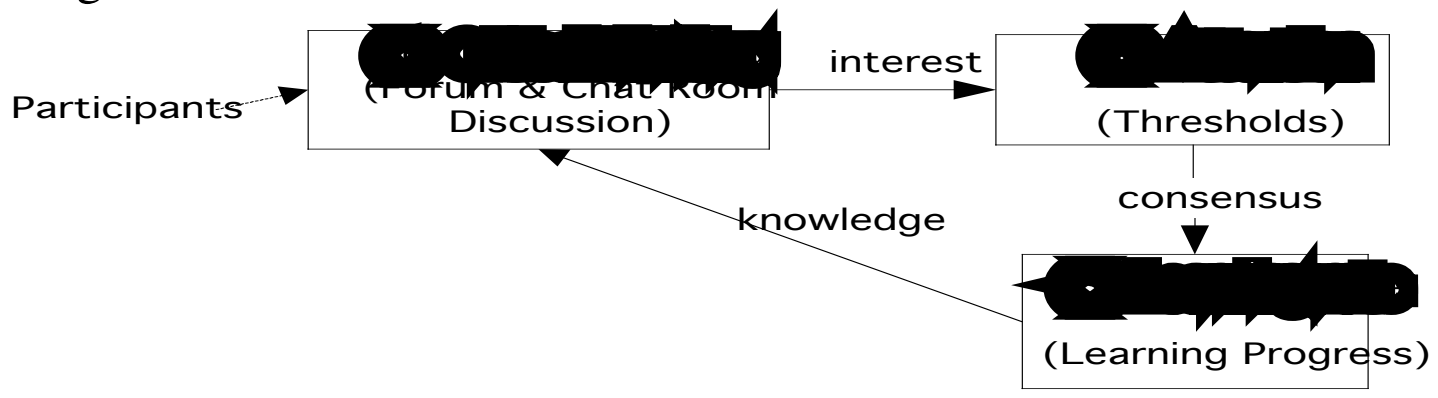

**Fig. 1.** The design model

Conceptually, the model is designed as an iterative three-stage cycle (Fig. 1) and the participants enter the cycle from the stage of critical thinking. In the *critical thinking stage*, participants post their problems and solutions as issues for discussion and judge others' problem-solution suggestions with their personal preference – a continuous value from –1 to +1. In the *attention stage*, the system will sum up the total participants' preferences with respect to the discussion issue and then use its learning thresholds to judge the sum-up-preference by:

i) ordering the importance of the issues according to the sum-up-preference
ii) highlighting or deleting the discussion-issue according to the sum-up-preference score

Finally, in the *learning-rate stage*, the system monitors the percentage of the discussion-issues that result in consensus and decides when to stop further discussion.

## 3. Using MALESAbrain on Problem-based Discussion

In this section, we use an example to explain problem-based discussion guided by MALESAbrain.

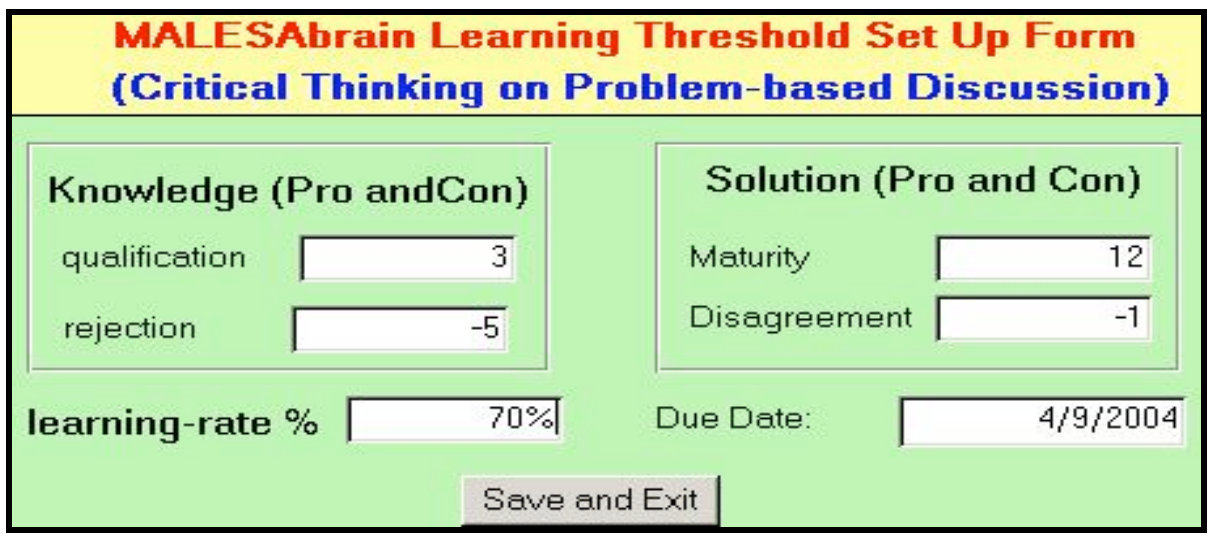

Fig. 2. Educator set up learning thresholds.

Before discussion, the educator needs to set up learning thresholds to enable MALESAbrain to recognize the importance of the discussion-issues. In Fig. 2, the educator has set up “3” as *knowledge qualification-threshold*, when knowledge-weight ≥ 3 then the knowledge becomes a qualified knowledge, which is the minimum requirement to join the competition for promotion to a higher order of discussion position; “-5” is a *knowledge rejection-threshold*, when knowledge-weight < *-5* MALESAbrain would then delete the knowledge; “12” is a *solution-maturity threshold*, when the solution-weight ≥ 12, then MALESAbrain would consider this solution is able to solve a discussion-problem; “-1” is a *solution-disagreement threshold*, if solution-weight < -1 then MALESAbrain would delete the solution. In the meantime, the educator also needs to set up *learning-rate* “70%” and due date “4/9/2004” to help MALESAbrain to understand when to suggest the stop discussion.

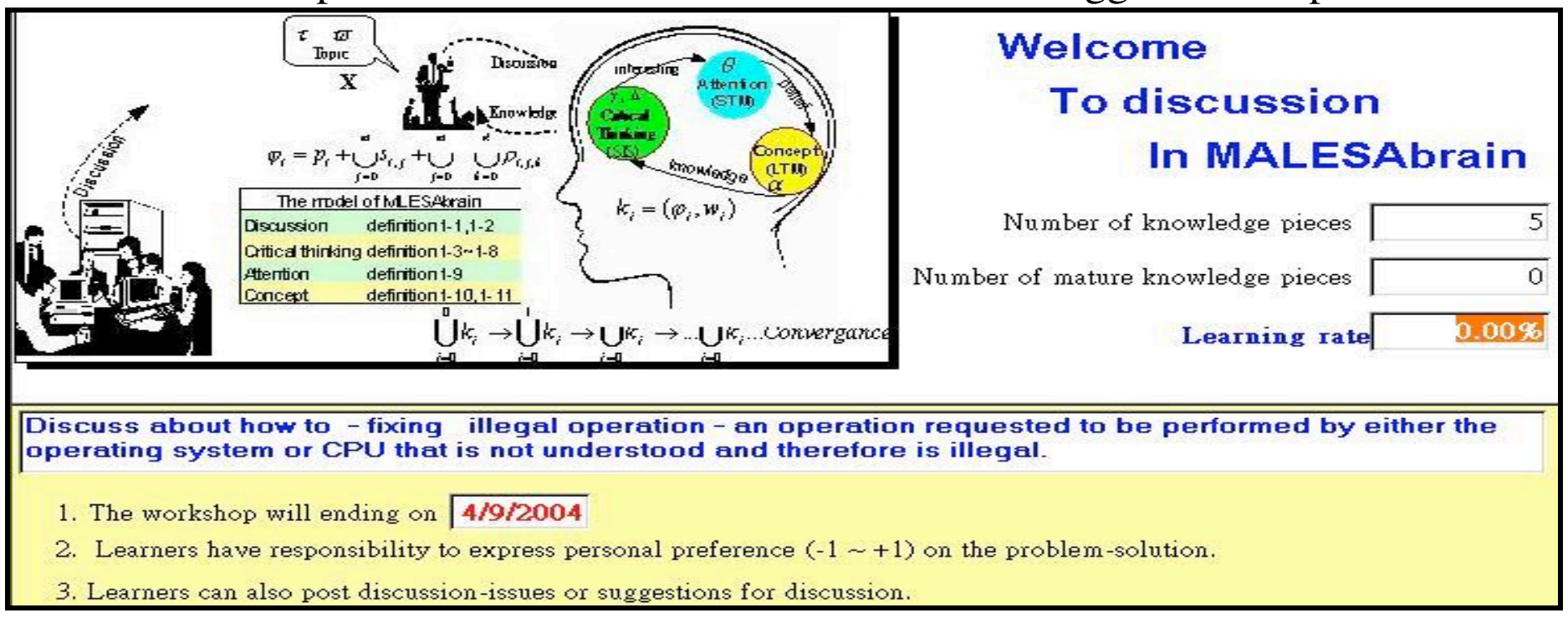

Fig. 3. Welcome page encourages participants to follow the learning rules for discussion

After the educator set up learning thresholds, learners can start their discussion. As shown in Fig. 3, MALESAbrain keeps assessing the learning-rate. During discussion, there are three figures shown on the welcome page, which help the educator to assess the current retained number of knowledge pieces “5”, the current matured number of knowledge pieces “0” and the current learning rate “0%”. They help the educator to decide whether to involve discussion, change the discussion-domain or to stop the discussion. Nevertheless, at the due date “4/9/2004”, if the learning-rate is still lower than the setup learning-rate 70% then the educator will need to decide whether to extend the workshop. The educator may calibrate the learning factors by re-setting the *knowledge qualification-threshold, knowledge rejection-threshold*, *solution-maturity threshold*, *solution-disagreement threshold* or *learning-rate* (see Fig. 2) and arrange another discussion.

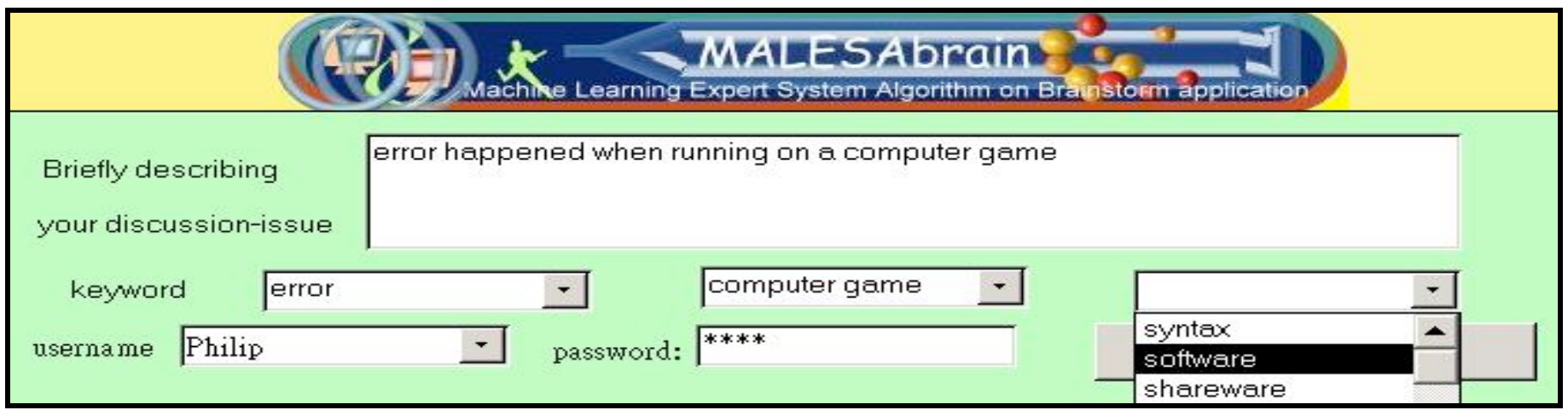

Fig. 4. A participant enters his discussion-issue for joining MALESAbrain discussion.

The Fig. 4 shows a participant enters his/her problem descriptions by completing a 'question enquiry form'. Once submitted, the query will be matched with MALESAbrain's retained knowledge according to the chosen keywords. Firstly the query will be matched by the keywords. If none is found, it will be seconded to match the problem description details. As a result, the location of the best-matched knowledge will be output where the participant is being advised to advance to.

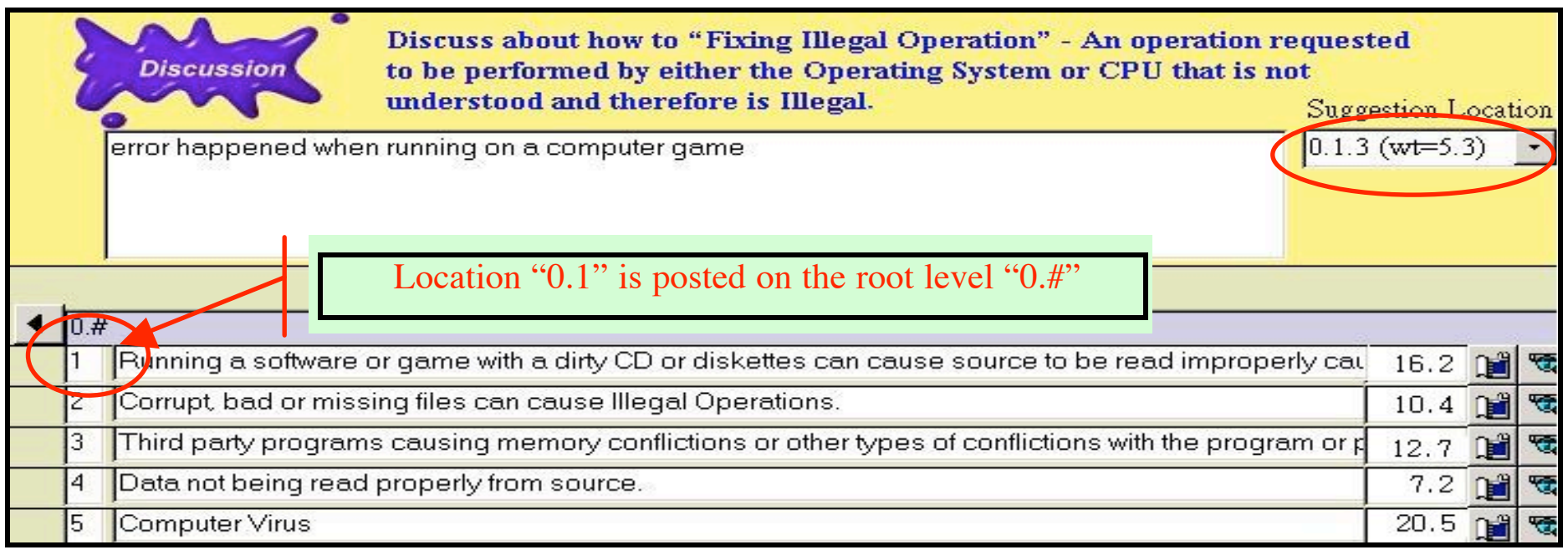

Fig. 5. After matching, the system suggests some locations for participants to discussion.

In Fig. 5, the system suggests some locations for learners to join discussion. The suggestion location "0.1.3"[2] weight "5.3" is on the lower/next level of location "0.1". Learners can click location "0.1" to browse next level "0.1.#" which includes location "0.1.3" or directly click suggestion location "0.1.3".

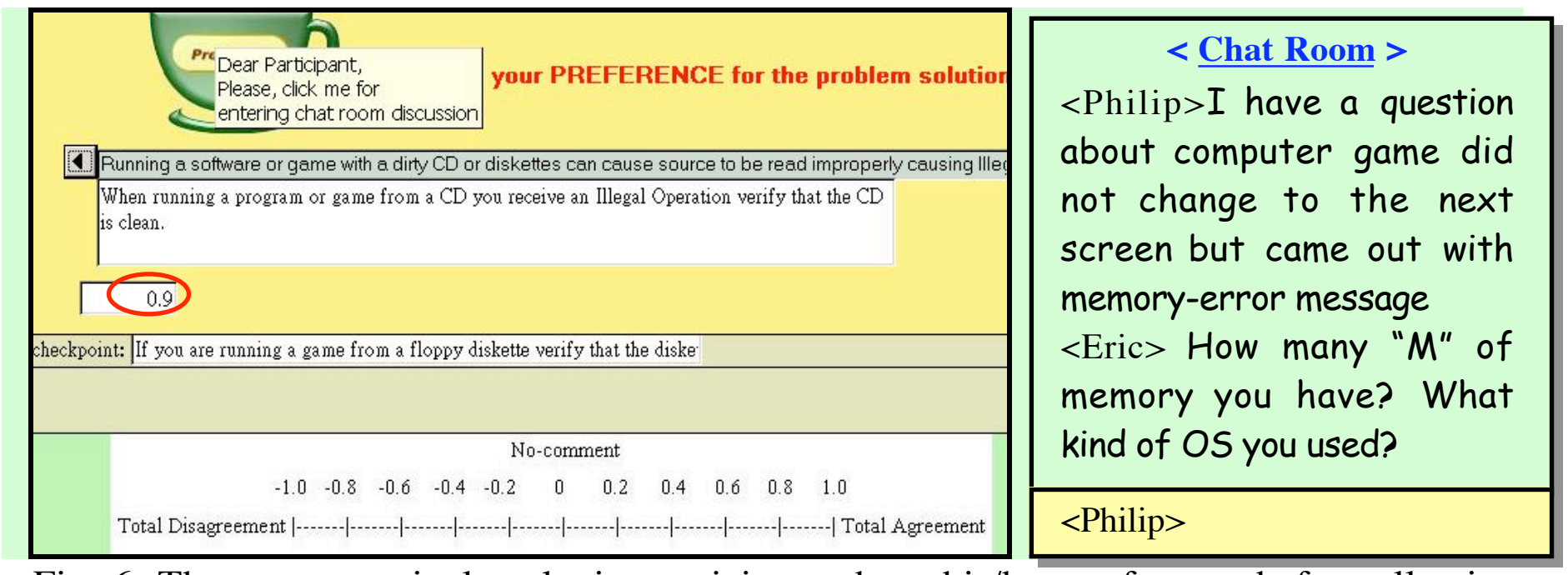

Fig. 6. The system actively asks its participant about his/her preference before allowing chat

During exploration, in Fig. 6, MALESAbrain actively questions the participants about their *preferences* - a numerical measure of the participant's degree of support (or not support) for a posted-solution to the problem. Participants should answer these questions prior to moving on to the next piece of content or chat room for discussion. The preference value ranges from 1 for total agreement, to 0 for no comment, and to -1 for total disagreement. Such a device provides a window of opportunity for individual participants to review other problems from different perspectives and subjectively evaluate the advantages and disadvantages of other people's works. Participants must judge or criticize another's proposed solution, for its ability to solve a problem. It is an important mechanism installed on the model, which encourages participants to critically think about a problem-solution from others' suggestions and carefully judge their own preference scores. In Fig. 6, the system asks its participant about his/her preference. The participant then express his/her preference score "0.9" from his own judgment before being allowed to enter chat room for discussion.

---

[2] is an example of a location address - about different levels separated with "."
Like the "dot" in Internet address, "0.1.3" is an address in MALESAbrain. "0" is the root level address a learner must choose 1 to go to "0.1" and then choose 3 to go to "0.1.3".

There is one important operation behinds Fig 7, which needs to be described. The knowledge base in MALESAbrain is the form of decision tree based on participants' preference judgments. A hierarchical structure of knowledge ranking from the knowledge-weights has been designed and based on the data structure AK-cell (see definition 4). The learning system fixes the number of knowledge-pieces on each level (see definition 5). MALESAbrain then sums up all solution-weights attached to a problem as the knowledge-weight (see definition 3). The knowledge-weights will then be compared with the set up learning thresholds and compete with other knowledge for promotion to the upper level (see definition 6).

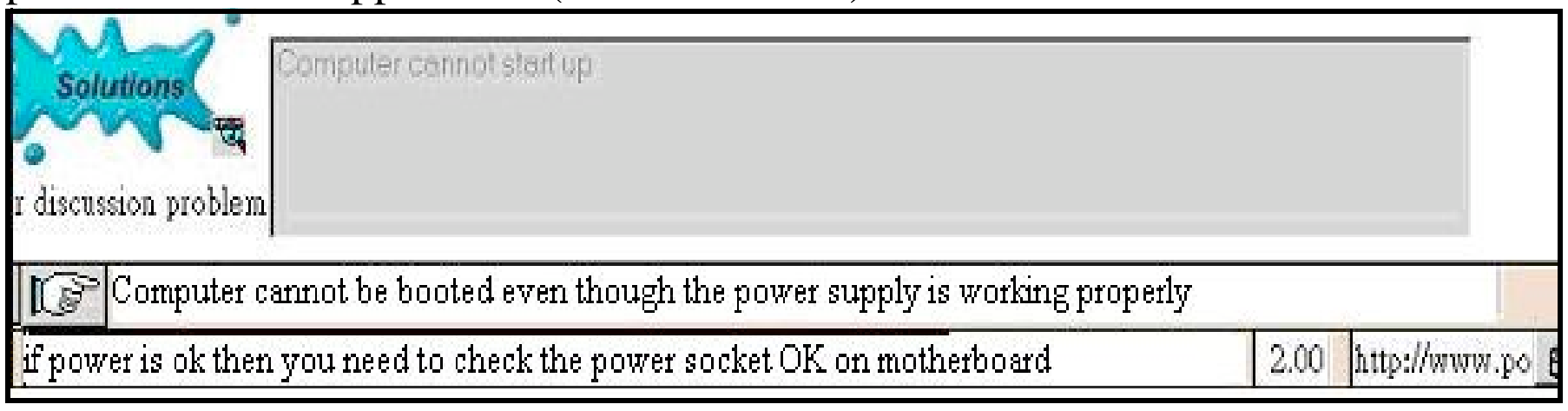

Fig. 7 shows a participant browsing deeper into a few levels from a suggested location.

In Fig. 7, the participant browses deeper into a few levels from a suggested location, and check an interested problem, such as, "Computer cannot be booted even though the power supply is working properly" which attached a solution of "if power is ok then you need to check the power socket OK on motherboard". By this kind of critical thinking and browsing others' posted problem-solutions, participants would build up their knowledge regardless of the previous learned or currently under developing.

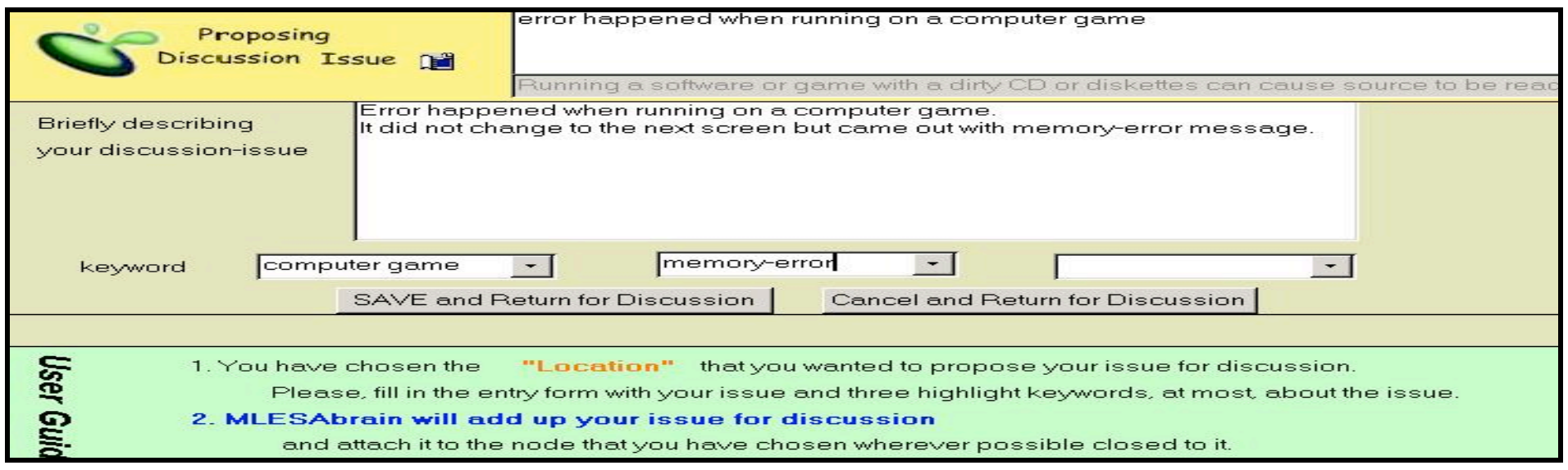

Fig. 8. After critical thinking, a participant posts his/her own problem for discussion.

However, if there is no suitable learning issues for their discussions, participants might post their own problems. Fig. 8 shows after criticizing with a preference-score and thinking about other participants' proposals. A participant posts his/her own problem, about "Error happened when running on a computer game. It did not change to the next screen but came out with memory-error message".

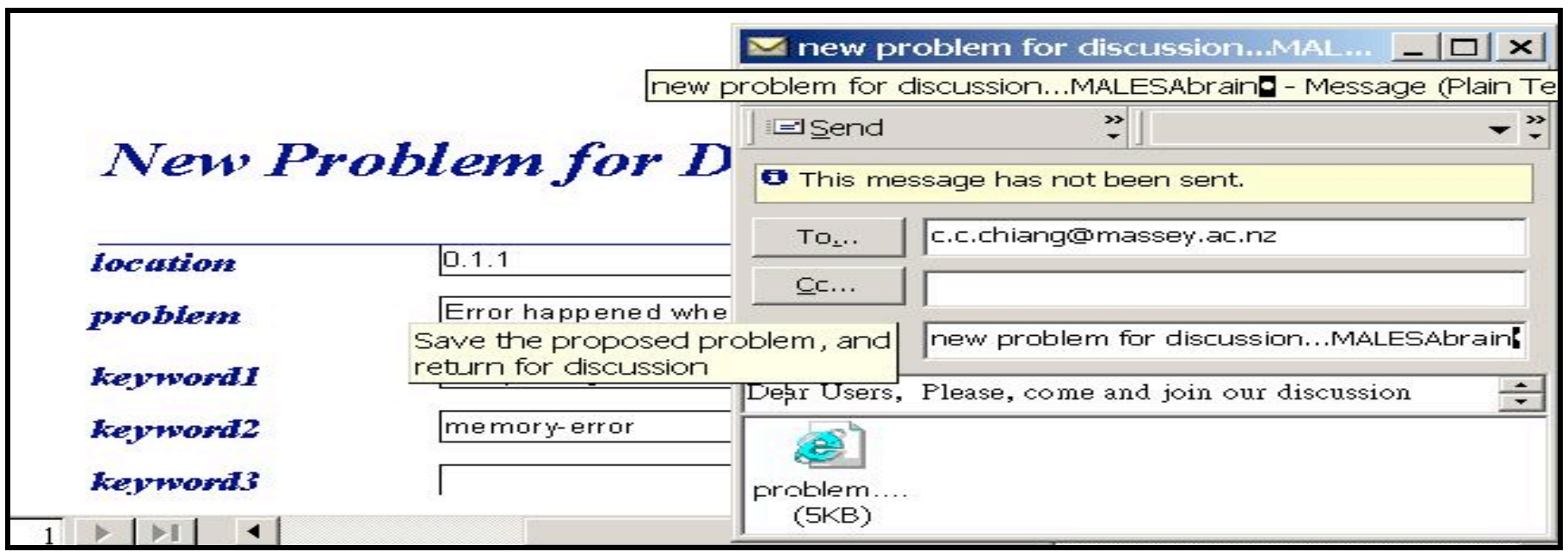

Fig. 9. MALESAbrain sends the participant's posted problem to other participants.

The Fig. 9 shown, the *broadcasting function* would give the posted problems a chance to be discussed from different viewpoints whenever different opinions exist. In the process of a problem-solution discussion, the participants who do not agree with a problem might propose another problem to clarify the original problem; and those who do not agree with a solution might contribute another solution to clarify or specify the original solution. However, when any new knowledge is added, MALESAbrain would notify other participants to encourage them to join the discussion.

The participant now needs to wait for the feedback from the others' judgments on his/her proposal. This broadcasting function can also be considered as a feedback mechanism for knowledge acquisition, which stimulates the participants to brainstorm more knowledge among discussants.

# 4 The Methodology

In this section, we explain the methodology of MALESAbrain. The design aims to capture knowledge from participants under the generic notion of questioning-responding. Under this notion, the discussion issue is initially posed as a problem by questions from the educator; a response or a counter-question is then suggested; another question follows; a subsequent response or another counter-question suggested and so on. This process continues until the solution(s) are narrowed down and found. Guided by the format of the *knowledge-content*, MALESAbrain launches its *critical thinking* function to help participants discuss their problems (and/or solutions) via an on-line tool.

---

**Definition 1**. A piece of *knowledge-content* $\varphi_i$ in MALESAbrain for problem-based discussion is defined as a pair of one problem and numerous solutions:
$\varphi_i = (p_i \, , \, \bigcup_j s_{i,j})$ where $p_i$ is a *problem* $\bigcup_j s_{i,j}$ is the collection of suggested solutions associated with $p_i$

---

In the learning system, the participants could submit a new problem or explore the knowledge base (see definition 1). If the participants choose to enter their discussion problems, they must complete a “question enquiry form”. Once submitted, the query will be matched with MALESAbrain’s retained knowledge according to the keywords and the problem descriptions

---

**Definition 2**. The knowledge *preferences* ***Pref*** in MALESAbrain is defined as a continuous function of real value ranging from -1 to +1

***Pref***:( $participant_k$ , $s_{i,j}$ ) °Ê$agreement_{k,i,j}$ (-1 ~ +1)

where ( $participant_k$ , $s_{i,j}$ ) is a pair such that

$participant_k$ = a participant in the workshop,

$s_{i,j}$ = a solution in MALESAbrain (see the defined solution $s_{i,j}$ in definition-1).

$agreement_{k,i,j}$ = the preference score of a participant’s, $participant_k$, judgment of a solution $s_{i,j}$ (value from –1 ~ +1).

---

During exploration, MALESAbrain actively questions the participants about their *preferences* - a numerical measure of the participant’s degree of support (or non-support) for a posted-solution to the problem. Participants must answer these questions prior to moving on to the next piece of content or chat room for discussion (see definition 2). The preference value ranges from 1 for total support, to 0 for no comment, and to -1 for total non-support. Such a device provides a window of opportunity for individual participants to review other problems from different perspectives and subjectively evaluate the advantages and disadvantages of other people’s works. Participants must judge or criticize other’s proposed solution, for its ability to solve a problem. Then new solutions may be confronted before they can chat or browse further knowledge-content. This is an important mechanism in the model, which enforces participants to think critically on a problem-solution from others’ suggestion and carefully judge their own preference scores.

**Definition 3** The *knowledge-weight* $w_i$ in MALESAbrain is defined as

$w_i = \sum_{j=1}^{m} w_{i,j} \cdot |s_{i,j}|$ (see definition-1 $\varphi_i = (p_i\ ,\ \bigcup_j\ s_{i,j})$), where $w_{i,j}$ is the summation of all participant $participant_k$ preferences towards $s_{i,j}$ : (For example, if we have 5 participants in the meeting then $_{k=1..5}$)

$w_{i,j}\ =\ \sum_k\ agreement_{k,i,j}$ (see definition-2 ***Pref***: < $participant_k$ , $s_{i,j}$ > °Ê $agreement_{k,i,j}$ )

Here, we define a symbol "| |" which will be used to test the existence of a solution, for transfer the existence of a solution into the value "0" or "1", to allow the knowledge-weight $w_i$ calculation. $|x| = \begin{cases} 0 & when\ (\neg\exists x) \\ 1 & when\ (\exists x) \end{cases}$ where χ is a solution

Note: see the knowledge weight $(\varphi_{2,1})$ calculation example at the end of this section

The *knowledge-weight* defined in Definition-3 is to tag the *knowledge-content* "$\varphi_i$" defined in Definition-1 with a numerical measure of its importance. Definition-3 sums up participants' preference-scores "$agreement_{k,i,j}$" toward each solution "$s_{i,j}$" as its solution-weight "$w_{i,j}$". Then it sums up (with the formula "$w_i$") all solution-weights and attaches it on a problem as the knowledge-weight "$w_i$". The summation provides indicators to the participants' willingness, interest, judgment and consensus on an individual discussion issue or problem.

**Definition 4**. An *Artificial Knowledge cell* (AK-cell) $k_i$ in MALESAbrain is a combination definition of definition-1 and definition-3, which is defined as,

$k_i = < \varphi_i,\ w_i >$ where

$\varphi_i$ is the knowledge-content (see definition-1 $\varphi_i = (p_i\ ,\ \bigcup_j\ s_{i,j})$) and

$w_i$ is the corresponding knowledge-weight

(see definition-3 $w_i = \sum_{j=1}^{m} w_{i,j} \cdot |s_{i,j}|$ )

In Definition-4, we define the basic elements of learning for MALESAbrain as *Artificial Knowledge cells* to combine the knowledge-content and the knowledge-weight into one single entity. An Artificial Knowledge cell (AK-cell) becomes the authorised representative for a piece of knowledge in MALESAbrain for self-learning.

**Definition 5**. The *growth-factor* $\gamma$, an integer number, in MALESAbrain is defined as the limit for constraining the posted number of AK-cells at each level, which converts the structure of knowledge base from linear structure to hierarchical structure, of γ-branch tree, in the forum.

In definition 5, the growth-factor normally is set up before discussion, however, it can also be changed after discussion whenever the educator wants to view from a different angle. It depends on the learning domain and target, if the growth-factor has been switched to three then the decision tree will be turned into three AK-cells on the top level and become a three-branch tree; if it has been switched to five then the decision tree will be turned into five best AK-cells on the top level for the workshop's learning decision; if set up as one AK-cell then there is only one best decision to be made.

To make a decision tree based on the participants' preference judgments, a hierarchical structure of knowledge ranking from the knowledge-weights has therefore been designed (see AK-cell in definition-4). Through limiting the number of AK-cells (see growth-factor in Definition-5) attached on each level, the learning issues posted in the forum of MALESAbrain can show as hierarchical in different important levels. In any form of discussion, it is natural that a question (or response) stimulates other questions or responses and therefore we model this phenomenon as the self-growing of an AK-cell. In other words, an AK-cell should be able to reproduce itself if required and branch out new connections in order to sustain these newly grown AK-cells. New AK-cells will also branch out whenever their parents can no longer sustain any new AK-cell, which might have grown beyond its setup limitation. This kind of AK-cell growth will keep occurring on and on, whenever the participants are posting new problems.

---

**Definition 6**. The *learning threshold* $\theta$ is defined as a collection of two decision pairs $\theta = \{<\theta_{kq},\theta_{kr}>, <\theta_{sm},\theta_{sd}>\}$, for comparing the retained AK-cells and their respective solutions, where

$\theta_{kq}$ is an *AK-cell qualification threshold*, when $w_i \geq \theta_{kq}$ then $k_i$ becomes a qualified AK-cell, which is the minimum requirement to join the competition for promotion to a higher order of discussion position

$\theta_{kr}$ is an *AK-cell rejection threshold*, when $w_i < \theta_{kr}$ then delete the AK-cell $k_i$

However, the system will not trigger any threshold and response the AK-cells within $\theta_{kr} \leq w_i < \theta_{kq}$

$\theta_{sm}$ is a *solution maturity threshold*, when $w_{i,j} \geq \theta_{sm}$ , the learning group agrees the solution $s_{i,j}$ is able to solve the problem $p_i$ .

$\theta_{sd}$ is a solution disagreement threshold, if $w_{i,j} < \theta_{sd}$ then delete the solution $s_{i,j}$

However, the system will not trigger any threshold and response the AK-cells within $\theta_{sd} \leq w_{i,j} < \theta_{sm}$

Whenever any of the thresholds are reached, MALESAbrain will trigger to re-organize the knowledge structure.

---

In Definition 6, according to the thresholds, MALESAbrain, is able to recognize the importance of the knowledge-weights among individual AK-cells. It then highlights the significant AK-cells and organizes the knowledge structure whenever the thresholds are reached. The hierarchical structure of AK-cells will be automatically arranged and ordered by the weights through the judgments of thresholds and the promotion competitions. Subsequently, some of the useful knowledge will be highlighted to the upper level for more discussions and the worthless knowledge will be deleted. From the ranking of the importance of the discussion problems, the educator can probe or challenge participants' thinking accordingly.

---

**<u>Definition</u>** 7. The knowledge based learning-rate [redacted] and *convergent-factor* "α" is defined as measuring *convergence* [redacted] $\geq \alpha$ , where

"α" is a real number between 0 to 1, called *convergent-factor*, is used to decide whether the percentage of mature AK-cells has achieved the educator's expectation.

$|\boldsymbol{M}|$ is the number of mature AK-cells, where $\boldsymbol{M} = \{k_i \mid k_i \in$ MALESAbrain $\wedge \exists s_{i,j}$ where $w_{i,j} \geq \theta_{sm}\}$

$|\boldsymbol{K}|$ is the number of current retained AK-cells , where $\boldsymbol{K} = \{k_i \in$ MALESAbrain$\}$

---

In Definition 7, MALESAbrain keeps assessing the learning-rate. During discussion, there are three figures shown on the learning system, the educator can check up the current learning rate (convergent or not), the current matured number of AK-cells and the current retained number of AK-cells to decide whether to involve discussion, change the discussion-domain or to stop the discussion. However, at the due date, if the learning-rate is still lower than the setup convergent-factor $\alpha$ then the educator will need to decide whether to extend the workshop. The educator can either calibrate the learning factors setup $\theta_{kq}$, $\theta_{kr}$ , $\theta_{sm}$ , $\theta_{sd}$ or $\alpha$ and/or even amend the learning topic/domain and arrange another discussion which follows.

## 5 The Learning System, Algorithm and Calculation Example

In this section we explain the details of the design features of MALESAbrain: (1) the working principles of the learning system, (2) the algorithm executed inside the system and (3) an example of the knowledge-weight calculation.

In the Internet learning workshop, with MALESAbrain facilitating discussion, the participants can link their computers to the MALESAbrain interface for discussing a problem domain. At the stage of critical thinking, they would contribute their knowledge regardless of the previous knowledge or current development, in terms of questions/responses, according to the knowledge representation format. The system then broadcasts the posted learning issue to the participants who are in the chat room to encourage discussion. The interested participants might join in with their viewpoints and/or add their learning issues accordingly. The system would actively ask the participants to show their preferences while participants browse their interested issues in the forum. Participants can judge the issues from different lines of reasoning based on their learning or understanding for solving the problem. As a result, they make their own judgments with individual preference-scores. However, if none of the posted suggestions are acceptable to a participant, he/she might also post his/her own problems or solutions on the forum for discussion, then wait for the feedback from others. Eventually, the scores will be summed up by the system to pass the decision onto the learning thresholds. These thresholds enable our participants to pay attention to highly-scored issues for more discussion and promote learning from peers in the workshop. At the last stage, the system indicates the learning-rate of what percentage of issues have arrived at consensus, which allows the educator or the system to understand the discussion progress and be in charge of when to stop the workshop for discussion.

The learning algorithm performs in the three stage cycling discussion sessions (see Fig. 1.) to achieve its learning-rate $\alpha$.

MALESAbrain($\theta_{kq}$,$\theta_{kr}$ ,$\theta_{sm}$,$\theta_{sd}$,$\gamma$,$\alpha$,due-date )
SET $\theta_{kq}$ ,$\theta_{kr}$ ,$\theta_{sm}$ ,$\theta_{sd}$ ,$\gamma$ ,$\alpha$ ,due-date /* def 6 */
REPEAT
/*** ***Critical thinking*** ***/
*COMPARE personal viewpoint "x" with retained knowledge pieces "$k_i$" in MALESAbrain* ***K*** = {$k_i \in$ MALESAbrain} /* def 4 */
Participants INPUT their ***Pref***:( *participant*$_k$ , $s_{i,j}$ ) °Ê*agreement*$_{k,i,j}$ (-1 ~ +1) /* def 2 */
IF (***Pref*** between –1 ~ +1 but not null) THEN /* def 2 */
allow ENTER chat-room or MOVE to next pieces of knowledge
MALESAbrain COMPUTE the *knowledge-weight* $w_i = \sum_{j=1}^{m} w_{i,j} \cdot |s_{i,j}|$ /* def 3 */
MALESAbrain DISPLAY (or POST) *participants' discussion issues based on "growth-factor $\gamma$"* /* def 5 */
/*** ***Attention*** ***/
IF $w_i \geq \theta_{kq}$ THEN /* def 6,7 */
INCREASMENT an AK-cell $k_i$ to knowledge base ***K*** ={$k_i \in$ MALESAbrain} by marking a "qualified' on this AK-cell
IF $w_i$ > parent($w_i$ ) THEN
SWAP($k_i$ ,parent($k_i$ ) )
END IF
END IF
IF $w_i < \theta_{kr}$ THEN /* def 6 */
DELETE $k_i$
END IF
IF $w_{i,j} \geq \theta_{sm}$ THEN /* def 6,7 */
DISPLAY a "mature" mark on solution $s_{i,j}$
INCREMENT a AK-cell $k_i$ to ***M*** = {$k_i$ | $k_i \in$ MALESAbrain $\wedge \exists s_{i,j}$ where $w_{i,j} \geq \theta_{sm}$} by marking a "mature" on this AK-cell
END IF
IF $w_{i,j} < \theta_{sd}$ THEN /* def 6)*/
DELETE $s_{i,j}$
END IF
IF (dd:mm:yy = due-date) THEN
IF ( [illegible] $\geq \alpha$) THEN /* def 7 */

```
      PRINT (MALESAbrain meeting with [redacted] % learning-rate)

      endDiscussion = TRUE  /* stop the meeting */
    ELSE
      endDiscussion = FALSE
      CALL MALESAbrain(θkq,θkr ,θsm,θsd,γ,α,due-date )    /* re-calibrate the
           learning thresholds and start another session of discussion whenever
           the learning-rate lower than α when time is due */
    END IF
  END IF
 UNTIL (endDiscussion = TRUE)  /*** learning-rate ***/
END MALESAbrain
```

**Algorithm 1**: the learning algorithm in MALESAbrain

***X:*** Discussion topic "How to Fix an Illegal Operation" - An operation requested to be performed by either the Operating System or CPU, which is not understood and therefore is Illegal.

- $K_1$ Running a software or game when memory shortage can cause Illegal Operations. ($w_1 = 3.1$)
- $K_2$ Running a source with a dirty CD or diskettes can cause data to be read improperly causing Illegal Operations. ($w_2 = 2.9$)
  - $K_{2.1}$ Corrupt, bad or missing files can cause Illegal Operations. ($\varphi_{2.1} = (p_{2.1}, \bigcup_j s_{2.1,j}), w_{2.1} = 1.4$)
    - $\bigcup_{j=1}^{1}$ It is recommended that you attempt to uninstall and or reinstall the program causing the Illegal Operation to verify that any corrupt, bad or missing files are replaced or repaired during the reinstallation. ($s_{2.1,1}$ , $w_{2.1,1}$=1.4)

**Fig. 10.** A snapshot of retained knowledge in MALESAbrain's forum

Fig. 10 shows a snapshot/example of knowledge pieces retained in the knowledge base. The knowledge posted by the participants has been organized in a tree-like manner according to the respective weights of individual nodes. In this example shown, the AK-cell $K_i = K_{2.1}$ includes a *knowledge-content* $\varphi_i = \varphi_{2.1}$ and a *knowledge-weight* $w_i = w_{2.1}$. For illustration the calculation on the knowledge-weight $w_i = w_{2.1}$ bonds with problem $p_i = p_{2.1}$ and solution $s_{i,j} = s_{2.1,1}$. Let us assume two visited participants have given their preferences on solution $s_{2.1,1}$ as 0.6, and 0.8 then $agreement_{k,i,j} = agreement_{1,2.1,1} = 0.6$ and $agreement_{k,i,j} = agreement_{2,2.1,1} = 0.8$

By Definition-3 $w_{ij}$, the solution-weight of $s_{2.1,1}$ (has two visitors' agreements ) is

$$w_{2.1,1} = \sum_{k=1}^{2} agreement_{k,2.1,1} = agreement_{1,2.1,1} + agreement_{2,2.1,1} = 0.6 + 0.8 = 1.4$$ …Ö@

$\varphi_i = \varphi_{2.1}$ includes one problem $p_{2.1}$ and one solution $s_{2.1,1}$ (because the problem only contained a solution currently, therefore the upper-limit of the summation, about the number of the solution, is set up as m = 1), so the knowledge-weight is:

$$w_i = w_{2.1} = \sum_{j=1}^{1} w_{2.1,j} \cdot |s_{2.1,j}|$$

$$= w_{2.1,1} \cdot |s_{2.1,1}|$$

$$= (1.4) \cdot |s_{2.1,1}| \text{ (ÅÊÖ@}$$

$$= (1.4) \cdot (1) = 1.4 \text{ (ÅÊ} |x| = \begin{cases} 0 & when\ (\neg \exists x) \\ 1 & when\ (\exists x) \end{cases} \text{ definition-3)}$$

Example 1: the calculation of knowledge weight ($w_{2.1}$)

This value of 1.4 represents the weight ($w_{2.1}$) of the knowledge-content ($\varphi_{2.1}$) in the knowledge (AK-cell $K_{2.1}$ ), which provides a quantitative measure of the synergic viewpoint on $\varphi_{2.1}$ obtained from the workshop discussion and it forms the basis for MALESAbrain's knowledge judgment capability. In the example, if the threshold $\theta_{kq}$ is set to 1, then any AK-cell weights higher than 1 point will be qualified to join the competition for promotion (see definition-6). This means at the moment the knowledge-weights of $K_1$ and $K_2$ should be greater than or equal to $K_{2.1}$ , otherwise the system will swap the positions of the lower-weighted AK-cells with the higher-weighted AK-cells (see Fig. 10).

## 6 Conclusion

In this paper, we have two contributions to advance the notion of cooperative learning via the Internet for knowledge acquisition.

- The first contribution is, the created data structure – coined as "AK-cell" – for cooperative learning, which combines knowledge-content and knowledge-weight. It allows the discussion-knowledge to become calculable in a threshold system. The knowledge-contents become mobile because of the combination of knowledge-weight and knowledge-content in the data structure. Knowledge-weight ranks AK-cells into different important locations, based on participants' judgments and the thresholds set up. Whenever the knowledge-weight moves the knowledge-content moves simultaneously. This helps participants pay attention to consensus knowledge for more discussion; and to think about why certain issues accumulate higher scores than others.

- The second contribution is, the dynamic structure of the knowledge base – coined as "MALESAbrain". A decision-making parameter called "growth-factor" (see definition 5) is set up to achieve the educator's training targets. It is the kernel function in the algorithm of MALESAbrain in the three stage cycling modules – critical thinking, attention and learning-rate (see Fig. 1). It helps the data structure AK-cell can be constructed according to the educator's viewpoint; whatever is right in his/her coaching for knowledge acquisition.

These two inventions, AK-cell and MALESAbrain, are adapted from PBL for sharpening critical thinking in Internet applications. This learning tool i.e. AK-cell and MALESAbrain encourages participants in teamwork as well as to share information and resources and support each other in the learning process. The learning discussions not only meet the objectives of the subject but also provide participants with a learning approach which gives them the opportunity to integrate and synthesize various knowledge components. There is a conceptual difference between forum and chat room discussion and MALESAbrain discussion. The former is more of a free style of chatting without constraining participants to judge others' suggestions as their joining for discussion; however, the latter is a training discipline, which will help participants to follow learning rules on the problem-based discussion for knowledge acquisition.

In our current experiments, we have prototyped MALESAbrain as a topic for "computer trouble shooting" discussion. With the prototype, we implemented the first version of MALESAbrain. We then invited our masters' students to discuss the question: "Is Java best for first-year computer science students or are there other suitable computer languages for the foundation paper"? We found this topic was not suitable for discussion by masters' students. Rather, it is more suitable for first or second-year computer science students, because it reflects a real-world problem which they will encounter. For this reason we changed the question to "How do I finish my masters' study in Computer Science within one year".

Forthcoming experiments will involve MALESAbrain with ill-structured problem discussion as PBL suggest. The same topic will also be discussed in a conventional PBL classroom to compare the quality of feedback from the learners.

# Towards Acquiring and Refining Class Hierarchy Design of Web Application Integration Software

Naoki Fukuta, Mayumi Ueno, Noriaki Izumi, Takahira Yamaguchi

[1]Department of Computer Science, Shizuoka University, 3-5-1 Johoku Hamamatsu Shizuoka 432-8011 JAPAN
fukuta@cs.inf.shizuoka.ac.jp
[2]Graduate School of Informatics, Shizuoka University, 3-5-1 Johoku Hamamatsu Shizuoka 432-8011 JAPAN
mayumi@ks.cs.inf.shizuoka.ac.jp
[3]Cyber Assist Research Center, National Institute of Industrial Science and Technology, 2-41-6 Aomi, Koto-ku Tokyo, 135-0064 JAPAN
niz@ni.aist.go.jp
[4]Depatrment of Administration Engineering, Keio University, 4-1-1 Hiyoshi, Kohoku-ku, Yokohama, Kanagawa, 223-8521, JAPAN
yamaguti@ae.keio.ac.jp

**Abstract.** Classes and their hierarchy are one of important artifacts of software development. In contrast of ontology development, the base concepts of classes do not exist even in the mind of software designer at the first time of development. Therefore, software development process needs not only to support externalization of concepts, but also to support the conceptualization process of the design. In this paper, we investigate a class diagram design support method that is suitable for web application integration software. We discuss features and drawbacks of the method through a case study in a development process of a business support system.

## 1. Introduction

Web applications have become sufficiently comprehensive and stable to provide various services. These Web applications include not only services to make the Web itself more accessible (e.g., Google [1], a Web search engine), but also emerging services that are closely related to the real world in terms of weather information (e.g., Weather.com [2]) and product merchandising (Amazon.com [3]). Web applications integration software(WAIS) is a kind of software that integrates such useful Web applications[16]. WAIS will enable us to develop more advanced applications to meet user needs. WAIS also offers cost reduction for application developments.

In WAIS development, respective Web applications have their own input and output parameter types. WAIS should treat such I/O types appropriately inside the system. Construction of a WAIS may impose strong constraints against the system design, such as the type of I/O objects accompanying a Web application. Design of WAIS requires that many elements that are essential as the software components (parameters required for the I/O of a Web application) be provided before starting to design its class diagram. Conventional class diagram design support methods have proposed a procedure for creating a class diagram from a use-case diagram with robustness analysis [4], and a measure for evaluating the design appropriateness of the finished class diagram [5]. The problem is that no methodology has been well studied for properly designing a class diagram that necessarily contains elements that is given before beginning the class diagram design. Accordingly, a design support procedure of a model should be developed that is applicable to design of a biased model affected by constraints of Web applications or similar frameworks.

Classes and their hierarchy are one of important artifacts of software development. In contrast of ontology development, the base concepts of classes do not exist even in the mind of software designer at the first time of development. Therefore, software development process needs not only to support externalization of concepts, but also to support the conceptualization process of the design.

In this paper, we investigate a class diagram design support method that is suitable for WAIS. We discuss features and drawbacks of the method through a case study in a development process of a business support system.

This paper is organized as follows: Section 2 proposes a design support method of a class diagram. Section 3 presents discussion of a business trip support system developed as a case study and applies the proposed class diagram design support method. In Section 4, we show some related studies and clarify the advantages of our proposed method. Section 5 concludes this paper and discuss about future work.

## 2. Class Diagram Design Support

This section proposes a design support method of a class diagram for WAIS development. Two types of support are proposed as the design support method of a class diagram: support by superficial information (syntactic information) and support using information on a semantic level. Lattice is used as the former support. We investigate the use of Lattice structure to by comparing distance and depth between elements.

The left end of Lattice is the shallowest and the right end of Lattice is the deepest. The depth is equivalent to the attribute number. We presume that it would be wrong to treat the attributes collectively in one class when the depth of each attribute of a certain class indicates a great difference. The difference in the depth was regarded as the difference in the attribute number. We inferred that the depth could be used as an index. Regarding the distance between elements, classes separated by a small distance were considered to be correlated. However, when few elements are available, a root element is routed and there is little actual correlation. Repeated up-and-down in small distances on Lattice also implies little correlation. For that reason, it was considered that two classes were correlated when they shared common multiple attributes, were in a parent-child relationship, and were located within a small distance. We assumed that this index would support the discovery of inclusion and succession relations.

Ontology is used for support using information on a semantic level. Ontology expresses the definition of conceptual relations. This study created Ontology according to DAML [6] and WordNet [7]. It was presumed that errors in a semantic level could be pointed out by matching Ontology and Lattice.

### 2.1 Model Design Support Method

This study employs the following two approaches of the class diagram design support method for software development incorporating a Web application:

**Approach A.**

Port information of a given Web application (service) is set as an initial class diagram.

**Approach B.**

The conceptual definition structure of region Ontology to the class diagram.

### 2.2 Approach A

Approach A has multiple Web applications, as shown in Fig. 1; it defines their I/O ports as respective classes to produce an initial class diagram. An I/O port is a set of elements required for input and output. For example, we define the input ports of the Web application A in Fig. 1 (a, c, d) and the output ports (b, d, f, k) as the input class $A_i$ and the output class $A_o$, respectively. For all Web applications, ports required for input and output are used as attributes, and the class is designed for each of them. The initial class diagram is defined as a unified class comprising all these classes and an attribute set required in order to acquire the attribute value for the I/O of a Web application. This study defines a Web application as a service provided on the Web. Such service released at one site having multiple functions is considered as one Web application. For example, TabinoMadoguchi, having the two functions of searching for a hotel and making a hotel reservation, is considered as one Web application.

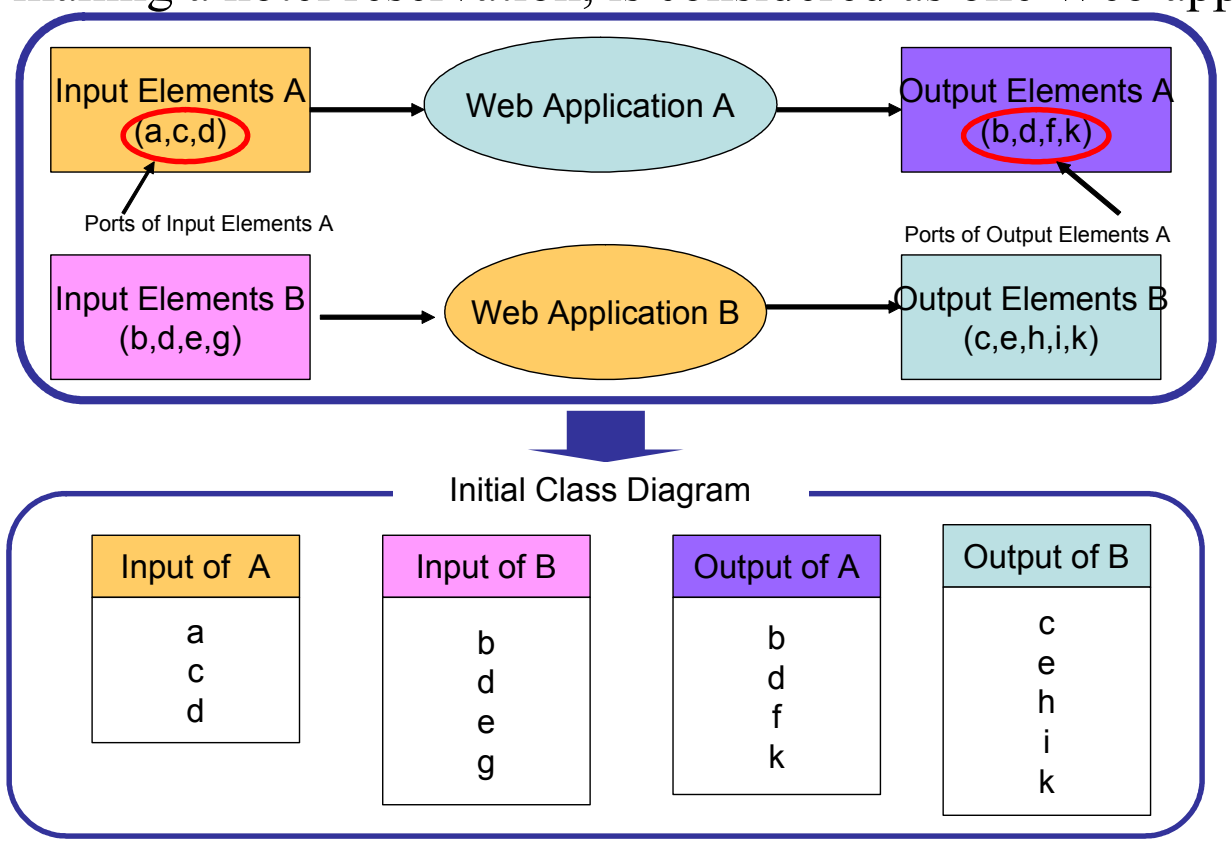

**Fig. 1.** Approach A

A structure that is derived from a lattice is employed to implement Approach A. This study defines lattice as one space that includes a whole process from an empty set to a universal set by the gradual increase of elements. Lattice is constituted using the I/O port of a Web application. A class of Lattice is created according to the following procedure:

1. All attributes of the class is added as elements of the lattice.
2. When other classes are referred as a type in the attribute of the class(e.g. has associations to another class), the attributes of the referred classes are also added as elements of the lattice..
3. Repeat 2 until all the attributes included in the class become an atom.

Then, a structure is constructed from the lattice by marking nodes as follows:

1. All nodes that only have elements as the same attributes of a class are marked.
2. All nodes that only have elements of a subset of a marked node are marked, and the nodes are linked. (Here , we call the newly marked node as a 'parent node'.)
3. Apply procedure 2 recursively.
4. The node that has no elements (empty node) is marked, and used as a root node of the structure.(Fig.2)
5. Here, when a parent node has only one child, the child node is unmarked. (Fig.3)

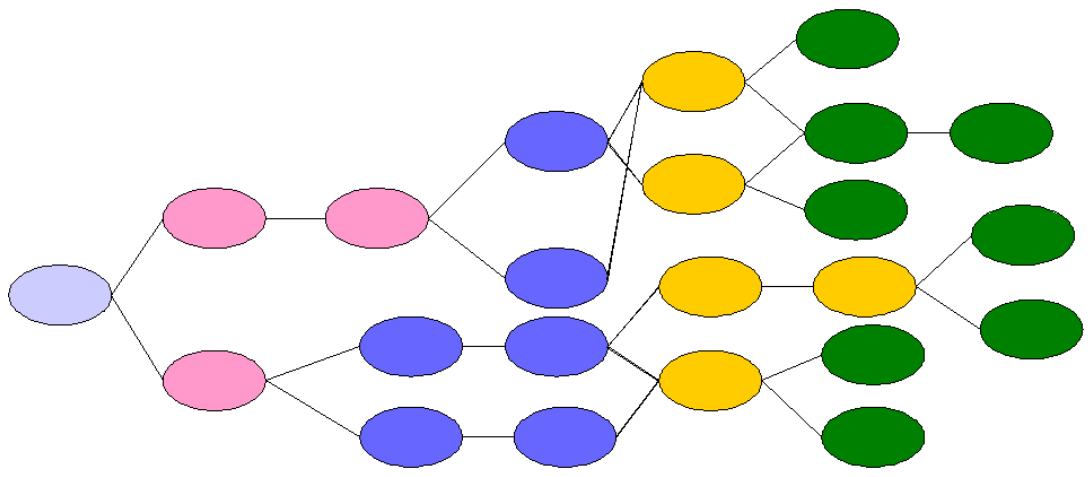

**Fig. 2.** Attribute lattice and marked nodes

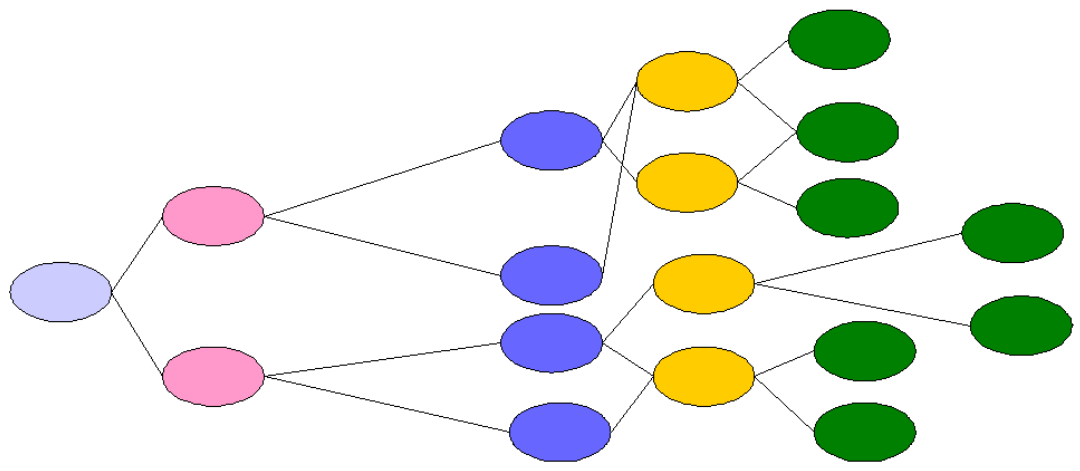

**Fig. 3.** A structure derived from the attribute lattice

The total attribute number in a class diagram usually reaches several tens or more. Therefore, when all attributes are used as lattice elements, they are incomputable if they are to be used after computing the whole lattice. This study comprehends lattice as a mode of expression that represents the degree of refinement of a class diagram. It limits computation only to the required section. Here, extracted within four levels from the common parent marked nodes as candidates for the new class, where the level is distance of links between marked nodes.

### 2.3 Approach B

Approach B compares the structure created in Approach A (upper part, Fig. 4) with an appropriate ontology (lower part, Fig. 4). For example, in the ontology at the bottom of Fig. 4, the element X is ranked higher than the element Y. At the top, the element Y is an upper concept of the element Y. Comparison between a class diagram and ontology can trigger a model designer to consider whether the conceptual definition is wrong or a wrong label is attached. Therefore it is supposed that comparison of a class diagram and ontology is effective in refinement of a class diagram.

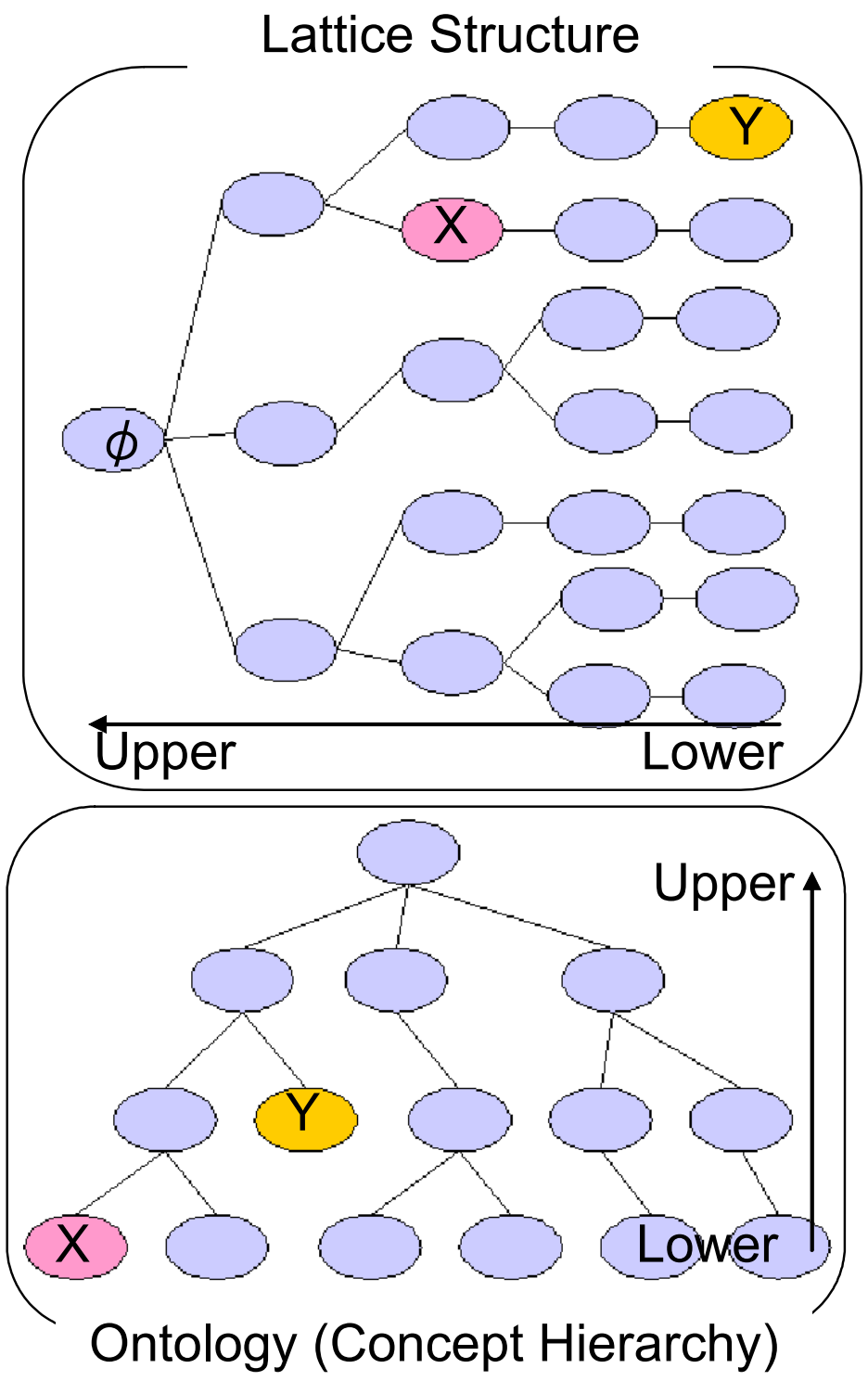

**Fig. 4.** Comparison with Ontology

## 3. Case Study

In the present business trip support system, when a user logs in, the system acquires his personal information. When the user enters a business trip schedule, the system proposes an appropriate train schedule and candidate hotels using his personal information; it makes necessary seat and hotel reservations. This business trip support system preserves, in its database, any changes made by the user against the system recommendation, such as changes in departure/arrival stations and departure/arrival times, alternative searches and new proposals, and previous use of the history of this system by the present user. This accumulated history enables the system to search again for a required item and thereafter propose a new schedule. The system employs a database to acquire the nearest station of origin or destination. When a previously visited place is input as the Origin or Destination, the system inputs its attribute value automatically using the database. This system cooperates with a business trip application system of a certain company. It has a mechanism for supporting the input of a schedule generated by this system to that business trip application system.

Search and reservation of trains or hotels are implemented using personal information such as the user's ID and password for using the Web application acquired at login, and using existing Web applications. The Web applications to be used include Jorudan[13] (train search), TabinoMadoguchi[14] (hotel search and reservation), JR Eki-net[15] (train reservation), and the above-mentioned business trip application system. Because this study aims at the design support of a class diagram when developing a system using a Web application, this business trip support system is presumed to be suitable as a case study.

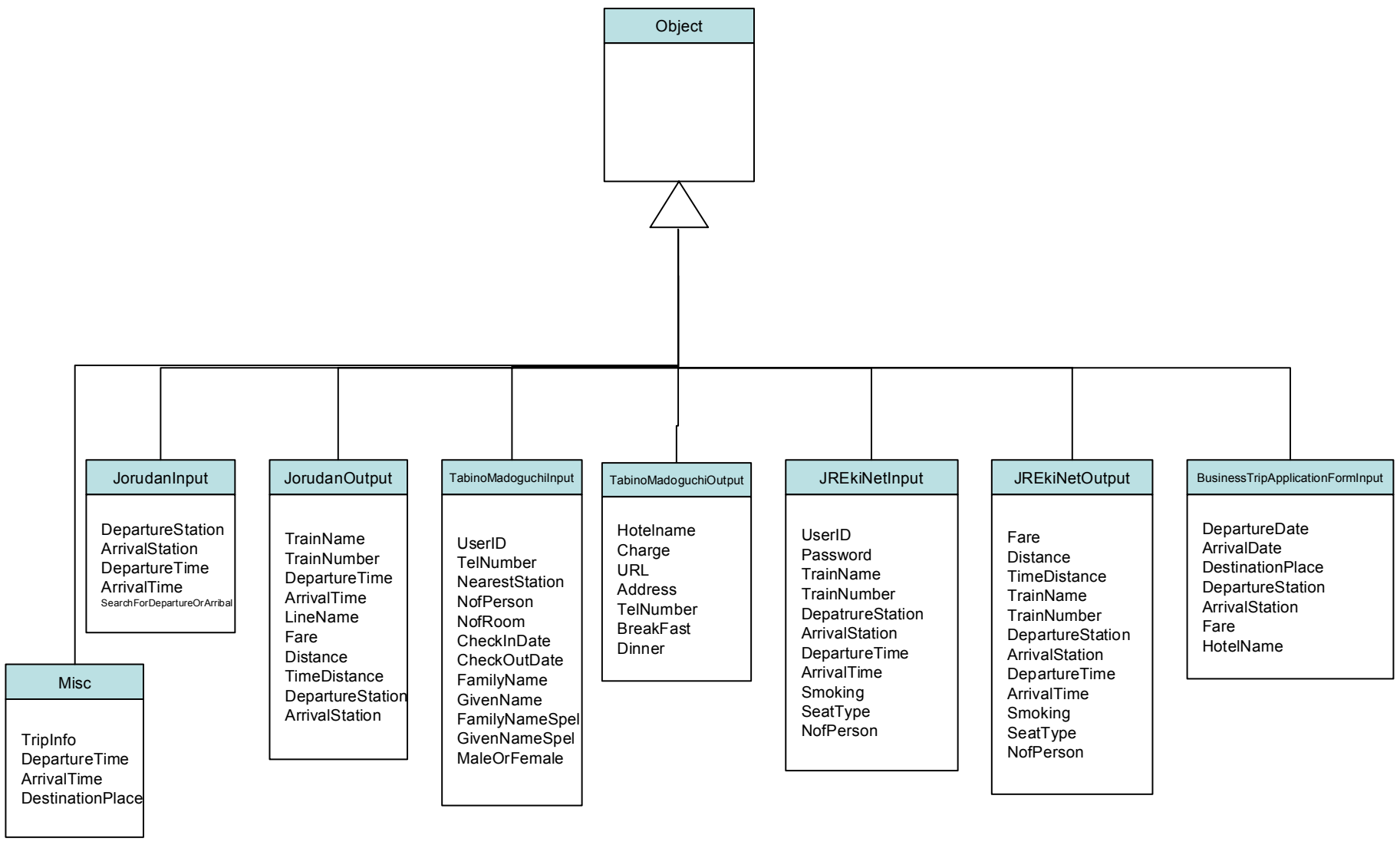

**Fig. 5.** The Initial Class Diagram

### 3.1 Application of Model Design Support Method

A class is assigned to each of the input and output of a Web application as an initial class diagram of the business trip support system mentioned above. The proposed method stated in Chapter 3 is applied to this initial class diagram. It is verified whether a refined class diagram can be generated.

In system development, I/O attributes are extracted for every required Web application, as shown in Table 1. In addition to Table 1, although it is not the I/O attribute of Web application, attributes used for acquiring input attribute values for Web applications are 1)Business trip description, 2)Starting date, 3)Ending date, and 4) Origin.
The initial class diagram is shown in Fig. 5.

**Table 1.** I/O Ports for Web Applications

| Web Application | Input Ports | Output Ports |
|---|---|---|
| Jorudan(Train Search) | DepartureStation, ArrivalStation, DepartureTime, ArrivalTime, SearchForDepartureOrArrival | TrainName, TrainNumber, DepartureTime,ArrivalTime, LineName,Fare,Distance, TimeDistance |
| TabinoMagoguhi (Hotel Search and Reservation) | UserID,TelNumber, NearestStation, NofPerson, NofRoom,CheckOutDate, FamilyName, GivenName, FamilyNameSpel,GivenNameSpel, MaleOrFemale, | HotelName, Charge, URL, Address, TelNumber, BreakFast, Dinner |
| JR Eki-net (Express Seat Reservation) | UserID, Password, TrainName, TrainNumber, DepartureStation, ArrivalStation, SeatType, NofPerson, Smoking | Fare, TimeDistance, Distance, TrainName,TrainNumber, DepartureTime, ArrivalTime, SeatType, Smoking, NofPerson |
| TripApplicationForm (IntranetBusinessApp) | DepartureDate, ArrivalDate, DestinationPlace, DepartureStation, ArrivalStation, Fare, StayingHotelName | TrainName, TrainNumber, DepartureTime, ArrivalTime, LineName, Fare, Distance, TimeDistance |

The system has 35 attributes, which include the I/O attributes of the above-mentioned Web applications and attributes for acquiring the input attribute values of the Web applications. Lattice was created with these 35 elements. Then, the initial class diagram was formulated for every Web application, defining each attribute set required for input and output as one class respectively. Fig. 7 shows in which position of Lattice these classes are located. Circles and triangles denote input classes and output classes, respectively. This Lattice indicates that classes are generally positioned in the left side of Lattice.

Next, attributes included in the final class diagram in the business trip support system (Fig. 6) were used as elements without using the above-mentioned method. Fig. 8 shows Lattice thus obtained. Squares and diamonds denote input classes and output classes, respectively. The section surrounded with a red curve is the class set up as the initial class diagram. The classes not surrounded with the red curve are classes related to the I/O of Web applications extracted out of the final class diagram designed at the development of the business trip support system was developed (Fig. 6). The figure shows where of Lattice these classes correspond. In some cases, multiple classes share the attributes of the input or output of a Web application, so that one service has multiple inputs or outputs.

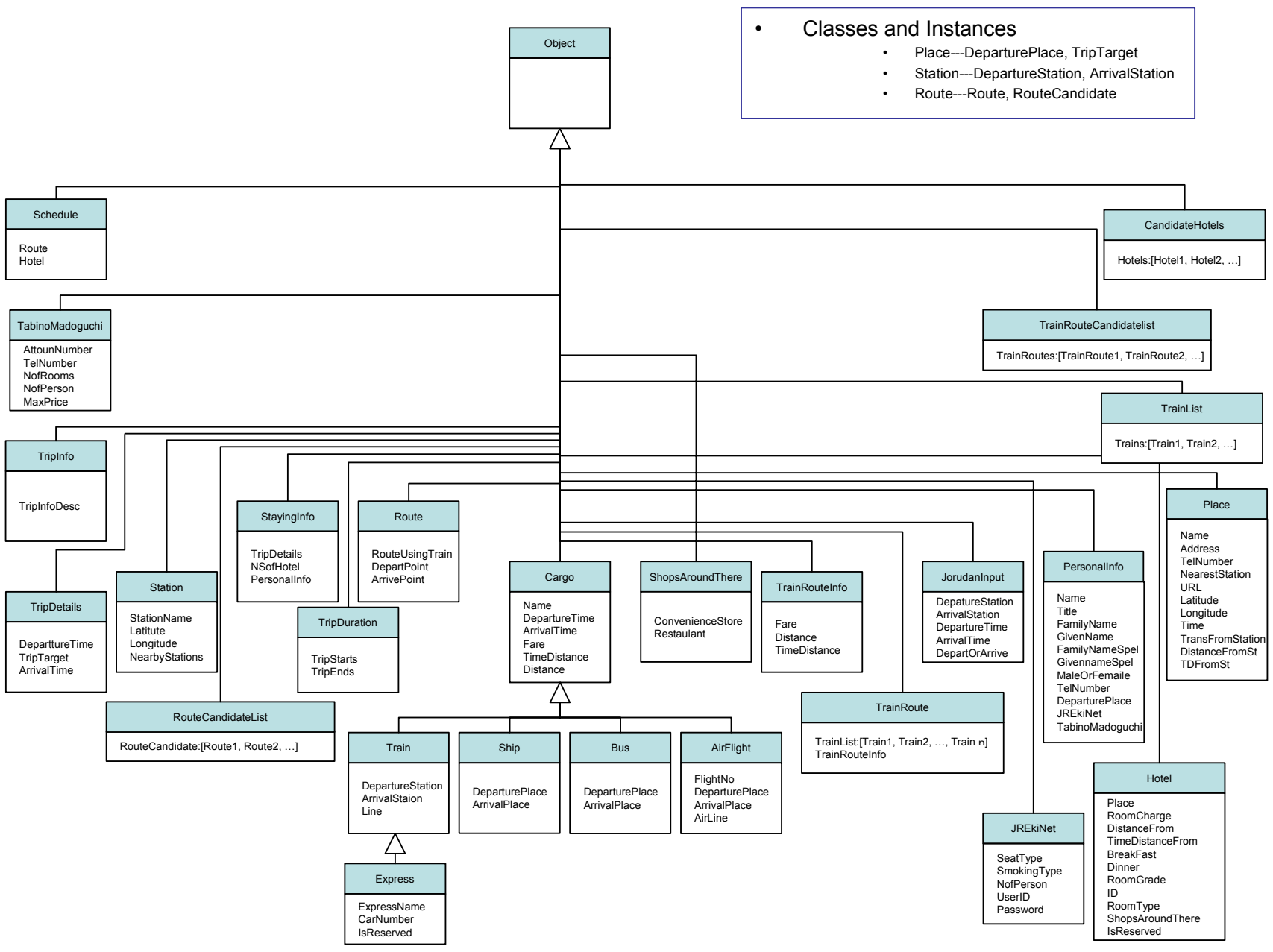

**Fig. 6.** Final Class Diagram in the Design

Comparison between the set of the initial classes (section surrounded by the red curve) with the set of the final classes (section outside the red curve) indicates that the former has fewer attributes per class than the latter. Therefore, the former tends toward the left side. That is, the last class has shifted to the right-hand side as compared with the first class. Fig. 7 implies that adding attributes to the initial class creates the final class. Because the initial class is a set of classes that are vital for use of a Web application, no attribute (element) would be deleted during refining into the final class diagram. If attributes (elements) on the interim class diagram are positioned on Lattice similarly, it is likely that attributes (elements) created during refinement from the initial to the interim may be judged unnecessary and deleted during refinement into the final class. However, we will treat only the change in refinement to the final class diagram from the initial class in this paper.

In output information on Jorudan, new elements were added by structuring on refining to the final class diagram from the initial class diagram. In addition, in the output of other Web applications such as TabinoMadoguchi and JR Eki-net, structuring caused the addition of new elements. For example, the Train class that was newly created serves as the next input of JR Eki-net. Thereby, coding was facilitated because only one class must necessarily be managed. Moreover, classifying outputs by type may improve maintainability and expandability. We conclude that elements that are added to the output information on each Web application are generated by structuring, which improves implementation efficiency.

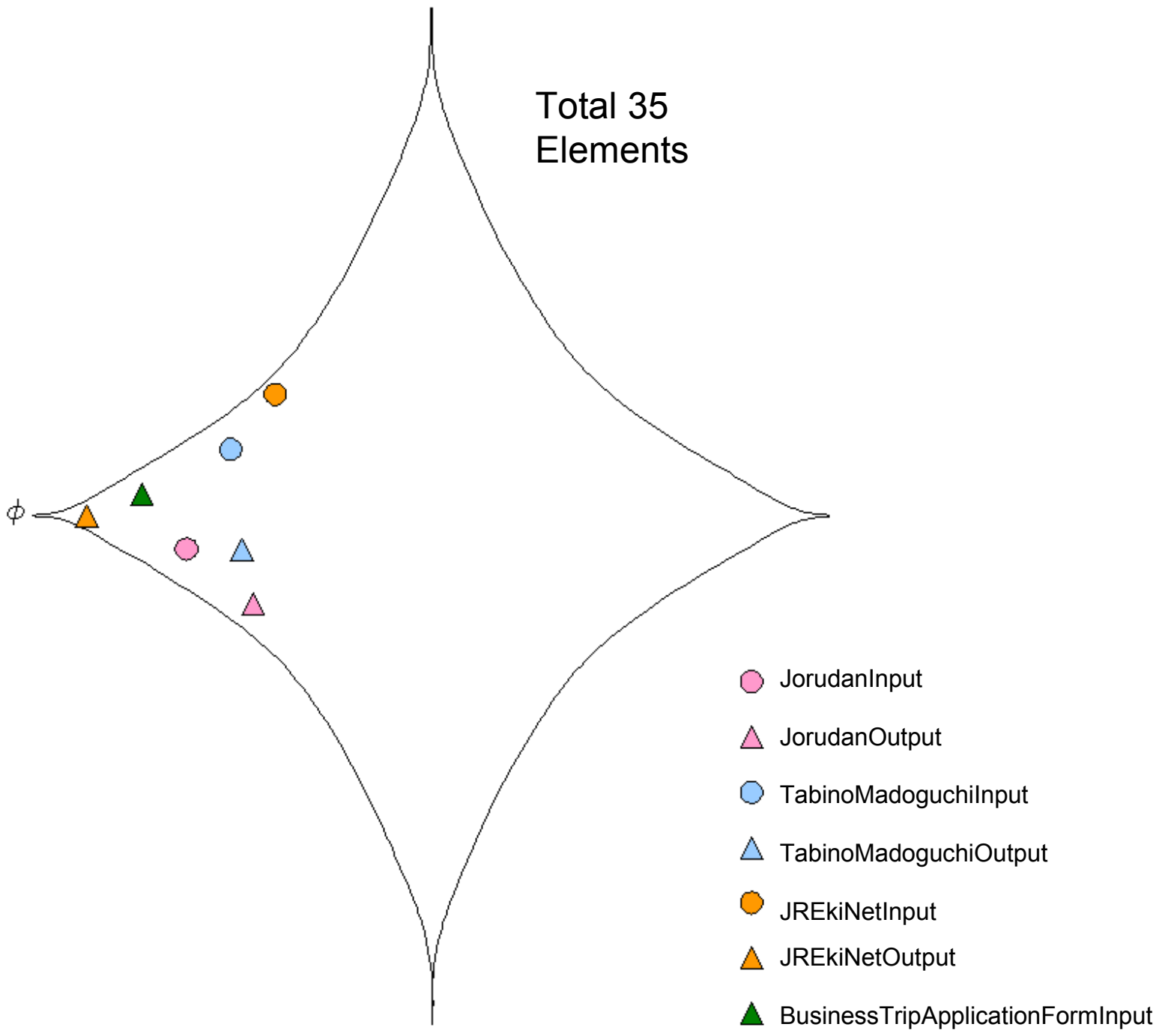

**Fig. 7.** Configuration of I/O Ports on Lattice

The initial class diagram (Fig. 5) is then expressed with Lattice. Fig. 9 represents Sub-Lattices extracted from the common parent node, based on the above-mentioned proposition method – "Sub-Lattices extracted within four levels from the common parent node are assumed as candidates for the new class." The pink nodes are common nodes and orange nodes are classes. The Jorudan output classes, such as Departure station, Arrival station, Departure date/time, Arrival date/time, Fee, Distance, Time required, and Route name are inferred to be the classes that represent a train. When this system was actually developed, the Express train class and the Standard train class were separated finally, as shown in Fig. 6. Although a train had its Train number and Train name and an express train had its Route name in reality, they were deleted at this time because they were unimportant at the time of search or reservation. However, as Fig. 9 indicates, as the Train name and Train number are classified in one common node, they can be separated into two classes. Therefore, the Train class was inherited to be the Express class (Train name, Train number). Because Fee, Time required, and Distance all express information on a train and its route, they were presumed to be separable as one class. Consequently, the class diagram was modified so that the Jorudan output class consisted of the Train class and Express class, and the JR Eki-net output related to the Express class. Because the JR Eki-net input had most attributes of the Express class as its own attributes, as shown Fig. 9, it was replaced by the Express class.

Ontology allows us to separate the TabinoMadoguchi input class (the left side in Fig. 11). Because the Family name, Given name, Family name reading, Given name reading, and MaleOrFemale are classified as subordinate concepts of personal information in the Ontology in Fig. 9, these attributes can be collected into one class as personal information. For that reason, the class diagram is modified as in the right-hand side of Fig. 11.

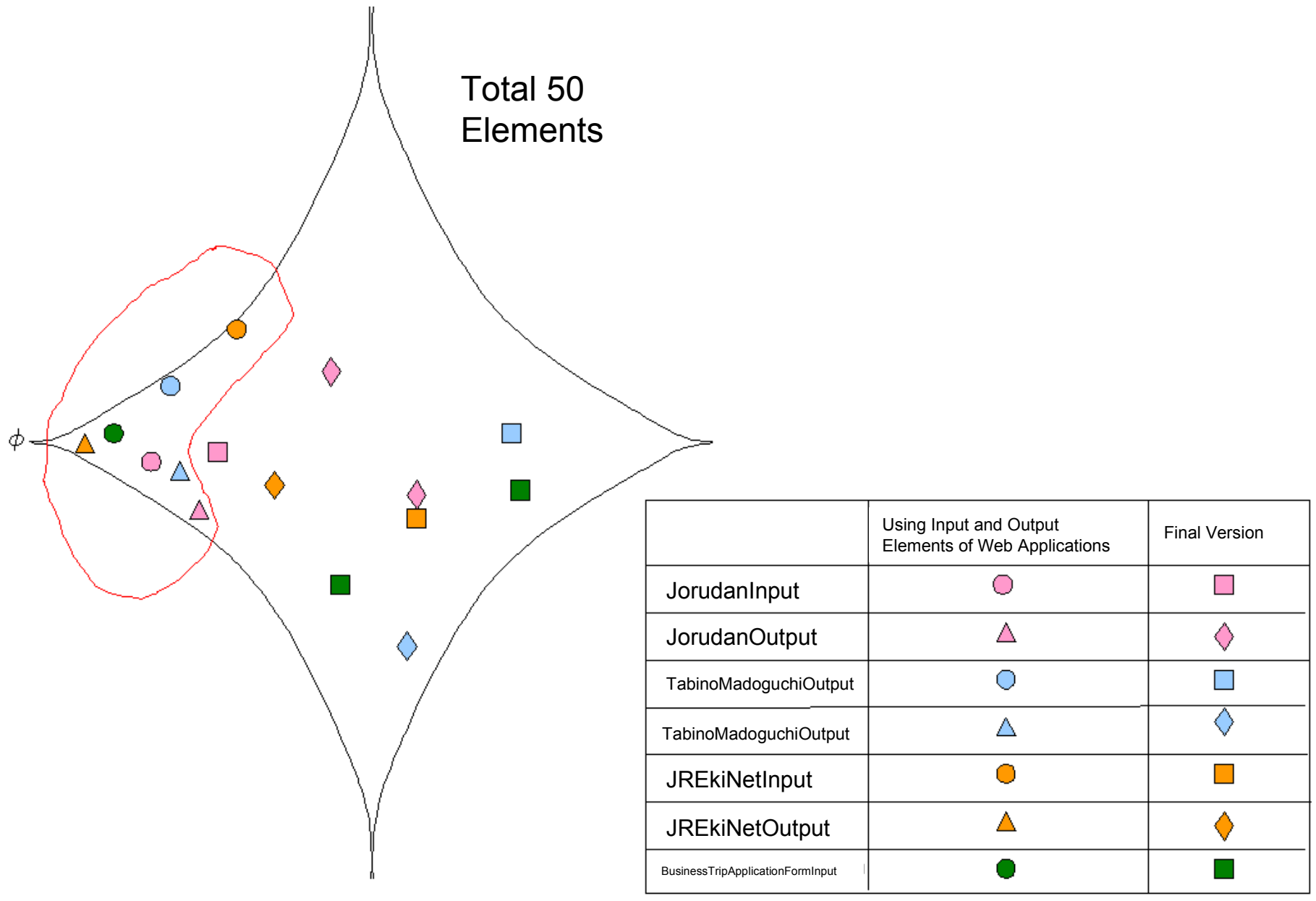

**Fig. 8.** Shift on Lattice by Refinement

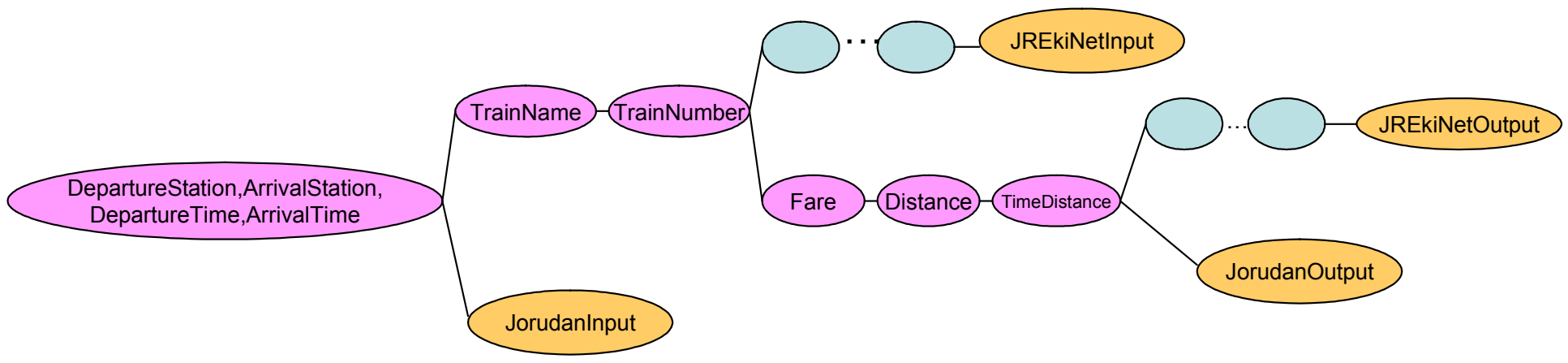

**Fig. 9.** Sub Lattices within Four Levels from the Common Parent Node

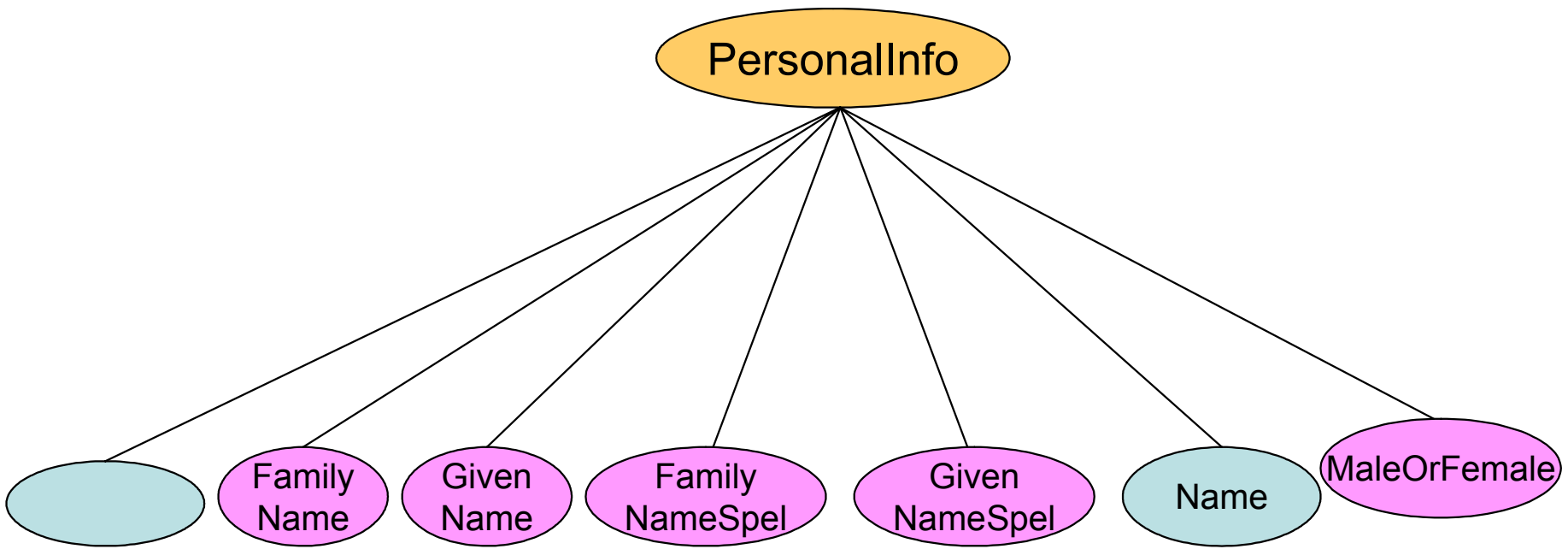

**Fig. 10.** Ontology on Personal Information

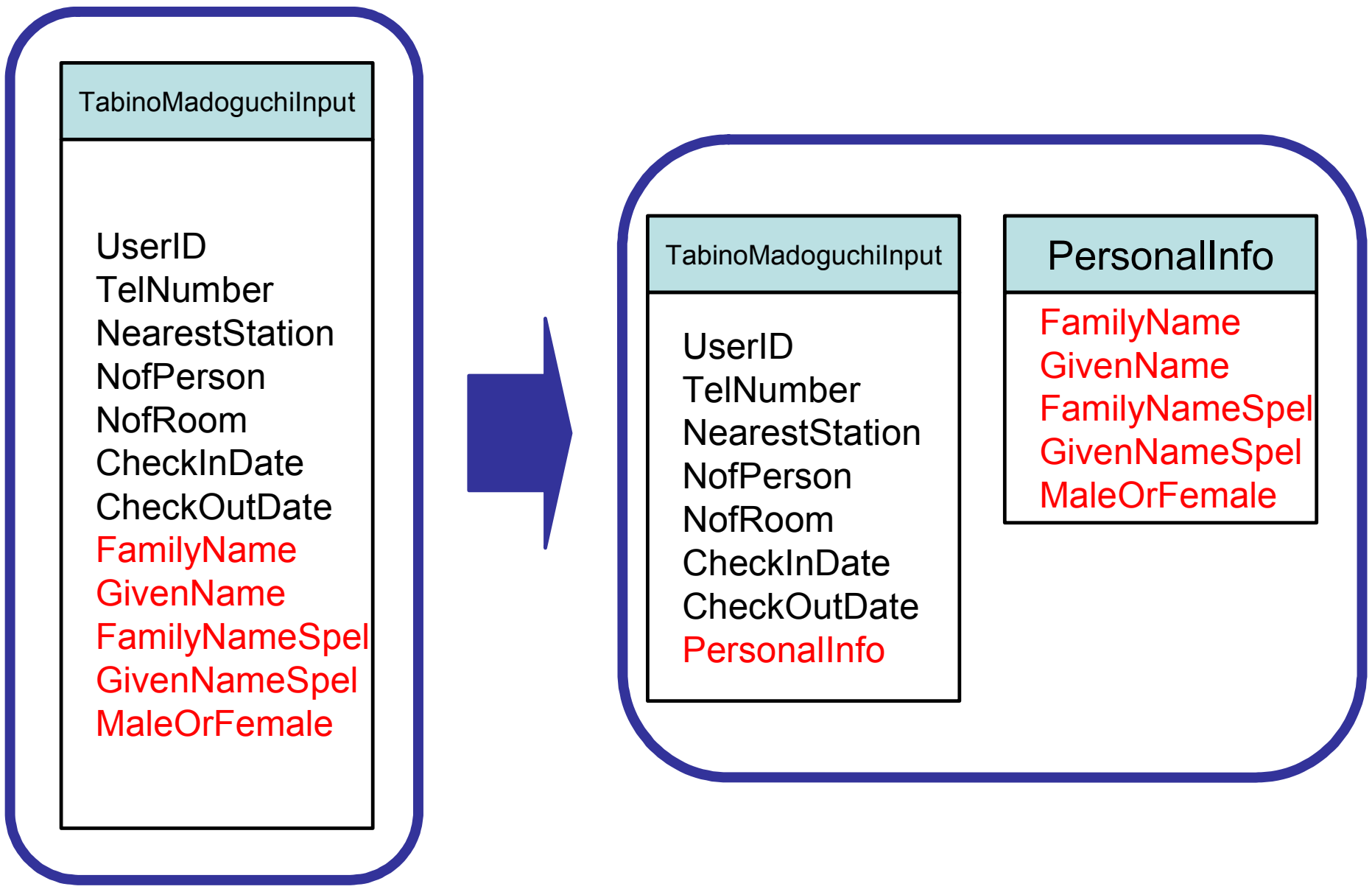

**Fig. 11.** Example of Class Diagram Restructuring

This study created Ontology as shown in Fig. 10 based on the ontology currently released by DAML (DARPA Agent Markup Language) [6]. DAML provides ontology on travel, which has no concepts regarding personal information. Instead, some ontology on a person was used to create the Ontology in this study. The Family name reading and Given name reading are expressions that are intrinsic to Japanese language: no currently-published existing ontology includes them. We added them considering that they are treated as a Family name or Given name.

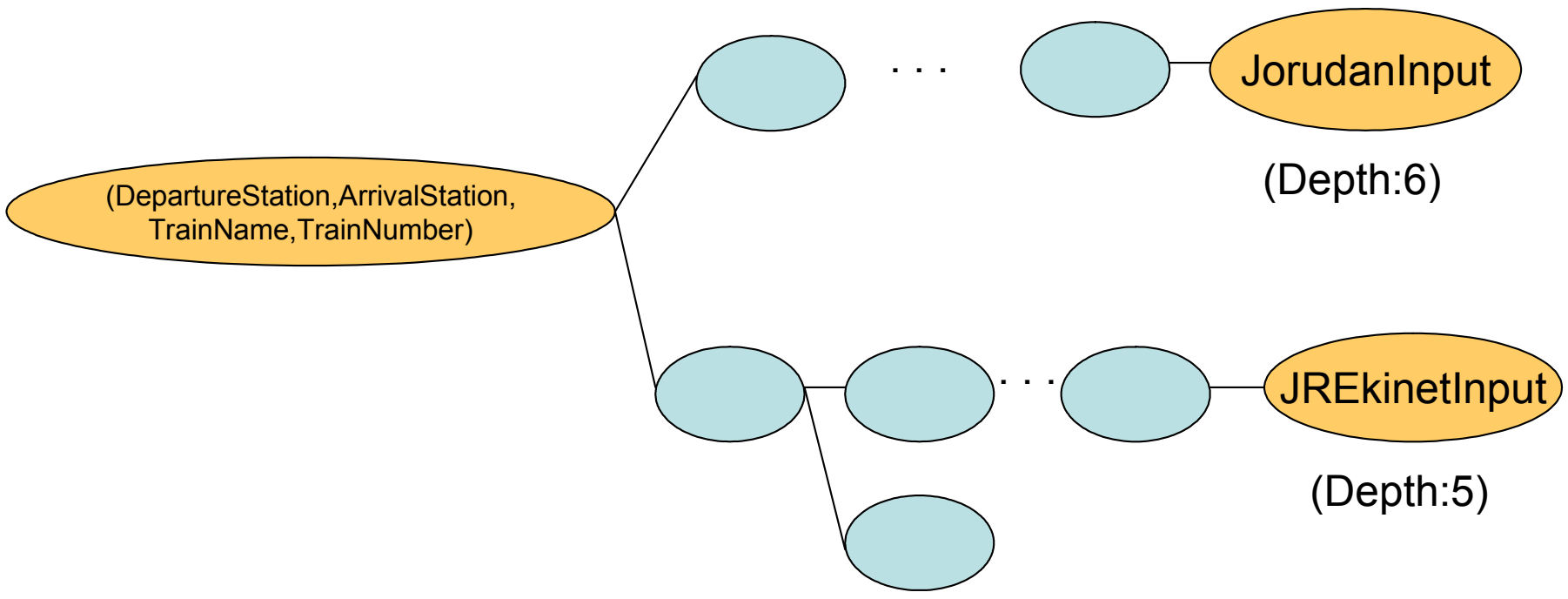

Fig. 12. Part of Lattice

When the output and input of cooperating Web applications (i.e., the output of train search and input of train reservation) have common attributes, they will be a part of a role. A role is a set of the I/O that defines one specific function when a Web application has multiple functions. Although not directly related to the refinement of a class diagram, a role, if known, may cancel the polysemy of a Web application.
The modified class diagram is shown in Fig. 13. Pink classes are newly created classes; attributes changed in connection are shown in red. This class diagram was compared with the final class diagram (Fig. 14) designed when the business trip support system was developed. Classes in pink and yellow in Fig. 14 denote classes completely and partially supported with the presently proposed method, respectively.

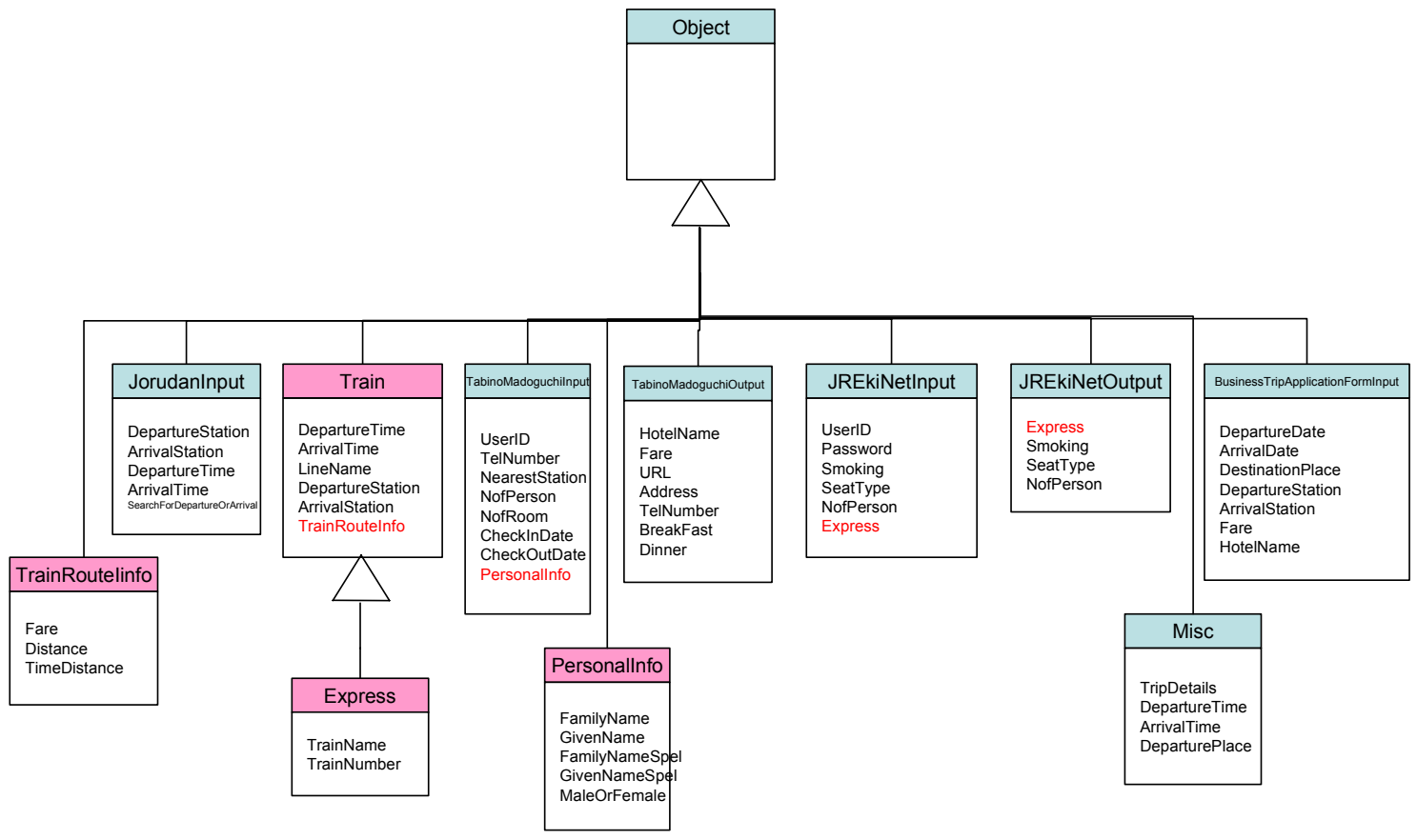

**Fig. 13.** Modified Class Diagram

Therefore, it is considered that the presently proposed method can support discovery of inclusion and succession relations and separation of attribute sets commonly shared by others or close in a semantic sense. On the other hand, it cannot separate information that is intrinsic to each Web application. Furthermore, although architecture matching the user interface and objects was constructed in this case study, no framework for it could be prepared.

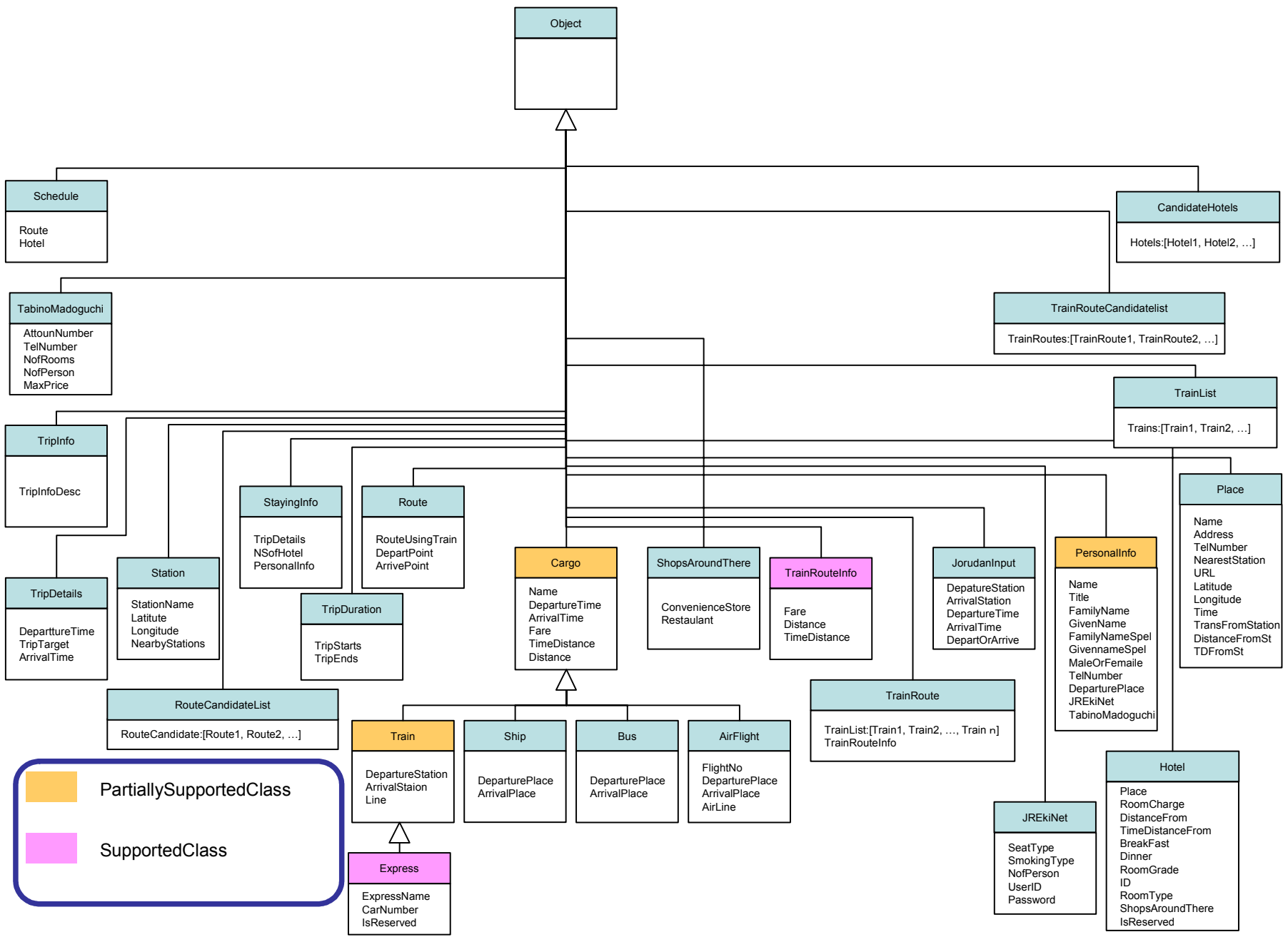

**Fig. 14.** Correlation with class diagram developed without using support method

## 4. Related Work

*Restructuring*[8] is a technique to refine the internal software structure. In the object oriented software development domain, we use the word *refactoring*[9] instead. In refactoring process, classes, variables, and methods are redistributed in order to facilitate future adaptations and extensions. Because it covers very wide area of incrementally development software process, our proposed method can be viewed as a variant of refactoring techniques. In last part of section 3, we demonstrated that our method can be applied to the refactoring process. But that is not our prior goal. Some methods and tools has been developed in the area of UML-based refactoring (for example, see [10]). Most of those tools tend to reduce design changing cost and to avoid errors caused by the change, not to improve the quality of refinement itself. We focus the improvement of quality of refinements for class diagram design.

Nytun et al.[11], proposed a method to enable modeling and consistency checking for legacy data sources. Because Web applications can be viewed as a type of legacy data sources, their work is closely related to our work. In Nytun's approach, OCL(Object Constraint Language) is used to check consistency of the current model. They assume the existence of objects' constrains that is well described using the OCL. Our approach can be applied even when there is not enough description of constraints. In that reason, our approach is mainly used in the initial stage of the development.

Egyed[12] proposed a method to abstract lower-level classes into higher-level classes by using relational reasoning approach. In Egyed's approach, relationships of inclusion among input and output parameters of a method, is used. Although using inclusion relationships of input and output parameters is a key feature of our approach, the goals are different. Egyed's method is developed for hiding complexity of class diagram that is designed in detail. In such diagram, because of implementation-related restrictions, one abstract class is often represented by several lower-level classes that contain same or similar features. For example, one distributed object is represented by three different classes in the implementation of J2EE framework. The goal of Egyed's method is to reproduce the higher-level class structure that is lost by the detailed design. Our method tends to be used in the initial stage of class diagram design where there is less detailed design.

## 5. Conclusions

This study specifically addressed the class diagram, especially in terms of the model design. It was assumed that many elements that are essential as the software components (attributes required for the I/O of a Web application) are provided by a Web application before starting to draw its class diagram. We proposed a class diagram design support method for designing the class diagram that implements these restrictions.

We adopted an approach to define the upper part of ports of a given application (service) into the initial class diagram. We then applied the conceptual definition structure of regional ontology to the class diagram. Specifically, the I/O ports of a Web application were treated as respective classes to create an initial class diagram. This class diagram was expressed using Lattice and Sub-Lattice extracted within four levels from the common parent node to be candidates for a new class. The semantic structure was verified by matching Lattice with ontological structure.

The business trip support system is itself supported using the presently proposed method in discovery of inclusion and succession relations, separation as a class of attributes shared commonly by others, and correction of the semantic structure with a wrong label attached. The presently proposed method cannot separate information that is intrinsic in each Web application. Furthermore, although architecture matching the user interface and objects was constructed in this case study, no framework for it could be prepared.

Future studies must examine whether the class diagram design support method proposed in this study can function as a support method in cases where there are other restrictions such as a framework or Web service. Moreover, it is necessary to examine how effective the existing ontology is when applied to other case studies. These will be our future subjects.

# Consideration on "Educationality" of Knowledge Acquisition Support Systems

Toshiro Minami

Kyushu Institute of Information Sciences, Faculty of Management and Information Sciences, 6-3-1 Saifu, Dazaifu, Fukuoka 818-0117 Japan
minami@kiis.ac.jp
http://www.kiis.ac.jp/~minami/

**Abstract.** In this paper we deal with the concept of "educationality" for knowledge acquisition support systems. A system with high educationality will help us effectively get knowledge by ourselves and develop learning skils as we use it. This concept is based on the learning model where the users actively think and learn by themselves, whereas in the typical CAI systems the users try to understand and memorize the materials prepared in advance. In order to investigate "educationality," we first take two example systems, the Web LEAP (Web Language Evaluation Assistant Program) and SASS (Searching Assistant with Social Selection), and analyze them in what way they are educational. We conclude that the most important aspect for educationality is to give appropriate materials in terms to the learners in terms of granularity of information, level and kind. This concept can be extensively applicable to other types of interactive systems as well.

## 1 Introduction

In this paper we deal with knowledge acquisition for human and machine parts in the framework of man-machine systems. Man-machine system is an integrated view for systems especially for interactive systems. A man-machine system consists of human user part and machine (or system in narrower sense) part. In this standpoint, both parts acquire necessary knowledge for achieving the goal of the whole man-machine system. We would rather put focus on the knowledge acquisition of human side in this paper. Thus, the major issue is how the user learns and acquires knowledge and skill as he or she uses the machine. Acquisition of knowledge of the system part consists of two aspects; (1) adaptation of user interface and other facilities for improving the ability of achieving its goal, and (2) facilities that support and accelerate the user's knowledge acquisition. The second aspect is the major concern in this paper. We call this aspect "educationality."

The major difference of knowledge acquisition styles between ordinary education systems and the interactive systems that are educational is shown in Fig. 1. In the figure, knowledge is considered as some information that is acquired from various material, facts, data, and etc., which is useful in solving problems, living better life, and other ways. In ordinary education systems, useful knowledge is extracted by teachers and experts in advance. Then such authorized knowledge is prepared as learning material in the form like texts and coursewares. Human learners get such knowledge by self-learning, by lecture, or through using CAI systems.

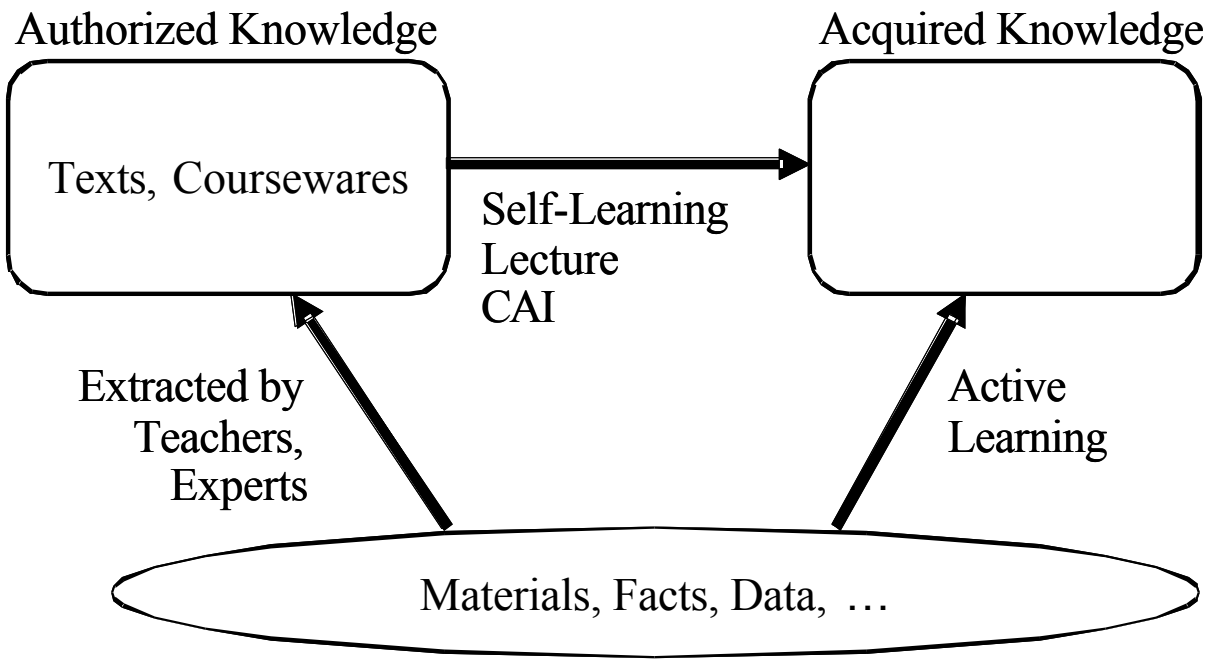

**Fig. 1.** Knowledge Acquisition Styles

On the other hand, in educational systems knowledge is acquired directly by the user. The user finds or creates useful knowledge and acquires them as he or she uses the system. For example, WebLEAP[13-16] displays the frequency data of English word sequences in a graphical way so that the users can analyze and compare them easily, and extract useful knowledge for acquisition. The detailed features of WebLEAP and its educationality will be explained and discussed in Section 2.

For considering educationality, we take attention to various issues relating to learning such as:

- Motivation:
  Motivation is the most important driving force to learning. Users will not even try to acquire knowledge without it. Other aspects like curiosity, learning mind or recognition of importance of learning, interest in the subject, etc. are also important for learning and knowledge acquisition.
- Learning Style:
  The typical learning style is to read/listen, understand, and learn by heart. This style is a typical one and good for learning authorized knowledge. We can call this style of learning also by knowledge transferring. On the other hand, research type of learning style is suitable in active learning. In active learning learners are supposed to investigate, analyze, and think based on their own motivation, curiosity, and others.
- Planning and Evaluation:
  In order to acquire knowledge more effectively and more efficiently, it is very important to make a learning plan in advance and then carry out the plan. It is also important to evaluate the result. From this point of view, understanding one's knowledge level is crucially important. One can make better plan if he or she knows oneself on what sort of knowledge should be learnt and what sort of learning strategy is best-suited to him or her.

If we put focus on active learning style, which is the major topic in this paper, the ability for analyzing the material, facts, data, and so on, are most important. For example, in WebLEAP, users are supposed to compare the frequencies of expressions and make effort to understand its meaning. In order to do such analysis, they have to think and make full use of lexical, grammatical, and other knowledge, they have already learnt. We will call this kind of thinking "active thinking" in this paper. The users can learn only through active thinking and active learning process.

## 2 WebLEAP and its Educational Aspects

In this section we take WebLEAP (Web Language Expression Assistant Program) as an example system that has facilities for educationality. This system has been developed for helping users who are non-native English speakers with writing documents in English by providing information about usage of English expressions. It provides the frequencies of documents in the Web (i.e. World Wide Web) that contain the given word sequence. Precisely, the user gives a sentence or an English expression to the system and the system displays the frequencies of all the subsequences of consecutive words that are included in the given expression, where the number of words of the subsequences is given in advance as a range of numbers.

### 2.1 The WebLEAP System

The WebLEAP system consists of several windows such as main window, draw window, KWIC window, browser window and some others[17]. In this section we will show two windows and explain general functions of the system from the active thinking point of view.

**The Draw Window**

A screen shot of draw window of WebLEAP is shown in Fig. 2. In the central part of the figure are two English expressions and their frequency displays. The frequencies are displayed graphically by using bars. The spans of the bars indicate the corresponding word sequences and the numbers in them indicate the frequencies in the Web. The frequency data is given by a Web search engine; actually Google[4,5] search engine in this case. Other search engines[2,6,12] can be applicable as well. The system uses the collection of Web documents like a huge corpus(e.g. [3]). So we call it Web-corpus. The colours of the bars suggest their magnitude; pink for small numbers and blue for big ones, for example.

When we use the system we first put an English expression to the text input field of the main window, and then push the "eval" button. Then the result is displayed in the draw window like in Fig. 2. We can specify the shortest and the longest length of subsequences of consecutive words of the given expression in the main window. We can also keep the old result in the draw window by pushing the down button in the main window so that we can make an empty space in the topmost part for another experiment.

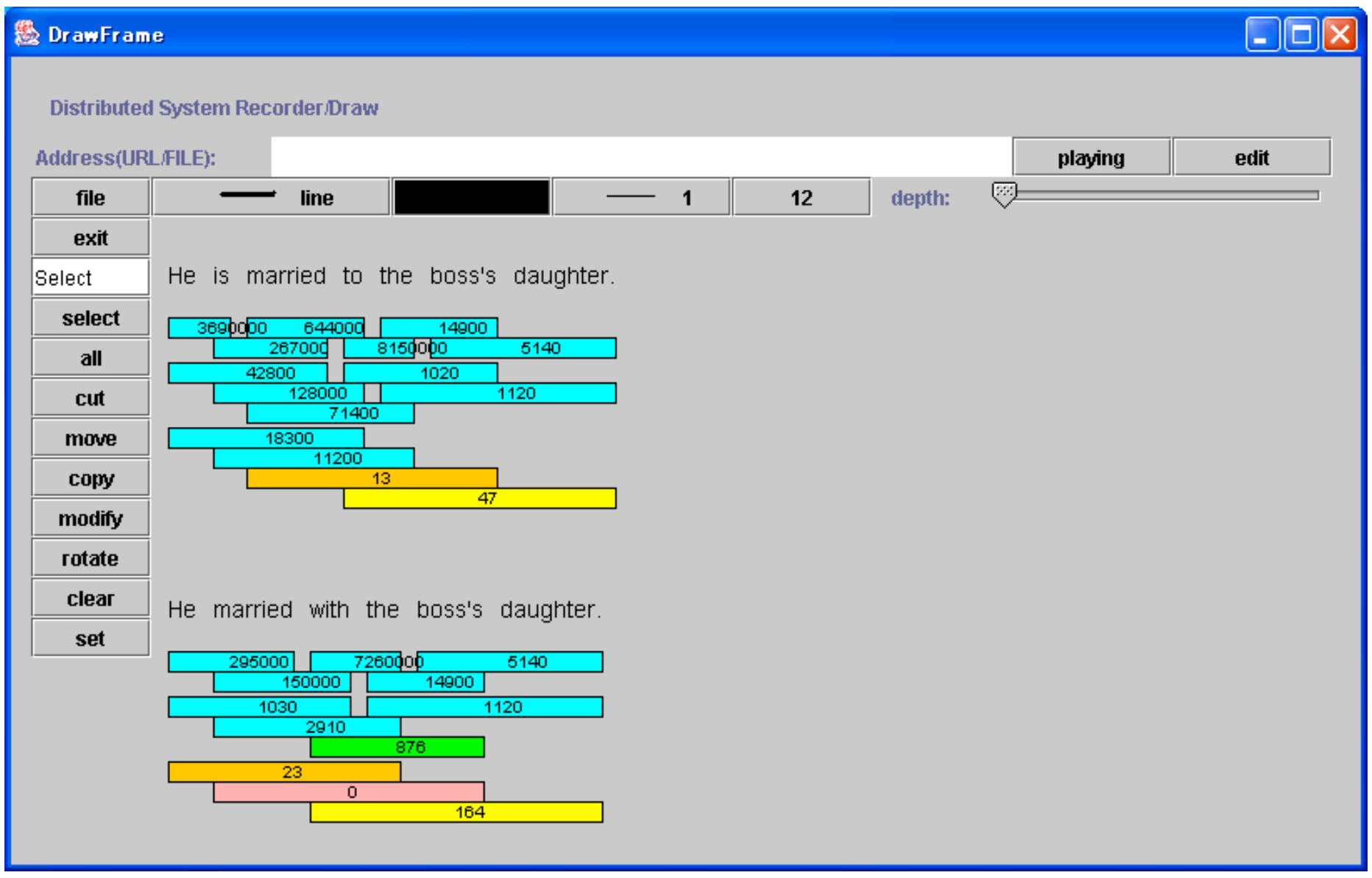

**Fig. 2.** Draw Window of WebLEAP.

The process of the WebLEAP system goes as follows:

1. Users give an English sentence or an expression to the system.
2. The system creates all the subsequences of consecutive words included in the given sentence or expression. The range of length, i.e. the number of words of the subsequence, is given in advance.
3. The system throws the subsequences of the given expression to a search engine, and gets the frequency numbers of them from the returned data.
4. The system displays the frequencies in a graphical way so that the users can easily read the numbers and compare them.

Fig. 2 is an example for comparing “He is married to the boss’s daughter” and “He married with the boss’s daughter.” The intention of this example is to find out which expression is better to use by comparing “is married to” and “married with.” In this case the frequency for “married to” is 640 thousand whereas that of “married with” is 150 thousand. Therefore former one is about four times as large. Further the frequency for “is married to” is about 130 thousand, which is close to that of “married with.” Considering there are expressions “are married to” and “am married to,” the number for “be married to” must be much bigger than “married with.” From these considerations we can estimate that "is married to" is more popularly used than "married with" and therefore better to use.

### The KWIC Window

WebLEAP has the KWIC (Key Word in Context) window, where we can see how the given expression or word sequence is used in the actual context; what words are used before and after the expression or the word sequence. This window is launched when we click the coloured bar of the draw window. The specified key words are displayed in bold letters in the sentences at the right column. A click on a URL column invokes the Web page display window for it.

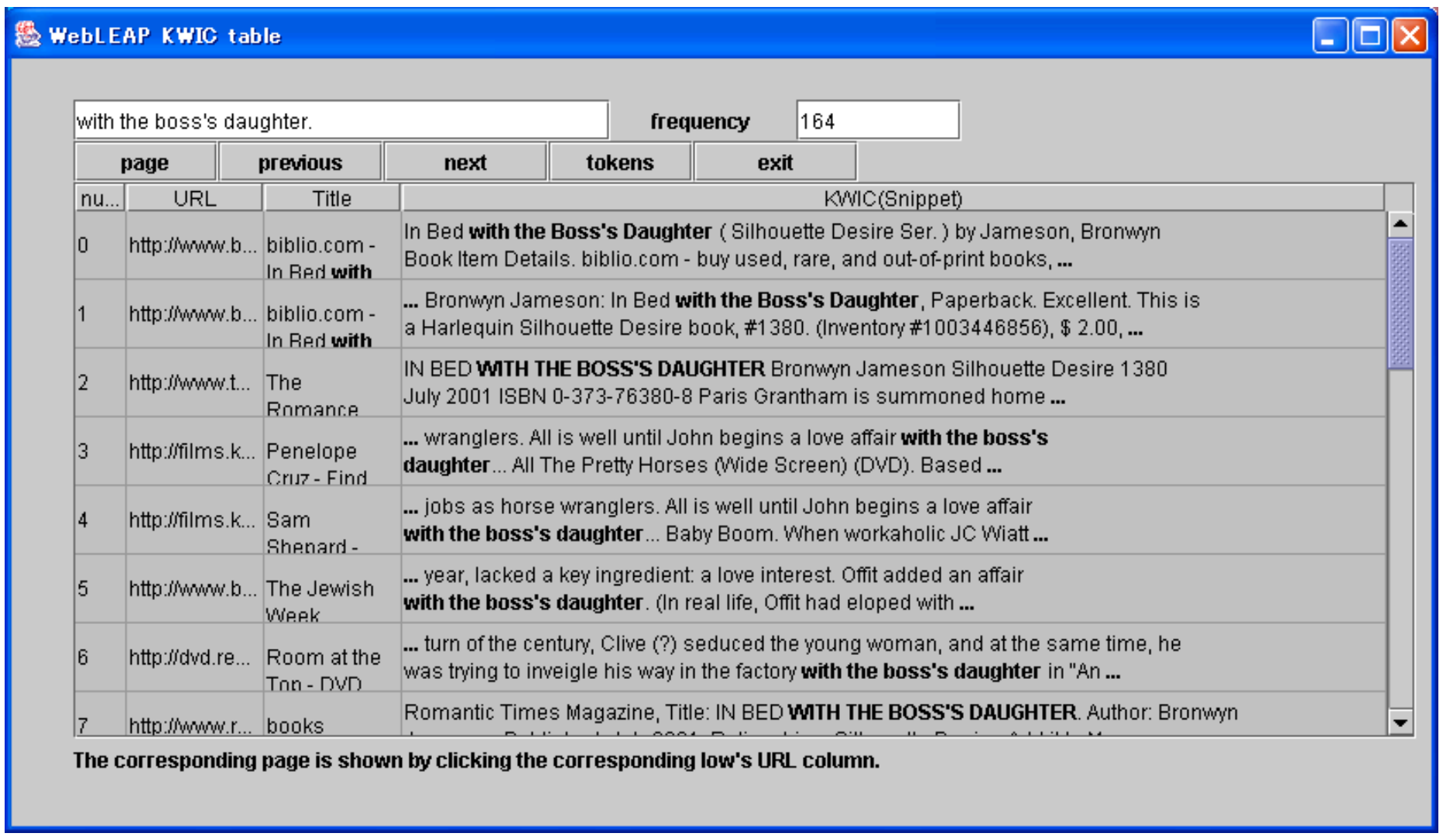

**Fig. 3.** KWIC Table Window

## 2.2 Example: Estimating the Most Appropriate Preposition

Choosing a preposition, which is best suited for an expression in a context that the user supposes, is one of the most difficult things for many Japanese people, and probably for other non-native speakers of English as well. Take, for example, the expression "your own risk." Our question is which preposition is the most appropriate for this. Will it be "by," "with," or "at"?

Fig. 4 is a result by WebLEAP for this question. Let us have a look at the sixth row that spans for 4 consecutive words. The frequencies are 41, 138, and 434,000 for the prepositions "by," "with," and "at," respectively. From this result it is easy to see that "at your own risk" is the most appropriate one.

Suppose further here that we do not have the preposition “at” in mind at first. By comparing the numbers only for “by” and “with,” we might conclude to choose “with” in this case. However if we are careful enough to see the number 456,000 for “your own risk,” we will recognize that 138 is too small for this large number. One of the good actions in such a case is to click the bar for “your own risk” and see the KWIC table window, which is shown in Fig. 5.

By the sentences, which are displayed in the table, we see that “at” is used most of the case. So we can speculate “at” is the most appropriate preposition for “your own risk.” Then we can ask the frequency for “at your own risk” and get the number 434, 000, which is 95% of cases for “your own risk.” So we can conclude with more confidence that the preposition “at” should be used here and thus “at your own risk” is the most appropriate expression.

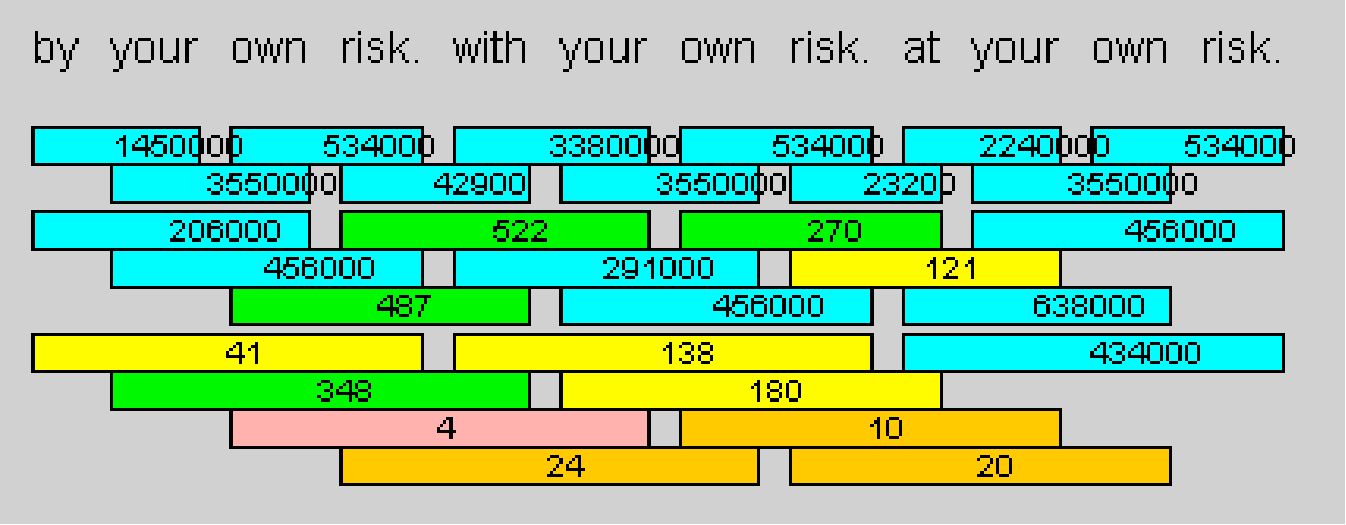

**Fig. 4.** by/with/at your own risk

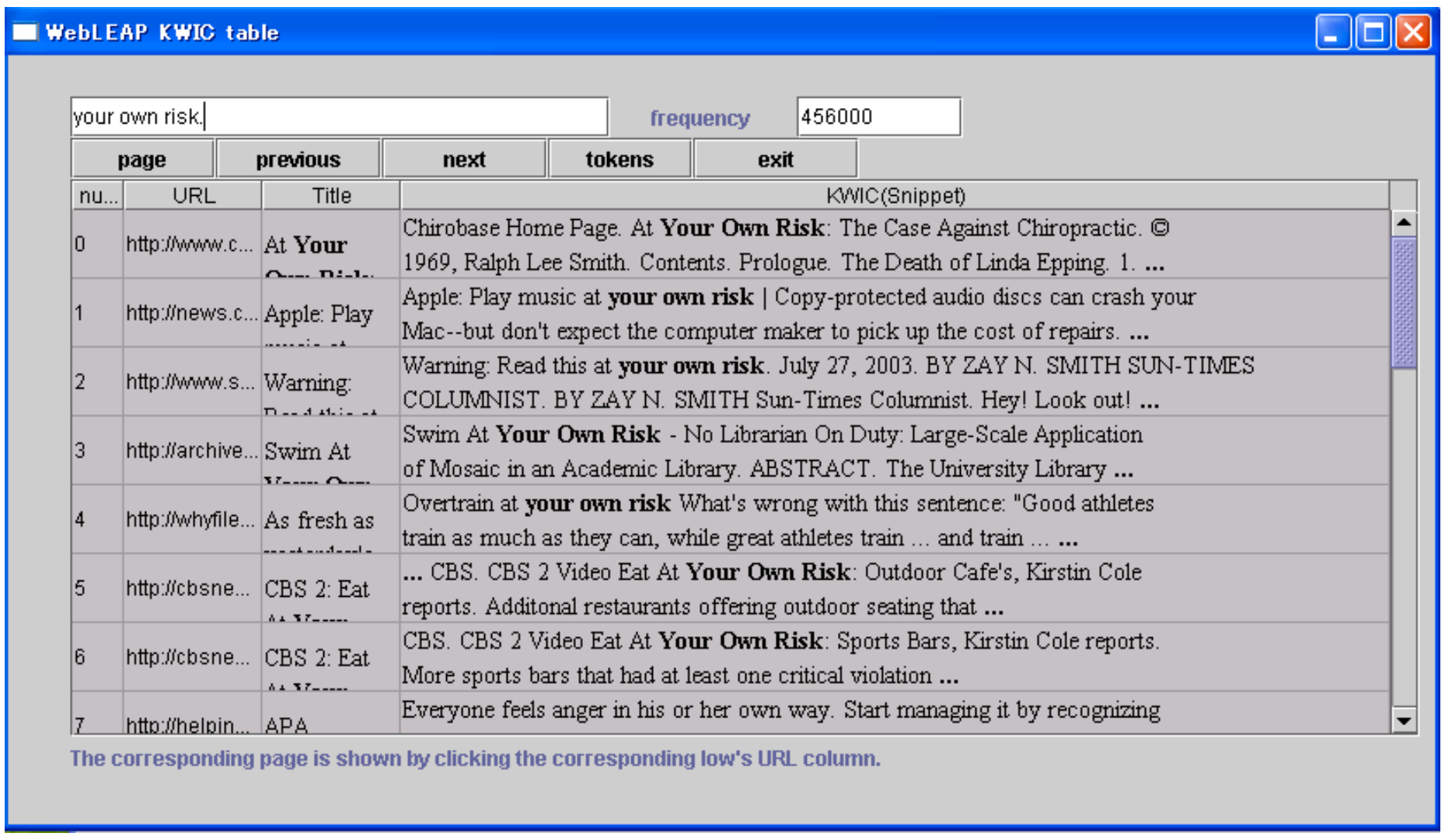

**Fig. 5.** Choosing a Preposition for “your own risk” in the Draw Window

## 2.3 Example: Comparing UK English and US English Expressions

Non-native English speakers are often confused when they are writing a sentence in a specific English dialect such as British English or American English. The Web-LEAP system has ability to filter the Web corpus by a domain name in the page's URL. With the specification of the domain, we can get the frequencies in the specified domain and will be able to analyze the differences of usage from domain to domain. This facility helps the user with writing an English sentence in a specific English dialect or to avoid using it.

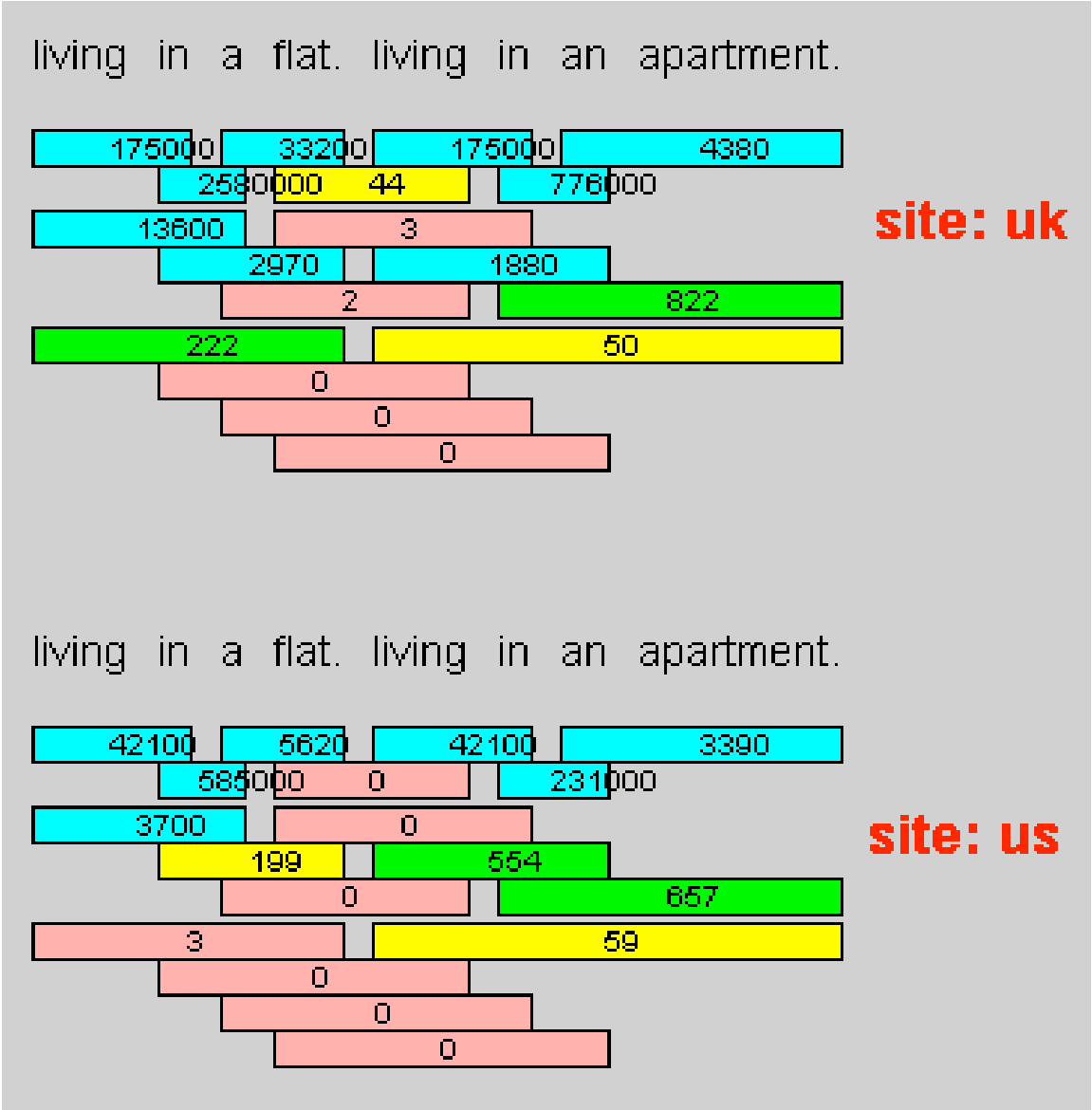

**Fig. 6.** Comparison of Usage of "flat" and "apartment" for "uk" and "us" Domains.

Fig. 6 shows WebLEAP outputs for comparing two sentences "living in a flat." and "living in an apartment." in the UK domain(site: uk) and the US domain(site: us). This figure shows that "living in a flat" is used much more than "living in an apartment" in the UK domain(to be precise, about 4 to 1), and "living in an apartment" is used much more than "living in a flat" in the US domain(about 20 to 1).

To investigate further, we compare the shorter expressions. In UK domain, the rate is rather stable; 4 to 1 for "in a flat" and "in an apartment" and 8 to 1 for "a flat" and "an apartment." From these results, we can conclude that "flat" is more preferably used to "apartment" in UK for expressing the living place.

On the other hand in the US domain, it is not as stable as in the UK domain. The ratio is 7 to 1 for "in an apartment" and "in a flat," and 1 to 2 for "an apartment" and "a flat" in the US domain. Thus, in a phrase with "living in," "apartment" is highly preferred to "flat" whereas as a raw word "flat" is more used than "apartment." One possible reason this result suggests is that English in the US domain consists of various types (or dialects) of English. Maybe “flat” is used much in other meanings, like in “flat tire.” We need to investigate further in order to know its true reason.

## 3 SASS and its Educational Aspects

In this section we take SASS (Searching Assistant with Social Selection)[8, 9] as another example system for educationality. SASS was developed as a keyword recommendation system for those who use search engines. It presents a list of search keywords that might be relating to the search intention of the users who does not have a clear search intension, i.e. in the level of conscious need[10]. One or a couple of recommended keywords are supposed to match to the keywords that the users are trying to find. We have found that this system is also educational soon as we use it as a trial.

This section is organized as follows. In Section 3.1, we give a brief explanation of the SASS system in its keyword recommendation feature. In Section 3.2, we describe the multi-path recommendation feature, which is one of the key features of educationality of SASS. And in Section 3.3 we discuss and summarize the educational aspects of SASS.

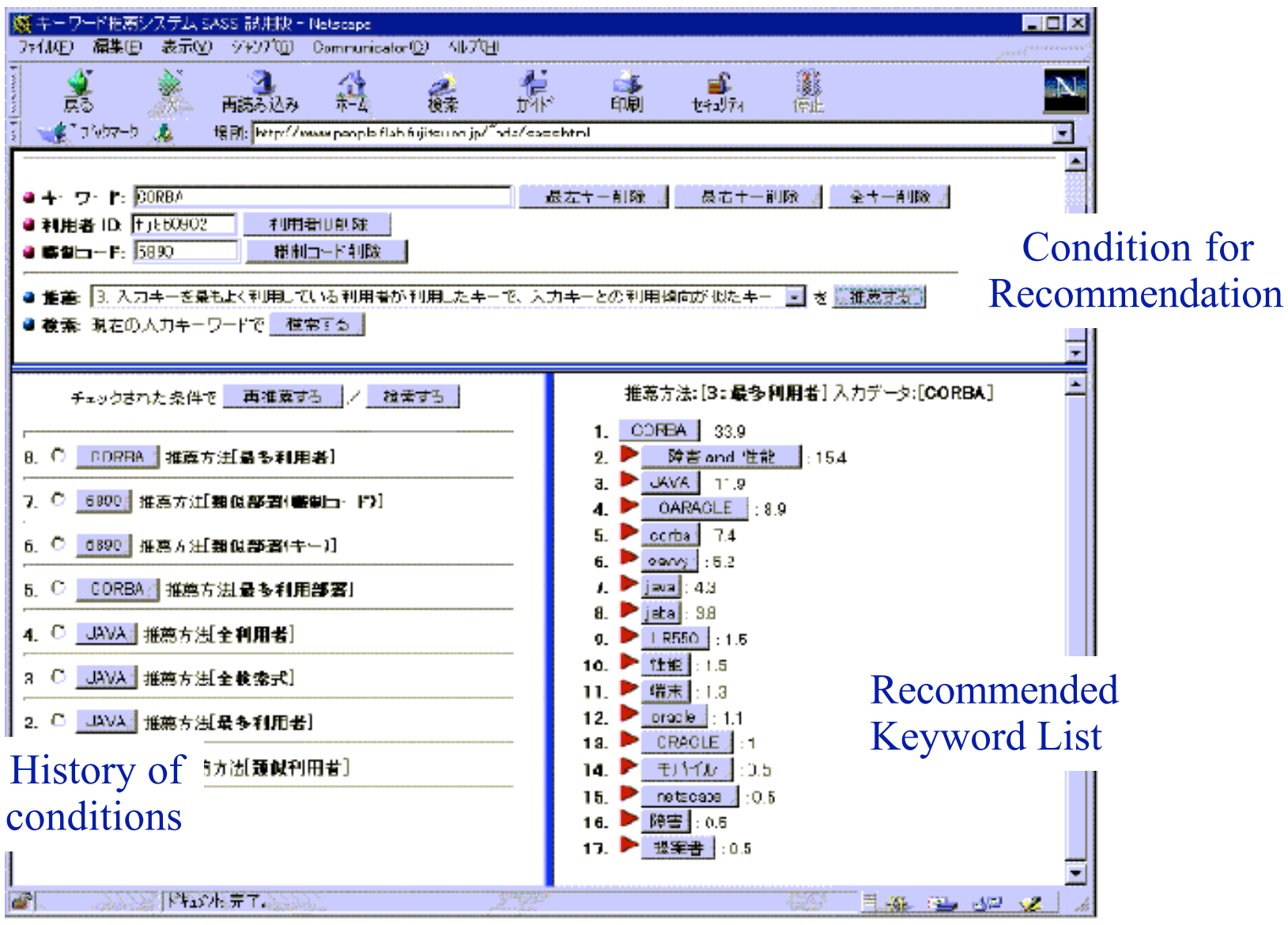

**Fig. 7.** Main Window of SASS System.

### 3.1 The SASS System

A screenshot of the SASS system is shown in Fig. 7. There are three panes in the window; the upper pane, the lower-right pane and the lower-left pane.

The upper pane is the area for giving conditions for recommendation, such as search formula, the user ID, its affiliation number, the specification of the recommendation path, and so on. Most of the cases the search formula is one or a couple of keywords, which is used as the original keywords for recommendation In this example, the keyword is “CORBA,” user ID is “fj860902,” affiliation number is “5890” of the company and the recommendation type is the one that uses the similarity between the users. We will take up and describe in detail about the multi-path recommendation in Section 3.2.

In the lower-right pane is the recommended keyword list. In this example the list starts from the original keyword “CORBA” followed by other keywords like “trouble and performance,” “JAVA,” “ORACLE” and so on.

The lower-left pane is for recording the history of the searching process so that the user can move back to the former status in the search session.

The keywords recommended by the system are chosen in a different way in SASS from other keyword recommendation systems. SASS system uses the database of search keywords that have been used by the users of SASS. The data can be collected from some search engine databases as well. In SASS system, the recommended keywords are chosen based on the co-occurrence in search formulas, whereas in other systems they use the document database and choose the keywords based on the co-occurrence in documents.

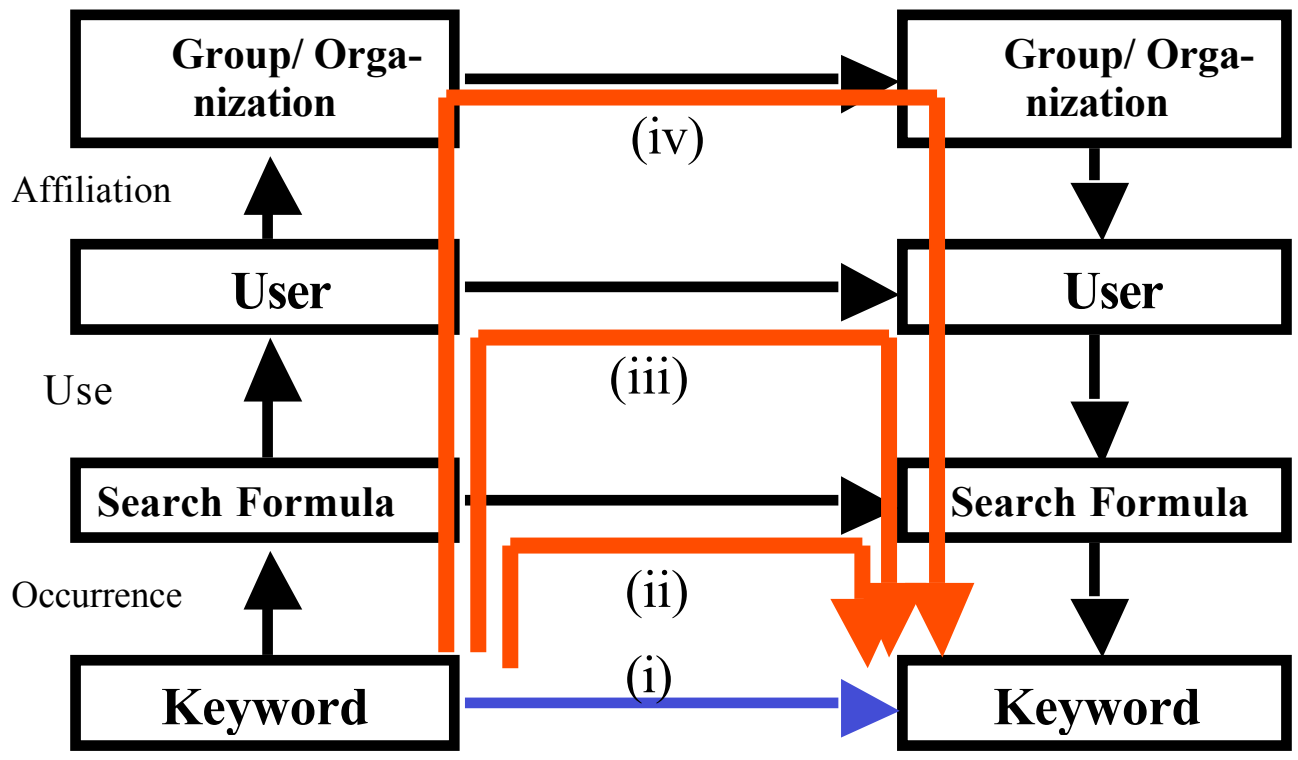

**Fig. 8.** Multi-Path Recommendation.

### 3.2 Multi-Path Recommendation

One of the characteristic features of the recommendation given by SASS is that it provides recommendations from various points of view. We will explain about this feature in this section.

The hierarchical structure of the components for the keyword recommendations in SASS is shown in Fig. 8. At the left columns are the "keyword," "search formula," "user" and "group/organization" from the bottom to the top. The relationship between "keyword" and "search formula" is the occurrence relation; the relationship if the keyword occurs in the search formula. The next one is if the search formula is used by the user. The last one is if the user belongs to the group or the organization.

The relationship between the left and right corresponding components are based on the similarity of keywords, search formulas, users and groups, respectively. The relationship between keywords is the similarity of these two, possibly the one based on the co-occurrence relationship. The similarity between the left and the right search formulas is the one is also based on how these formulas include similar keywords. For the users, similarity comes from how they are similar in terms of the keywords they use in their searching activities. Similarity between groups and organizations also comes in the same way.

By using Fig. 8, we describe how to realize the multi-path recommendation. By having a look at the figure we recognize that there are four paths beginning from the lower-left keyword to the lower-right one. The first path is the direct link from keyword to keyword. We can take this path by using any similarity definition, including the document-based similarity definition. The second one is the path that goes through the similarity of search formulas. This means that the right keyword is in the recommendation list if the search formula that contains the left keyword and the ones that contain the right keyword are similar in its nature. Similarly, for the third path, the recommended keywords are chosen among the ones from the users who are similar in the use of keywords to the user who uses the left keyword. The similar rule can be applied to the fourth path as well.

### 3.3 Educational Aspects of SASS

The SASS system is not only a system that helps us users with finding appropriate search keywords but also is educational in some aspects such as:

(1) Know Your Knowledge Level
(2) Make Study Menu
(3) Know Who

Suppose we want to find some information that relates to the keyword A. We put A as the root keyword to SASS. Let B1, B2, …, Bn are the keywords returned from the system. These keywords are chosen because they have been used together with the keyword A in one reason or another by many users. If you can guess why B1, B2, …, Bn are used in relation with A, you can say that you know fairly well about the field relating to A. On the other hand, if you can guess the relationship of Bs except B1 and B2, it probably means that your next target to be learned should be B1 and B2. This is an interesting way of using the keyword recommendation system SASS as an educational system.

In this way we can use SASS for making a study menu for a field. We choose some typical technical words from the field and put them in SASS. Then we have a list of words relating to the given words, which might be other relating words in the field. We can use this list for checking out our knowledge level. If we already know most of the words then we can say that we know fairly well about the field. If not, we will recognize that we have to learn more about the field.

By using the multi-path recommendation facility, we can get several lists of related words. Even if we know what it means for a specific word, we may not be able of explain why it is in the recommended words. Why is it in the list from the path of similar search formulas, users, or groups? It is a good study to try to find good explanations to these questions.

The words that we don't know indicate what knowledge we lack, i.e. what we have to learn in the field. We can make our study menu by opening a textbook and find the chapters that explains the words what we don't know.

We can use the hierarchy of Fig. 8 in different ways. For example, we start with the left bottom keyword and go up to the user and go right to the similar user. Then we have a mapping from a keyword to a user. This mapping can be used for "know who." If we choose a typical group of words from a specific field, we can get a list of user names who relate to the field. Maybe we can ask some of them about the field. Some of them may give us lectures on some topics of the field.

## 4 Discussion

The concept of educationality and active thinking is a different criterion in evaluating education systems from the evaluation method taken in most of education systems. The essential difference comes from the existence of intended knowledge that is supposed to be learnt by the users.

In the ordinary education system the teachers or system designers prepare a set of knowledge and organize them in advance. In such systems, one of the biggest issues is how to let the users learn such specific knowledge more effectively and more efficiently.

In educationality point of view, the most important issue about the system is how to support users effectively and efficiently find useful knowledge by themselves and acquire it so that such knowledge can be used in appropriate situations.

Originally the aim of developing the WebLEAP system is to provide the users with frequency data so that they are able to know if some expressions are popularly used or not. Soon we found that the system has another feature as an educational system.

One of the big issues in educationality is the granularity of data or information given to the user. The frequency data are good examples of appropriate data for the users to think from it. If the system gives the authorized knowledge (see Fig. 1) the users tend to memorize it without thinking further. If the system just gives the raw data, a lot of English documents for example in this case, users might not be able to extract any useful knowledge from them.

Profoundness, or the flexibility of understanding level, is another important feature for the data/information given to the user. Any user can use the frequency data according to their purposes and their level on English skill. If you are in the beginner level of English, you can just compare two expressions and know which one is more popularly used. If you are in the middle or advanced level, you can investigate further and get better knowledge by fully utilizing the system.

Also the SASS system can be used by any user in a variety of knowledge level. You can learn differently according to what words you know and how much you know.

Importance of such systems is growing as we are in the IT age and therefore we are supposed to keep learning as long as we live; i.e. life-long learning[. The useful knowledge in these days varies from time to time and occasion to occasion. Even a good teacher cannot prepare all the useful knowledge in advance. We have to find or create such knowledge from our everyday job, which is called OJT (on the job training)[7]. This is an essential skill in addition to the learning skill in the ordinary sense in order to live in these days and in the future.

## 5 Concluding Remarks

In this paper we have dealt with the educationality of knowledge acquisition support systems. The models of learning styles that are assumed in most educational systems in the field of CAI (Computer Assisted Instruction) or CAL (Computer Assisted Learning) are either knowledge transfer or simulation-based learning[1]. In either case, the major concern of education is how efficiently the learners can acquire the intended knowledge.

We have presented a new model type of learning style. The learners are supposed to actively think and achieve their goals under the support from the system like WebLEAP and SASS. The knowledge that is learnt by the users is not prepared in advance. It is focused and acquired by the users, thus the acquired knowledge varies from one person to another.

The actual active learning styles vary from system to system and situation to situation. In WebLEAP system the users have to think and have many trials in order to find out rules and theories behind the frequency numbers of word sequences. In SASS system the users have to think hard and have additional learning and search in order to find out relationships between word and word, word and people, people and people, and so on.

The key issue is the granularity, or the level, of materials given to the users by the system. In this model, the system that gives too high or low level of abstraction to the learners is not a good system for education. Materials in too high abstract level are knowledge themselves and thus users are easy to try to just memorize them. Those in too low abstract level are kind of raw data so that it is quite difficult for the user to interpret what these data are trying to say.

In this respect, the frequency data by WebLEAP and relationship data by SASS are in the middle abstract level and are appropriate for the users to think and try to extract some rules from them. We have to investigate further what sort of materials are in good abstract level and thus effective for active learning. If the system can drive the learners to think more and think deep, it becomes better educational system.

This point of evaluation of a computer system can be extended widely to other interactive systems as well. An interactive system is called educational if it is designed to intend to help its users learn more efficiently and more effectively than without it. This gives a new standpoint, which we call "educationality", to software evaluation. Ideally all the interactive systems should be evaluated from this view and the users would learn as they use such systems with high educationality. Such a system can be called a knowledge acquisition support system or a computer assisted knowledge acquisition system.

## Acknowledgment

The author appreciates his coworkers Professor Yamanoue of Kagoshima University and Professor Ruxton of Kyushu Institute of Technology for their contributions and discussions on WebLEAP. He thanks to his daughter Mariko Minami for her valuable comments on the draft of the paper.

# A Comparative Study on Cost and Benefit of Capturing Design Rationale

Yoshikiyo Kato[1] and Koichi Hori[2]

[1] The Institute of Space Technology and Aeronautics
Japan Aerospace Exploration Agency
2-1-1 Sengen, Tsukuba, Ibaraki 305-8505, Japan
kato.yoshikiyo@jaxa.jp

[2] Research Center for Advanced Science and Technology, The University of Tokyo
4-6-1 Komaba, Meguro-ku, Tokyo 153-8904, Japan
hori@ai.rcast.u-tokyo.ac.jp

**Abstract.** Although the importance of recording design rationales has been recognized, it has not been widely accepted in practice. The authors has been proposing the concept of knowledge recycling to overcome the capture bottleneck problem, and developed a system to capture design rationale from e-mail communication by annotation.
This paper reports a user study which was conducted to compare the cost-effectiveness of different design rationale notations, including the annnotation scheme we have developed in implementing the system.

## 1 Introduction

The modern society relies on various large complex systems. In order to design, construct, and operate such systems efficiently and reliably, all of those who engage in such activities should share the knowledge about the systems. One of the important knowledge which should be employed when one is engaged in any activities related to a large complex system is *design rationale.*

Design rationale is a theoretical basis or designer's intention behind a design decision of a system. Not only is it useful for system designers to understand the designs done by other designers, but also is it important for those who engage in the phases after design, i.e. construction, test, operation, to better understand the system they are dealing with.

Although the importance of recording design rationales has been recognized, it has not been widely accepted in practice. The major factors hindering the acceptance of the practice of recording design rationales are 1) the high cost for recording design rationales, 2) the effect of recording design rationales not being readily perceivable, 3) the difficulty of utilizing recorded design rationales.

The capture and use of design rationale has become a major topic in design research since 1970's [1]. However, it has been reported that there is a tremendous cost associated with capturing design rationales formally [2–4]. Similar problem is recognized in the domain of knowledge management as *capture bottleneck* [5]. Capture bottleneck problem in the knowledge management domain refers to the lack of enough incentives for the users to invest time for sharing their knowledge or resources.

The authors have been proposing the concept of *knowledge recycling* as an approach to overcome the capture bottleneck problem in knowledge management, especially targeting at development of large complex systems [6]. The basic idea of knowledge recycling is to utilize available information produced in the course of development process in order to minimize the input cost, thus relaxing the capture bottleneck.

Along this approach, we developed a system to capture design rationale from e-mail communication [7]. Nowadays, a lot of communication is done via e-mails. As a result, there is a good chance that some design rationale be included in e-mails. However, if it was stored in its raw form, there would be a limited use for such information. Instead of having users produce a separate representation apart from the original e-mails, we took an approach to let users annotate e-mails to represent design rationales.

In implementing the system, we had to devise an annotation scheme. In designing the annotation scheme, we have decided to make it as simple as possible, so that the input cost is minimized. Since our focus was on capturing issues on the system's design or design decisions appearing in e-mail communications, the annotation scheme was named *Issue/Decision model*.

Of course, annotating documents still requires cost. The hypotheses we had were that annotation requires less cost than creating a whole new design rationale representation does, and that the resulting representation of annotations is as useful as that of creating a separate one.

This paper reports the study for the comparison of cost-effectiveness of the annotation scheme the authors devised and existing design rationale notations, i.e. IBIS and QOC. Section 2 briefly describes each design rationale notation to be compared. Section 3 explains the user study, and its results are presented in Section 4. In Section 5, the results is discussed, and we conclude in section 6.

## 2 Design Rationale Notations

In this section, overview of two design rationale notations, i.e. IBIS and QOC, and the proposed annotation scheme is described.

### 2.1 IBIS

IBIS stands for *Issue-Based Information System*. It was first proposed by Kunz and Rittel [8] as an argumentative problem solving model for decision-making in a complex problem setting with multiple stakeholders involved. Various software tools have been developed based on IBIS: gIBIS for capturing design rationale [9, 2] or HERMES for supporting decision-making [10], for example.

IBIS uses three types of nodes to represent design rationale: issue, position and argument. Issue node represents an issue to be discussed or a question about a topic. For an issue, there can be several alternative *positions* which solves or answers to the issue. Each position has *arguments*, which can either support or oppose to the position with regard to the issue in concern. Relationships between nodes are represented as links with label such as `generalizes`, `specializes`, `responds-to`, `questions`, `is-suggested-by`, `supports` and `objects-to`. An example of design rationale represented with IBIS is shown in Figure 1.

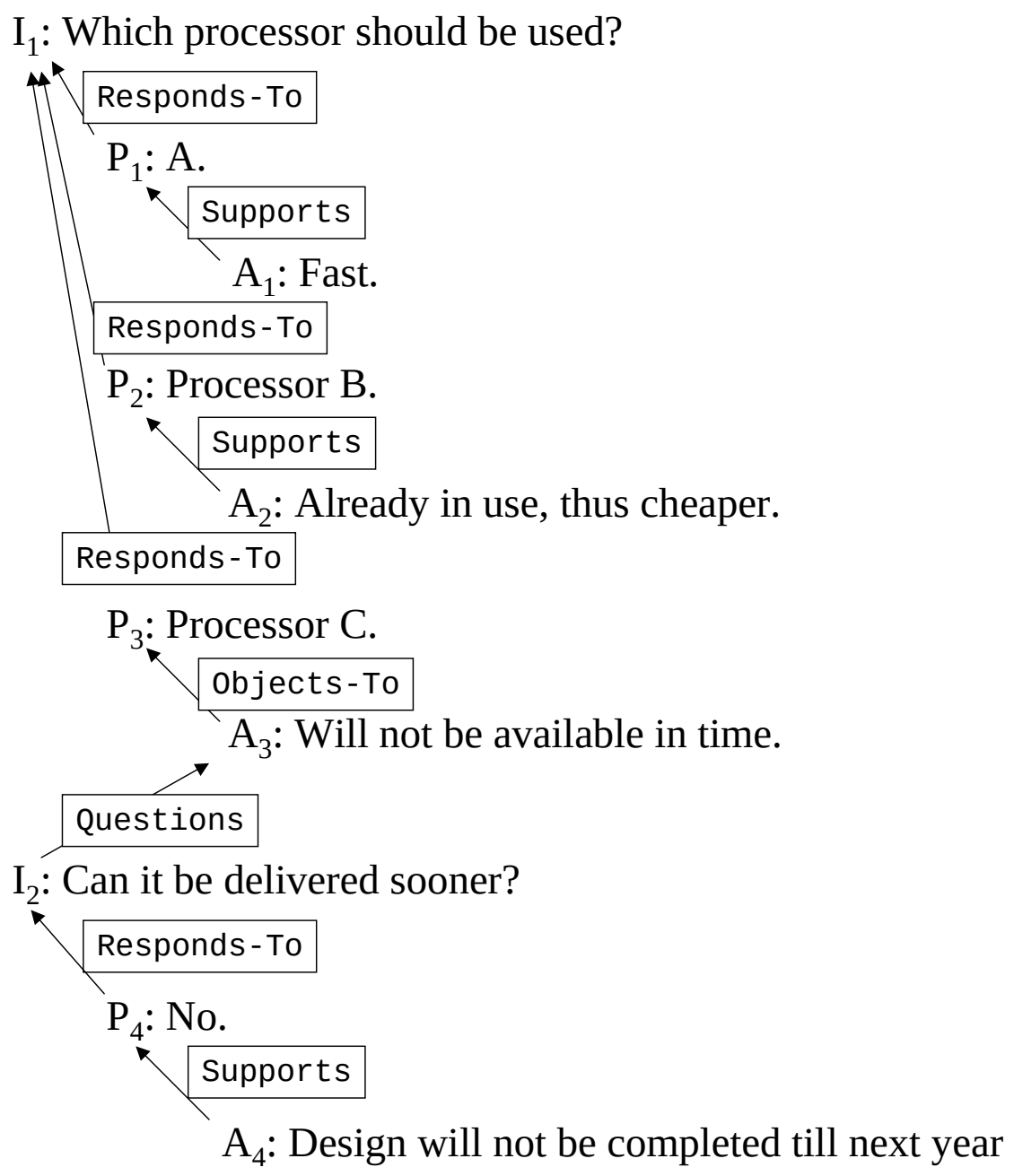

**Fig. 1.** An example of design rationale represented in IBIS. Adapted from [2].

### 2.2 QOC

QOC stands for *Questions, Options, and Criteria.* QOC was proposed by MacLean et al. [11] as a semiformal notation for design rationale. It was introduced to be used in a style of analysis called *Design Space Analysis*, in which one considers possible alternative designs of an artifact and understands the artifact in terms of its relationship to the alternative designs.

QOC, as its name suggests, uses three elements: questions, options, and criteria. *Questions* have to be asked so that alternative *options* be answers to the question. Criteria are used to evaluate and choose among the options. For a question, there is a set of options and a set of criteria. To denote evaluation of an option on a criterion, a line is drawn between them, where a solid line indicates positive assessment and a dashed line indicates negative assessment. While IBIS records the process of design, QOC is thought of as a *coproduct* of design. An example of design rationale represented with QOC is shown in Figure 2.

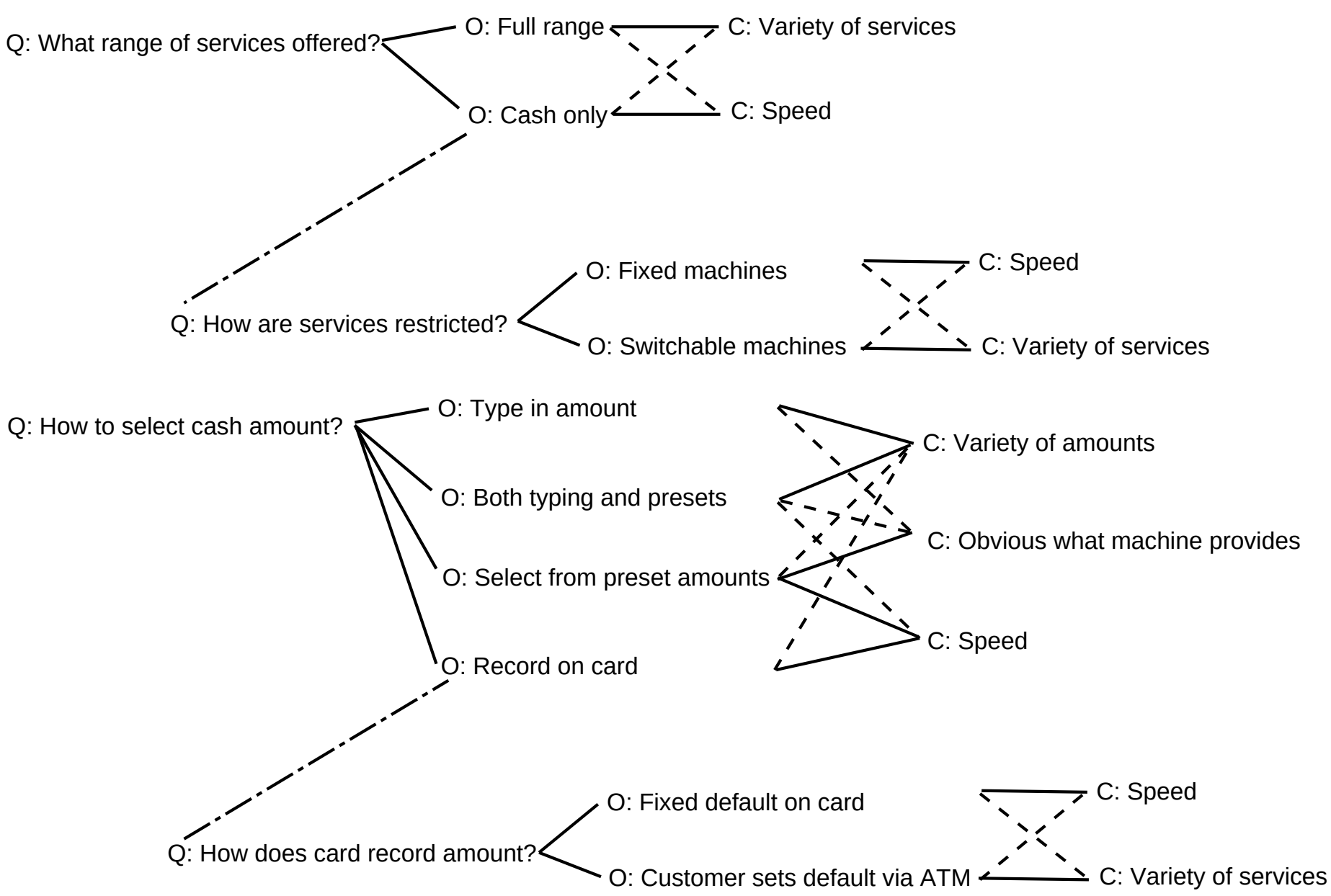

**Fig. 2.** An example of design rationale represented in QOC. This example represents an ATM design. (Adapted from [11].)

### 2.3 Issue/Decision Model

The annotation scheme devised in this study is called Issue/Decision (I/D) model. I/D model can be thought of as a simplified version of existing design rationale notations, such as IBIS [8, 13] or QOC [4]. Simple representation was adopted to mitigate the cost of formalizing argumentation. I/D model is intended to capture argumentation perspective of design rationale [12].

I/D model consists of two basic elements, which is obvious from its name: *issue* and *decision*. *Issues* are problems or concerns about the design or implementation of the developing system. *Decisions* are resolutions to issues such as trade off between conflicting design parameters. Issues and decisions are represented as annotation to documents.

Issues and decisions can be linked, constructing an argumentation structure. An example of an argumentation structure is shown in the left pane of Fig. 3. Although links between issues and decisions have no labels, we do assume semantics of links according to its source and destination (Table 1). A link from an issue to a decision indicates that some kind of decision has been made on the issue. Inversely, a link from a decision to an issue would mean that a new concern has emerged on the decision that has already been made. A link from an issue to another issue would mean the latter is a sub-issue of the former, or the latter induced the former. A link from a decision to another decision indicates the latter is an auxiliary of the former.

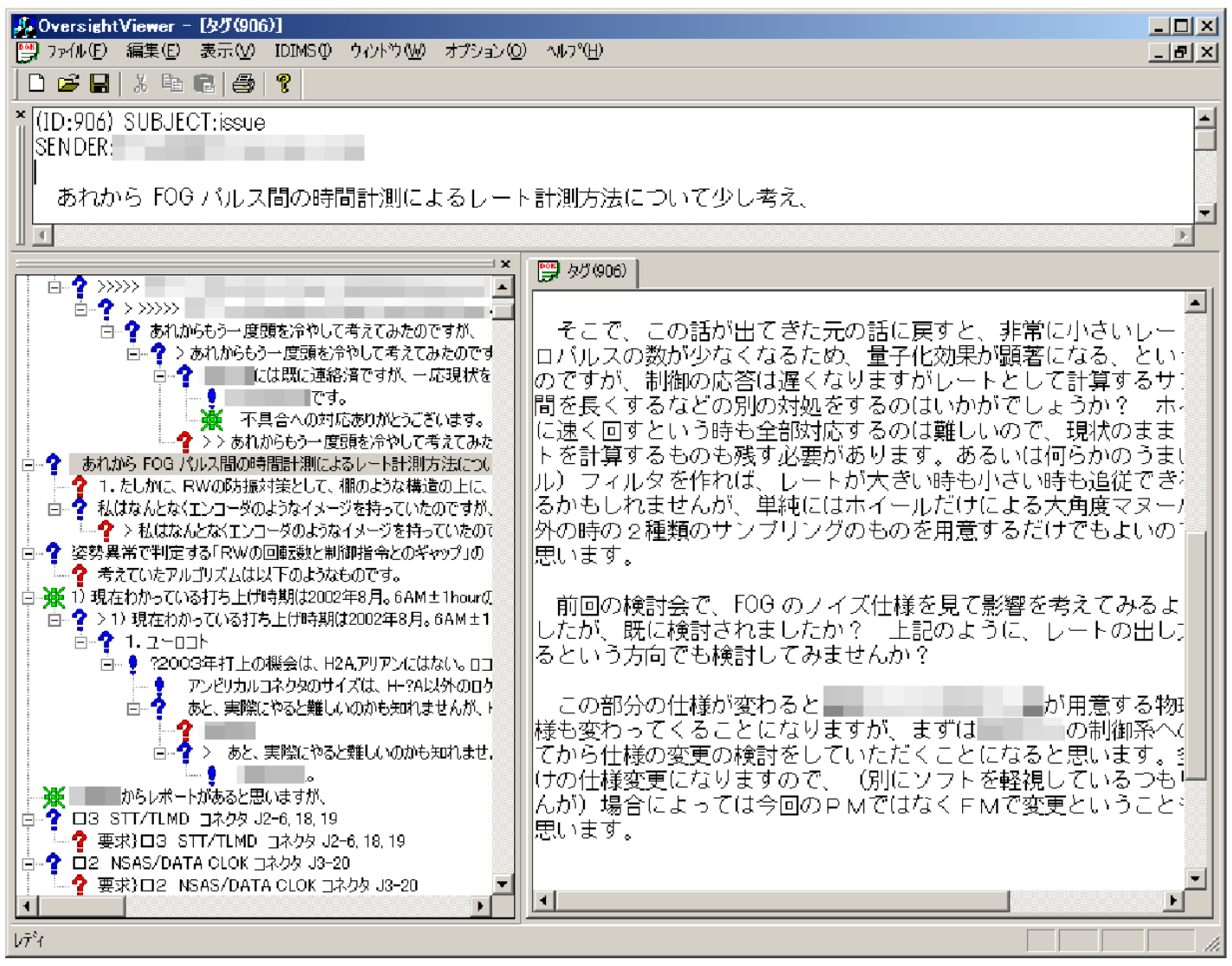

**Fig. 3.** A screen shot of Issue/Decision repository viewer. Tree structure of issues/decisions is shown on the bottom left pane.

**Table 1.** The implications of links in the Issue/Decision model.

| Link Type | Description |
|---|---|
| $I \rightarrow D$ | A decision was made on an issue. |
| $D \rightarrow I$ | A new issue arose on a decision which has already been made. |
| $I_1 \rightarrow I_2$ | The latter issue ($I_2$) is an sub-issue of the former ($I_1$). |
| $D_1 \rightarrow D_2$ | The latter decision ($D_2$) is an auxiliary decision to the former ($D_1$). |

## 3 User Study

### 3.1 Purpose of the User Study

The purpose of the user study is to measure the cost of describing design rationale with different design rationale notations and the effectiveness of them, in order to give justification to the use of I/D model in the system to demonstrate the concept of knowledge recycling.

Issue/Decision model was introduced for cost-effective description of design rationale. The aim of this study is to verify whether the proposed model is indeed cost-effective compared to the existing design rationale notations. For the purpose of comparison, we used IBIS [8, 13] and QOC [4] as they widely appear in the design rationale literature.

The study aims at justifying the use of Issue/Decision model by showing that it is more cost-effective than the other two notations.

### 3.2 Hypotheses

In order to compare the cost-effectiveness of design rationale notations, we measured the cost in terms of the time to construct design rationale, and effectiveness in terms of the number of issues captured by design rationale.

The user study was designed so that the hypotheses described in this section be verified. Through the process of verifying these hypotheses, the cost-effectiveness of the three design rationale notations were evaluated.

**H1 (Cost of description):** When $C_D(\mathcal{D}, \mathcal{N})$ is a cost of describing design rationale for a design $\mathcal{D}$ with a notation $\mathcal{N}$, the following relationship holds:

$$C_D(\mathcal{D}, ID) < C_D(\mathcal{D}, IBIS) < C_D(\mathcal{D}, QOC) \tag{1}$$

where $ID$ represents Issue/Decision model, $IBIS$ represents IBIS notation, and $QOC$ represents QOC notation.

**H2 (Effectiveness of DR notation):** When $N_{risk}(\mathcal{D}, \mathcal{N})$ is a number of risks in a design $\mathcal{D}$ which can be found by inspecting the design rationale of $\mathcal{D}$ represented with a notation $\mathcal{N}$, the following relationship holds:

$$N_{risk}(\mathcal{D}, ID) < N_{risk}(\mathcal{D}, IBIS) < N_{risk}(\mathcal{D}, QOC) \tag{2}$$

**H3 (Efficiency of risk discovery):** When the efficiency of risk discovery of a notation $\mathcal{N}$ for a design $\mathcal{D}$ is defined as:

$$\eta_{risk}(\mathcal{D}, \mathcal{N}) = \frac{N_{risk}(\mathcal{D}, \mathcal{N})}{C_D(\mathcal{D}, \mathcal{N})} \tag{3}$$

the following relationship holds:

$$\begin{cases} \eta_{risk}(\mathcal{D}, ID) > \eta_{risk}(\mathcal{D}, IBIS) \\ \eta_{risk}(\mathcal{D}, ID) > \eta_{risk}(\mathcal{D}, QOC) \end{cases} \tag{4}$$

In this study, benefit of design rationale is measured by number of risks, or issues, captured by design rationale.

### 3.3 Subjects

Nine graduate students majoring in aerospace engineering were selected as subjects. They all had experience of designing a satellite and were actually involved in a satellite development project.

### 3.4 Task and Materials

Material for the study was selected from the aerospace domain. More specifically, e-mail communication between developers of a satellite was used as a base on which design rationale to be constructed. The e-mail was taken from the CubeSat project at the University of Tokyo [14]. The contents of e-mail were concerned with the specification of the radio equipment to be used in the satellite, and several developers from the project team and the manufacturer of the equipment were involved in the discussion.

Given a discourse in the form of e-mail messages, subjects were asked to construct a design rationale using the specified notation. Subjects were given three different discourses and asked to use different notation, i.e. ID, IBIS and QOC, for each of them. This is to exclude the effect of subjects getting familiar with the discourse if they work on the same discourse with different notations.

### 3.5 Evaluation

Cost of description $C_D$ was measured by the time to complete construction of design rationale. Number of discovered risks $N_{risk}$ was measured by a number of issues in the case of I/D or IBIS, and a number of questions in the case of QOC.

### 3.6 Procedure

Subjects had to construct design rationale for three different mail sets A, B, and C, in different design rationale notations. Each subject was assigned mail sets in different order, and was asked to construct design rationale of each discourse in ID, IBIS and QOC, in this order. The order of mail sets given to each subject was arranged so that no mail set-notation pair was worked on twice. Thus, there were 3 mail set, 3 design rationale notations, resulting in 9 instances of design rationale.

Before beginning construction of design rationale, subjects were given instruction on the design rationale notations.

## 4 Results

### 4.1 Data Obtained from the Experiment

The results is shown in Figure 4, 5, 6, and Table 2. All figures show the average values for each mail set and the overall average for each DR notations. Figure 4 shows the average time required to describe design rationale given a mail set. Time is shown in minutes. Figure 5 shows the average number of issues in design rationale described for each mail set and DR notation. For ID and IBIS, nodes described as *issue* were counted as issues. For QOC, *questions* were counted as issues. Figure 6 shows the average minutes per issue for each mail set and DR notation. Table 2 shows p values of t-Test on the average values for each possible combination of three DR notations. Hatched cells indicates the average value for the DR notation pair statistically showed significant difference.

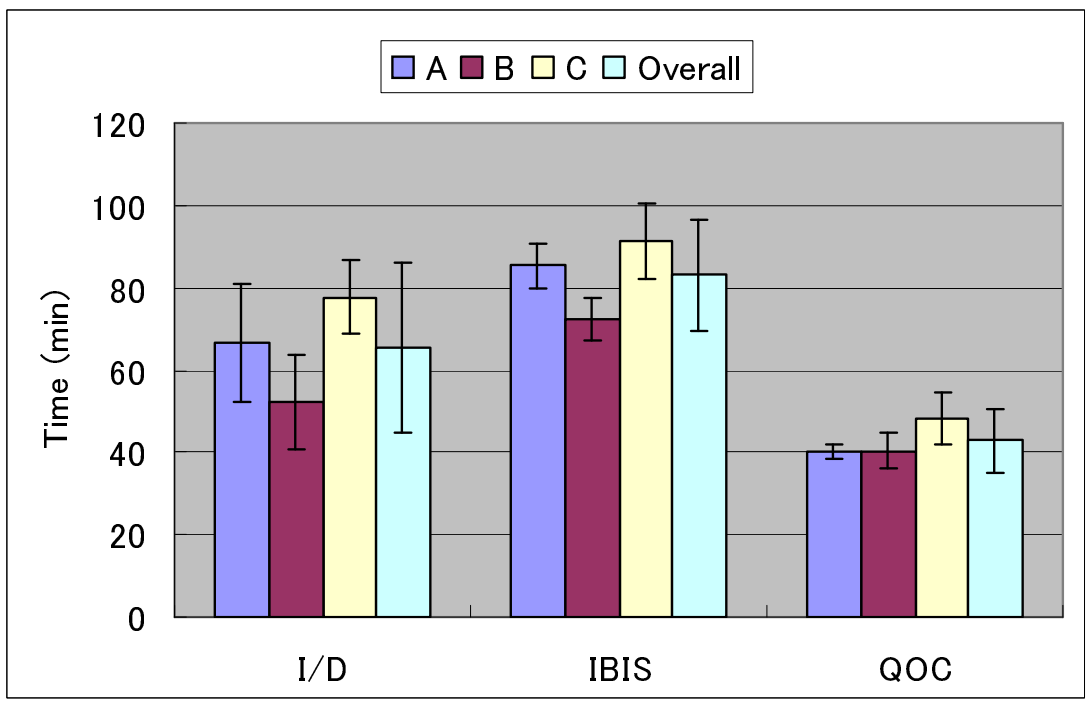

**Fig. 4.** Comparison of average time required to describe design rationale for the same material with different DR notations.

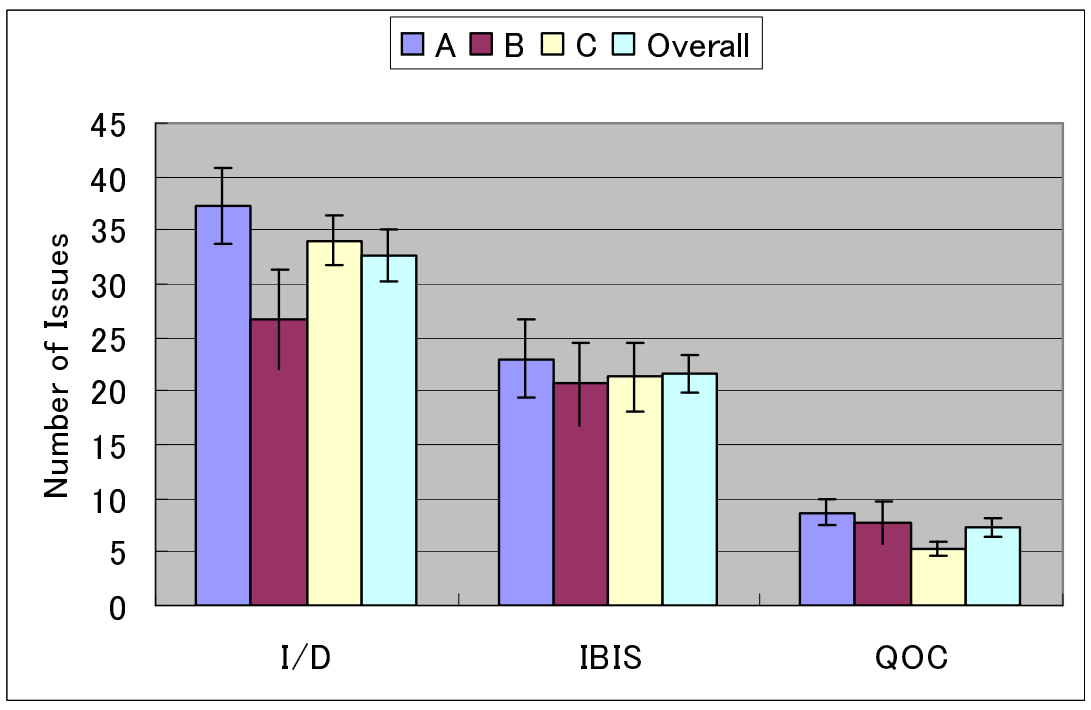

**Fig. 5.** Comparison of average number of issues described for the same material with different DR notations.

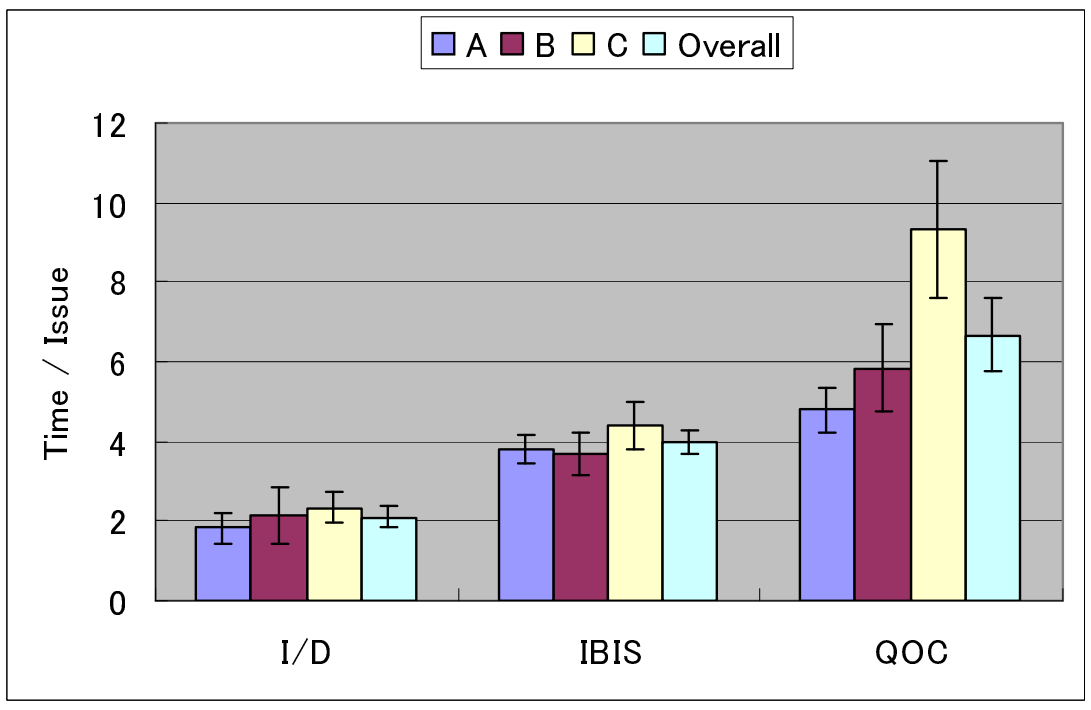

**Fig. 6.** Comparison of average time per issue required to describe design rationale for the same material with different DR notations.

p values

| Pair | Time | # of Issues | Time/Issue |
|---|---|---|---|
| I/D-IBIS | 0.053 | 0.023 | $1.9x10^{-4}$ |
| I/D-QOC | 0.012 | $1.5x10^{-6}$ | $9.1x10^{-4}$ |
| IBIS-QOC | $3.0x10^{-6}$ | $1.4x10^{-5}$ | 0.020 |

: Showed significant difference

**Table 2.** Statistics of the result. The table is showing the p-values for unpaired, two-side t-Test conducted on average time, number of issues, and minutes per issue for each pair of the three DR notations.

### 4.2 Verification of the Hypotheses

*H1: Cost of description.* Equation 1 of Hypothesis H1 was rejected. Instead, the following inequalities were shown to hold.

$$\begin{cases} C_D(\mathcal{D}, ID) > C_D(\mathcal{D}, QOC) \\ C_D(\mathcal{D}, IBIS) > C_D(\mathcal{D}, QOC) \end{cases} \tag{5}$$

*H2: Effectiveness of DR notations.* The effectiveness of DR notations was measured by using the number of issues described. Equation 2 of Hypothesis H2 was rejected. Instead, the following inequalities were shown to hold.

$$N_{risk}(\mathcal{D}, ID) > N_{risk}(\mathcal{D}, IBIS) > N_{risk}(\mathcal{D}, QOC) \tag{6}$$

*H3: Efficiency of risk discovery.* Equation 4 of Hypothesis H4 was shown to hold.

## 5 Discussion

From the experiment, it was shown that Issue/Decision model was the most cost-effective notation among the three DR notations. The result justifies the design of Issue/Decision model as a light-weight design rationale notation, and its use in a system to overcome capture bottleneck.

Despite the hypothesis with which the user study started out, QOC required less cost than IBIS, or even ID. However, actually looking at the time/issue figures, QOC was the worst in its performance. The reason for the QOC's poor performance can be attributed to the fact that QOC is not the process-oriented notation like I/D or IBIS. QOC can capture the structure of design problems, but cannot describe the flow of argumentation which process-oriented notations can. Thus, when subjects were asked to represent DR of discussion in email, they had to reconstruct the design space from the thread of discussion, or had to simply ignore arguments or details that did not fit in the model of QOC. As a result, although less time was required to describe DR with QOC compared to other notations, the number of issues described was the smallest as well.

Although the effectiveness of DR notations was measured by the number of issues described for the same material, this measure captures only one aspect of the effectiveness of DR notations. It should be noted that other measure could be used to capture other kind of aspects. Especially, we need to look into the difference in topic coverage between design rationale notations. It can be easily imagined that process-oriented notations and design space oriented notations will cover different topics.

As the experiment was paper-based, it is well expected that the result would be different if the computer interface was involved. Another study would be required, if the cost-effectiveness, or efficiency, of DR notations was to be compared in a manner that the effect of human-computer interaction is taken into consideration.

## 6 Conclusion

This paper reported the user study of comparing the cost and benefit of capturing design rationale. The study showed that the annotation scheme we have devised was the most cost-effective. However, we admit that the measure of design rationale benefit is insufficient, as the number of issues indicate just one aspect of the benefits. A future work is to study on benefit measure of design rationale.

## Acknowledgment

This study was conducted at the University of Tokyo. The authors are thankful to Shinichi Nakasuka for providing the resources of the CubeSat project at the University of Tokyo. We are also grateful to members of the CubeSat project, especially those who participated in the experiment as subjects.

# Using Neural Network to Weight the Partial Rules: Application to Classification of Dopamine Antagonist Molecules

Sukree Sinthupinyo[1], Cholwich Nattee[1], Masayuki Numao[1], Takashi Okada[2], and Boonserm Kijsirikul[3]

[1] Department of Architecture for Intelligence, The Institute of Scientific and Industrial Research, Osaka University,
8-1 Mihogaoka, Ibaraki, Osaka, 567-0047, Japan
{cholwich,sukree,numao}@ai.sanken.osaka-u.ac.jp
[2] Department of Informatics, School of Science and Technology, Kwansei Gakuin University
okada@kwansei.ac.jp
[3] Department of Computer Engineering, Chulalongkorn University
boonserm.k@chula.ac.th

**Abstract.** In this paper, we propose an approach which can help Inductive Logic Programming (ILP) in multiclass domains and also its application to a real world domain, Classification of Dopamine Antagonist Molecules. When we classify an example by using the unordered rules constructed by standard ILP systems in multiclass domains, an example may match with the rules from different classes or may match with no rule in the rule set. Thus, using the rules alone is insufficient. We present the approach which utilises some matches in the rule to classify such examples. First, we extract the pieces of knowledge from the original rules, called partial rule. Then, we apply Neural Network to assign the importance to each partial rule. Finally, we use the weighted partial rules to classify unseen examples. Furthermore, the weights from Neural Network also show the importance of the piece of knowledge which is different from the knowledge originally represented in the form of First Order rules.

## 1 Introduction

In recent years, Inductive Logic Programming (ILP) has been widely used to discover knowledge from various real-world domains, especially in chemistry domains [1–4]. The knowledge obtained from ILP is explained in the form of First Order Logic which is rich in comprehensiveness and expressiveness. However, the ordinary ILP's systems are two-class classifier. They construct the rules which cover all positive examples and none of the negatives. The rules are then used to classify unseen examples. The example which is covered by some rules are classified as the positive class, otherwise classified as the negative class. Some problems arise when we need to use the rules from ILP in multiclass domains.

An example may match with the rules from different classes or may match with no rule in the rule set. Thus, ILP's rules alone are insufficient.

In this paper, we present the approach which can help ILP in such cases. Our aim is to use only the ordinary rules which are constructed as the unordered rules in multiclass domains. Our approach is based on the idea of using the partial matches from the original rule to collaboratively classify unseen examples. When an example is covered by the rules from different classes or covered by no rule in the rule set, we can make use of some partial matches in the rule to classify the example. To find the partial matches, we extract the partial rules from the original rules, and then compare them to the example. Since the partial rules are some part of the origial rule, they are more general than the original one. Then, the partial rules are used collaboratively to classify the example. Hence, the problem of using ILP's rules is transformed into how to assign the importance to each partial rule and determine which class is most suitable for the example when there are serveral partial rules cover the example.

In our previous work, we employed Winnow-based algorithm to assign the importance to each partial rule [5]. However, in this paper, we present the use of Neural Network, instead of Winnow algorithm, to determine the importance in the form of the weights. Then, we represent the knowledge hidden in the network structure and the weights of the links as *Weight Partial Rules* which not only keep the comprehensiveness and the expressiveness of First Order rule but also represent the importance of each partial rule as the attached weights.

In this work, we applied our approach to a real-world problem, classifying the activity of dopamine antagonist molecules. The dopamine antagonist molecule is a kind of molecules which can block the binding between the dopamines and dopamine receptors in signal transfer process in the brain. The excessive levels of the dopamine have been implicated in schizophrenia. Hence, for the medical treatment of schizophrenic patients, the dopamine antagonist molecules are used to reduce the signal transfer level which can limit the effect of the high density of the dopamines. The knowledge discovered from this domain may be useful for schizophrenic drug development.

The paper is organised as follows. In the next section, we present the concept of ILP and the obstacles when ILP is applied to the multiclass problems. The strategy of Weight Partial Rules is expressed in Section 3. The details of the experiments are presented in Section 4. The paper ends with the conclusion in Section 5.

## 2 Inductive Logic Programming (ILP)

As mentioned in the previous section, the ordinary ILP systems are two-class classifier. ILP's search strategies aim to seek for the hypothesis which covers all positive examples and none of negative examples. The constructed hypothesis is represented as the First Order rule set which is rich in comprehensiveness and expressiveness. Then, the rule set is used to classify unseen examples. The examples which match with some rule(s) are classified as the positive class, while the examples which do not match with any rule are classified as the negative class. However, when we need to use ILP in multiclass domains, we must employ other methods to help ILP construct the rule and determine the class of examples. As described before, ILP constructs the rules which cover all positive examples and none of the negatives. While in multiclass problems, we must construct the rules for each class, so that another method is needed. Moreover, after we obtain the rules of each class, the problem of selecting the class for examples may arise. An example may match with the rules from different classes or may not match with any rule in the rule set.

There have been some works which can be employed to help ILP in such cases. Dietterich and Bakiri [6] proposed the method which employs the error correcting code to represent the class of examples and tries to predict the most suitable class for test examples. Round Robin Rule Learning proposed by Fürnkranz [7] focuses on training examples rearrangement. The training examples from each class are used to train the learner several times. A test example is tested with all trained classifiers. The most winning class is selected as the class of the test example. Eineborg and Bostr [8] proposed the method for selecting the class for the uncovered examples, Rule Stretching. The method aims to deal with an uncovered example by generalising the original rules to cover the example and select the most accurate rule as the rule which best matches with the example.

The proposed approach is different from the works described above. Our approach utilises the unordered ILP's rules which are constructed by using the common algorithm, the one-against-all method in which the $k$-class problem is reduced to $k$ two-class problems. The rules of class $i$ are constructed by using the training examples of class $i$ as positive examples and the training examples of class $j$ where $j = 1, ..., k$ and $j \neq i$ as negative examples. For example, our data set contains 4 classes, i.e. D1, D2, D3, and D4 (as will be described in Section 4). We use the training examples of class D1 as the positive examples and use those of classes D2, D3, D4 as the negatives for learning rules of class D1. Using this strategy, the obtained rules are unordered, so that when an unseen example matches with no rule or with multiple rules from different classes, we cannot select which rule should be applied.

In this work, we select an ILP system, Aleph [9], to construct the rules. Aleph employs the Inverse Entailment algorithm which was previously used by Progol [10] to generate the most specific clause, called *bottom clause*. Then, to seek for the best generalised clause, Aleph provides many search algorithms which users can select the most suitable one for their domain. In our experiments, we selected the randomised search method using an altered form of the GSAT algorithm [11] that was originally proposed for solving propositional satisfiability problems. The GSAT algorithm provided by Aleph is modified to suit the sequence of literals searching process in ILP fashion.

## 3 Weighted Partial Rules

Our approach is based on the idea that some partial matches of the rule can be used to classify unseen examples. The partial matches of the rule are represented in the form of the partial rules which are extracted from the original rule. Then, we utilise all extracted partial rules to classify the unseen examples. The class of a test example should be the class from which the important partial rules cover the example more than those from other classes. In this paper, to determine the importance of each partial rule, we train a neural network to assign the importance to the partial rule in the form of the weights of the links.

### 3.1 Partial Rule Extraction

A *partial rule* is *a rule whose body contains a valid sequence of the literals, from the body of the original rule, which starts with the literal consuming the input variables in the head of the rule.* Our partial rule extraction algorithm is based on the idea of the new introduced variables, similar idea as the feature extraction in BANNAR [12]. A literal will be added to the sequence if it consumes some new variables introduced by previously added literals in the sequnce. The extraction procedure starts with an empty sequence, and uses variables in the head of the rule as new variables. Then, the literal which consumes the new variables as input variables is gradually added to the sequence. The new variables introduced in the newly added literal are again used as the new variables for searching other literals to be added. The search stops when the newly added literal introduces no new variable or cannot find any literal which consumes the new variables in this newly added literal. Finally, we make all possible combinations of the two sequences which have the common variables not occurring in the head. The partial rule extraction algorithm is shown in Figure 1.

### 3.2 Neural Network Training

As mentioned earlier, when we use the unordered rules to classify examples in multiclass fashion. The examples may be not covered by any rule or may be multiple covered by the rules from different classes. In the previous subsection, we can see that the literals in the body of partial rules are subset of the literals in the body of the original rule. This causes the partial rules are more general than the original one. Hence, when we use the partial rules instead of the original rule, the number of the uncovered examples decreases while the number of the multiple covered examples increases. So that the problem is now transformed into how to select the class for the example which is covered by the rules from different classes. The class of the example should be the class from which the number of the covering important partial rules is more than those from other classes. We thus employ neural network algorithm to assign the weight to each partial rule.

Since we must finally extract the knowledge from the network, we decide to use the simple network structure which consists of only two layers, i.e. *input layer*

```
function ExtractPartialRule(rule)
returns a sequence list
  inputs:    rule as original rule
  variables: literals, remained_literals as sequence of literals
             output, combined, partial_rules as sequence list
             literal as literal
  output ← empty list
  literals ← the body of rule
  for each literal, in literals, which consumes the variables in the head of
  rule as input
    remained_literals ← remove literal from literals
    partial_rules ← SearchPartialRule(literal, literal, remained_literals,
                    output)
    output ← add partial_rules to output
  combined ←  make all possible combination of sequence of literals in
              output, which have common variables that do not occur
              in the head of the rule
  output ← add combined to output
  remove redundant sequence of literals from output
  return output


function SearchPartialRule(input_literal, partial_rule, literals, output)
returns a sequence list
  inputs:    input_literal as literal
             partial_rule, literals as sequence of literals
             partial_rules as sequence list
  variables: literal as literal
             remained_literals, new_partial_rule as sequence of literals
             unfinish as sequence of literals
             new_partial_rules as sequence list
             found as boolean, initially false
  new_partial_rule ← add input_literal to partial_rule
  if no new argument in input_literal then
    partial_rules ← add new_partial_rule to partial_rules
    output ← add partial_rules to output
    return output
  for each literal, in literals, which consumes the new variables
  in input_literal as input
    unfinish ← add literal to new_partial_rule
    remained_literals ← remove literal from literals
    partial_rules ← SearchPartialRule(literal, unfinish,
                    remained_literals, output)
    output ← add partial_rules to output
    found ← true
  if found then
    partial_rules ← add new_partial_rule to partial_rules
    output ← add partial_rules to output
  return output
```

**Fig. 1.** Partial Rule Extraction Algorithm

and *output layer*. Each input node represents the truth value of the body of each partial rule and each output node represents a class. The input nodes are fully connected to the output nodes. The number of input nodes is the number of the partial rules extracted from the whole original rules from all classes. The number of output nodes is the number of the classes. All output nodes are sigmoid unit.

In training process, an input vector contains elements each of which is the truth value of each partial rule evaluated against a training example and the corresponding output vector is the vector which represents the class of the example. The input value of the input node representing the *true* partial rule is 1, while the input value of the node representing the *false* partial rule is $-1$. For the output vector, we assign 1 for the node that represents the class of the example and 0 for the others. The structure of our neural network is shown in Figure 2. The example of generating a training vector for the neural network is shown below.

Consider the following rules $R_1, R_2, R_3, R_4$, for classes D1, D2, D3, and D4, respectively.

$R_1$: `molecule(A,d1) :- atm(A, B, C, D, E, F), F=1.4, E=2.3, atm(A, G, H, I, J, K), J=1.5.`

$R_2$: `molecule(A,d2) :- atm(A, B, C, D, E, F), bond(A, G, H, I, J, K), E=2.8, C=n, gteq(K, 1.5).`

$R_3$: `molecule(A,d3) :- link(A, B, C, D), atm(A, E, F, G, H, I), I=3.8, H=3.6, D=4.8.`

$R_4$: `molecule(A,d4) :- link(A, B, C, D), atm(A, E, F, G, H, I), I=1.4, H=8.5, D=2.9.`

The following rule $R_iC_j$ is a partial rule extracted from $R_i$.

$P_1$: $R_1C_1$: `molecule(A,d1) :- atm(A, B, C, D, E, F), F=1.4.`
$P_2$: $R_1C_2$: `molecule(A,d1) :- atm(A, B, C, D, E, F), E=2.3.`
$P_3$: $R_1C_3$: `molecule(A,d1) :- atm(A, G, H, I, J, K), J=1.5.`
$P_4$: $R_1C_4$: `molecule(A,d1) :- atm(A, B, C, D, E, F), F=1.4, E=2.3.`

$P_5$: $R_2C_1$: `molecule(A,d2) :- atm(A, B, C, D, E, F), E=2.8.`
$P_6$: $R_2C_2$: `molecule(A,d2) :- atm(A, B, C, D, E, F), C=n.`
$P_7$: $R_2C_3$: `molecule(A,d2) :- bond(A, G, H, I, J, K), gteq(K, 1.5).`
$P_8$: $R_2C_4$: `molecule(A,d2) :- atm(A, B, C, D, E, F), E=2.8, C=n.`

$P_9$: $R_3C_1$: `molecule(A,d3) :- link(A, B, C, D), D=4.8.`
$P_{10}$: $R_3C_2$: `molecule(A,d3) :- atm(A, E, F, G, H, I), I=3.8.`
$P_{11}$: $R_3C_3$: `molecule(A,d3) :- atm(A, E, F, G, H, I), H=3.6.`
$P_{12}$: $R_3C_4$: `molecule(A,d3) :- atm(A, E, F, G, H, I), I=3.8, H=3.6.`

$P_{13}$: $R_4C_1$: `molecule(A,d4) :- link(A, B, C, D), D=2.9.`
$P_{14}$: $R_4C_2$: `molecule(A,d4) :- atm(A, E, F, G, H, I), I=1.4.`
$P_{15}$: $R_4C_3$: `molecule(A,d4) :- atm(A, E, F, G, H, I), H=8.5.`

$P_{16}$: $R_4C_4$: `molecule(A,d4) :- atm(A, E, F, G, H, I), I=1.4, H=8.5.`

We now have 16 partial rules of 4 classes, so that our neural network contains 16 input nodes and 4 output nodes. Assume that we are constructing the input and output vector of `molecule(m06497)`, an example of class D1. We first evaluate the truth value of `molecule(m06497)` when applying it to each partial rule. Then, the obtained truth values are organised as an input vector. For example, if the truth values of all partial rules from $R_1$ and only the second one from $R_2$ are *true* while the others are *false*, the input vector of this example will be $(1, 1, 1, 1, -1, 1, -1, -1, -1, -1, -1, -1, -1, -1, -1, -1)$. As this is an example of class D1, the output value of the output node representing class D1 is only activated while the others are zero. Thus, the output vector is $(1, 0, 0, 0)$.

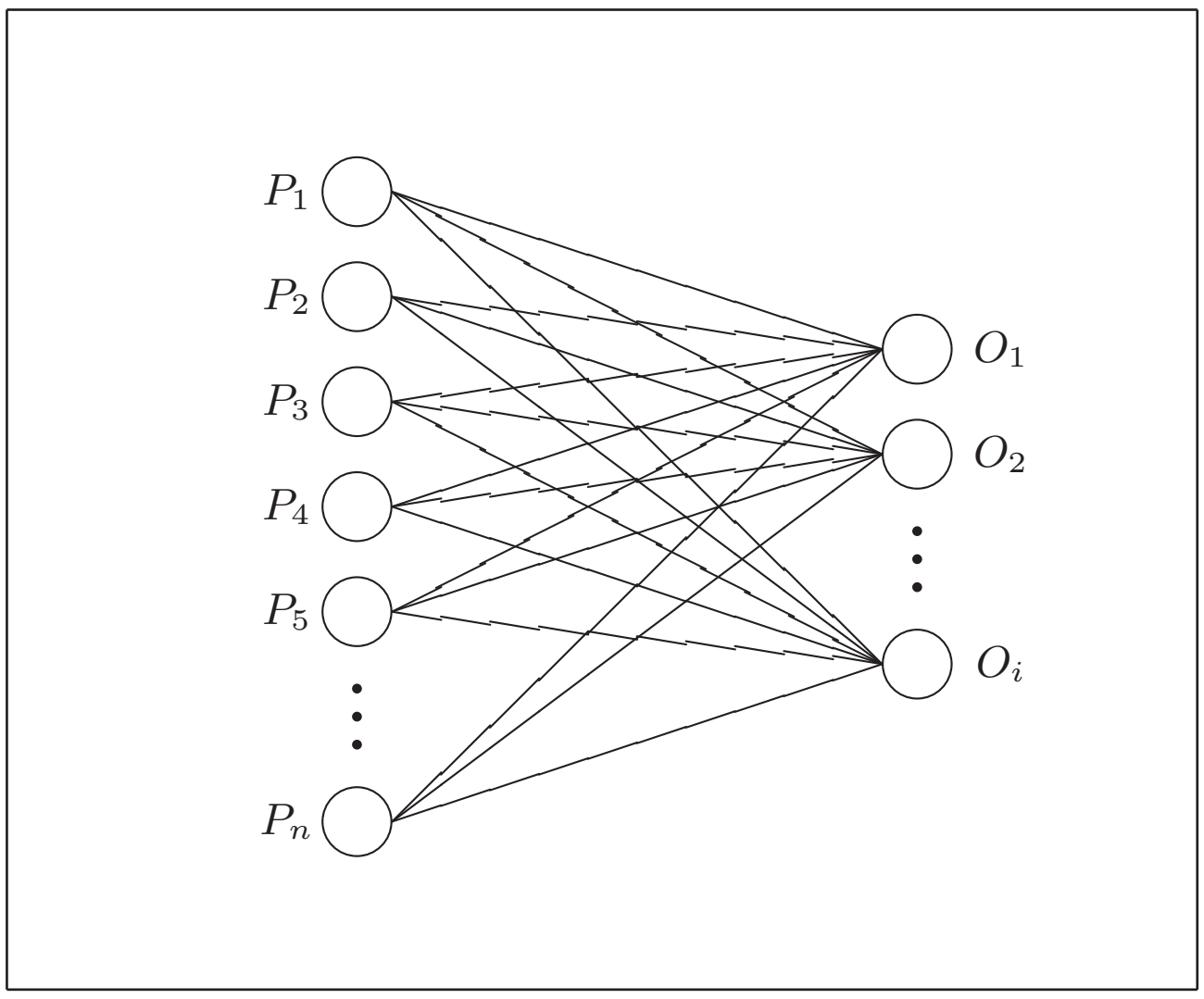

**Fig. 2.** The Neural Network Structure

### 3.3 Weighted Partial Rules

After the neural network is trained, the importance of each partial rule is hidden in the weights and the structure of the network. This makes the comprehensiveness disappear, if we directly classify a new example by using the network instead of the original rules. We thus propose a knowledge extraction technique to convert the weights and the structure of the neural network into the human-understandable form. The strategy used in our approach is described as follows.

Given a problem with $n$ partial rules and $m$ classes. $W_i$ is a vector of length $m$, where the element $w_{i,j}$ of $W_i$ is the weight of the link between input node $i$ and output node $j$, $P_i$ is the partial rule represented by input node $i$, $W_0$ is a vector of the bias of output nodes, and $P_0 = true$. The *Weighted Partial Rules*, $WPR$, is a vector of length $n + 1$, where the $i^{th}$ element of $WPR$ is a pair ($W_i$, $P_i$).

To classify example $e$, we use the following strategy.

- Let $V$ is a vector of length $m$, initially with $V = W_0$.
- Compare $e$ with $P_i$ in all elements of $WPR$. If $e$ matches with $P_i$, $V \leftarrow V + W_i$ else, $V \leftarrow V - W_i$.
- Let $v_l$ be the maximum in $V$, return class $l$.

In above strategy, we represent the knowledge hidden in the neural network in the form of a vector of the pairs ($W_i$, $P_i$), where $W_i$ is a vector of the weights of the links from the input node representing the partial rule $P_i$ to all output nodes. The class which has the maximum output value is selected as the class of the example.

In the feed forwarding step of our backpropagation neural network, the output node is composed of two consecutive units , i.e. the *summation unit* and the *activation unit*. The inputs of the summation unit are the multiplication of the input values and the weights of the links from all input nodes to that output node and the output of the unit is the summation of all inputs. The output of the summation unit is sent through the activation unit which uses the sigmoid function as the activation function. Hence, the final output of the output node is limited to the interval $(0, 1)$. However, in our approach, the information we only need is which output node gives the maximum value. Thus, we can ignore the usage of the sigmoid function because the sigmoid function, $sigmoid(x) = \frac{1}{1+e^{-x}}$, is an increasing function where $sigmoid(x_i) > sigmoid(x_j)$ when $x_i > x_j$. The output of the function is higher for the higher input value. Therefore, only for comparison purpose as in our approach, it is not necessary to use the activation unit, and only the summation unit is sufficient.

For example, let $w_{0,i}$ be the bias value of output node $i$ and $w_{i,j}$ is the weight of the link between input node $i$ and output node $j$. From the previous example, the $WPR$ which represents the original rules and their importance is shown below:

$$
\begin{aligned}
WPR = (\quad & ((w_{0,1}, w_{0,2}, w_{0,3}, w_{0,4}), true), \\
& ((w_{1,1}, w_{1,2}, w_{1,3}, w_{1,4}), P_1), \\
& ((w_{2,1}, w_{2,2}, w_{2,3}, w_{2,4}), P_2), \\
& ((w_{3,1}, w_{3,2}, w_{3,3}, w_{3,4}), P_3), \\
& ((w_{4,1}, w_{4,2}, w_{4,3}, w_{4,4}), P_4), \\
& ((w_{5,1}, w_{5,2}, w_{5,3}, w_{5,4}), P_5), \\
& ((w_{6,1}, w_{6,2}, w_{6,3}, w_{6,4}), P_6), \\
& \ldots \\
& ((w_{16,1}, w_{16,2}, w_{16,3}, w_{16,4}), P_{16}) \quad )
\end{aligned}
$$

## 4 Experiments

The data set used in the experiments contained 1366 molecules of dopamine antagonist molecules of 4 classes, D1, D2, D3, and D4 [13]. The information of the molecules was originally described in the form of the position in three dimension space of atoms, types of atoms, types of bonds, and dopamine antagonist activity of molecules. However, the position in three dimension space was not useful for discriminating examples because a molecule could rotate or move to other positions in the space. Hence, we converted the positions of atoms to the relations between atoms and bonds. We instead represented the information of atoms, bonds, and distances between atoms in term of 3 predicates, `atm/6`, `bond/6`, and `link/4`, respectively. The details of these three predicates are described below:

- `atm(A,B,C,D,E,F)` represents that the atom `B` is in molecule `A`, is type `C`, forms a bond with oxygen atom if `D` is 1, otherwise it does not link to any oxygen atom, has distance `E` to the nearest oxygen atom, and has distance `F` to the nearest nitrogen atom.
- `bond(A,B,C,D,E,F)` represents that the bond `B` is in molecule `A`, has atoms `C` and `D` on each end, is type `E`, and has length `F`.
- `link(A,B,C,D)` represents that in the molecule `A`, the distance between atoms `B` and `C` is `D`.

### 4.1 Compared Approaches

We compare our approach with other three approaches, i.e. Winnow-Based algorithm [14], Majority Class [15, 16] method and Decision Tree Learning algorithm. The brief review of each approach is described below.

#### Winnow-Based Algorithm

Each partial rule can be viewed as an expert which can vote for the class of test examples. The weights of the partial rules which cover the example are summarised and the class with the highest summation is selected as the class of the example. The weights are obtained by using the training strategy described below:

Given a problem with $n$ partial rules, $m$ classes, and promotion factor $\alpha$. $P$ is a vector of length $n$, where element $p_i$ of $P$ is a partial rule. $W_i$ is a vector of length $m$, where element $w_{i,j}$ of $W_i$ is the weight of class $j$ of partial rule $p_i$. $V$ is a summation vector of length $m$, where $v_i$ of $V$ is the summation of the weights of class $i$. The weight vector $W_i$ are updated by using the following procedure.

- Initialize all $w_{i,j} = 1$
- Until termination condition is met, Do
  - For each training example $e$, Do
    - Initialize all $v_i = 0$ and $c$ as the class of $e$

- For all partial rules $p_i$ which match with $e$, add corresponding $W_i$ to $V$,

$$V = V + W_i$$

- Let $v_k$ be the maximum element in $V$, predict the example $e$ as class $k$
- If $c = k$, no update is required; otherwise the weight $w_i$ corresponding to $p_i$ which matches with $e$ is updated by,

$$w_{i,j} = \begin{cases} \alpha w_{i,j} & \text{if } j = k, \\ \\ \alpha^{-1} w_{i,j} & \text{if } j = c. \end{cases}$$

For more details please refer to [5].

### Majority Class Method

In the Majority Class method, we selected the class which had the maximum number of examples in training set as the default class. An example which matched with only rule(s) from one class was classified as that class, while an example which could not match with any rule was classified as the default class. In case of the examples which matched with the rules from two or more classes, we selected the class of which the matched rules covered maximum number of examples.

### Decision Tree Learning Algorithm

Decision Tree Learning (DTL) is a well-known propositional Machine Learning technique which employs the Information Theory to guide in searching for the best theory. DTL has been successfully applied to many attribute-value real-world domains [17–19]. The decision tree learner used in our experiments was C4.5 system [20].

We trained C4.5 using the features obtained by comparing the partial rules with an example, and these features were a set of truth values (either $true$ or $false$). The features were the same as those used as the input vector of the neural network. The reason that we selected C4.5 to apply to the same feature set is to compare the efficiency of the weights from the neural network which is employed in the proposed method with the decision tree.

## 4.2 Experimental Results

We ran 10-fold cross validation experiment using four approaches, the original ILP system with the Majority Class method (ILP+Majority Class), Partial Rules and C4.5 (PR+C4.5), Partial Rules and Winnow (PR+Winnow), and our approach, Weighted Partial Rules (WPR).

Table 1 shows the accuracy of each approach on the test examples. The accuracy of ILP+Majority Class approach is 79.11%. The accuracy of PR+C4.5 is 85.71%, higher than ILP+Majority with 99.5% confidence level using the standard paired t-test method. The accuracy of PR+Winnow is 88.65%, higher than ILP+Majority and PR+C4.5 with 99.5% and 99.0% respectively, The accuracy of WPR is 92.03%, higher than other approaches with 99.5% confidence level using the same comparing method.

**Table 1.** The accuracy of the compared approaches.

| Method | Accuracy (%) |
|---|---|
| ILP+Majority Class | 79.11±4.37 |
| PR+C4.5 | 85.71±3.41 |
| PR+Winnow | 88.65±3.85 |
| WPR | 92.03±3.14 |

Furthermore, in Table 2, we show the ratio between the number of examples correctly classified and the number of examples for each portion. Exactly Covered column indicates the number of the examples covered by the rule(s) from only one class, Multiple Covered column indicates the number of the examples covered by the rules from different classes, and Uncovered column indicates the number of the examples which are not covered by any rule. For exactly covered examples, WPR correctly classified 991 of 1049 examples, whereas 965 were correctly classified by the rules from ILP. For the examples covered by multiple rules from different classes, WPR correctly classified 81 of 97 examples, more 32 examples than the Majority Class method. Finally, 185 of 220 examples were correctly classified by WPR, while only 68 examples were correctly classified by the Majority Class method. These results show that WPR much improved the accuracy in Multiple Covered and Uncovered, and slightly improved in Exactly Covered portion.

**Table 2.** Improvements of WPR over the original rules with Majority Class method, reported according to exactly covered examples, multiple covered examples, and uncovered examples.

| Method | Exactly Covered | Multiple Covered | Uncovered |
|---|---|---|---|
| ILP+Majority Class | 965/1049 | 49/97 | 68/220 |
| WPR | 991/1049 | 81/97 | 185/220 |

An example of some partial rules which are highly weighted is shown below.

```
molecule(A) :- atm(A, B, C, D, E, F), bond(A, G, B, H, I, J),
   bond(A, K, H, L, M, J), atm(A, L, C, D, E, S), F=3.3.
[2.88, -1.95, -0.97, -0.27]
```

```
[The original rule is
molecule(A) :- atm(A, B, C, D, E, F), F=3.3, bond(A, G, B, H, I,
   J), bond(A, K, H, L, M, J), bond(A, N, L, O, M, P), bond(A, Q,
   O, R, M, P), atm(A, L, C, D, E, S).]

molecule(A) :- atm(A, E, F, G, H, I), bond(A, N, E, O, P, M),
   atm(A, O, F, G, Q, R), H=2.4.
[-1.42, 4.12, -1.11, -2.16]
[The original rule is
molecule(A) :- link(A, B, C, D), atm(A, E, F, G, H, I), D=5.6,
   H=2.4, gteq(I, 3.8), bond(A, J, B, K, L, M), bond(A, N, E, O,
   P, M), atm(A, O, F, G, Q, R), lteq(Q, 2.9), lteq(M, 1.4),
   bond(A, S, C, T, P, U).]
```

The Weight Partial Rules in the above example show another advantage of our approach. We can see that when an example matches with these highly weighted partial rules, the example has the high probability of being classified as the class whose weight is very high. This provides us some knowledge which can be discovered from the dataset, different from the original rules which sometimes are too specific and not useful. Our approach can seek for some pieces of knowledge which are more important than the others in the original rule.

## 5 Conclusion

We have proposed an approach which not only improves the accuracy of ILP's rules in multiclass problems but also extracts some pieces of knowledge from the rules. The approach is based on the idea that the importance of the partial matches of the rules is different. In this work, we train the neural network to assign the weights, which represent the importance, to the partial rules. The experimental results on classifying the activity of the dopamine antagonist molecules show that our approach was successfully applied to the domain by yielding 92.03% accuracy. Moreover, the weights of the partial rules also show some important pieces of knowledge which are previously hidden in the original rules.

# Future Directions in Adapting GAs using Knowledge Acquisition

J.P. Bekmann[1,2] and Achim Hoffmann[1]

[1] School of Computer Science and Engineering,
University of New South Wales, NSW 2052, Australia.
[2] National ICT Australia (NICTA), A.T.P, NSW 1430, Australia.

**Abstract.** We present an incremental knowledge acquisition approach that incrementally improves the performance of a Genetic Algorithm as the provided knowledge tailors the Genetic Algorithm towards a given application domain. While it has been established that such an approach can be useful for building problem solvers for industrially relevant problems [1], this paper discusses in more detail what challenges lie ahead on the path towards a broader applicability of the approach.

It is generally known that adapting probabilistic search algorithms, such as genetic algorithms (GAs), to a given problem domain is critically important to make the probabilistic algorithm efficient. In our approach we build domain specific knowledge bases that control two parts of the genetic algorithm: the fitness function and the mutation operators. The knowledge bases are built by a human who has at least a reasonable intuition of the search problem and how to find a solution. The human monitors the probabilistic search algorithm and intervenes when he/she feels that produced candidates have only a small chance to lead to an acceptable solution or the human helps by providing rules on how to generate candidates with high chances of success. Our framework is inspired by the idea of (Nested) Ripple Down Rules (NRDR) where humans provide exception rules to rules already existing in the knowledge base, using concrete examples of inappropriate performance of the existing knowledge base.

In this paper we discuss a number of avenues that can be taken to make our approach more effective and applicable to a broader range of domains, This includes a discussion on how the specific needs for integrating domain knowledge into probabilistic search algorithms relates to the new way of utilising ideas of traditional Ripple Down Rules.

We also briefly present some experimental results on industrially relevant domains of channel routing as well as switchbox routing in VLSI design.

## 1 Introduction

Genetic Algorithms (GA) is a popular technique to solve combinatorial search problems. While GAs generally can find solutions due to their probabilistic nature, finding the solution may take substantial time. Indeed, in many cases the GA needs to be substantially optimised in order to produce satisfactory solutions within an acceptable time [4].

For most applications, developers take a relatively unstructured 'trial and error' approach to tailor a generic GA to a problem domain. Major aspects of a generic GA that can be modified to suit a domain include the representation of solution candidates (genomes), the fitness function to be used, as well as the chosen operators that modify the genomes, such as mutation or cross-over operators. Other aspects, such as the population size, criteria for selecting parents, rate of modification of genomes, etc. can be varied and may result in performance improvements, depending on the nature of the problem.

In [1] an incremental knowledge acquisition approach was introduced that allows for the incremental development of a fitness function as well as specialised mutation operators that can be incrementally tailored towards the requirements of a given problem domain.

The knowledge acquisition approach was inspired by Ripple Down Rules [3]. However, since a number of aspects are different when it comes to acquiring knowledge for GAs as opposed to more traditional fields of domain expertise, the knowledge acquisition approach for GAs looks quite different from RDRs in a number of ways.

In this paper we present perspectives on expanding our framework. We also present a detailed discussion on how the new knowledge acquisition framework for GAs relates to traditional scenarios where RDRs, MCRDRs or NRDRs have been used. These differences are not only in regards to the required representation language, but also in regards to the nature of the expertise itself and the abilities of the expert to provide definite rules.

The overall idea being that GAs provide a basic mechanism for searching for a solution, which can be gradually improved by providing hints of what search direction, i.e. how to create a new individual from one or more parents, is more likely to lead to good solutions. Since the performance of GAs is already improved if less successful search directions are taken a little less often, knowledge which changes the probability of taking a given search direction can make all the difference, without needing to be 'correct in an absolute sense'.

This paper is organised as follows: In the following section 2 the general architecture of our GA framework is briefly sketched. Section 3 discusses the effort required by the expert or system engineer to adapt the GA to a given problem domain. In section 4 it is discussed in what way individual operators can be evaluated and what further automatic methods are likely to be useful to assist human expert in improving operators further or in refining the conditions of an operator's application. This is followed by section 5 about controlling the evolutionary process of a GA. The penultimate section 6 discusses relationships of our approach to more traditional forms of Ripple Down Rules. Section 7 contains the conclusions.

## 2 Existing Architecture

Our framework HeurEAKA (Heuristic Evolutionary Algorithms using Knowledge Acquisition) consists of a genetic algorithm and a knowledge base component. The knowledge base is built up incrementally by an expert reviewing past performance and recommending improvements.

The evolutionary process of the GA can be monitored by the human and individual problem instances can be evaluated. If a particular individual is generated that appears undesirable or suboptimal, the human could enter a new rule that prevents such behaviour in future or provide an improved alternative action. The user might also add a rule which imposes a fitness penalty on such individuals. More generally, the user formulates rules based on characteristics of selected individuals, and these are applied in the general case by the GA.

### 2.1 The Genetic Algorithm

Evolutionary algorithms are loosely based on natural selection, applying these principles to search and optimisation. Basic GAs are relatively easy to implement. A solution candidate of the problem to be solved is encoded into a genome, and a collection of genomes makes up a population of potential solutions. The GA performs a search through the solution space by modifying the population, guided by the evolutionary heuristic. When a suitable solution has been identified, the search terminates.

In HeurEAKA, genome encoding and manipulation is treated by the GA module as opaque. All manipulations take place indirectly via a *primitives interface*. This interface is designed to isolate domain specific information from the overall architecture. It is formulated in such a way that it can easily be extended to different domains.

In order to generate new individuals, the GA has to select parents from the current population and then generate offspring by mutation and/or by crossover. Further, some individuals of the current generation should be selected for removal and replaced by newly generated individuals.

Offspring of selected parents are either created via a crossover copy operation or as a mutated copy. A parameter determines which operator will be applied. The crossover operator mimics natural evolutionary genetics and allows for recombination and distribution of successful solution sub-components in the population.

In order to select individuals either as a parent for a new individual or as a candidate to be removed from the population, the knowledge base is invoked to determine the fitness of an individual as explained below. In order to generate suitable offspring, another knowledge base is invoked which probabilistically selects mutation operators.

### 2.2 The Knowledge Base

Our knowledge acquisition approach for building the knowledge base for fitness determination and the knowledge base for selecting operators for offspring generation is based on the ideas of Ripple Down Rules (RDR) [3]. RDR builds a rule base incrementally based on specific problem instances for which the user explains their choices. An extension of RDR allows hierarchical structuring of RDRs - "nested RDR" (NRDR) [2]. NRDR allows re-use of definitions in a KB, and the abstraction of concepts which make it easier for the expert to describe complex problems on the knowledge level and also allow for more compact knowledge bases for complex domains.

**Ripple Down Rules Knowledge Base** We use single classification RDRs (SCRDRs) for both types of knowledge bases. A SCRDR is a binary tree where the root node is also called the default node. To each node in the tree a rule is associated, with a condition part and a conclusion which is usually a class - in our case it is an operator application though. A node can have up to two children, one is attached to an *except* link and the other one is attached to the so-called if-not link. The condition of the default rule in the default node is always true and the conclusion is the default conclusion. When evaluating a tree on a case (the object to be classified), a *current conclusion* variable is maintained and initialised with the default conclusion. If a node's rule condition is satisfied, then its conclusion overwrites the *current conclusion* and the except-link, if it exists, is followed and the corresponding child node is evaluated. If the node's rule condition is not satisfied, the if-not link is followed, if it exists, and the corresponding child node is evaluated. Once a node is reached such that there is no link to follow the *current conclusion* is returned as a result.

In typical RDR implementations, any KB modification would be made by adding exception rules, using conditions which only apply to the current case for which the current knowledge base is inappropriate. By doing this, it is ensured that proper performance of the KB on previous cases is maintained.

Nesting RDRs allows the user to define multiple RDRs in a knowledge base, where one RDR rule may use another, NRDR tree in its condition, and in HeurEAKA as an action, i.e. the NRDR tree is evaluated in order to determine whether the condition is satisfied. A strict hierarchy of rules is required to avoid circular definitions etc.

For the purpose of controlling the Genetic Algorithm, in our approach all conclusions are actually actions that can be applied to the case, which is an individual genome. The rules are formulated using the Rule Specification Language as detailed below.

**Fitness Knowledge Base** The user specifies a list of RDRs which are to be executed when the fitness of a genome is determined. The evaluation task can thus be broken into components as needed, each corresponding to a RDR. The evaluator executes each RDR in sequence.

**Mutation Knowledge Base** For determining a specific mutation operator a list of RDRs provided by the user is consulted. Each RDR determines which specific operator would be applied for modifying an individual. Unlike for the evaluation, for mutation only one of the RDRs for execution will be picked probabilistically using weights supplied by the user.

**Rule Specification Language (RSL)** Conditions and actions for a rule are specified in a simple language based loosely on "C" syntax. It allows logical expressions in the condition of a rule and a list of statements in the action section.

**The Knowledge Acquisition Process in HeurEAKA** The knowledge acquisition process goes through a number of iterations as follows: on each individual the fitness KB is applied and a set of rules executed. The user can select some of the individuals that appear to be of interest and then review the rules that were involved in the creating of those individuals.

The genetic algorithm can be started, stopped and reset via a graphical user interface. A snapshot of the GA population is presented, from which the user can pick an individual for closer inspection.

As shown in Fig. 2, an individual solution can be inspected. It was found necessary to have a good visualization and debugging interface to be able to productively create and test rules.

A user can step back, forward and review the application of RDR rules to the genome, and make modifications to the respective KB by adding exception rules.

Since the evaluation of a set of RDRs can cause a number of actions to be executed, it is necessary to allow the user to step through an execution history. Given non-deterministic elements of operator selection, the interactive debugger has to maintain complete state descriptions, i.e. genome data, variable instantiation and values used in random elements, to make it feasible for the user to recreate conditions for repeated testing.

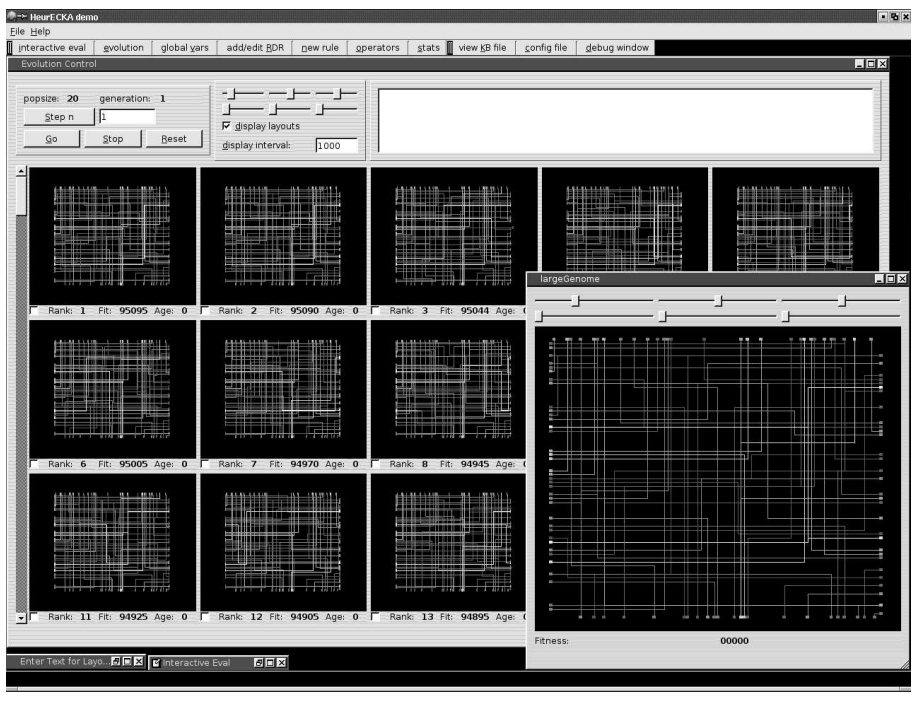

**Fig. 1.** Taken from an application of the HeurEAKA framework to the domain of switchbox routing: The GA window displays snapshots of the population at given intervals. The user can start, stop and step through evolution. An individual can be selected for closer inspection and evaluation, leading to Fig. 2.

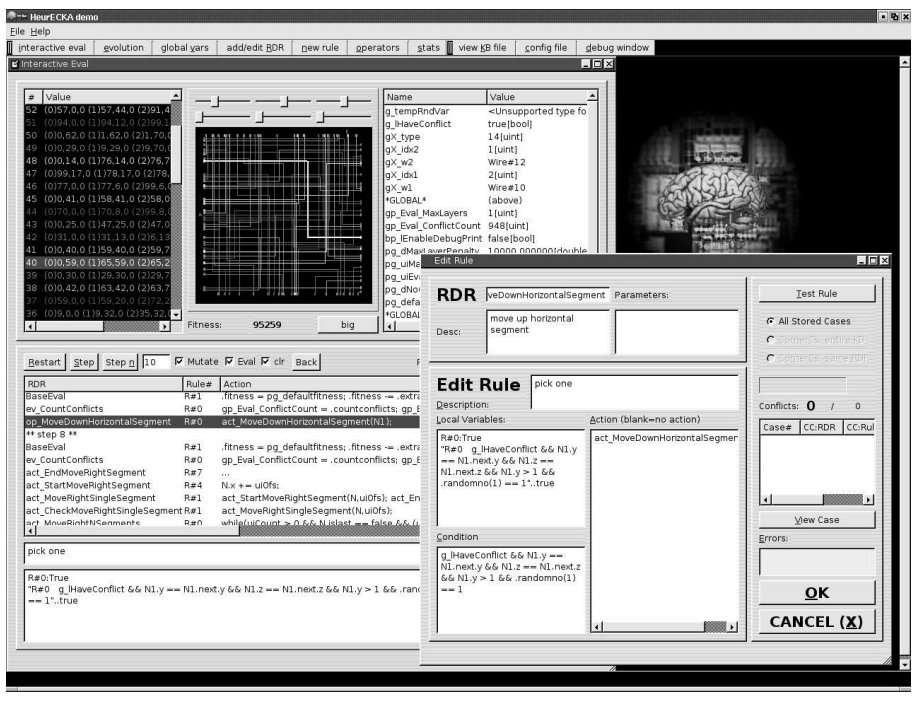

**Fig. 2.** The interactive screen allows the user to inspect the individual, step through rule application, and amend the KB as needed.The two figures are intended to give an overall impression only, detailed use is described in our PRICAI paper [1].

If the user does not agree with the performance of a rule $R$ in the knowledge base, there are two ways of addressing it: Either the addition of an exception rule to $R$ can be made or a modification of $R$ is possible.

**Experiments**

In order to demonstrate that our approach is useful in developing algorithms for solving complex combinatorial problems, detailed channel routing as well as switchbox routing were chosen to demonstrate the approach. These problems are found in the VLSI domain and are known to be NP-hard combinatorial problems. In the design of modern VLSI chips much attention has been given to finding algorithmic solutions for these problems. A number of effective techniques have been developed over the years, and these are used widely in industrial deployments. We have tackled these problems to show that with our framework it is possible to solve such problems with algorithms developed with far less effort. Since well known commercial solutions exist, we wish to measure ourselves against these to show the viability of our approach. Ultimately, the HeurEAKA approach is best used in domains where little is known about how the problems are to be solved. Our philosophy of exploration and incremental improvement, guided by the intuitions of an expert, is designed to address such a scenario.

A channel routing problem (CRP) is given by a channel of a certain width. On both sides of the channel are connection points. Each connection point belongs to a certain electrical net and all connection points of the same net need to be physically connected with each other by routing a wire through the channel and, of course, without two nets crossing. The width of the channel determines how many wires can run in parallel through the channel. The length of the channel determines how many connection points on both sides of the channel there may be. Furthermore, the layout is done on a small number of different layers (e.g. 2 to 4 layers), to make a connection of all nets without crossing possible at all. It is possible to have a connection between two adjacent layers at any point in the channel. Such a connection is also called a *via*.

The switchbox routing problem is similar to the CRP, but does not deal with only a two-sided channel, but rather a rectangle with connections on all sides. Since wires can originate and terminate on the sides of a switchbox, and the width of a channel generally being fixed (due to the fixed wire terminals on either side) the SRP is more difficult to solve.

A knowledge base containing 49 RDRs and 167 rules was created. A fair number of the RDRs were for low-level rules describing the manipluation of nodes and wires. This work was fairly technical and required repeated editing and debugging to ensure operator correctness. Incorrect operators at the low level would often create invalid layouts (diagonal wires or vias, moved wire terminals) or undesirable wiring (double-backing, repeated nodes etc.). After the low level rules were defined, the KA process became easier since most rules could be defined at a higher level of abstraction. Because this was a more intuitive abstraction level, it was easier to formulate rules and required less revision of conditions or actions.

The formulation of effective CRP and SRP algorithms has been the subject of much study and industry-standard algorithms took many years to develop [6]. In our case KA was done by a novice in the field of VLSI, using mainly intuition and being able to incrementally specify rules in a natural way on the knowledge level. On average rules took approximately 10 minutes each to formulate, taking about 30 hours for the formulation of a viable knowledge base. The effort and expertise required was significantly less than commercial routing solutions.

We were able to find results comparable to some of the existing benchmarks [7], in some cases even well surpassing them. The choice in representation for the CRP in our initial experiments was, however, not suitable for a direct comparison to more of the common benchmarks. Currently we are working on addressing this and are confident in being able to favourably measure our results against a wider selection.

## 3 Effort of Adapting to the Domain

The thrust of our proposed approach is the ability to rapidly find algorithmic solutions to problems in a new domain. There exist many approaches for developing algorithms. We will note the traditional approach of "manually" designing an algorithm by a domain expert guided by experience and trial and error. Solutions developed in this way often take many months if not even years of development, and are usually very domain dependent. This means that they are likely to be very efficient, but not readily applicable to other domains. Arguably the effort required to deploy in another domain is akin to starting from scratch.

Using conventional genetic algorithms can require less effort than traditional approaches since operators do not encode a complete solution and the search strategies embodied in GAs provide a powerful and flexible framework. Existing GA solutions use some variety of the basic GA algorithm and build on top of this. The effort expended in developing the basic GA component (or use of libraries) is typically minor compared to that of formulating the domain specific problem encoding, operator formulation and the evaluation function. Another major effort is the "tuning" of GA parameters such as population sizes, mutation and crossover probabilities, selection strategies and many more specific to different GA architectures.

While GAs can thus be an easier approach, it does still require a large effort in adapting it to the problem domain. While numerous methods exist in helping the developer hone the behaviour of the GA itself, there is still a significant shortfall in support for the development and evaluation of domain specific crossover, mutation and fitness functions chosen for any particular problem. Anecdotal accounts suggest that this difficult part of deploying a GA can take many months and even years for complex problems. Furthermore, a problem-specific solution will not be very useful in another domain.

HeurEAKA addresses this problem by supporting the process of mutation and fitness function formulation. As outlined in the previous sections, it provides comprehensive support through the use of a knowledge based system. Our framework is designed such that the main modules are domain- independent. i.e. the knowledge base management and genetic algorithm components do not require modification between problems. A basic problem representation needs to be provided by the knowledge engineer. This part would be similar to that of other GAs, it might be noted however, that no specific encoding such as binary string encoding etc., needs to be done.

HeurEAKA can be used in two different modes, one supporting a rule specification language (RSL). This language needs to incorporate primitives which relate to the problem specific encoding chosen by the knowledge engineer. This entails adding support for these primitives to a LEX / YACC specification of the RSL, an easy task completed in a matter of minutes. Adding accessor functions to the underlying primitives to be called from the parser requires a little more work, but is also a simple task taking maybe one hour. Arguably, the additional work over what is needed over conventional GAs is thus very small.

To provide an intuition about the effort involved, the following example was taken from the CRP domain. A solution is encoded as a *layout* object, containing an array of *wire* objects, which in turn each contain a vector of *node* objects. The parser specification contains the following (simplified for this paper; the HeurEAKA implementation uses a far more expressive grammar and produces a compiled version):

```
expression: wireaccessor | ...;
wireaccessor: wiretype optional_node;
wiretype: "wire[" index_expression "]"
              { CurrentWire = CurrentLayout.Wires[$2]; }
          | variable { CurrentWire = $1; };
optional_node: ".node[" index_expression "]"
              { CurrentNode = CurrentWire.Nodes[$2]; }
          | /* E */ { CurrentNode = NULL; }
```

HeurEAKA can also be run in a compiled mode, where the KB, including rules is generated in C++ code and requires no primitives extension for RSL. This mode does not yet, however, support integrated debugging tools, so is not as practical for KA purposes.

The user interface is re-usable for different problem domains. The only part that needs to be changed is the visualisation window. The GUI is implemented in Qt (a portable GUI toolkit by Trolltech), and the visualisation is wrapped in a single graphical widget object. This can readily be replaced by something that translates the genome into graphical format. Current visualisation code consists of a few dozen lines of QtOpenGL.

The overhead of defining the primitives interface and a simple visualization of the genome, is arguably quite small when compared to the development of mutation and evaluation functions. Thus the difference in overall effort of working with HeurEAKA and conventional GA lies in the development of the KB versus the traditional trial and error “manual” approach of finding useful mutation and evaluation functions. Due to the intuitive nature of our KA techniques we believe it to be the easier approach. For complex domains where traditional approaches require sophisticated mutation and evaluation functions, we believe our approach would be comparatively even stronger as the large amount of detailed knowledge to be encoded would be hard to handle in a conventional 'trial and error' approach. Where it usually proves to be rather difficult to adapt a general purpose GA design to a particular problem type at hand [8, 5].

## 4 Evaluation of operators

The KB in HeurEAKA describes operators for the evaluation and mutation function of the GA. While the evaluation KB is the accumulation of rules describing what is desirable about genomes, the mutation KB contains a collection of operators which modify the genome.

As outlined in section 2.2, the interactive debugging interface is effective in supporting the formulation of rules based on genome instances. Lower level operators can be evaluated and refined in this context as they have relatively simple interactions with the genome.

In order to be able to evaluate more abstract high level operators, manually stepping through a small sample of genomes is not sufficient for gaining a good overall impression of how they would scale to the general case; i.e. being able to judge their role in the larger context of contributing strategically to the GA's search.

Since the GA operates at high speed with a large number of genomes, some automated measures help the expert identify whether defined operators are useful and functioning correctly.

### 4.1 Existing Operator Evaluation

One way this is done is the collection of statistics on a per operator basis as the GA executes. This gives some indication on whether they contribute usefully to finding good solutions. Currently, the number of times an operator is used in the GA execution is tracked: each time the operator is applied to a genome, the operators counter is increased. There is also a counter associated with each genome, which counts the frequency of each operator's application over the genome's lifetime. Offspring genomes inherit the parent's counters, thus at any point in the GA execution, each genome contains the total accumulated applications since the initial population. (The lifetime of a genome is measured from the initial population until it, or a descendant, is removed). Another measure is the change in fitness that the operator application caused - each operator has a counter that records the cumulative fitness deltas over the GA's execution. This is done by taking the difference between a genome's fitness before and after operator application and adding it to the operator's total.

These current statistics give some idea of operator usefulness when tracked over time. Given the GA's search strategy, overall population fitness rises over time. Changes in the distribution of operator application counts over time show which operators are more frequently applied in the successful genomes.

The fitness delta for operators is more difficult to interpret. In past experiments with the CRP, it was easier to fix conflicts at the beginning of the evolution process, whereas for layouts already close to completion it was difficult to find changes that lead to successful solutions. In order to arrive at solutions, sometimes short-term fitness decreases have to be accepted in order to move out of a local minimum. Often operators that are not used or do not actually change the genome will attract less/no penalties, whereas useful operators could accumulate a net penalty in finding a solution.

The cumulative fitness delta as it currently exists was thus hard to use as a criteria for judging operator usefulness.

In order to improve the evaluation of operators, it would be useful to allow the expert to judge the application of those operators within the context of their application. An expert may select a genome and step through the application of operators manually. This is a useful tool in the evaluation of correct operator application, but is not as useful in evaluating it for usefulness towards the overall goal of achieving a final solution. The GA's search heuristics make up for this to some extent, since they will identify the better genomes. For complex domains this may not be taking full advantage of expert interaction. Since it is not feasible for an expert to review all genomes in a GA's execution, some automated methods should identify instances of operator application to be reviewed for their merit by the expert.

Based on such automatic means of selecting examples for review, the expert may modify an operator or its condition for application so that it can perform better. One could use these conditions in two different ways: either as part of a probabilistic weighting for selecting the mutation operator, or as part of the definition of a higher level operator.

When using NRDR, a natural progression is for the initial formulation of relatively simple operators to form the basis of the KB that is then extended by the expert to more abstract operators. This allows the expert to work at a more intuitive knowledge level when specifying new rules. In the context of operator evaluation, cases presented to the expert can be used to devise a higher level operator that takes into account under which conditions the operator is best used. This more abstract operator can combine a number of scenarios in one concept and thus be more reliable when used in an algorithm.

Here is an example from channel routing: Cases where the operator *move_horizontal_wire_up()* appears to be a good choice is where there is no vertical wire causing the wire conflict, while it usually leads to no improvement (reducing conflicts) or worse (increase in conflicts) if there is one. A higher level operator could be created called *fix_horizontal_wire()* which uses the condition *is_there_is_a_vertical_wire()* and the action *move_horizontal_wire_up()* if false, and *raise_horizontal_wire()* if true (*raise* in this context means placing the wire on a higher layer).

Alternatively, the expert may decide to use the knowledge gained from the examples to provide probabilistic weighting for operators. This would be appropriate should it not be clear to the expert what strategy would be best, however intuition might suggest one option would be better than another. In this case, the choice of operators would be left to a random selection, using weightings specified by the expert. The weight could also be associated with a pre-condition to decide which weightings to apply. When the mutation function of the GA is to be applied, a random selection based on these weights would be made.

The following example is a variation of the one given above. If the expert can't work out what condition is a good determinant for deciding between the two operators, but feels or notices that *raise_horizontal_wire()* has a better success rate, he might specify that 70% of mutations are to be *raise_horizontal_wire()* while 30% are to be *move_horizontal_wire_up()*. If a pre-condition is to be used, the expert might say that if *a_wire_on_next_layer()* is true, we would only with probability of 30% use *raise_horizontal_wire()* and with probability of 70% use

*move_horizontal_wire_up()* (if there is a wire on the next layer, it's not certain that it will cause a conflict after the move), otherwise 80% and 20% respectively are to be used.

### 4.2 Specification of Operator Probabilistic Weight

It is likely that the expert will not be able to give specific weightings to the probabilistic weightings. Instead, a simple scale such as "high", "medium" and "low" might be used, or assigned relative weights such as 1:3:4 and normalised. In the former case, one might even be able to let a learning mechanism such as co-evolution decide on exact percentage weightings, depending on empirical performance measures. The shape co-evolution could take would be to try and improve the operator application conditions from a large number of evolutionary processes on individual problem instances.

### 4.3 Identifying Candidate Operator Applications

Given that it is infeasible for the expert to review all operator applications in the course of a GA's execution, an automated procedure should identify candidates to be reviewed. In order to present an "interesting" selection to the expert, some criteria of candidacy need to be established. The candidate operator applications would be presented to the expert with a 'before' and 'after' genome to show the effect. The expert can then decide whether it is indeed an interesting case, and use it as a basis for KB refinement.

**Fitness Changes Due to Operator Application** One possibility for identifying candidates would use changes of fitness as a result of applying the operator. As previously mentioned, the context of such a fitness change is very important. For example, while an operation might incur a short-term fitness penalty, it might move the solution out of a local minimum with more long-term benefits. Alternatively, a short-term gain might be at the expense of some other operation which would prove better in the long term. An operation with no fitness change may have created no benefit, or may have cleared the way for a subsequently beneficial action.

In order to take account of these possibilities, a window can be established after operator application. This considers the fitness variation over a number of steps. Thus if there is a longer term benefit/disadvantage as a result of the operation, this would be taken into account. One might establish a threshold relative to the original fitness. Should this be exceeded in the positive or negative direction after $n$ steps, the candidates would be considered "good" or "bad" examples to be reviewed by the expert. In the extreme case, the window could extend over the lifetime of a genome. This would mean that it would be quite apparent whether a genome was successful in the (very) long term, but might potentially generate too many candidate operation applications for review.

**Counting Operator Applications** Another possibility of identifying operator application candidates builds on the existing algorithm described in section 4.1, counting the number of times an operator is applied over a genome's life span. When a genome is eliminated from the GA population, one could identify the point where it diverged from its last "relative" (i.e. it's last parent or where it "sired" an offspring). Assuming the "relative" is still in the population with a higher fitness value, one might review all operator applications since divergence as potentially "bad", and vice versa.

Another way an expert might use the operator application count is when the expert is unsure modifying an operator. The expert creates two versions of an operator, the original and one with the modification. Each is assigned a 50% probabilistic weighting. After running the GA, the version with the highest number of applications in the final population (i.e. those genomes that proved to be the fittest over time) would be the more useful one.

**Possible problems with Crossover** A caveat for these two approaches exists: The HeurEAKA GA uses either asexual reproduction, where a parent is copied verbatim and a mutation introduced, or sexual reproduction, where there is a crossover copy of two parents. The crossover operation could introduce modifications to the genome that are unrelated to mutation. Crossover usually causes a large change in the genome, potentially radically changing its fitness. When comparing to a "relative" produced by sexual reproduction, this could drown out any fitness effects that operator applications might have.

Since we might be looking at a specific operator used among $n$ others, the other operators would cause fitness change extraneous to the one we're interested in. If these operators are high level and potentially cause extensive genome changes, we could argue that sexual reproduction is just a special case of operator and should thus not be singled out.

(In experiments for the channel routing problem, only 20% of genomes were produced sexually, so there would still be sufficient data if we decided to exclude these cases).

### 4.4 Using relational constructs for operator conditions

In section 4.3 we discussed that a number of operations could be considered at a time in order to view their effect in a larger context. This entailed setting a threshold for a cumulative fitness change over a number of operations. When reviewed by the expert, it might be apparent that certain operators have a synergistic effect on each other. Should this be the case, the expert could encourage the coordination of these operators by creating a higher level operator that incorporates combines them. Another, more flexible, option would be to create a condition boosting the probabilistic weight of one operator should the other one be present. This would take the form of a relational expression in the precondition for probabilistic weighting considering what the past operations were. So, for example, *Move_Horizontal_Wire_Right()* would be assigned a weight of, say, 0%, if the previous operation was *Move_Horizontal_Wire_Left()*.

## 5 Control of GA Parameters and Evolutionary Phases

It can be useful to vary aspects of the GA's operation during execution. Currently HeurEAKA has the ability to access overall GA population fitness and generational information. This supports a strategy tried for the channel routing problem: limit initial layout solutions to a single layer, later add a second layer once the first one has been optimized. The decision to add the second layer is taken based on the last change in average population fitness. If the last change in average population fitness lies back sufficiently far, e.g. 1000 generations (determined empirically), we have probably arrived at population convergence with an inability to further improve fitness. At this point, we add a second layer to the channel routing solution, clearing the way to further conflict resolution by "raising" wires to the second layer.

Our experiments found that this "incremental" layer assignment approach turned out to take slightly longer (due to waiting for convergence) than simply allowing access to both layers from the beginning. In the experiments we ran, we did not notice a difference in the ability of solving problems.

While these experiments were somewhat limited and did not appear to be too promising, it would nonetheless be interesting to pursue other strategies, possibly in other domains, that alter the GA dynamics as the search progressed.

Examples include: changing the ratio of crossover operations depending on population convergence or selecting more aggressive mutation operators to move the population out of a local minimum. As with the channel layout problem, there may be domain-specific examples where different strategies could be adopted at different evolutionary stages.

## 6 Heuristic RDR vs. other forms of RDR

In order to differentiate our approach to KA using the RDR philosophy, we will introduce the term "heuristic" RDR (HRDR).

### 6.1 HRDR: Essential difference in RDR Usage

HRDR works as an integral part of a probabilistic search, where an expert's heuristic knowledge is encoded. The way it is deployed differs to other kinds of RDR, which are deterministic and usually used for classification type applications. It is also different in that actions in HRDR form part of an algorithm. Each rule can contain a number of actions for a single condition. (In Multiple Classification RDR one could have a similar situation, however each rule would contain one action and the conditions would all be the same). Each of the HRDR actions can be a reference to a nested RDR which could cause the execution of further actions.

### 6.2 Differing Assumptions on Expert Knowledge

Our approach to knowledge acquisition is slightly different to traditional ones. As noted previously, we support a philosophy of exploration where the expert provides heuristic knowledge as guidance. In other RDR frameworks, the acquisition of knowledge proceeds by eliciting specific knowledge from the expert, usually in the form of a classification based on a presented case. The assumption behind this is that the expert already knows what the correct answer is, and "justifies" it in terms of referring to the selected case.

In our approach, we aim to 'discover' a way of solving the problem. This means that the expert guides the search strategy by incrementally adding to/refining the search operators, without initially knowing what the correct approach will be. In this context, it is more the intuition and heuristic knowledge of the domain that the expert will draw on, rather than definitive 'factual' knowledge. Furthermore, we expect that rules entered by the expert might be incorrect and subject to revision (in other RDR approaches this is also possible, but is handled in a different way). We also allow the expert to be speculative, by providing a probabilistic rule selection mechanism, in which the expert can supply a weighting. The expert can then adjust the weighting based on empirical performance of the algorithm if so needed.

In the current approach the justification of using GAs and probabilistic elements is that our KB is assumed to be "probably approximately correct". This means we do not demand that the KB is necessarily correct or complete. Rules will not always be applied correctly since there is not always supervision by the user. Also the selection of rules when the GA is executed is left to chance, so there is no guarantee that an 'optimal' rule will be picked. Nonetheless, the heuristic search will be able to cope with an incomplete ruleset and still find useful solutions. The idea is that one provides generally useful operators and heuristic knowledge (e.g. in the form of operator weightings) and take advantage of the flexibility of a GA in coping with any "noise".

### 6.3 Revising rules already in the KB

Modification of rules in the KB is not allowed with other implementations of Ripple Down Rules, as it is assumed every rule entered into the KB is correct. This is done due to rule editing having undesirable side effects which are not easy to control. Under this scheme if an expert wants to make any modification in behaviour, these need to be made in the form of exceptions. The reason for the no-edit policy is to ensure integrity and correctness of the KB - subsequent modification of rules in an established tree can cause conflicts. In RDR the tree is not restructured once built, so these conflicts will not be fixed.

A problem presents itself when a form of nesting is introduced into the KB. In repeat inference RDR (RIRDR) and nested RDR (NRDR) conditions of a rule can reference a definition found elsewhere in the KB. This definition takes the form of a RDR structure (NRDR and RIRDR differ in implementation, but essentially share this quality). In this way, when the referred to RDR is modified by the addition of a rule, the referring condition is indirectly modified. This means that previously classified cases (e.g. the one used to compose the rule) might no longer match the condition. To combat this, RIRDR forces a strict schema on the tree traversal by using a form of version control to ignore changes made to the KB after the referring condition was added.

NRDR as proposed by Beydoun [2] could have used a similar approach as RIRDR, but instead uses a different strategy. Each rule has a set of cases associated with it, these are the ones classified correctly under the supervision of the expert. Whenever a modification to the NRDR is made, these cases are re-evaluated and any discrepancy in classification causes the system to prompt the expert to perform a *secondary refinement*. This *secondary refinement* means that NRDR conditions have to be modified to ensure the case's classification remains correct.

While these are generally valid reasons for not modifying rules, our initial experiments suggest that it is very useful and sensible to modify rules at the beginning of building a knowledge base. In particular, for the definition of new actions of modifying genomes, it proved useful to have this option. In the channel routing experiments, the low level operators contained complex sequences of actions which needed to be debugged. The assumtion that the expert will get a rule correct on the first attempt would have severely hampered this process.

Experiments were done where exceptions needed to be added into the NRDR for each modification. The operators became convoluted and brittle when attempts were made by the expert to review their behaviour. This is exacerbated by the fact that rule 'conclusions' in HRDR are snippets of RSL code, potentially containing control structures, variables, RDR references etc. It would be quite impractical to force the expert to add a new exception with equal conditions and a minimally modified action body for each edit.

We found that when an expert added more high level rules, the need to edit rules in the RDR went down significantly. This is due to the reduced complexity of the rules and the more intuitive knowledge level.

Our algorithm is "approximately correct" and robust when it comes to incomplete or inconsistent conclusions by the reliance on probabilistic search and GA heuristics. We also do not, currently, store cornerstone cases (see discussion below), and also allow existing rules to be edited in contrast to existing RDR approaches. Given these characteristics we have not enforced a version control schema for the KB. Our experiments suggest that our approach is nonetheless effective.

### 6.4 RDR cornerstone cases

In regular RDR a case is stored in the KB when a rule is added. This case is classified under the supervision of the expert and serves as the context in which a new rule was composed. These cases are called *cornerstone cases* and are used for assisting the expert in the subsequent tasks such as addition of further exception rules.

Traditional RDR systems can suggest conditions for new rules based on cornerstone cases and the conditions of rules that did not fire. In our current framework this is not supported for two reasons. Firstly, the rule space is very large - it extends over expressions containing GA attributes, Rule Specification Language variables and nested RDRs with parameters. Coming up with useful suggestions based on this large set is quite hard - one might consider machine learning methods for suggesting conditions. The second reason is that we are dealing with non-deterministic choices in the probabilistic algorithm. These choices are not reflected in the attributes tested by conditions, therefore past cases are not usually sufficient for determining how one arrived at a certain conclusion.

The approach described in section 4 allows the system to identify example cases where operators were applied. These are presented to the expert who can then decide whether existing rules need to be revised, new operators introduced, or probabilistic selection added. If the KB is modified, the previously selected examples can be re-evaluated to see what effect the changes had. In this way, the expert is assisted in a similar way to the use of cornerstone cases in traditional RDR. One caveat is that when selected cases are re-evaluated, care needs to be taken in making all probabilistic elements deterministic so as to match the previous evaluation. Only this way can the expert be sure that variation in outcome is only as a result of modifications made. The currently implemented debugging environment in HeurEAKA already does this to make testing of individual cases possible. (Note that this is useful only for debugging, since general use would rid us of an important feature of HRDR).

## 7 Conclusion

We have briefly discussed our framework for improving the performance of GAs by using knowledge acquisition techniques in the form of Ripple Down Rules. Our approach is used for the development of heuristic algorithms to be used in solving complex combinatorial problems. We argue that our techniques allow for the faster development of solutions than conventional techniques. Our experiments (briefly described in this paper, more detail is available in [1]) show that we are able to develop an algorithm in about a week. This algorithm produces results approaching those of industrially used algorithms which took many years to develop. With some planned minor modifications, we should be able to evaluate it against a wider range of benchmarks.

We pointed out why using our approach of adapting GAs to a problem domain is preferable than traditional 'trial and error' approaches. Furthermore, we presented arguments why we believe that our techniques should be readily transferable to other domains, and that the amount of effort should be less than if done with conventional GAs. We plan to apply our techniques in other domains, such as scheduling, to provide evidence of that.

In order to improve our knowledge acquisition yield from an expert, we have discussed more advanced techniques in evaluating an existing knowledge base. We intend to provide the expert with automated methods of identifying and reviewing scenarios in which operators can be improved. The identification of examples is based on tracking operator applications on genomes in terms their ability of contributing to successful solutions. Through a process of reviewing these examples, the expert can suggest new operators or strategies under which to try the existing operators.

A novel technique for controlling GA operation from within a knowledge base is also presented. This is only briefly discussed, but could have far-ranging implications for controlling GA deployments using a KB if extended to encompass more aspects of GA construction. After the choice of representation and fitness and mutation functions, the next biggest problem of using GAs is the appropriate configuration of a sizable array of parameters determining the operation of the GA. This paper touches on how this might be addressed with KA.

Finally, we outline differences of our approach to RDR when compared to other forms of RDR such as 'traditional' use of SCRDR, MCRDR, RIRDR and NRDR. These are partly due to a different approach on the kind of knowledge they encode and how this is used - characterised by the name, HRDR. Other differences are of a more practical nature such as controlling how rules may be revised and maintenance of case databases.

We are confident that further research and experimentation in the directions outlined in this paper will allow us to make our framework more effective in capturing heuristic and domain expertise. We also believe that it will be more portable and applicable to a wider range of domains by making the architecture more domain-neutral, allowing the expert flexibility in reducing search space dimensionality (more structure by identifying higher level operators, and the use of strategic knowledge through probabilisitic weights) and supporting more speculative exploration by the expert (easier revision of rules, hierarchical RDR and potentially introducing limited machine learning in probabilistic selection).

# Incremental Learning of Control Knowledge for Lung Boundary Extraction

Avishkar Misra, Arcot Sowmya and Paul Compton

School of Computer Science and Engineering
University of New South Wales
NSW, Australia
amisra@cse.unsw.edu.au
sowmya@cse.unsw.edu.au
compton@cse.unsw.edu.au

**Abstract.** The goal of this work was to develop an adaptable computer vision system that refines itself to the specific task of extracting lung boundary in High Resolution Computed Tomography (HRCT) scans. We have developed an incremental learning framework called ProcessRDR that allows the underlying procedures of a computer vision system to learn knowledge pertaining to their control. This approach to learning control knowledge provides a systematic mechanism to customisation of the procedures for a domain, whilst the system is in operation.

## 1 Introduction

Computer vision is generally formulated as a two step process. Firstly, image analysis for *feature extraction* processes the input image(s) to extract the feature of interest to the system. This is followed by *recognition and classification* of the extracted features according to a semantic model of what is expected in the scene or image(s).

The success of computer vision systems depends on the strength of feature extraction and classification processes. It is generally accepted that these processes work in tandem and complement each other. A complex and sophisticated feature extraction process would focus on the important features of interest and provide features that can accurately delineate the object in the scene. Therefore a simple classification process would do the job quite well. A system with a simplistic feature extraction, however, would require a significantly more complex and powerful classification system.

Computer vision has seen various approaches to improve both the classification and feature extraction stages. The initial focus was on improving the underlying algorithms or procedures used for feature extraction and using machine learning and pattern recognition to improve the classification processes. The learning for classification processes was directed to accommodate *domain knowledge* of what is expected in the scene.

However, by the last decade, computer vision experts had amassed a variety of underlying procedures and were faced with a new problem of how and where to use these procedures. Some procedures are only applicable under certain circumstances or for certain types of images. For example, image enhancement and restoration will work to improve a poor quality image but will adversely affect a high quality image in terms of loss of information. Other procedures, designed to be image independent, have to be specialised for a particular domain via their parameters.

The selection of optimal parameters for a procedure within a specific domain requires expertise in the field of computer vision and often in the specific domain of application. These experts draw upon their knowledge about computer vision and the domain of application to try and find the optimal parameters for the task. This process often ends up being rather an ad-hoc process of trial and error, at the end of which there is no guarantee that the chosen set of parameter values would be optimal in all situations.

This problem of learning *control knowledge* [20] to use vision procedures is complicated through the high degree of variability in the optimal solution for different applications. Not only do we wish to determine the optimal set of parameters for a procedure in a domain, but also how to combine a number of procedures to solve a larger and more complex feature extraction task.

In this paper we present a scheme for incremental learning of control knowledge for computer vision systems. We have built a system, which learns from the expert how to extract the lung boundary from High Resolution Computed Tomography images. The following section will provide an overview of relevant computer vision problems and the need for appropriate learning mechanisms. In section 3 we will introduce ProcessRDR and discuss a prototype lung boundary extraction system in section 4. The paper concludes in section 5.

## 2 Background

The advent of medical imaging technologies such as X-Ray, Magnetic Resonance Imaging, and Computed Tomography [41] provides doctors with a non-invasive alternative to looking inside a patient.

As a result Computer Aided Diagnosis (CAD) systems which use medical imaging technologies, have been an area of active research in the last two decades [1] [2] [4] [5] [6] [7]. [3] provides a good overview of the various techniques and developments within the field of Pulmonary (Lung) Imaging and Analysis, used by various CAD systems designed for the Lung.

The identification of lung boundary (pleura) within Lung CAD systems is an important step towards detecting diseases and abnormalities for a patient. An accurate delineation of the lung boundary is especially important when detecting diseases affecting the walls of the lung and its neighbouring regions.

A number of systems similar to [8] were developed to extract lung boundary, relying on computer vision procedures and hand-crafted heuristics that combine these procedures together. Noisy data or variations within the norm, can affect the results and true location of the lung boundary. Approaches like [11] and [12] look at ways to improve the feature extraction algorithms themselves. Here the responsibility of developing accurate feature extraction algorithms resides with the computer vision expert. In both cases the researchers have sought to improve the underlying feature extraction algorithm manually.

As the domain becomes more complex, hand-crafted feature extraction algorithms must be supplemented with heuristics and generalisations. Often these heuristics are developed on a trial-and-error basis and make amendments difficult. The development of these algorithms themselves is a difficult and time consuming process, with no assurance that the developed algorithm using an expert's control knowledge would be sufficiently flexible in all circumstances.

In order to provide the needed domain knowledge, [9] and [10] have used an anatomical model for segmentation. Their process involves generating a model of expectation, which is used to support their lung boundary segmentation algorithms. Though this approach incorporates greater degree of domain knowledge, the models developed for such systems need to address the large variability in the lung shape and size across patients and within a patient. For example the lung boundary in the top third of the lung has a remarkably different shape and size to the lung boundary in the bottom third of the lung. Even if the variability was accounted for during the stage of generating the model, we have to question the ability of these model-based approaches to accommodate and learn from their failures.

Such model-based computer vision systems have been of interest in medical and non- medical applications of computer vision. [13] [14] [15] [16] [17] [18]. The limitation of model-based techniques is that they are generally applied either at the classification stage or the later stages of feature extraction. Here these techniques do not learn control knowledge to guide the feature extraction and instead use domain knowledge to select features that best suit the model.

A number of researchers have looked at learning control knowledge, through Parameter Tuning and Task or Goal based expert shells for computer vision.

Work by [21], [22], [23], [24] and [25] learn the optimal set of parameters or values for the procedures, by using statistical measures of success. These approaches are not successful when we have limited amount of data, or when the measure of success cannot be defined in statistical terms, which is often the case.

In other approaches, [26], [27], [28], [29] and [30] have developed computer vision systems to try to learn from human experts and expect their teachers to manually describe the solution to a problem in a structured way. In [26] for example, the system requires a computer vision expert to explicitly define all concepts of a problem and its associated solution. This form of learning requires the Computer Vision expert to pre-empt all possible outcomes and define a complete solution to the problem. This is not only impractical for many domains but also fails to capture all possible knowledge for a particular domain.

Research in Knowledge Acquisition, [31] [32], has shown that though human experts maintain an internal structure to their knowledge, they lack the ability to communicate the complete knowledge in a structured way. Instead they can justify a decision made, by using their knowledge. So irrespective of the level of expertise of a person, he or she cannot completely articulate that knowledge.

Ripple Down Rules (RDR), [33] [34], propose an approach to knowledge acquisition that addresses this problem. RDR and its variants, [35] [36] [37], are Knowledge Bases, composed of a tree or chain of rules. Each rule and its parent rules define the context or conditions which must be met in order for the consequent of the child rule to classify or act on the queried input. If RDR's rules make a mistake, the expert can teach the system by creating a child rule as an exception to the rule where it failed. Here, the problem of experts articulating their knowledge is addressed, by asking the expert to justify his/her reasoning. This justification, which is intuitive and highly effective, forms the context of an RDR rule.

This approach can also extract tactical and non-factual knowledge which is even harder to articulate using traditional methods. RDR's ability to learn incrementally on a case-by-case basis is an additional benefit to applications where the data is sparse, making it difficult to learn using statistical measures or traditional machine learning.

A number of alternatives have been proposed to the aforementioned Single Classification RDR techniques. Nested-RDR[35] for example, works to describe complex concepts within separate RDRs of their own. Here each RDR learns to describe a specific concept and the connected RDRs work together to solve a more complex task.

Multiple Classification RDR (MCRDR)[36], provides the capacity of the same RDR structure to maintain differential classification for the same case. It allows for the possible alternatives in classification for a given case. Park et al, [38] have used MCRDR to carry out lung boundary extraction from X-ray images. The knowledge-based method used works only at the classification and region selection level. The MCRDR is never used to guide the underlying processing.

Evaluation of various lung extraction techniques has posed a dilemma of its own. Due to the high degree of variability across patients, there is no correct lung boundary against which segmented boundaries can be evaluated. There have been a number of attempts to describe and measure the level of successful boundary extraction, [39] [40], but there is no technique which provides an automated statistical measure. This problem is exacerbated when even radiologists themselves cannot always agree upon where the lung boundary should be drawn. This makes statistical evaluation of a lung boundary difficult, and we have to rely upon radiologists deeming a boundary to be acceptable or not.

## 3 ProcessRDR

We have already alluded to the difficulty of learning for feature extraction procedures within computer vision, which requires a great deal of expertise and refinement. The manual customisation of feature extraction algorithms for a particular domain can often lead the development in an ad-hoc way. Though some might argue that a formal approach to software engineering can address this ad-hoc nature, it can equally be argued that these do not work well unless the programmer has complete knowledge of all types of inputs within the domain. Considering that computer vision systems are developed by experts to solve problems in other highly specialised domains, it is unreasonable to expect the expert to understand all the nuances of that domain.

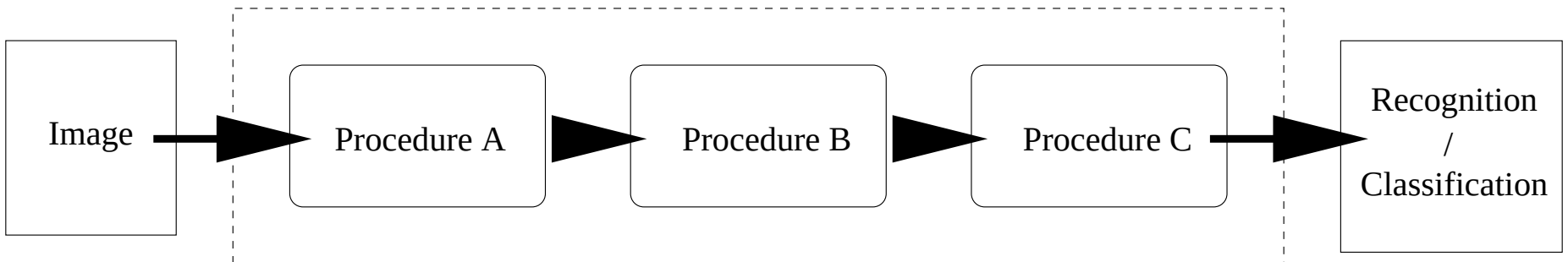

**Fig. 1.** Generalised Computer Vision system.

At the same time, work in Knowledge Acquisition [31] points out that even with complete knowledge, the programmer would have problems articulating a solution which cover all conceivable cases. Clearly, we need to develop computer vision systems that learn control knowledge to refine themselves during operation, and continue to improve beyond the initial training stage.

We propose a solution by extending RDR to feature extraction processes in computer vision. Here we are using RDR to learn control knowledge from the expert and subsequently use it to guide the underlying procedures during operations. Hence the so called ProcessRDR. RDR's learning mechanism provides a systematic approach to knowledge acquisition and maintenance, even when used in an ad-hoc method of operation.

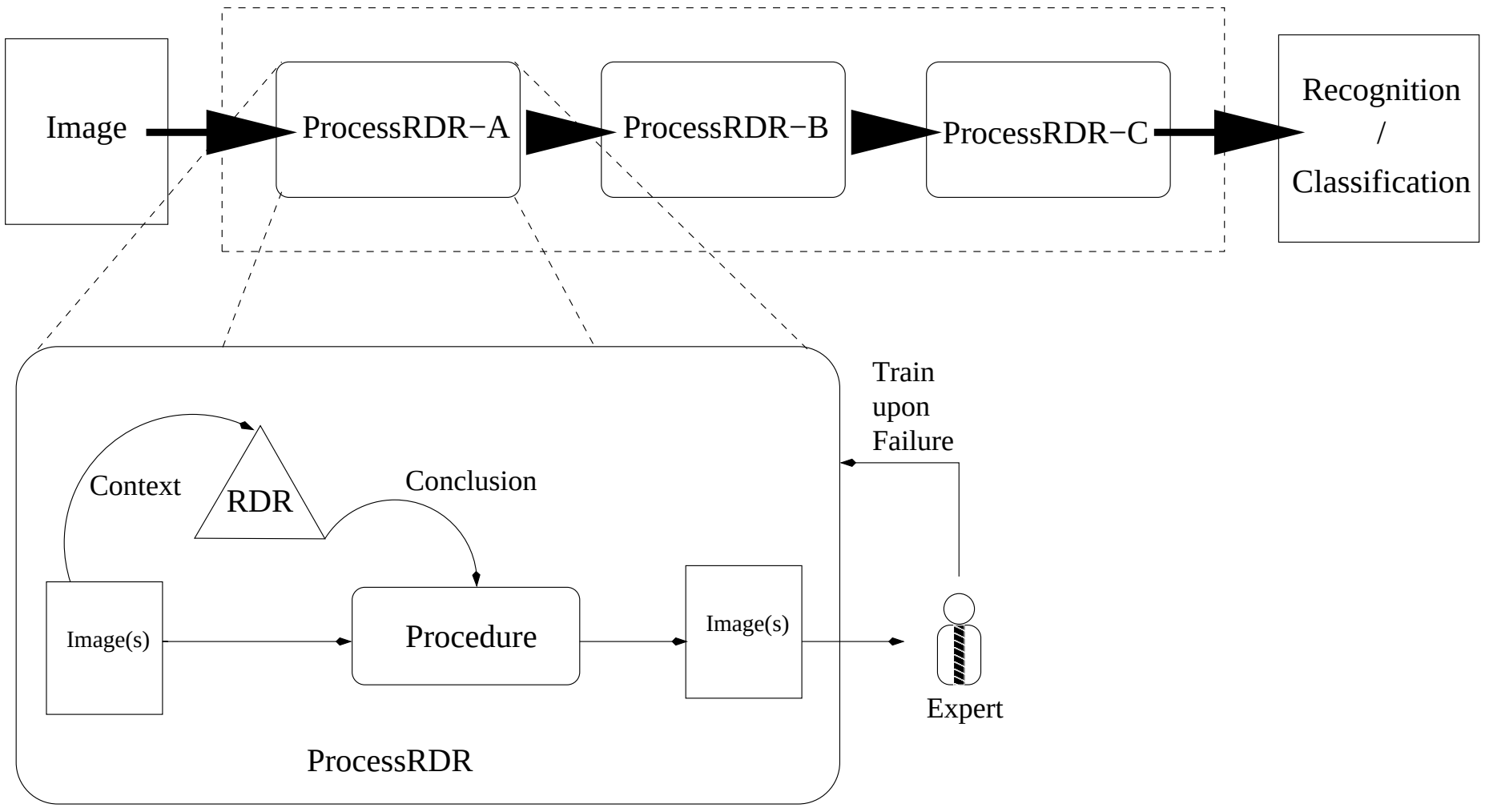

**Fig. 2.** ProcessRDR for a Computer Vision system.

A computer vision system, as shown in Figure 1, can be thought of as a sequence of procedures, connected up in a specific order by a programmer to solve the task at hand. ProcessRDR involves isolating important and configurable processes within a system, and attaching an RDR knowledge learner to it. Since complex computer vision systems have a number of procedures to carry out the necessary feature extraction, we would end up with a number of separate RDRs. A new image would pass through each of the ProcessRDR procedures in the sequence. Here the RDR controlled procedures carry out their specific processing on the image. For example in Figure 2, an image passing through ProcessRDR A would be transformed by the procedure. The transformed image would serve as the input for ProcessRDR B. The sequence of ProcessRDRs will conclude with the recognition and classification, which was the system's overall objectives.

RDR has been applied to a range of problems [37]. The closest of these is the use in configuration, also knowns as parameter optimisation. The difference with ProcessRDR is that, rather than producing some appropriate combination of parameter values in a static configuration, parameter values are selected for a sequence of processes and the output is the result of the processes, not the set of parameter values.

The learning of control knowledge involves learning of the optimal parameters for a specific case or domain, as well as the learning of sequencing of underlying computer vision procedures to solve a complex task. The ProcessRDR mechanism allows an expert to teach each of the procedures, the optimal operational control of that procedure or module which comprises of smaller atomic procedures.

The fine-grain control of learning with a specific task means that we can potentially teach a ProcessRDR using separate experts. This allows computer visions systems spanning a number of different domains to learn from each of the domain experts separately. For example, in a lung boundary extraction system the ProcessRDR would allow both computer vision experts and radiologists to collaborate on their domain-specific knowledge.

As the ProcessRDRs for a complex system are connected in a sequence, the conclusion for a case in a ProcessRDR, would effect the decisions and localised context of subsequent ProcessRDR. This is because the conclusion from one ProcessRDR serves as the input for the next ProcessRDR.

Therefore any correction made in ProcessRDR A in Figure 2, would warrant a re-evaluation of the cornerstone cases in ProcessRDR A as well as all the cornerstone cases in B and C. While any correction in ProcessRDR C will only require re-evalutation of local cornerstone cases.

## 4 ProcessRDR Application

In order to test our ideas, we applied ProcessRDR to address an important problem in medical imaging. The objective was to develop a system using ProcessRDR to learn control knowledge for the individual vision procedures to extract the lung boundary. We compared the performance of a non-ProcessRDR system against a system build using ProcessRDR. The non-ProcessRDR system was an existing simple lung boundary extraction system developed by Po, [8], known as Po Boundary Extraction. We compared the same cases and their results against an ProcessRDR Boundary Extraction system.

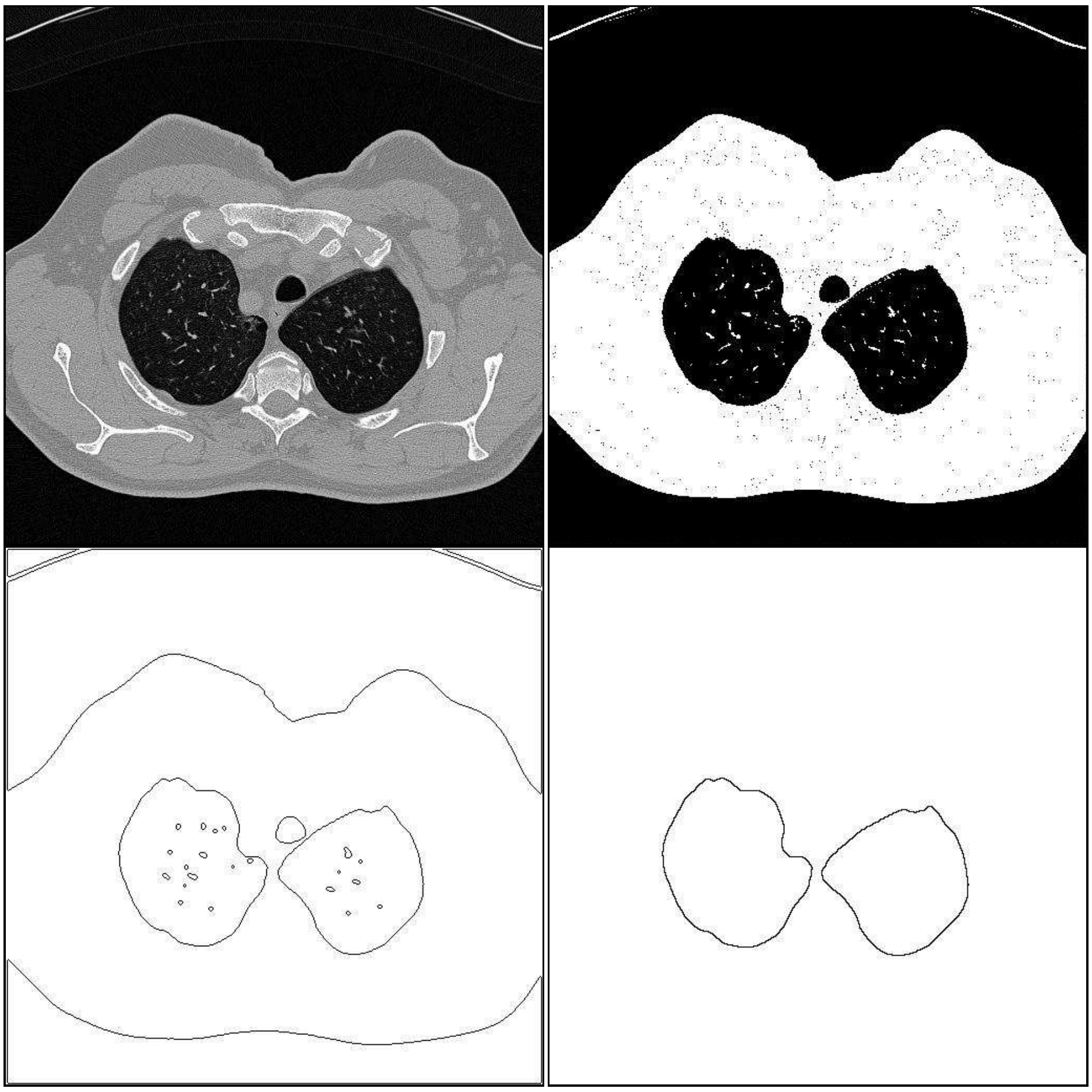

**Fig. 3.** A Generic Boundary Extraction Process (Top, Left to Right):(a) Original image, (b) after thresholding. (Bottom, Left to Right):(c) After morphological cleanup and (d) final lung boundary. Note that this sequence represents an example of expected behaviour, which Po Boundary Extraction does not achieve, and ProcessRDR Boundary Extraction converges towards.

### 4.1 Po Boundary Extraction

Po Boundary Extraction, hence forth known as Po-BE, uses standard Computer Vision procedures in an approach similar to existing boundary extraction systems mentioned earlier. The underlying algorithm has 3 main Vision Procedures and the results to each step can be seen in Figure 3.

1. Thresholding - carries out a grouping of similar pixel intensities and binary separation according to a defined threshold value. The pixels inside the lung have a lower pixel value compared to regions outside the lung. The thresholding process attempts to get the best separation between regions inside the lung and outside the lung as shown in Figure 3(b). The selection of the optimal threshold value can vary from image to image.
2. Morphological Operations - a series of erosion, dilation, opening or closing operators are applied to remove as much of the noise as possible, while preserving the integrity of the lung boundary. The order and number of times these operators are applied are important and can also vary from case to case. Once finished, an outline of the candidate regions is carried out as shown in Figure 3(c).
3. Connected-Component and Boundary Selection - Here the procedures find the most likely candidate for regions defining the boundary and present the extracted boundary. This is often very hard to define in explicit terms or even non-conflicting heuristics. Figure 3(d) shows an extracted lung boundary from the original input as shown in Figure 3(a).

Po developed this algorithm manually and through trial-and-error. In doing so, he produced optimal set of parameters for thresholding, sequences for morphological operators and boundary selection.

### 4.2 ProcessRDR Boundary Extraction

The ProcessRDR Boundary Extraction, hence forth known as ProcessRDR-BE, was developed to have the similar underlying procedures as Po-BE. The only difference was that we attached a RDR learner to each of the configurable procedures of the system. The procedure with the associated RDR forms a ProcessRDR bundle. The complete system is a connected sequence of ThresholderRDR, MorphologyRDR and RegionSelectionRDR.

The context for each of the ProcessRDRs to evaluate a case can be broken into the following categories:

1. Image properties - defines the properties of either the original Image or the image which is the immediate input to the ProcessRDR. It includes:
   - Statistical measures such as pixel intensity, mean and variance.
   - Objects of interest and their properties (i.e. number of objects)
   - Textures and region properties
2. HRCT or scan properties - defines the properties of the scan itself which are generally available from the HRCT header:

- Patient orientation during scan. i.e. prone vs. supine
   - Slice location and range. i.e. image number 3 of 20.
   - Image reconstruction properties (scaling and algorithms)
3. Patient properties - defines the properties of the patient. This information is also available from the HRCT header, but may eventually include intermediate diagnosis as a result of querying external modules. Examples are:
   - Personal details such as age and sex.
   - Previously diagnosed disease or presence of other disease processes.

Though many of the context evaluations are common and can be treated in global fashion, we opted for a specific, localised evaluation within each RDR. The reason is that different ProcessRDRs might interpret the concepts in different terms. For example, the concept of mean in ThresholderRDR is seen as the mean of the pixel intensities, whilst the mean in RegionSelectionRDR refers to the mean area of candidate regions. Here the context evaluation changes due to the nature of the image data.

The conclusions within each ProcessRDR are the parameters to use for the processing and depend purely on the type of procedure we are dealing with. For example, ThresholderRDR's conclusion is the minimum and maximum thresholding values. The conclusions in MorphologyRDR, however are sequences and iterations of morphological operators.

The experts or users of the systems are responsible for training the RDR. Clearly there are two domains which interact and overlap - computer vision and medical imaging. In order to facilitate learning directly from users in medical imaging (i.e. radiologists) we developed appropriate graphical user interfaces, which allow the non- vision expert to articulate expected behaviour of the underlying procedure through a mouse. The designed interface allows a person who is not an expert in computer vision to actually define the context and processing conclusions for a procedure. For example, Figure 4 shows the interface to select the optimal threshold value. The user moves the scrollbar until the desired thresholding is achieved. We are able to build such interfaces for procedures that allow some form of mapping from the low- level procedure control/response to high-level visualisation.

### 4.3 Comparision/Results.

The ProcessRDR-BE system is still a sequence of vision procedures as with Po-BE and other similar techniques. The difference in ProcessRDR-BE is that each of these procedures acquire control knowledge of optimal operation within their domain, over the life-time of the application. ProcessRDR-BE will continue to expand its control knowledge and refine the underlying algorithm, directly from end user interactions.

Since there is no gold-standard in lung boundary evaluation, we could only evaluate the results of Po-BE and ProcessRDR-BE visually. Here the boundary is deemed acceptable or unacceptable, which automatically warrants a correction by the expert.

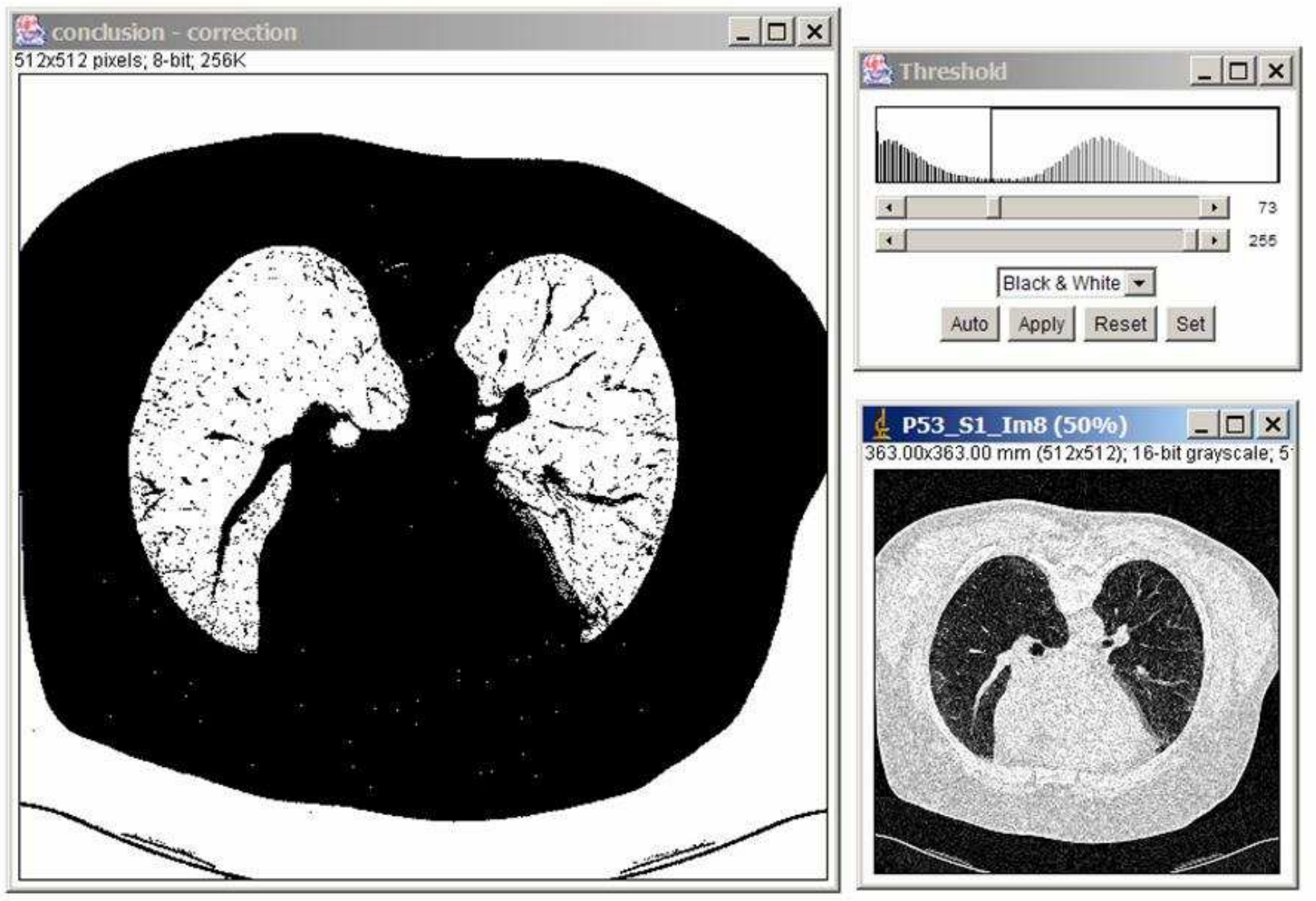

**Fig. 4.** GUI for teaching correct threshold value. Bottom-Right pane shows the original image. Left pane shows the affects of thresholding, which is controlled by the slider in the Top-Right.

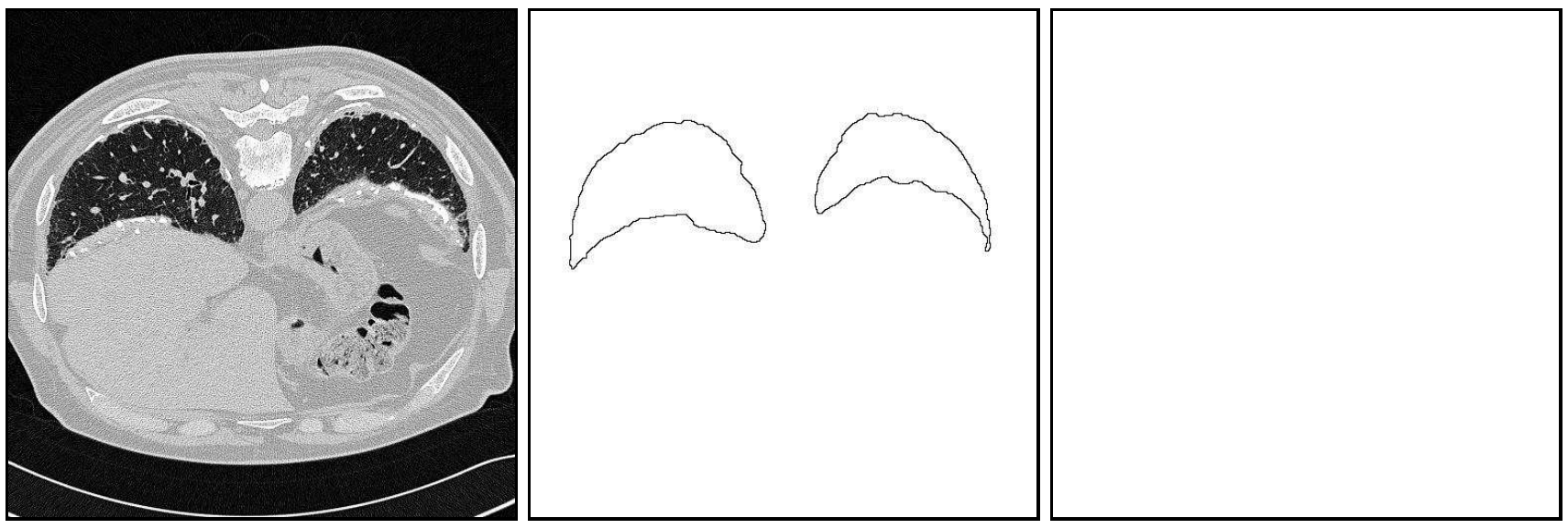

**Fig. 5.** Prior To Training (Left to Right): Original image, Po Boundary Extraction result and ProcessRDR Boundary Extraction with no result.

Prior to training, the ProcessRDR-BE's default rule settings were clearly not optimal and ProcessRDR-BE could produce no result, as shown in Figure 5. After training the performance of the ProcessRDR-BE started to converge and was comparable to Po-BE. Figure 6 shows a failure of both systems. Po-BE included a smaller circular region, known as the trachea, into the lung boundary. The trachea is often mistaken to be a part of the lung by Po-BE if it lies close to the boundary. This is due to selection of morphological operators and region selection procedures in the system. The ProcessRDR-BE's MorphologyRDR was able to keep the trachea isolated from the lung boundary, but the RegionSelectionRDR failed to eliminate the smaller region. The expert was able to define the appropriate rule to correct this behaviour as shown in Figure 7. Though it might seem that RegionSelection procedures within Po-BE could be directly improved, the reason why Po-BE failed to exclude the trachea from the boundary was a failure during Morphological cleanup of the image. Po-BE makes a single generalisation in order to accomodate cases where 'dilation' must be applied multiple times. ProcessRDR-BE gets around this problem by allowing the system to treat those cases differently.

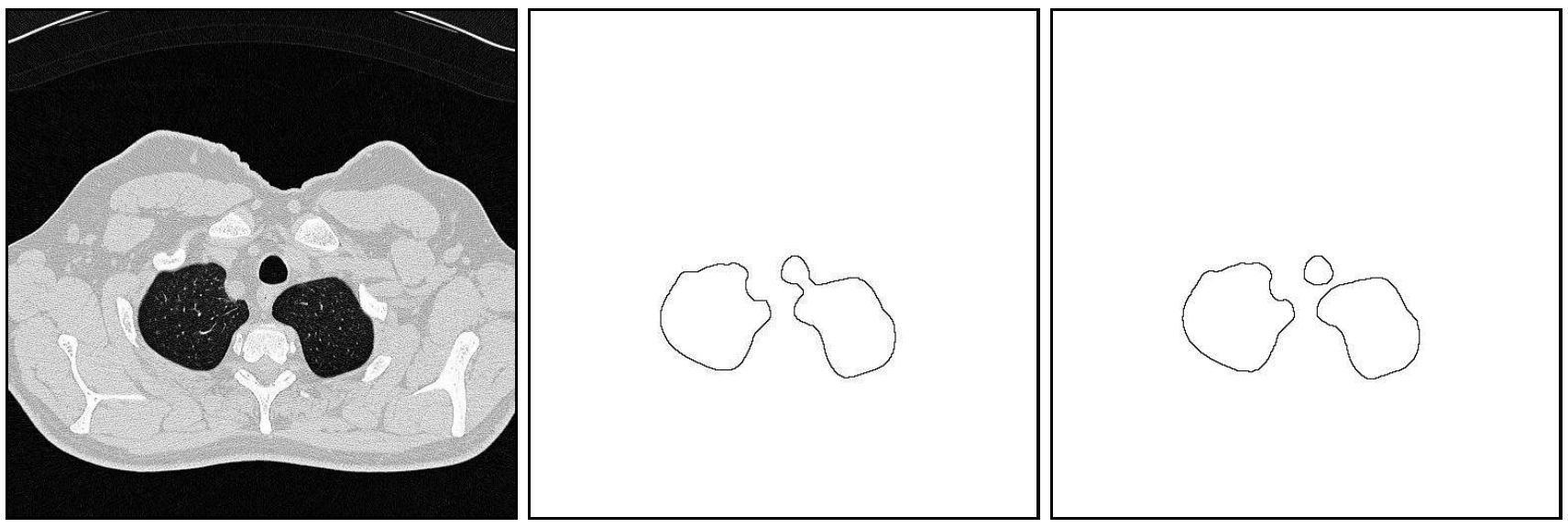

**Fig. 6.** Failure During Training (Left to Right): Original image, Po Boundary Extraction result and ProcessRDR Boundary Extraction result.

## 5 Conclusion

We have presented an incremental learning technique for Computer Vision systems that uses Knowledge Acquisition to refine the control knowledge. In this work we have defined and used ProcessRDR, that is taught directly by the radiologists on how to extract the lung boundary.

The ProcessRDR framework allows for continual learning and refinement of control knowledge which guides the underlying procedures in a complex computer vision systems. Traditionally, these procedures are customised by experts via modification to the algorithm itself, development of supporting heuristics or a selection of parameters. The approach presented eliminates the problems associated with what is inherently an ad-hoc refinement process, and instead offers a structured approach to learning of knowledge to guide the processing.

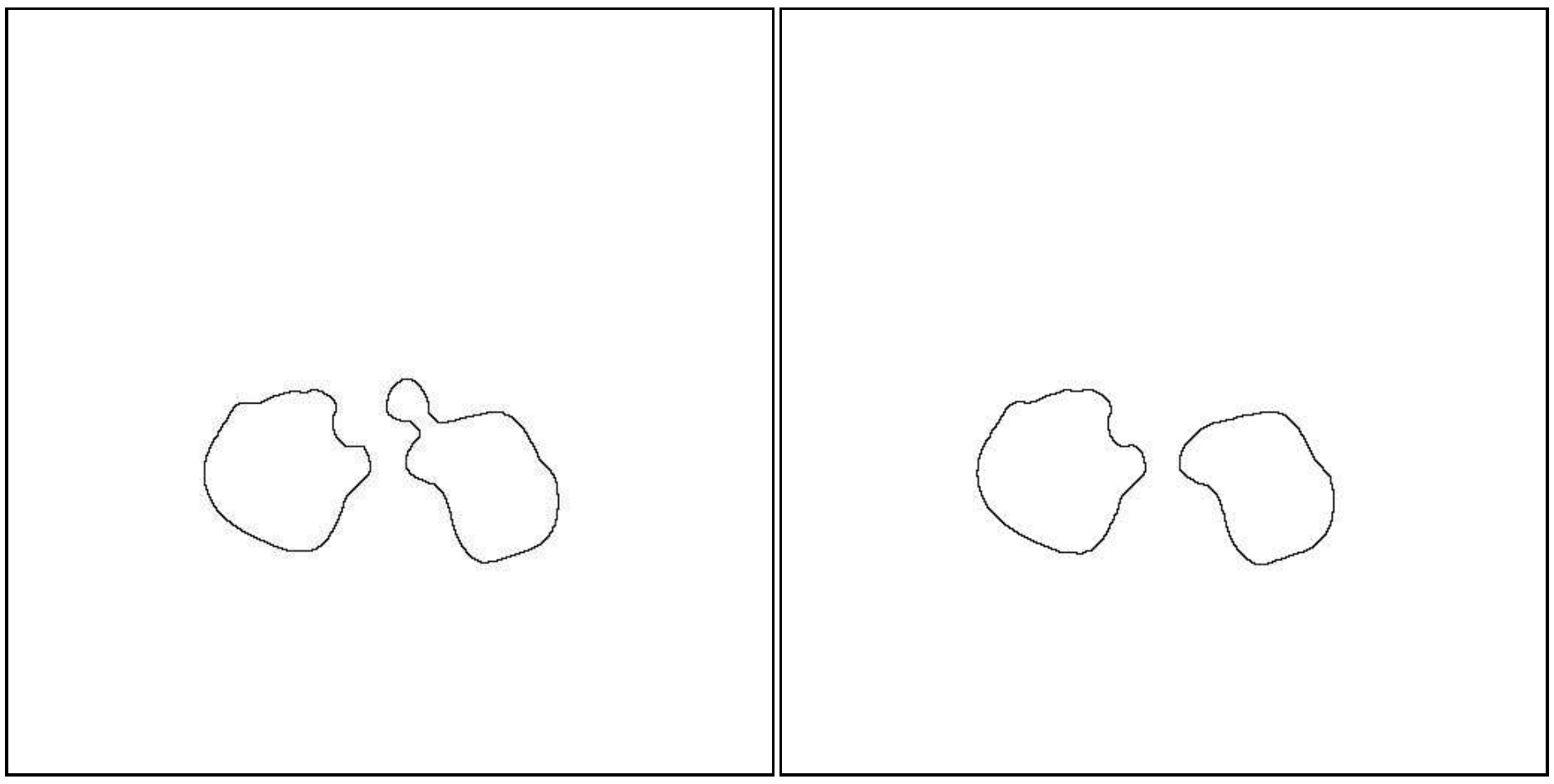

**Fig. 7.** After Correction of case shown in Figure 6 (Left to Right): Po-BE remains unchanged. ProcessRDR-BE's process is corrected.

ProcessRDR uses RDR's knowledge acquisition technique to learn control knowledge for computer vision procedures. The same technique can be applied to other forms of processing systems, that often require expert customization for a specific application.

We are currently applying ProcessRDR to more complex computer vision procedures to validate the flexibility of the ProcessRDR framework. In addition to this, we are also trying to address the issues of dependence between individual ProcessRDR and the significant number of cornerstone evaluations under some worst-case scenarios.

# E-Mail Document Categorization Using BayesTH-MCRDR Algorithm: Empirical Analysis and Comparison with Other Document Classification Methods

Woo-Chul Cho, Debbie Richards

Department of Computing
Macquarie University, Sydney, NSW 2109, Australia
{wccho, richards}@ics.mq.edu.au

**Abstract.** The research suggests the BayesTH-MCRDR algorithm for effective classification of E-mail documents. This is a compound algorithm which combines a naïve Bayesian algorithm using Threshold and the MCRDR (Multiple Classification Ripple Down Rules) algorithm. The significant feature of document classification using the BayesTH-MCRDR algorithm is the achievement of higher precision by first establishing a knowledge base of optimally related words generated from the document training set before going on to classify the set of test documents. Further, we demonstrate the system we have developed in order to compare a number of classification techniques.

## 1 Introduction

The amount of E-mail in usage is increasing by geometric progression with network growth, and e-mail is the favourite program of Internet users. With ongoing development of the Internet, e-mail, the representative communication instrument, costs little, and enables users to exchange information in real time so that many people choose to use it as a communication tool. At the moment private users and companies use it for marketing. This results in the problem of memory shortages for Internet service providers, and requires users to continually spend time removing numerous emails which they do not want to get, and to classify those documents that they are interested in [1][2].

Existingresearch for automatic document classification by machine learning uses a range of techniques such as probability [3][4], statistical methods [5][6], vector similarity [4] and so on. Among these techniques, Bayesian document classification is the method achieving the most promising results for document classification in every language area [7]. However, the naïve Bayes classifier [8] fails to identify salient document features because it extracts every word in the document as a feature. Further, it calculates a presumed value for every word and carries out classification on the basis of it. The naïve Bayes classifier produces many noisy (stop-word) and ambiguous results, thus affecting classification. This misclassification lowers the precision. So in order to increase precision TFIDF (Term Frequency Inverse

Document Frequency) is suggested which uses the Bayesian classification method [9][10]. This produces less misclassification than the naïve Bayes classifier, but does not reflect the semantic relationships between words and fails to resolve word ambiguity. Therefore it cannot resolve misclassification of documents. In order to solve this problem, we have developed an e-mail system which combines both the Bayesian Dynamic Threshold algorithm and the MCRDR algorithm, to produce what we refer to as the BayesTH-MCRDR algorithm. This system applies both the Bayesian algorithm using Dynamic Threshold in order to increase precision and the MCRDR algorithm in order to optimise and construct a knowledge base of related words.

In short, our system first extracts word features from e-mail documents by using Information Gain [11]. Then the documents are classified temporarily by the Bayesian Algorithm, optimised by the MCRDR algorithm and then finally classified. In order to evaluate this system, we compare our approach to E-mail classification with the naïve Bayesian, TFIDF and Bayesian-Threshold algorithms.

## 2 Algorithms for E-Mail Document Classification

In this section we briefly introduce the key concepts underlying the BayesTH-MCRDR algorithm: Naïve Bayesian, Naïve Bayesian with Threshold, Term Frequency Inverse Document Frequency and Multiple Classification Ripple Down Rules. The final subsection describes how we have combined the two techniques.

### 2.1 Naïve Bayesian

Naïve Bayesian classification [12][13] uses probability based on Bayes Theorem. This system inputs a vector model of words ($w_1, w_2, \ldots\ldots w_n$) for the given document ( $d$ ), and classifies the highest probability ( $p$ ) as the class ( $c$ ) among documents that can observe the given document. That is, as shown in formula (1) the system classifies it as a highest conditional probability class.

$$\begin{aligned}\arg\max_{c\in C} P(c \mid d) &= \arg\max_{c\in C} P(c \mid w_1, w_2, \ldots., w_n \mid c) \\ &= \arg\max_{c\in C} \frac{P(w_1, w_2, \ldots., w_n \mid c) p(c)}{P(w_1, w_2, \ldots., w_n)} \\ &= \arg\max_{c\in C} P(w_1, w_2, \ldots., w_n \mid c) p(c)\end{aligned} \quad \textbf{(1)}$$

If we are concerned with only the highest probability class, we can omit Probability ( $P$ ), because it is a constant and normalizing term. Also, this approach applies the naïve Bayesian assumption of conditional independence on each '$w_t$' which is a feature belonging to a same document (see Formula (2)) [12].

$$P(w_1, w_2, ...., w_n \mid c) = \prod_{t=1,n} p(w_t \mid c) \quad \textbf{(2)}$$

So, the naïve Bayesian Classification method decides the highest probability class according to formula (3).

$$\arg\max_{c \in C} P(c) \prod_{t=1,n} p(w_t \mid c) \quad \textbf{(3)}$$

### 2.2 Naïve Bayesian with Threshold

In the definition in section 2.1, the Threshold value of Naïve Bayesian algorithm is fixed. It results in lower precision when Naïve Bayesian algorithm classifies documents with low conditional probability. The Naïve Bayesian Threshold algorithm is able to increase the precision of document classification by dynamically calculating the value of the threshold as given in formula (4).

$$\begin{aligned}
&\textit{Category (Class) Set } C = \{c_0, c_1, c_2, c_{3,....,} c_n\} \quad , \quad C_0 = \textit{unknown class} \\
&\textit{Document Set } D = \{d_0, d_1, d_2, d_{3,....,} d_i\} \\
&\Re(d_i) = \{P(d_i \mid c_1), P(d_i \mid c_2), P(d_i \mid c_3), ...., P(d_i \mid c_n)\} \\
&P_{\max}(d_i) = \max \{P(d_i \mid C_t)\} \quad , \quad t = 1, ...., n \\
&C_{best}(d_i) = \begin{bmatrix} \{c_j \mid P(d_i \mid c_j) = P_{\max}(d_i), \ \textit{if } P_{\max}(d_i) \geq T\} \\ \textit{where} T = 1 - \dfrac{P_{\max}(d_i)}{\sum_{t=1}^{n} P(d_i \mid c_t)} \\ c_0 \quad , \quad \textit{otherwise} \end{bmatrix}
\end{aligned} \quad \textbf{(4)}$$

### 2.3 TFIDF (Term Frequency Inverse Document Frequency)

TFIDF [5], traditionally used in information retrieval, expresses a weight vector based on word frequency of the given document 'd'. In this case, each word weight ( $W$ ) is calculated by multiplying the Term Frequency ( $TF$ ) in a given document 'd' and its reciprocal number, Inverse Document Frequency ( $IDF$ ), of all documents having the word feature. This means that the higher the IDF, the higher the feature (see Formula (5)). That is, if there is a word which has a higher frequency in a certain document, and a lower frequency in other documents, then the word can express the document very well.

$$W_i = TF_i \bullet IDF_i \quad \textbf{(5)}$$

For document classification we require a prototype vector expressing each class. The prototype vector ( $c$ ) of each class is calculated as the average of the weight vector of its training document. Only if each class is expressed in a prototype vector, the similarity is calculated by applying the cosine rule between the weight vector of a given document 'd' and each class prototype vector as shown in formula (6).

$$\arg\max_{c\in C} \cos(c,d) = \arg\max_{c\in C} \frac{c}{\| c \|} \bullet \frac{d}{\| d \|} \qquad (6)$$

### 2.4 MCRDR (Multiple Classification Ripple Down Rule

Kang [14] developed Multiple Classification Ripple Down Rules (MCRDR). MCRDR overcomes a major limitation in Ripple Down Rules (RDR), which only permitted single classification of a set of data. That is MCRDR allows multiple independent classifications. An MCRDR knowledge base is represented by an N-ary tree [14]. The tree consists of a set of production rules in the form "If Condition Then Conclusion".

#### 2.4.1 Creation of Rule

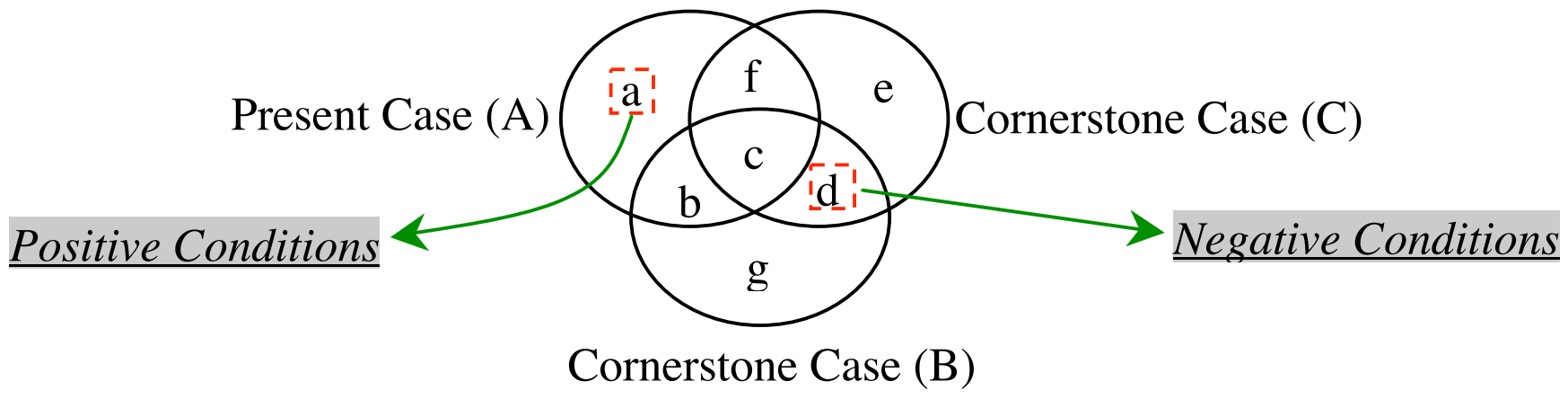

**Fig. 1.** Difference list {a, not d} are found to distinguish the Present Case (A) from two Cornerstone Cases (B) and (C) [14]

We consider a new *case (present case) A* and two *cornerstone cases B* and *cornerstone cases C*. The cornerstone case is the case that prompted the rule being modified (that is, the rule that currently fires on the present case but which is deemed to be incorrect) to be originally added. The present case will become the cornerstone case for the new (exception) rule. To generate conditions for the new rule, the system has to look up the cornerstone cases in the parent rule. When a case is misclassified, the rule giving the wrong conclusion must be modified. The system will add an exception rule at this location and use the cornerstone cases in the parent rule to determine what is different between the previously seen cases and the present case. These differences will form the rule condition and may include positive and negative conditions (see Formula (7)).

Positive Condition : (7)

Present Case (A) - (Cornerstone Case (B) $\cup$ Cornerstone Case (C))

Negative Condition :

(Cornerstone Case (B) $\cap$ Cornerstone Case (C)) – Present Case (A)

Figure 1 shows a difference list {a, NOT d} between the present case and two cornerstone cases. After the system adds a new rule with the selected conditions by the expert or system, the new rule should be evaluated with the remaining cornerstone cases in the parent rule [14]. If any remaining cornerstone cases are satisfied with the newly added rule, then the cases become cornerstone cases of the new rule [14].

### 2.4.2 Inference

The inference process of MCRDR is to allow for multiple independent conclusions with the validation and verification of multiple paths [14]. This can be achieved by validating the children of all rules which evaluate to true. An example of the MCRDR inference process is illustrated in Figure 2.

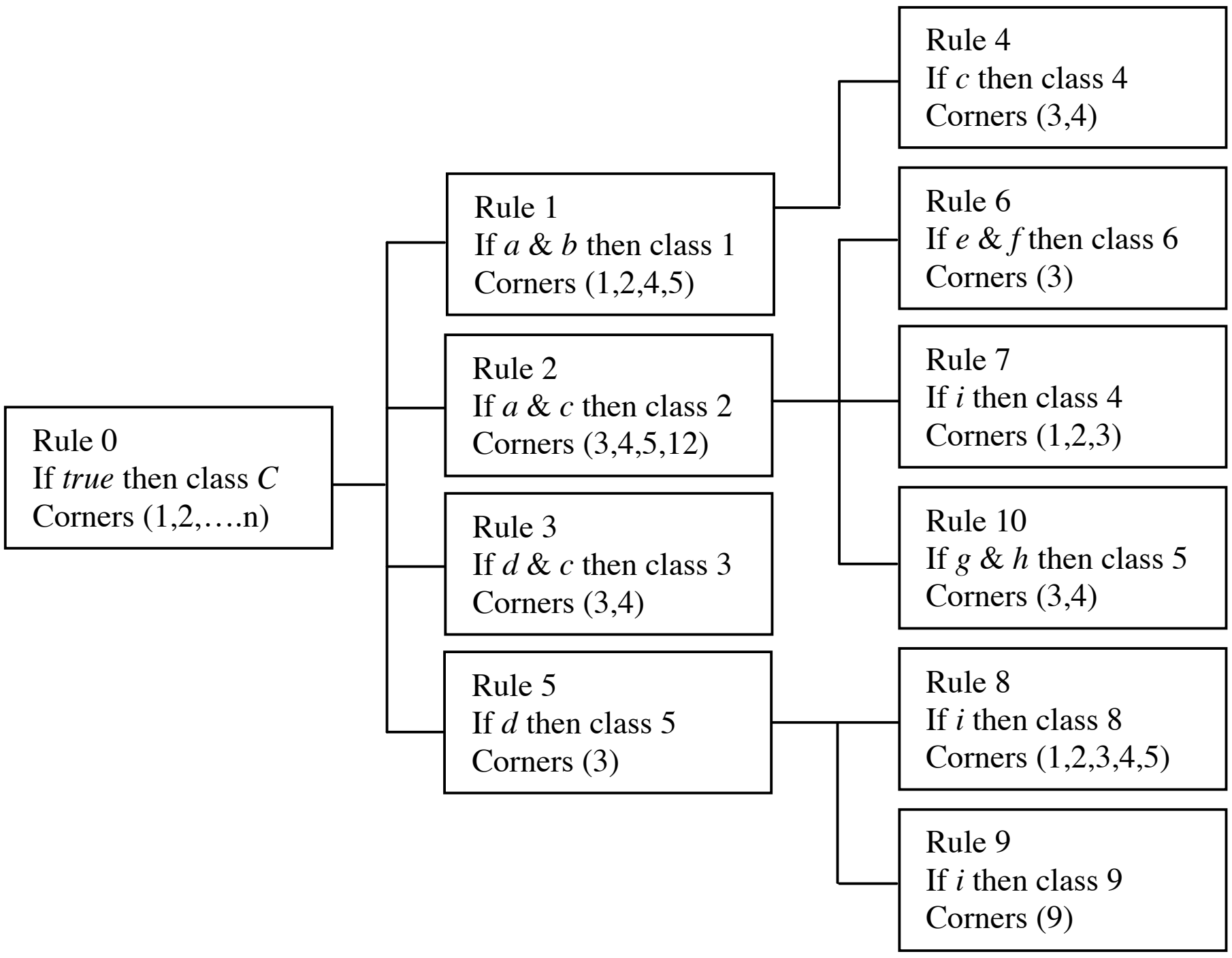

**Fig. 2.** Knowledge Base and Inference in MCRDR, Attributes: {a, c, d, e, f, h, k} [14]

In this example, a case has attributes {a, c, d, e, f, h, k} and three classifications (conclusion 3, 5 and 6) are produced by the inference. Rule 1 does not fire. Rule 2 is validated as true as both “a” and “c” are found in our case, Now we should consider the children (rules 6, 7, and 10) of rule 2. From comparison of the conditions in children rules with our case attributes, only rule 6 is evaluated as true. Hence, rule 6 would fire to get a *conclusion 6* which is our case classification. This process is applied to the complete MCRDR rule structure in Figure 2. As a result, rule 3 and 5 can also fire, so that *conclusion 3* and *conclusion 5* are also our case classifications.

## 2.5 BayesTH-MCRDR

The BayesTH-MCRDR algorithm combines the merits of both the Naïve Bayesian using Threshold (BayesTH) and MCRDR algorithms. As shown in figure 3, a new document can be extracted from feature keywords which are obtained through the Information Gain method (see Section 3.1.2). And then, the document is classified by the BayesTH algorithm into a temporary knowledge base (Table 1.1). At this moment a document is classified, that is assigned a class. The MCRDR algorithm creates new rules based on the feature keywords in the document. In the BayesTH algorithm, the feature keywords are independent of one another. The MCRDR rules represent the semantic relationships between feature keywords. In BayesTH-MCRDR, rules stand for a condition for a case to be classified, class stands for a conclusion of a case.

**Table 1.1.** Table of Temporary Knowledge Base by BayesTH algorithm

| Category | Class | Document No | Keyword |
|---|---|---|---|
| Database | mySQL | 1 | A, B |
| | pgSQL | 2 | X, Y, Z |
| | …. | …. | …. |

**Table 1.2.** Table of Knowledge Base by MCRDR algorithm

| Step | Document | Algorithm | Rules (Keywords) | Class |
|---|---|---|---|---|
| 1 | 1 | Bayesian Threshold | A, B | MySQL |
| 2 | 1 | MCRDR | A&C | MySQL |
| 3 | 1 | BayesTH-MCRDR | A, B, A&C, A&B, A&B&C | MySQL |

For example, the learning process for document 1 using the BayesTH-MCRDR algorithm into MySQL class is as follow:

Step 1: Document 1 creates rule A and rule B through BayesTH algorithm. In the BayesTH algorithm, the feature keywords are independent of one another and its created rules. That is, “If Rule A then Class MySQL” or “If Rule B then Class MySQL”. Step 2: Document 1 creates rule A&C according to the creation rule process of MCRDR algorithm described above (see section 2.4.1). Step 3: Document 1 creates new rules by combining rules from Step 1 and Step 2. And then, the created rules get a Rule ID and document 1 is classified into MySQL according to the inferencing process described above (see section 2.4.2).

## 3 E-Mail Classification System

We now introduce the system and accompanying process that have been developed. Section 3.1 describes the preprocessing performed on the documents (email messages). Section 3.2 describes the implemented system.

### 3.1 Data Pre-Processing

Data preparation is a key step in the data mining process. For us this step involved deletion of stopwords, feature extraction and modeling and document classification. We describe how these were achieved next.

#### 3.1.1 Deletion of Stopwords

The meaning of 'Stopwords' refers to common words like 'a', 'the', 'an', to', which have high frequency but no value as an index word. These words show high frequencies in all documents. If we can remove these words at the start of indexation, we can obtain higher speeds of calculation and fewer words needing to be indexed. The common method to remove these 'Stopwords' is to make a 'Stopwords' dictionary in the beginning of indexation and to get rid of those words. This system follows that technique.

#### 3.1.2 Feature Extraction and Document Modelling

The process of feature extraction is that of determining which keywords will be useful for expressing each document for classification learning. Document modelling is the process of expressing the existence or non-existence, frequency and weight of each document feature based on a fixed feature word [15]. Feature extraction and document modelling are the most important factors affecting document classification efficiency when applying a classification-learning method. We note that there has been a lot of research into both feature extraction and document modelling due to their suitability for Information Retrieval, Information Filtering and Fusion.

The most basic method to choose word features which describe a document is to use a complete vocabulary set which is based on all words in the document sets. But this requires extensive computation due to a greater number of word features than the number of given documents, and the inclusion of a number of word features which do not assist classification but instead reduce classification power. Some words offer semantics which can assist classification. Selecting these words as word features from the complete word set for the set of documents will reduce effort. In this way we consider Feature Extraction to be Feature Selection or Dimension Deduction. There are various ways to achieve feature selection, but our system uses the well-known Information Gain approach [11] that selects words that have a large entropy difference as word features based on information theory.

$$V = \{w_1, w_2, w_3, w_4, w_5, \ldots, w_n\} \quad \textbf{(8)}$$

$$InforGain(w_k) = P(w_k)\sum_i P(c_i \mid w_k)\log\frac{P(c_i \mid w_k)}{P(c_i)} \quad \textbf{(9)}$$

$$+ P(\overline{w_k})\sum_i P(c_i \mid w_k)\log\frac{P(c_i \mid \overline{w_k})}{P(c_i)}$$

When the complete set of vocabulary ( $V$ ) consists of rules (formula (8)) and n words, formula (9) shows the calculation of the information gain for each word $w_k$. Those words which have the largest information gain are included in the optimized set of word features ( $K$ ) as in formula (10).

$$K = \{w_1, w_2, w_3, w_4, w_5, ...., w_L\}, K \subset V \quad \textbf{(10)}$$

### 3.1.3 Learning and Classification

In order to do supervised learning and evaluate the accuracy of e-mail document classification based on BayesTH-MCRDR we must provide classified documents as input. Our system uses the naïve Bayesian learning method as it is a representative algorithm for supervised learning. The Naïve Bayesian classification learning method classifies each e-mail document with the highest probability class. Where the conditional probability of a given document is low or there is a conflict the system asks the user to choose the most appropriate classification. In situations where either the difference between the two or more highest conditional probabilities is small or the highest conditional probability is low (for example, the highest conditional probability is 0.2 ~ 0.3 and less) we ask the user to intervene. Since precision and trust are closely related, we don't want the system to give an incorrect classification, resulting in the users loss of faith in the system. Hence, when the system can not clearly assign a class, the system assigns the document to 'Others' for the user to deal with (see Formula (4)). In our system the user is able to set the probability threshold 'T' (see Figure 1), above which the system will assign its own conclusion.

### 3.2 Implementation

The screen dump in Figure 3 displays the key elements of our system, which has been developed to evaluate the performance of the implemented algorithms. The screen consists of three parts; the top panel is for choosing which classification rule to apply to the set of e-mail documents, the second panel allows selection of the class (mySQL, pgSQL, PHP and so on) of the data and whether training (learning) or testing (experiment) data is to be used. The third section on the screen (large lower panel) is used to display the contents of the data for the purposes of evaluating and confirming that the data has been classified into the correct class.

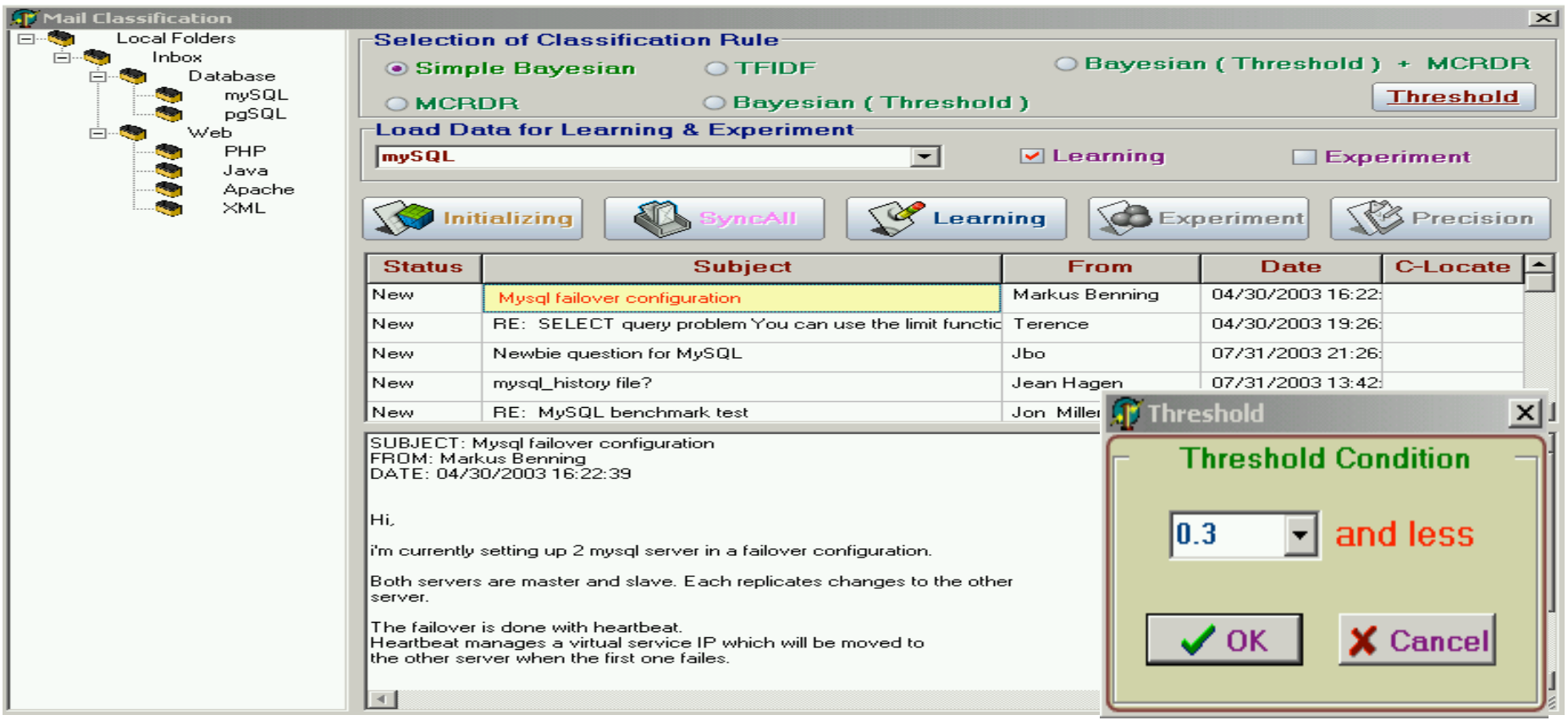

**Fig. 3.** E-Mail Classification System and Control of Threshold value

# 4 Experiment

## 4.1 Aims

A key goal of any classification system is to avoid misclassification. Therefore to validate the precision of the BayesTH-MCRDR algorithm for e-mail classification, we carried out some experiments. And through the experiments, we compared the classification precision across four different learning methods.

## 4.2 Data Collection and Setup

We used a commercial FAQ (Frequently Asked Questions) E-mail archive as our experimental data in order to ensure fairness. This E-mail archive is available at the website called "Geocrawler.com[2]" and is owned by Open Source Development Network, Inc. We selected two categories, database and web, in order to evaluate the capability of our system. The 'Database' category has two subcategories, 'mySQL' and 'pgSQL', and the 'Web' category has four subcategories, 'PHP', 'Java', 'Apache' and "XML' (see Figure 4). We conducted five experiments for each of the six classes. We gave input learning data 100, 200, 300, 400, 500 into each class (total of 1,500 per class). For evaluating precision we used test sets of 500 experimental data at each experiment. The total number of Learning data and Experiment data was 9,000 and 3,000 each.

| **Category** | **Class** |
|---|---|
| Database | MySQL |
| | PgSQL |
| Web | PHP |

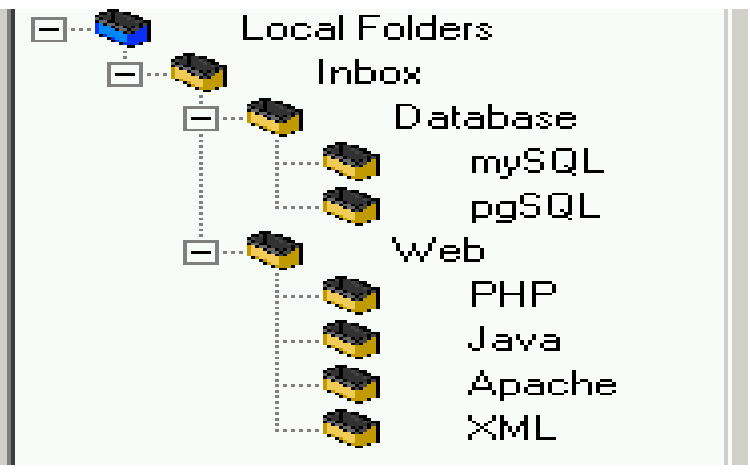

2 http://www.geocrawler.com/ (viewed 20/4/2004)

| | |
|---|---|
| | Java |
| | Apache |
| | XML |

**Fig. 4.** Experimental Category and Class

**Table 2.** Data Set for Experiment

| Algorithm Name | Class | Learning Data | Experiment Data | Correct Data | Precision |
|---|---|---|---|---|---|
| | mySQL | 100,200,300,400,500 | 500 | | % |
| | pgSQL | 100,200,300,400,500 | 500 | | % |
| | PHP | 100,200,300,400,500 | 500 | | % |
| | Java | 100,200,300,400,500 | 500 | | % |
| | Apache | 100,200,300,400,500 | 500 | | % |
| | XML | 100,200,300,400,500 | 500 | | % |
| Total | 6 Class | 9000 | 3000 | | |

### 4.3 Results

Figure 5(a) shows the formatting of E-mail text data provided by the system. To assist evaluation of the precision of each algorithm the user is provided with the Precision check function as shown in Figure 5(b).

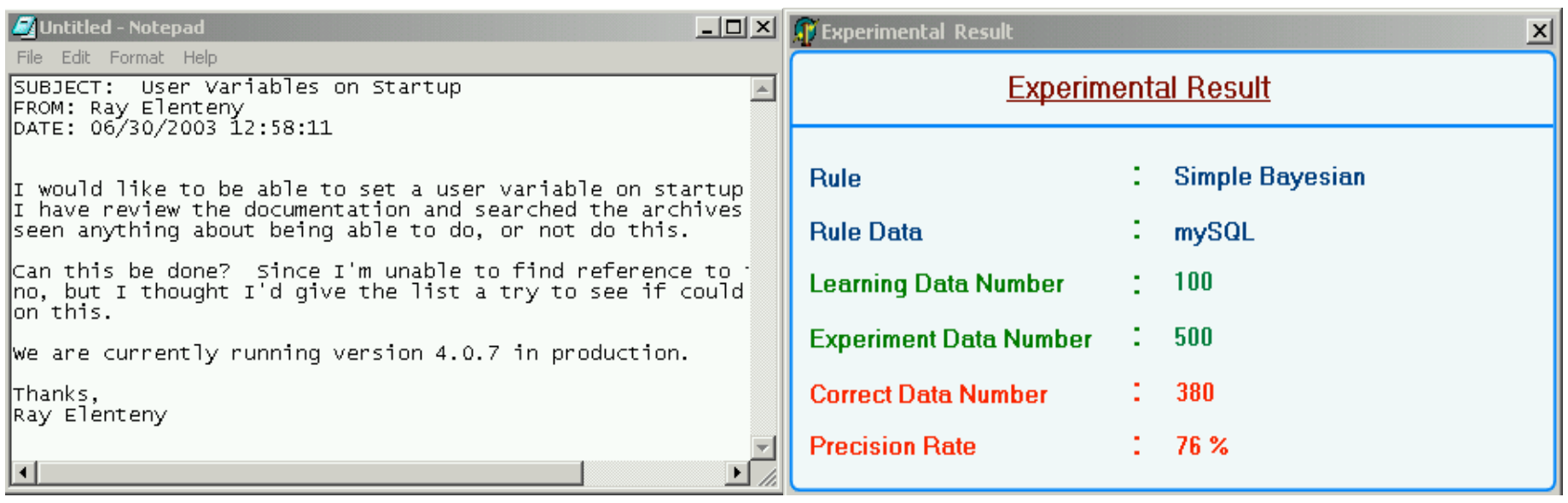

(a) (b)

**Fig. 5.** E-Mail Document Data Format

Figures 6-10 provide the precision results for each of the five algorithms: simple naive Bayesian, TFIDF, Bayesian Threshold, MCRDR and BayesTH-MCRDR, respectively. Averages for all algorithms are given in Figure 11.

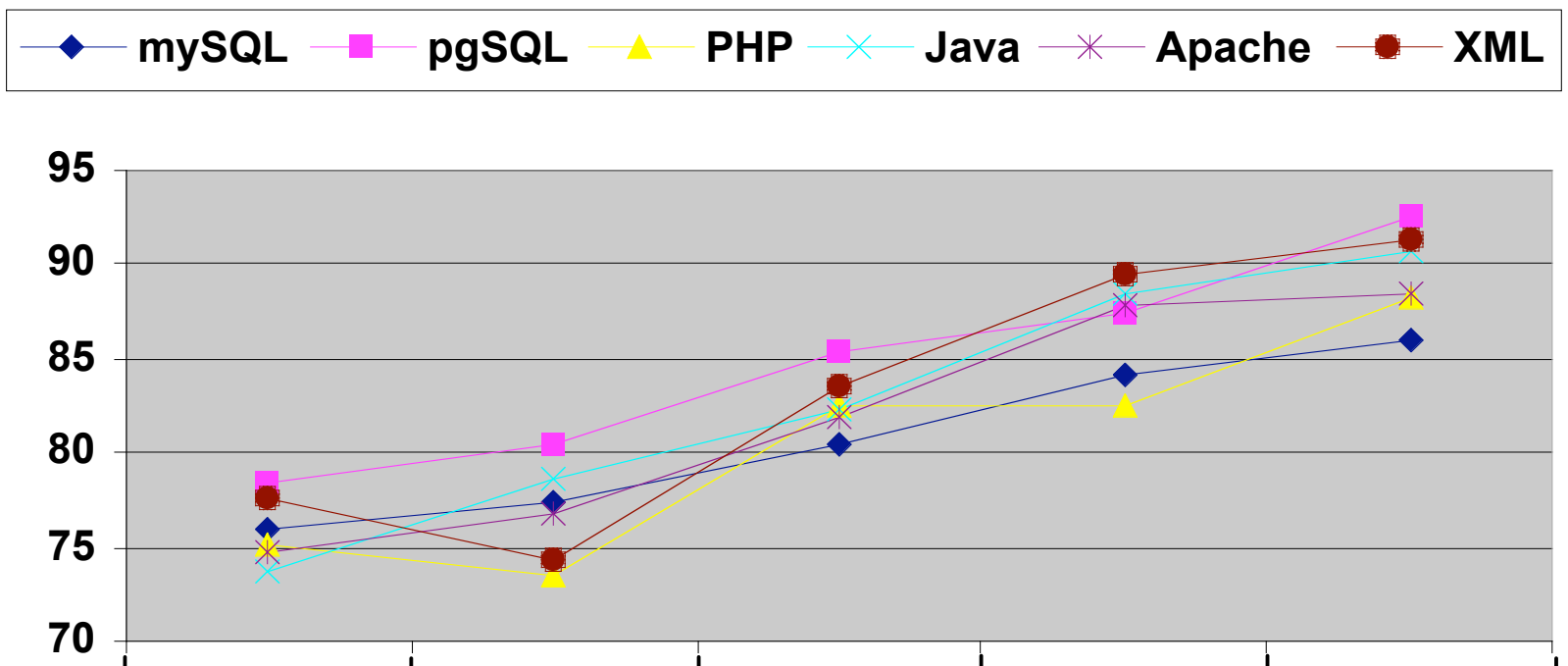

| | 100 | 200 | 300 | 400 | 500 |
|---|---|---|---|---|---|
| **mySQL** | 76 | 77.4 | 80.4 | 84.2 | 86 |
| **pgSQL** | 78.4 | 80.4 | 85.4 | 87.4 | 92.6 |
| **PHP** | 75.2 | 73.4 | 82.6 | 82.6 | 88.2 |
| **Java** | 73.6 | 78.6 | 82.2 | 88.4 | 90.6 |
| **Apache** | 74.8 | 76.7 | 81.8 | 87.9 | 88.4 |
| **XML** | 77.6 | 74.4 | 83.6 | 89.4 | 91.4 |

**Fig. 6.** Results of Experiment using simple naive Bayesian Algorithm

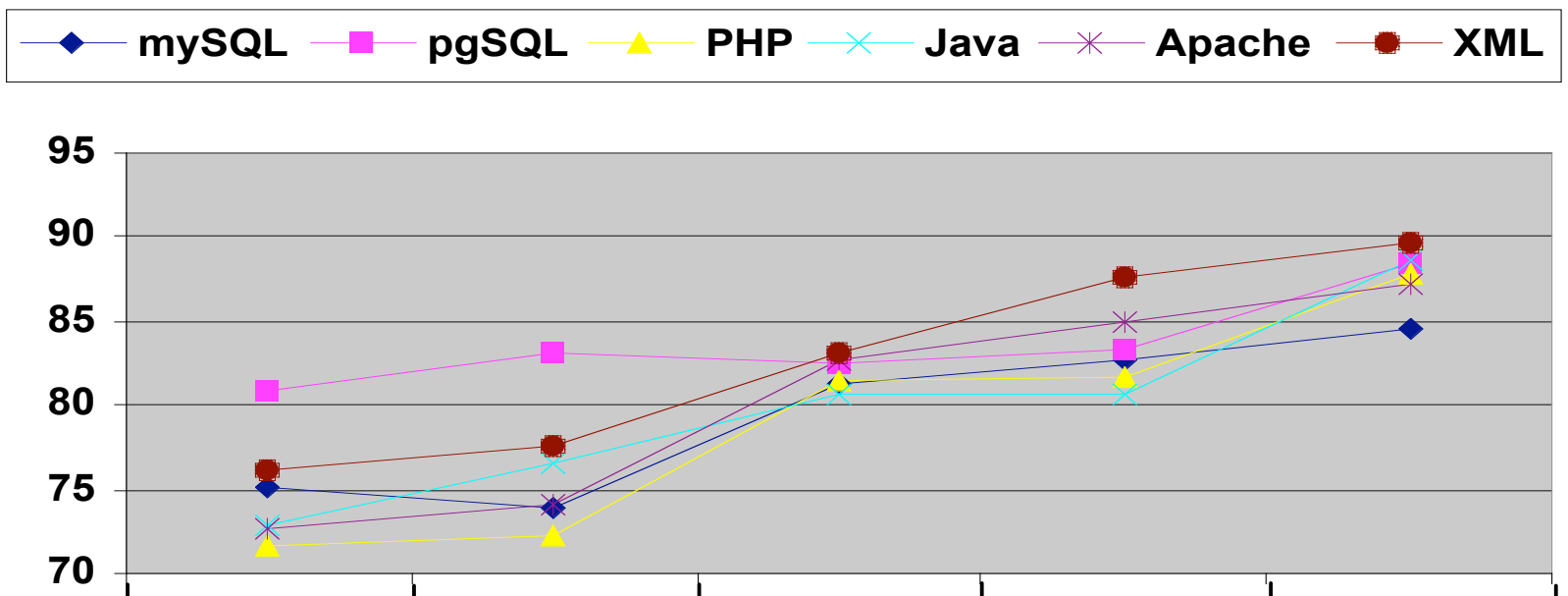

| | 100 | 200 | 300 | 400 | 500 |
|---|---|---|---|---|---|
| **mySQL** | 75.2 | 73.8 | 81.2 | 82.8 | 84.6 |
| **pgSQL** | 80.8 | 83.2 | 82.4 | 83.4 | 88.4 |
| **PHP** | 71.6 | 72.2 | 81.4 | 81.6 | 87.8 |
| **Java** | 72.8 | 76.6 | 80.6 | 80.6 | 88.6 |
| **Apache** | 72.6 | 74.2 | 82.8 | 85 | 87.2 |
| **XML** | 76.2 | 77.6 | 83.2 | 87.6 | 89.6 |

**Fig. 7.** Results of Experiment using TFIDF Algorithm

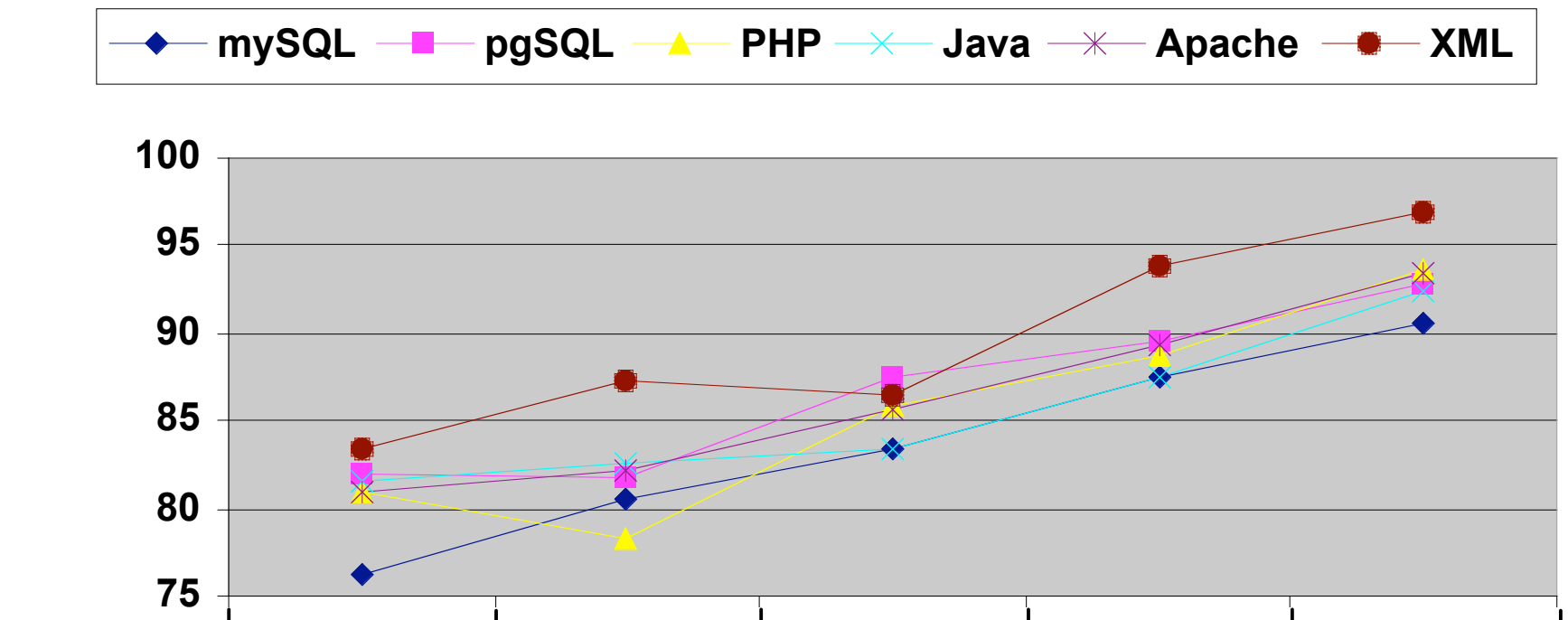

| | 100 | 200 | 300 | 400 | 500 |
|---|---|---|---|---|---|
| **mySQL** | 76.2 | 80.6 | 83.4 | 87.4 | 90.6 |
| **pgSQL** | 82 | 81.8 | 87.6 | 89.6 | 92.8 |
| **PHP** | 81 | 78.2 | 85.8 | 88.8 | 93.6 |
| **Java** | 81.6 | 82.6 | 83.4 | 87.6 | 92.4 |
| **Apache** | 81 | 82.2 | 85.6 | 89.4 | 93.4 |
| **XML** | 83.4 | 87.2 | 86.4 | 93.8 | 97 |

**Fig. 8.** Results of Experiment using naive Bayesian Threshold Algorithm

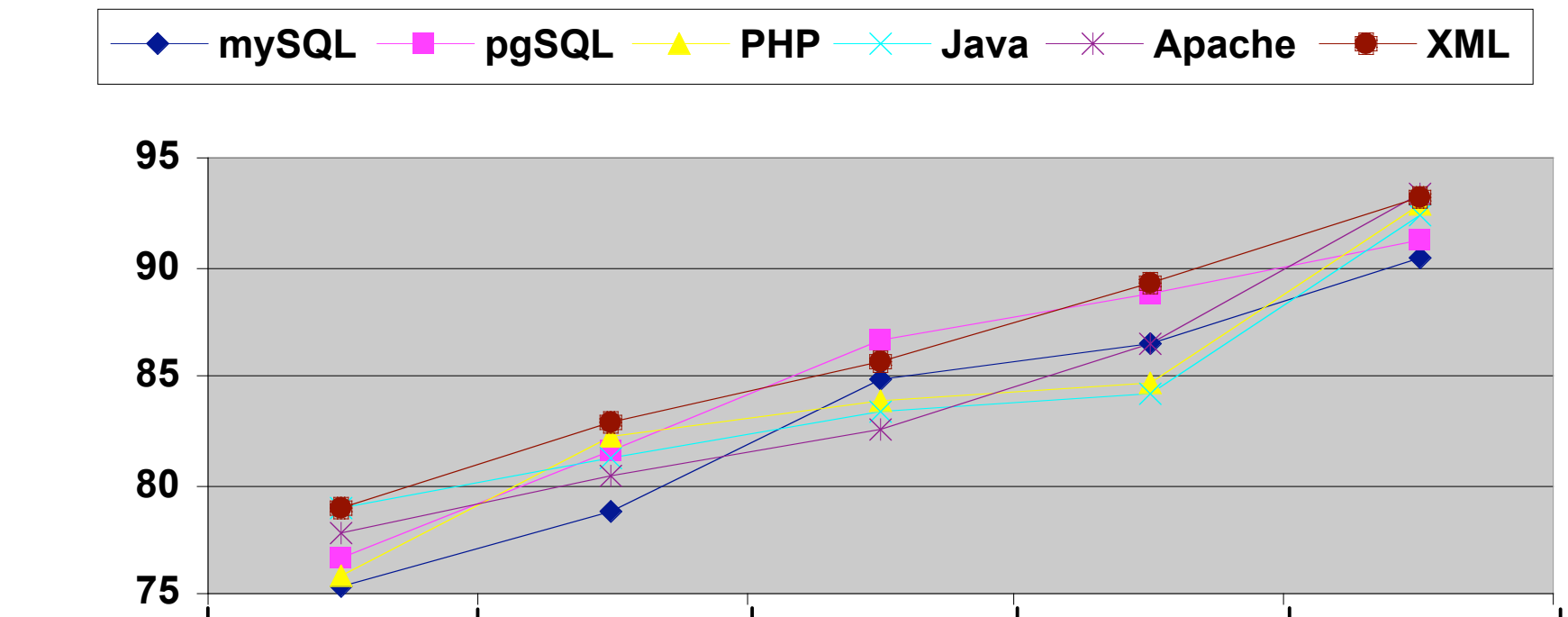

| | 100 | 200 | 300 | 400 | 500 |
|---|---|---|---|---|---|
| **mySQL** | 75.4 | 78.8 | 84.8 | 86.4 | 90.4 |
| **pgSQL** | 76.6 | 81.6 | 86.6 | 88.8 | 91.2 |
| **PHP** | 75.8 | 82.2 | 83.8 | 84.6 | 92.8 |
| **Java** | 79 | 81.2 | 83.4 | 84.2 | 92.4 |
| **Apache** | 77.8 | 80.4 | 82.6 | 86.4 | 93.4 |
| **XML** | 79 | 82.8 | 85.6 | 89.2 | 93.2 |

**Fig. 9.** Results of Experiment using MCRDR Algorithm

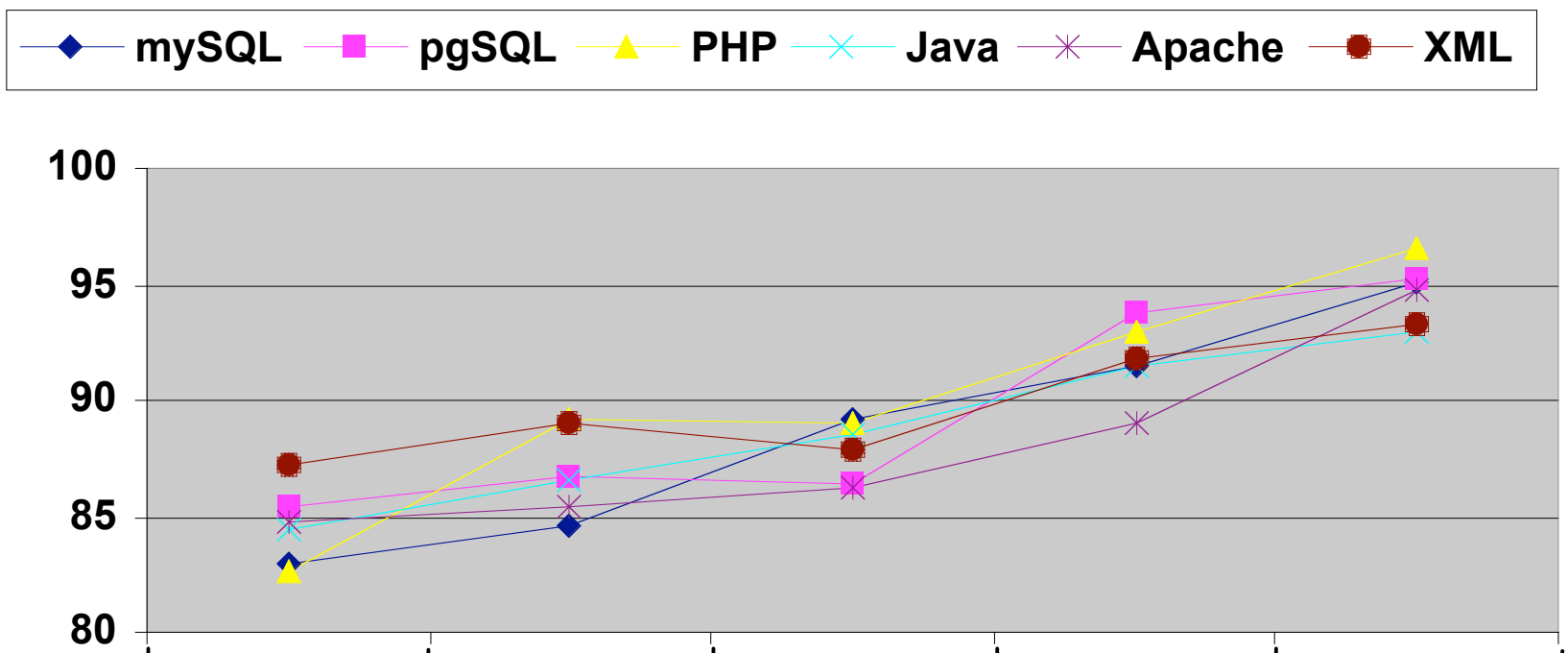

| | 100 | 200 | 300 | 400 | 500 |
|---|---|---|---|---|---|
| **mySQL** | 83 | 84.6 | 89.2 | 91.4 | 95.1 |
| **pgSQL** | 85.4 | 86.8 | 86.4 | 93.8 | 95.2 |
| **PHP** | 82.6 | 89.2 | 89 | 93 | 96.6 |
| **Java** | 84.4 | 86.6 | 88.6 | 91.4 | 93 |
| **Apache** | 84.8 | 85.4 | 86.2 | 89 | 94.2 |
| **XML** | 87.2 | 89 | 87.8 | 91.8 | 94.7 |

**Fig. 10.** Results of Experiment using BayesTH-MCRDR Algorithm

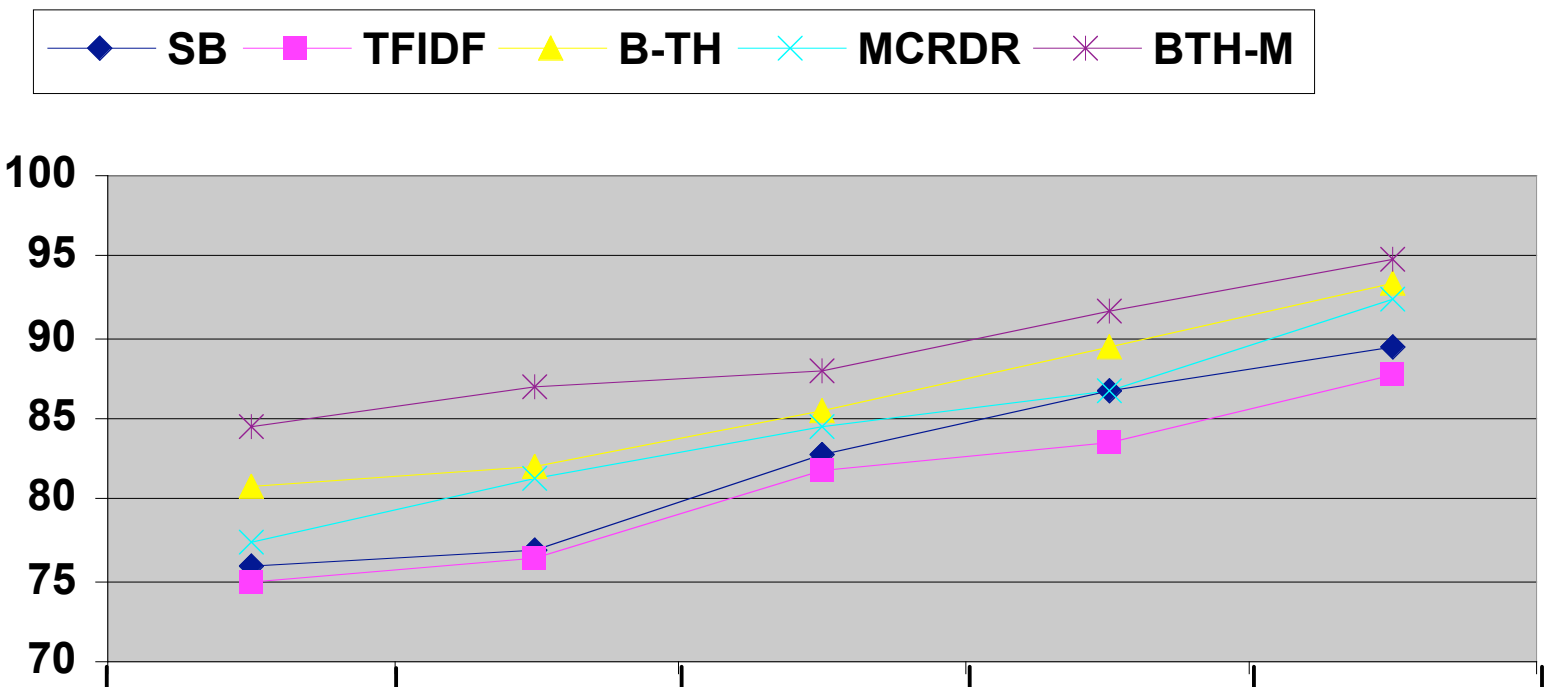

| | 100 | 200 | 300 | 400 | 500 |
|---|---|---|---|---|---|
| **SB** | 75.9 | 76.8 | 82.7 | 86.7 | 89.5 |
| **TFIDF** | 74.9 | 76.3 | 81.9 | 83.5 | 87.7 |
| **B-TH** | 80.9 | 82.1 | 85.4 | 89.4 | 93.3 |
| **MCRDR** | 77.3 | 81.2 | 84.5 | 86.6 | 92.3 |
| **BTH-M** | 84.6 | 86.9 | 87.9 | 91.7 | 94.8 |

(a)

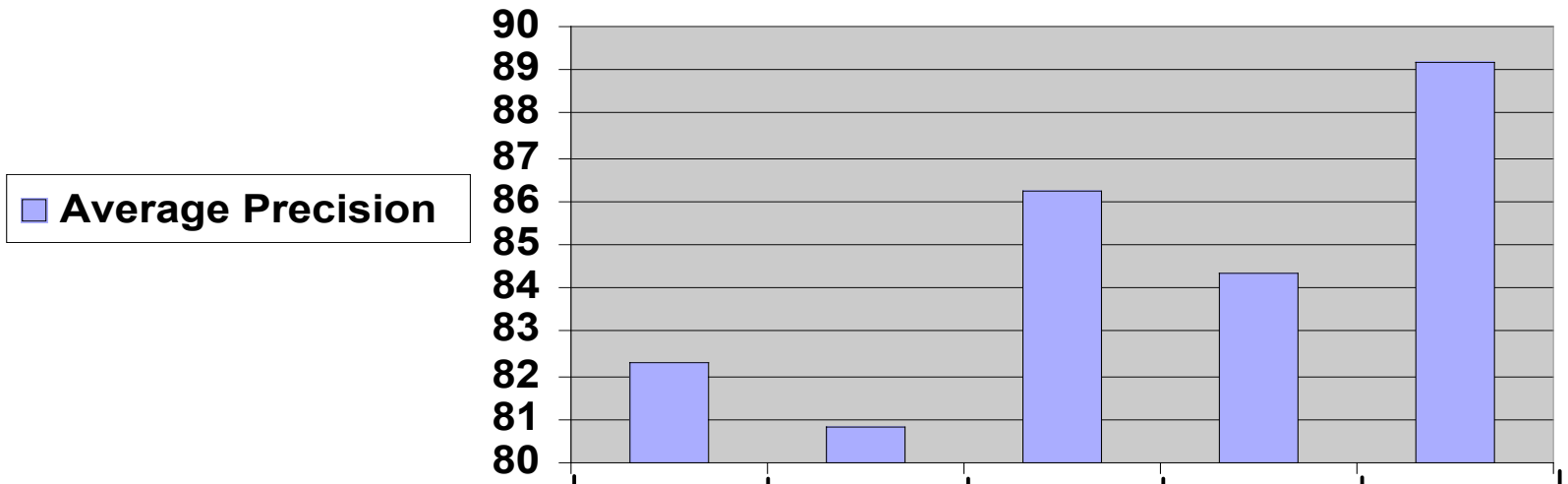

| | SB | TFIDF | B-TH | MCRDR | BTH-M |
|---|---|---|---|---|---|
| **Average Precision** | 82.32 | 80.86 | 86.22 | 84.38 | 89.18 |

(b)

**Fig. 11.** (a) Results of Average Precision for each experiment, SB: Simple Bayesian; B-TH: Bayesian Threshold; BTH-M: Bayesian Threshold and MCRDR (BayesTH-MCRDR).
(b) Results of average precision for each algorithm

The experimental results show high overall precision 80% - 89% for all algorithms even though there are some differences according to the method of classification learning. Specifically, the more documents used in training the higher the classification accuracy, as we expected. Also there are clear differences in classification accuracy among classification learning methods. The system, BayesTH-MCRDR shows the highest precision 89.18%. On the contrary, TFIDF shows the lowest precision 80.86%. And TFIDF, naïve Bayesian, and MCRDR show 80.86%, 82.32%, and 84.38% respectively. We also note, that BayesTH-MCRDR outperforms all the other algorithms for all sizes of training sets and matures more quickly, achieving accuracy levels after 100 cases similar to the accuracy levels achieved by the other algorithms after seeing 300 cases. Looking at the individual results (in Figures 6-10), rather than the average precision (figure 11), we note that the two methods using MCRDR tend to have a smaller spread of results across classes. That is the standard deviation of results across the six classes is smaller (for example MCRDR had a range of 90.4-93.2 for 500 cases and BayesTH-MCRDR had a range of 93-96.9) than for the other techniques. In contrast, the Bayesian Threshold algorithm achieved the highest precision rate of 97 for XML using a training set of 500 cases but only achieved 90.6 accuracy for the mySQL class.

## 5 Conclusions and Future Work

The development of the Internet enables us to exchange many e-mail correspondences but also to receive many messages that we are not interested in and must expend time and energy to filter out. To make matters worse, the filtering process can result in the loss or misplacement of messages that we did need to respond to. To alleviate the amount of human effort involved, we suggest the BayesTH-MCRDR algorithm for effective e-mail classification. As presented in the paper, we have achieved higher precision by using the BayesTH-MCRDR algorithm than existing classification methods like simple Bayesian classification method, TFIDF classification method and simple Bayesian classification method. The specific feature of this algorithm which enables it to achieve higher precision is the construction of a related word knowledge base from the learning documents before applying the learnt knowledge to the classification of the test set of documents. Other research has shown in general that the Bayesian algorithm using a 'Threshold' has better results than the simple Bayesian algorithm. But this paper shows that the BayesTH-MCRDR algorithm has 3% higher precision than the Bayesian Threshold algorithm. If we can construct a related word database through the learning documents, we can get much higher accuracy of document classification.

# Detecting the Knowledge Frontier: An Error Predicting Knowledge Based System

Richard Dazeley and Byeong-Ho Kang

School of Computing, University of Tasmania, Hobart, Tasmania 7001, Australia.[1]
Smart Internet Technology Cooperative Research Centre, Bay 8, Suite 9/G12
Australian Technology Park Eveleigh NSW 1430[1]
{rdazeley, bhkang}@utas.edu.au

**Abstract.** Knowledge Based Systems (KBS) have long wrestled with the problem of incomplete knowledge that occasionally causes them to make ridiculous conclusions. Knowledge engineers have searched for methodologies that allow for less brittle systems. Additionally, KBS systems for general knowledge have been developed to try and build background information that a system can fall back on when they cannot find a conclusion in their specific domain. However, it is next to impossible to include all the required knowledge to completely eradicate the inherent brittleness of these systems. This paper presents a method for predicting when a case being presented to the KBS is outside its current knowledge. When the system notices such a case it provides a warning allowing the user to investigate the case further. This preliminary study of the system has been tested using a simulated expert with randomly generated data sets. It shows that this system has great potential for predicting almost every error and rarely issuing warnings for correct conclusions. Such a system could significantly reduce the knowledge acquisition task for an expert. These results clearly show the potential of such a system and that further investigation with recognised data sets and real user tests should be performed.

## Introduction

Regardless of how intelligent we think we are, there will occasionally be that slip up - that embarrassing moment when we do or say something that reveals an extreme lack of knowledge about one particular detail that *everyone* else knows. This, fairly *rare* human ailment, has long been recognised as being a *frequently* occurring flaw in knowledge based systems (KBS). While, such an error may make the person that made the mistake uncomfortable, it can usually be covered up or laughed at and then simply put aside. However, when a knowledge base makes this error it highlights a major error in a system's knowledge base and causes the system's user to lose faith in the computer's ability to give an accurate and meaningful conclusion. The famous apocryphal example is the expert system that diagnosed a man as being pregnant [1].

---

[1] Collaborative research project between both institutions

This *brittleness*, often associated with KBSs, is founded in the system's inability to realise when its knowledge base is inadequate to provide an accurate and meaningful conclusion. The cause of such inadequacies is generally recognised as being due to the concentration of specialised knowledge in the target domain for the particular system [2]. Within this particular domain they may perform exceptionally well, however, the moment some form of knowledge is needed from just outside of this domain their *competence* drops off quickly to complete *incompetence*. As the introductory paragraph alluded, people are also subject to this difficulty. However, they have the ability to fall back on layer upon layer of general knowledge providing a much less precipitous slope [2].

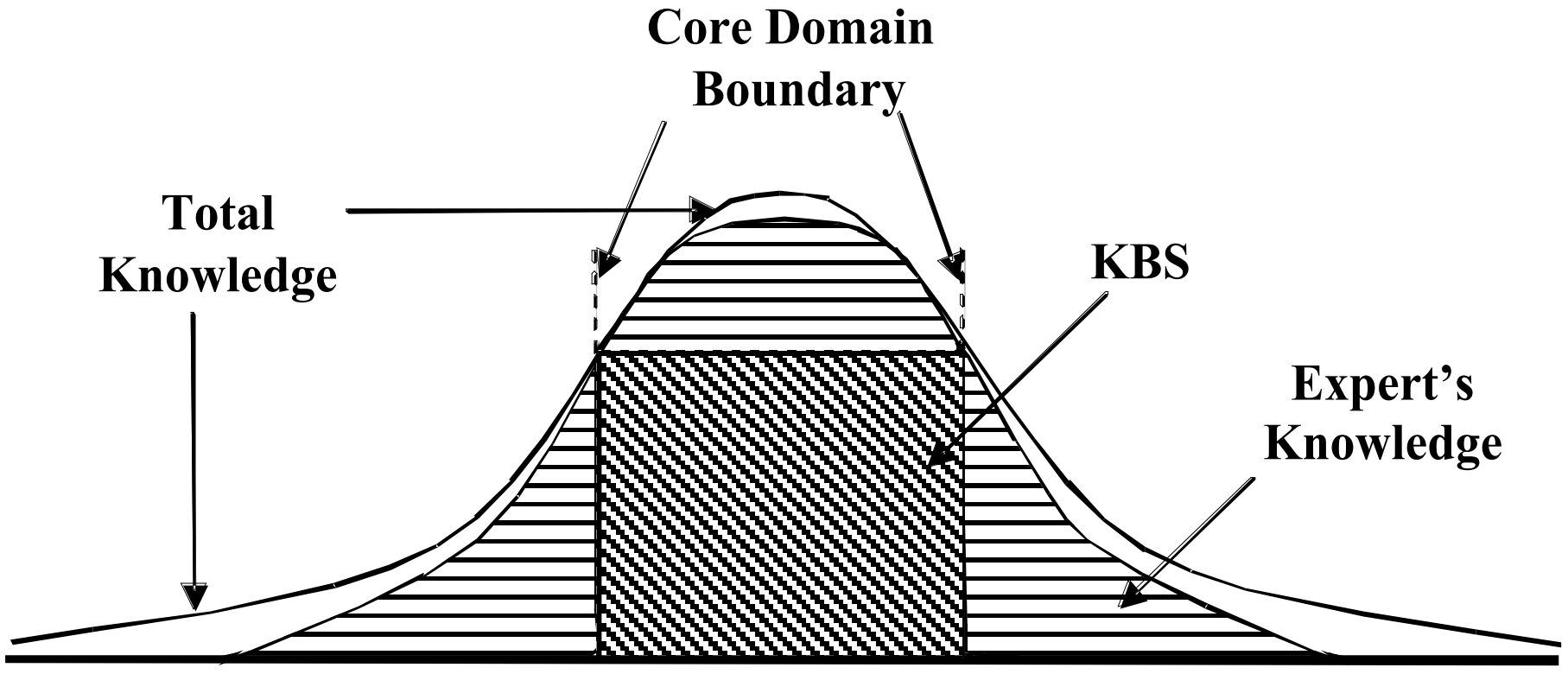

**Figure 1:** Conceptual diagram of knowledge distribution within a particular domain.

This can be viewed in the conceptual diagram, Figure 1, where the full knowledge needed for a particular domain is represented by the normal distribution. The choice of the normal distribution is to highlight that, while the vast majority of the knowledge required is in the core of the domain, the potential knowledge needed is actually infinite. Therefore, at no time can either a person or computer know everything required. It also shows that the further you move away from the core domain knowledge the less relevant and less likely that knowledge will actually be required. It can be seen in this diagram that the KBS generally has a very poor coverage of the general knowledge beyond the core domain, as well as, an incomplete knowledge base for the domain itself. This is primarily due to the inherit difficulties of current KA methodologies being unable to extract all of the experts knowledge.

While researchers and designers have sought methods of improving the knowledge coverage of KBSs, it should be realised that a *perfect* KBS in a domain may never be achieved. Therefore, researchers have been simultaneously investigating methods for checking if a knowledge-base is *complete* (as complete as possible for a KBS), commonly referred to as validation and verification [3]. However, they tend not to be particularly useful or cover a large range of data patterns. The larger the data pattern the more frequently the expert will be asked to speculate about hypothetical situations that they may have no experience [1, 4, 5] and that may never occur in reality.

Rather than attempting to investigate the completeness of the knowledge base as a whole; this paper presents a rarely attempted approach of identifying inadequacies in the knowledge base for only the case currently being presented. Therefore, when a case is presented, which the system detects as being beyond its current ability to accurately assess, it provides a warning. This is a highly useful tool in systems where the knowledge base is being constructed incrementally by the expert. Multiple Classification Ripple Down Rules (MCRDR)[6-8] is one such system.

MCRDR has previously been shown to be a highly effective incremental learning Knowledge Based System (KBS)[6-8]. It allows a domain expert to add rules online by providing justifications identifying the differences between cases within the context provided. The one major inherent problem with MCRDR is that the human expert must be in a position to review each and every case to ensure that it is classified correctly. This can be excessively time consuming, especially in later stages of development when there are only a tiny percentage of misclassifications requiring correction. Therefore, the aim of this work was to reduce the need for case review by the expert without reducing the system's accuracy.

Thus, this paper describes an augmented hybrid system, referred to as Rated MCRDR (RM) and how it can be applied to Knowledge Acquisition Warnings (KAW). In this system, the conclusions and their justifications from MCRDR for each case are fed into a purpose built resource-allocating radial basis function (RBF) neural network and trained using the single-step-Δ-update-rule [9]. RM is capable of classifying cases into single or multiple classifications. Additionally, RM has been shown to be able to provide a prediction or evaluation for continuous value ranges [9]. However, this paper will specifically investigate RM's ability to predict the likelihood that a classification or value prediction by the system is correct.

Previously, a system was developed for predicting errors in the single classification predecessor to MCRDR, known as Ripple Down Rules (RDR)[1]. The approach taken was based simply on the idea of keeping a record of every case that passed through the system and then issuing a warning when a case presented was sufficiently different from all the previously seen cases within the particular classification path followed [1]. The system was tested using a simulated expert on three datasets. The best results were found on the Garvan dataset, where the system was able to predict approximately 92% of errors. However, it achieved this by provided a warning on 17% of cases, where there were actually only 2.6% of errors [1]. While this reduces the load on the expert significantly, there is still much opportunity for further improvement. In conclusion the system was found to be a "…major advance" [1], and potentially a "…useful aid in incremental knowledge acquisition"[1]. However, its performance was not sufficient by itself for most applications.

As will be discussed further in this paper, preliminary testing of RM has been able to significantly improve on these results, with the added advantage of not being required to store, retrieve and compare every case; significantly reducing the system's memory and computational load on the computer. The assumption behind RMs Knowledge Acquisition Warning (RM-KAW) technique is that if a case presented to the system follows a significantly different pattern-of-paths (POPs) through the MCRDR structure, then the expert should be warned of a potential error. Therefore, the system is using the structure of the MCRDR n-ary tree itself to determine when a case is outside its experience and, thus, requiring the expert to verify the conclusion.

Effectively, RM-KAW is keeping track of the knowledge the system has for a particular domain and, thus, tries to identify when a case being classified or rated is outside that area. For instance, it identifies when a case is using knowledge from outside the KBSs area of knowledge, figure 1. This allows the system to slowly expand its knowledge frontier into very specific specialised areas of a domain, along with gathering general background information. This provides a method for significantly reducing the brittleness of a knowledge based system.

## Multiple Classification Ripple Down Rules (MCRDR)

MCRDR uses an n-ary tree where each node contains a rule. Inference is performed by passing each case to the root node, which in turn feeds it on to any children with rules that evaluate it to *true*. Thus, the case continues to ripple down, level by level, until either a leaf node is reached or all of the child rules evaluate to false. Due to the fact that any, or all, of a node's children have the potential to fire, the possibility exists for a number of conclusions or classifications to be reached by the system for each case presented [8]. The system then lists the collection of classifications and the paths they followed.

Knowledge Acquisition is achieved in the system by inserting new rules into the MCRDR tree when a misclassification has occurred. The new rule must allow for the incorrectly classified case, identified by the expert, to be distinguished from the existing stored cases that could reach the new rule [10]. This is accomplished by the user identifying key differences between the current case and one of the earlier cornerstone cases. Where, a cornerstone case is any case that was used to create a rule and was also classified in the parent's node, or one of its child branches, of the new node being created. This is continued for all stored cornerstone cases, until there is a composite rule created that uniquely identifies the current case from all of the previous cases that could reach the new rule. The idea here is that the user will select differences that are representative of the new class they are trying to create [10].

## Rated MCRDR (RM)

Individual classifications in MCRDR, however, are all uniquely derived with no consideration for what other classification paths may have also been followed. Thus, there is no cohesion between any of the classifications found, however, the fact that this case was classified in these classes; means there must be either a conscious or subconscious relationship between these cases in the experts mind. The intention of RM is to try to capture these relationships between various classifications that may exist, either consciously or otherwise. If we can identify a set of relative values for the various relationships, this information could be used to improve the functionality available to the user. In the case of RM-KAW this value can be used to identify a confidence-*like*-factor. This could then be used to identify when the system should issue a warning that the case is unfamiliar and the conclusions should be checked by the expert.

## Implementation

Firstly, looking at what RM must accomplish mathematically, it can be seen that the output from the MCRDR methodology is essentially a set of classifications, denoted *C*, where $C \in \wp(C^*)$, and $C^*$ is the set of all possible classifications. The output from the RM engine is a set of values, $\bar{v}$, to provide one or more varying results in applications where dissimilar tasks may need to be rated differently. For instance, $v_0$, may identify the desirability, importance or confidence in its own classification for the case presented. Therefore, a mapping must be found from the set $C_\bar{v}, \forall C \in \wp(C^*)$. Additionally, RM should be able to learn this mapping for both linear and non-linear sets of classifications quickly and be able to generalise effectively.

Thus, RM needs to identify patterns of classifications and then associate a value for each pattern. While there are a number of techniques used for pattern recognition, in this implementation of RM an Artificial Neural Network (ANN) was selected, primarily because of its adaptability, ease of application to the problem domain, and because pattern recognition is one of the dominating areas for the application of ANN's [11].

The neural network was integrated into MCRDR by linking each possible rule or class to an input neuron. Then, for each rule or classification found by the MCRDR system, an associated neuron will fire. In this implementation of RM a purpose built resource-allocating radial basis function (RBF) network was used. The output nodes use the standard sigmoid thresholding function, equation 1, using a *modified* generalised delta rule. A subset of the input nodes is selected by the hidden layer by using the Gaussian function, equation 2, where the distance measure *r* is taken to be Euclidean, equation 3.

$$f(net) = 1 \Big/ \left(1 + e^{-k\,net}\right) \tag{1}$$

$$\phi(r) = e^{\left(-r^2\right)} \tag{2}$$

$$\|x - y\| = \sum_i \left(x_i - y_i\right)^2 \tag{3}$$

There are three possible methods for the association of neurons to the MCRDR structure: the Class Association method (CA), the Rule Association method (RA) and the Rule Path Association method (RPA). The different methods arise from the possibility of many paths through the tree that result in the same class as the conclusion. The *class association* method, where each unique *class* has an associated neuron, can reduce the number of neurons in the network and potentially produce faster, but possibly less general, learning. The *rule association* method, where each *rule* has an associated neuron and only the terminating rule's neuron fires, allows for different results to be found for the same class depending on which path was used to generate that class as the conclusion. Therefore, it is more capable of finding variations in meaning and importance within a class than may have been expected by the user that created the rules. The *rule path association* method, where all the *rules* in the *path* followed, including the terminating rule, cause their associated neuron to fire, would be expected to behave similarly but may find some more subtle results learnt through the paths rules, as well as being able to learn meaning hidden within the paths themselves.

Thus, the full RM algorithm, given in pseudo code in Figure 2 and shown diagrammatically in Figure 3, consists of two primary components. Firstly, a case is pre-processed to identify all of the usable data elements, such as stemmed words or a patient's pulse. The data components are presented to the standard MCRDR engine, which classifies them according to the rules previously provided by the user. Secondly, for each rule or class identified an associated input neuron in the neural network will fire. The network finally produces a set of outputs, $\bar{v}$ , for the case presented. The system, therefore, essentially provides two separate outputs; the case's classifications and the associated set of values for those classifications.

For example, in Figure 3, the document {a b b a c f i} has been pre-processed to a set of unique tokens {a, b, c, f, i}. It is then presented to the MCRDR component of the RM system, which ripples the case down the rule tree finding three classifications: *Z*, *Y*, and *U*; from the terminating rules: 1, 5, and 8. In this example, which is using the RA method, the terminating rules then cause the three associated neurons to fire and feed forward through the neural network producing a single value of 0.126. Thus, this document has been allocated a set of classifications that can be used to store the document appropriately, plus a value, which in the case of RM-KAW, can be used as a measure of confidence in the systems conclusion.

1. **Pre-process Case**
   - Initialise Case *c*
   - *c* ← Identify all useful data elements.

2. **Classification**
   - Initialize *list l* to store classifications
   - Loop
     - If child's rule evaluates Case c to *true*
       - *l* ← goto step 2 (generate all classifications in child's branch).
   - Until no more children
   - If no children evaluated to true then
     - *l* ← Add this nodes classification.
   - Return *l*.

3. **Rate Case**
   - $\bar{i}$ ← Generate input vector from *l*.
   - NN ← $\bar{i}$
   - v ← *NN output value.*

4. **Return RM evaluation**
   - Return list *l* of classifications for case *c* and
   - Value *v* of case *c*.

**Figure 2:** Inference Algorithm for RM.

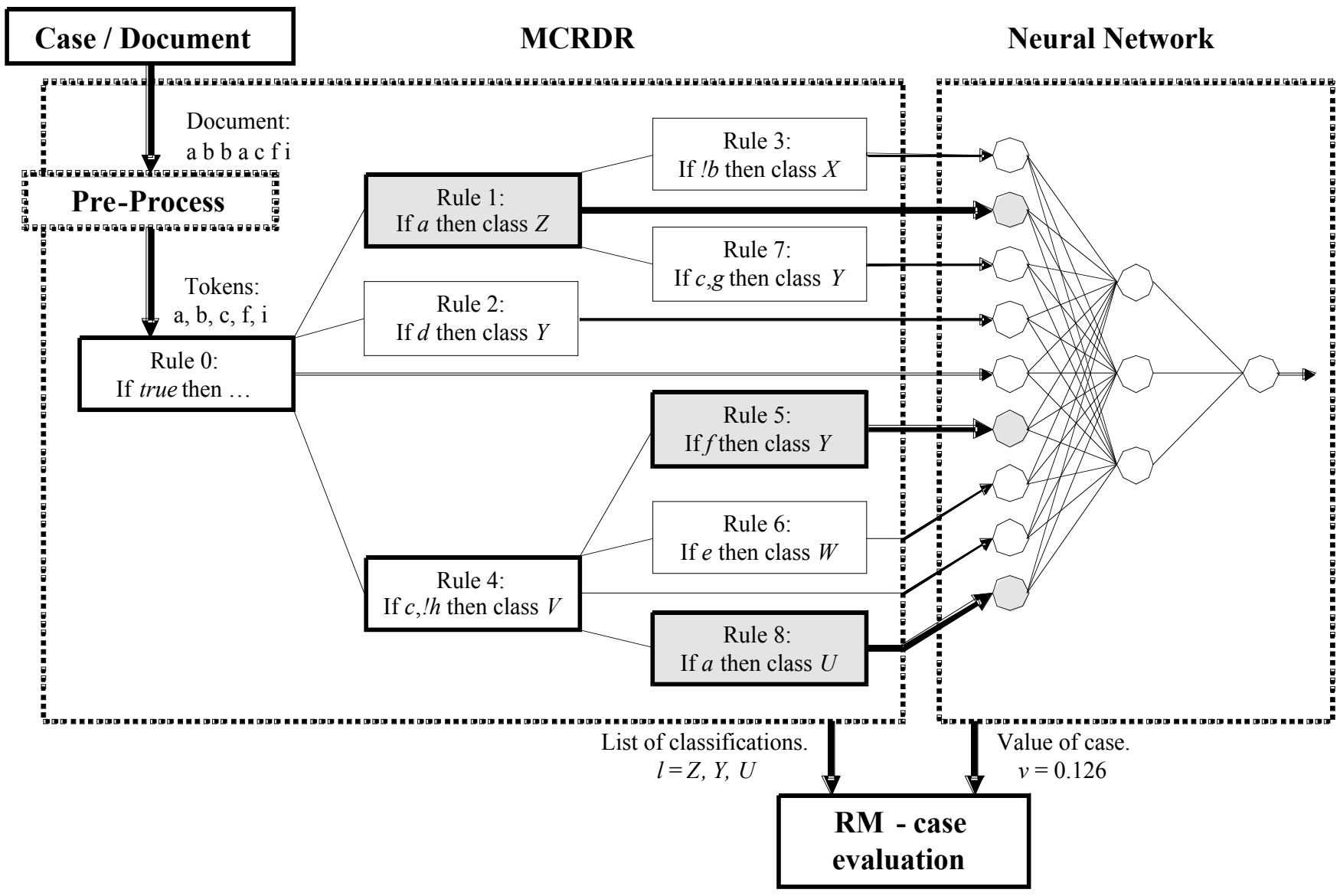

**Figure 3:** RM illustrated diagrammatically.

### Learning in RM

Learning in RM is achieved in two ways. Firstly, the rating component receives feedback from the environment concerning the accuracy of its predicted rating. Thus, a system using RM must provide some means of either directly gathering or indirectly estimating the correct rating. For example, in RM-KAW feedback with a high value may be given if the conclusion produced by the system is not changed by the expert. Secondly, the MCRDR component still acquires knowledge in the usual way; by the user identifying incorrect classifications and creating new rules and occasionally new classifications.

### Neural Network Structure and Learning

The neural network selected in developing RM for KAW was the standard radial basis function where the output nodes use the standard sigmoid thresholding function, equation 1, and the hidden layer uses the Gaussian function, equation 2, with a Euclidean distance measure, equation 3. Further experimentation still needs to be carried out to determine if other functions would provide better results. For instance, due to the discrete inputs currently used, it is expected that a broader fitness function at the hidden layer may provide better generalisation.

Some variations to the standard RBF network were required, due to the creation of additional rules and classifications, described earlier. Primarily, the capability to increase its number of input nodes to ensure one input node for each possible rule or class. Likewise, due to the growing nature of the input space, it can be expected that the number of significant patterns would also increase. Thus, a system was also developed for automatically allocating addition resources, in the form of hidden nodes.

A new input node is added to the system only when the user has corrected a conclusion. This, therefore, also means the user has identified a new significant pattern (whatever the new input vector is after the correction), which should automatically be captured in the hidden layer. Thus, a new hidden node is added with appropriate input weights assigned that will produce a Euclidean distance of zero only for that input sequences. However, the weight assigned to determine the contribution of this pattern to the overall confidence of future cases also must be found.

The simplest approach is to assign a random start up weight in the same fashion used when first initialising the network. However, we already have an accurate measure of the confidence for the new pattern, because the expert just created the new conclusion. Thus, we can be reasonably sure it is correct and assign the maximum confidence level for the new pattern.

Furthermore, in this implementation, additional hidden node resources are also added even when input nodes aren't added. This is done when a significant error is found, generally when a warning was made and the user decided that the conclusion was correct. If, when such a situation occurs, the system will check to see whether there are any nodes providing a reasonably close representation of the input pattern and if not a new hidden node is added with a Euclidean distance of zero. Once again a valid estimate can be made from the user's behaviour, providing a recommended value for the new patterns contribution to the system's overall confidence.

**Single-Step-_-Update-Rule**

In order to calculate the weight needed for a new connection from a just created hidden node and each output node, $w_{no}$, the system must first calculate the error in the weighted-sum, _$_{ws}$. This is then divided by the input at the newly created hidden node, $h_n$, which is always one in this implementation due to it being assigned a Euclidean distance of zero for the given input vector, where there are $n>0$ hidden nodes and $o>0$ output nodes.

$$w_{no} = \frac{\delta_{ws}}{x_n} \quad \textbf{(4)}$$

_$_{ws}$ is calculated by first deriving the required weighted-sum, $R_{ws}$, from the known error, _, and subtracting the actual weighted-sum, denoted by *net*.

$$\delta_{ws} = R_{ws} - net \quad \textbf{(5)}$$

The value of *net* for each output node, *o*, was previously calculated by the network during the feed forward operation, and is shown in Equation 6, where there are $n>1$ hidden nodes and the $n^{th}$ hidden node is our new input node.

$$net_o = \left( \sum_{h=0}^{m} x_h w_{ho} \right) \qquad \textbf{(6)}$$

$R_{ws}$, can be found for each output node, by reversing the thresholding process that took place at the output node when initially feeding forward. This is calculated by finding the inverse of the sigmoid function, Equation 1, and is shown in Equation 7, where *f(net)* is the original thresholded value that was outputted from that neuron.

$$R_{ws_o} = \frac{\log\left[\frac{f(net)_o + \delta_o}{1-\left(f(net)_o + \delta_o\right)}\right]}{k} \qquad \textbf{(7)}$$

Thus, the full *single-shot-_-update-rule*, used for each new hidden node's connection with each output node is given in Equation 8. Due to the use of the sigmoid function, it is clear that at no time can the system try and set the value of the output to be outside the range *0 > (f(net) + _) > 1* as this will cause an error.

$$w_{no} = \frac{\left(\log\left[\frac{f(net)_o + \delta_o}{1-\left(f(net)_o + \delta_o\right)}\right] \Big/ k\right) - \left[\sum_{h=0}^{m} x_h w_{ho}\right]}{x_n} \qquad \textbf{(8)}$$

## Testing with a Simulated Expert

The problem with testing a system, such as RM, is the use of expert knowledge that cannot be easily gathered without the system first being applied in a real world application. This is a similar problem that has been encountered with the testing of any of the RDR methodologies [6, 12, 13]. Thus, these systems built their KB incrementally through the use of a simulated expert. The simulated expert essentially provided a rule trace for each case run through another KBS with a higher level of expertise in the domain than the RDR system being trained [6, 8].

Thus, in order to test the rating component of the system, while still being able to create a KB in the MCRDR tree, a simulated expert was also developed for RM, with the ability to both classify cases, as well as form an opinion as to a case's overall importance. Basically, the simulated expert randomly generates weights, representing the level that each possible attribute in the environment contributes to each possible class, which is used to define rules for the MCRDR tree.

The environment then created many sets of documents consisting of randomly generated collections of attributes and passed each set to the RM system for classification. When classifying each case the case also gauged the level of confidence the system has in its conclusion. The test carried out at this stage has not used this confidence directly. Instead, it simply gathered statistics on how accurate its predictions actually were. Thus, the simulated user actually still checked every case, ignoring any warnings, and created new rules when ever it found a case incorrectly classified. Rewards were given to the RM system depending on what it predicted and the action taken by the expert. In testing the system, assumptions were made that the expert's interest in a case could be accurately measured and that the behaviour was constant, without noise or concept drift.

## Results and Discussion

In the preliminary test discussed in this paper, one of the following four details where recorded for each case as it was processed:

- If the conclusion was correct and RM gave a warning then a *False Positive* was recorded.
- If the conclusion was correct and RM did not give a warning then a *True Negative* was recorded.
- If the conclusion was incorrect and RM gave a warning then a *True Positive* was recorded.
- If the conclusion was incorrect and RM did not give a warning then a *False Negative* was recorded.

In Figure 4, the black line shows how many times the user corrected rules overtime. While the white line shows when a rule that the user corrected was warned about by RM; *true positives*. It can clearly be observed that RM was able to identify nearly all, approximately 97.73%, incorrect classifications. Therefore, these results indicate that an expert using this system to identify when a case is misclassified may be confident that the majority of errors would generate a warning.

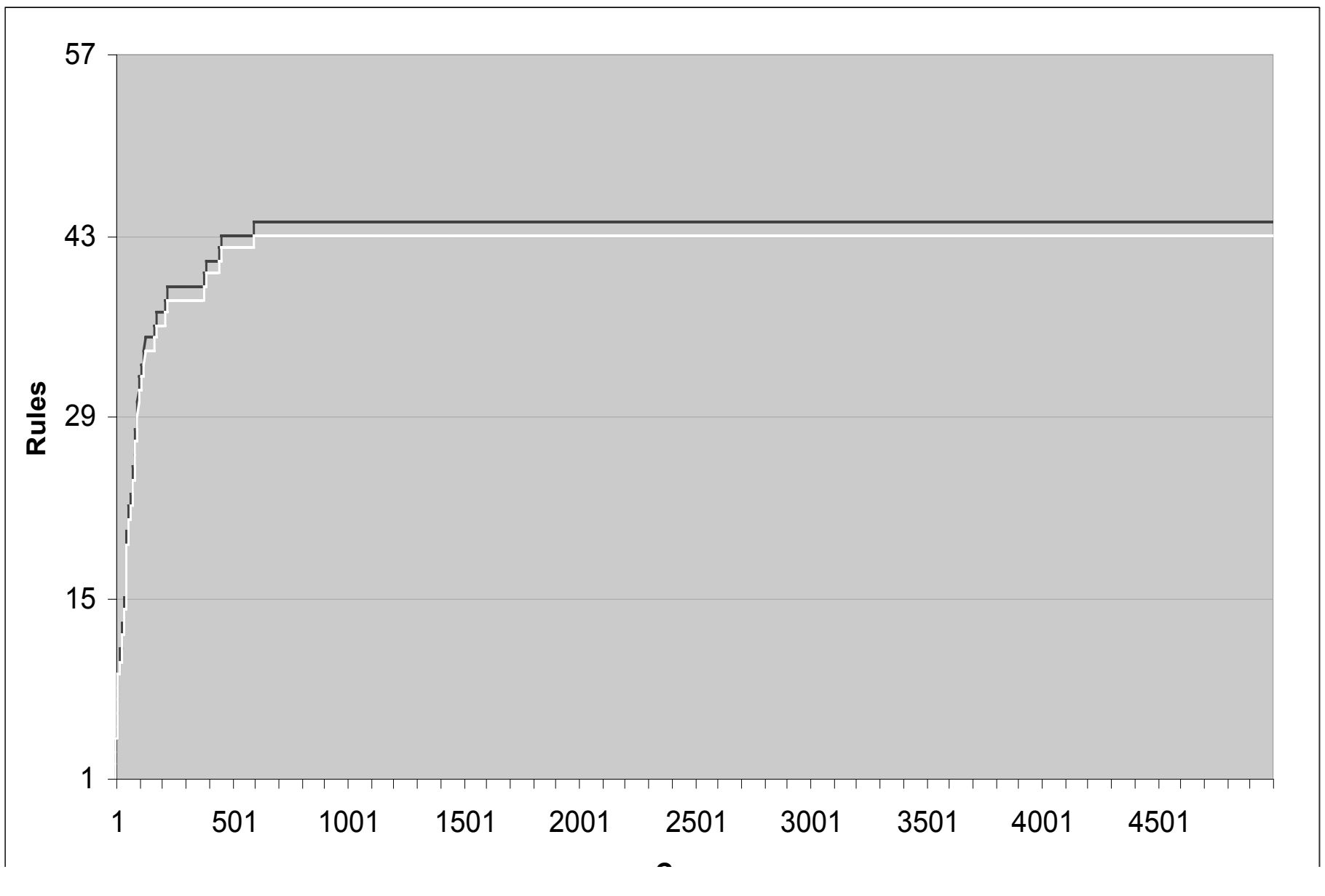

**Figure 4: Comparison of the total rules created and the total rules first warned about and then created.**

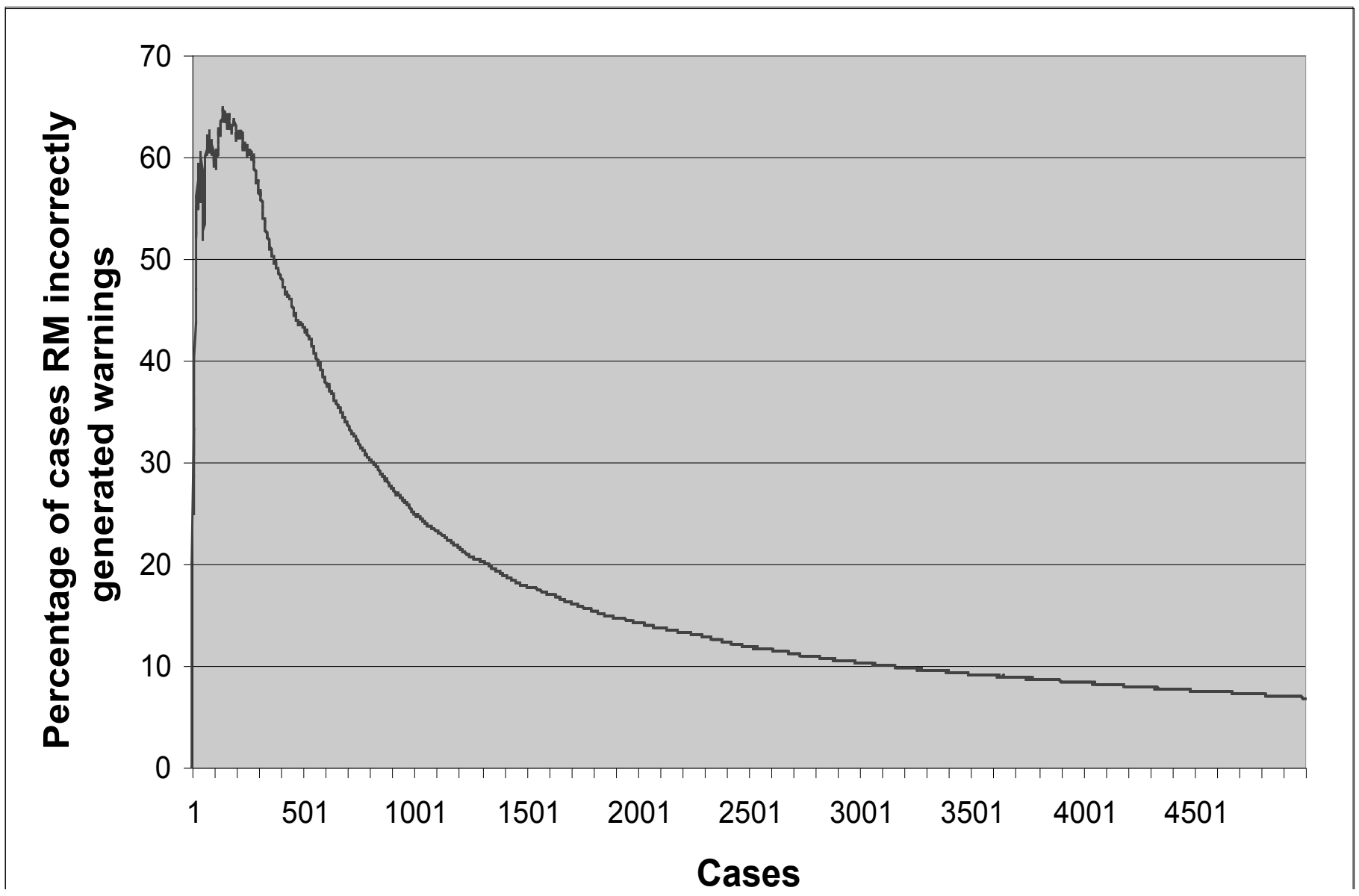

**Figure 5: Percentage of cases that recieved a warning over time.**

However, if the system simply provides a warning for every case then this result is meaningless. Therefore, RM would also need to minimise the amount of warnings generated when a case is correct, without reducing the amount of correct warnings. Figure 5, illustrates how RM is highly successful at reducing the amount of incorrect warnings, down to only 6.8% after 5000 cases, being generated as the knowledge base grows and RM learns. Furthermore, it has accomplished this without loosing its ability to identify the incorrect cases.

While these results are only preliminary, and direct comparisons with Compton et al's [1] work of identifying warnings in RDR cannot be made at this stage, they do show that RM has lots of potential when used for generating warnings. With 97.73 % accuracy and only generating warnings in less than 8 % of cases shows that it could possibly be used for knowledge acquisition and reducing the load on the user having to review every case. It would be particularly useful after the majority of the knowledge base has been formed and we are primarily interested in only identifying those rare cases that require knowledge from outside the core domain already known to the system.

## Conclusion and Future Work

The system described in this paper was developed to provide a means for identifying knowledge that sits outside the current knowledge held by the knowledge based system. The system was designed to learn which patterns of paths followed were likely to be correct and which were unusual. When an unusual classification pattern was found then the system provides a warning bring the case to the user's attention, allowing them to verify the correctness of the conclusions found. The system used Multiple Classification Ripple Down Rules (MCRDR) as its incremental Knowledge Based System. The patterns of rules followed in generating a set of classifications, then fed into a purpose built resource allocating radial basis function neural network using the single-step-Δ-update-rule for faster learning.

The system has undergone preliminary testing with a simulated expert using a randomly generated data set. These tests were done primarily to show that the system was able to learn quickly and to be used for parameter tuning purposes. Clearly, a more rigorous testing regime needs to be used in order to fully justify the algorithm's ability to learn. Additionally, testing using data sets used by Compton et al's work and real world user tests need to also be performed.

## Acknowledgements

The authors would like to thank Dr Ray Williams and Mr Robert Ollington for their helpful discussions.

# Author Index